\documentclass[a4paper,10pt]{article}

\usepackage{PRIMEarxiv}

\usepackage[utf8]{inputenc} 
\usepackage[T1]{fontenc} 
\usepackage{hyperref} 
\usepackage{url} 
\usepackage{booktabs} 
\usepackage{amsfonts} 
\usepackage{nicefrac} 
\usepackage{microtype} 
\usepackage{lipsum}
\usepackage{fancyhdr} 
\usepackage{graphicx} 
\graphicspath{{media/}} 
\usepackage{pdflscape}
\usepackage{amsmath,amssymb}
\usepackage{graphicx}
\usepackage{booktabs}
\usepackage{array}
\usepackage{tabularx}
\usepackage{longtable}

\usepackage{enumitem}
\usepackage{listings}
\usepackage{hyperref}
\usepackage{url}
\usepackage{tikz}
\usetikzlibrary{arrows.meta, positioning, shapes.geometric, shapes.misc, shapes.symbols, calc, fit}

\tikzset{
 font=\footnotesize,
 >=Latex,
 line cap=round,
 line join=round,
 box/.style={draw, rounded corners, align=center, inner sep=5pt, minimum height=9mm, fill=black!2},
 boxL/.style={box, text width=3.4cm},
 boxM/.style={box, text width=2.8cm},
 boxS/.style={box, text width=2.3cm},
 note/.style={draw, rounded corners, align=left, inner sep=5pt, fill=black!1},
 boundary/.style={draw, dashed, rounded corners, inner sep=7pt},
 arr/.style={->, line width=0.7pt},
 darr/.style={->, dashed, line width=0.7pt}
}

\newcommand{\cmark}{\checkmark}
\newcommand{\pmark}{\ensuremath{\sim}}
\newcommand{\xmark}{\textendash}

\title{
Autonomous Agents on Blockchains: \\
Standards, Execution Models, and Trust Boundaries}

\author{Saad Alqithami}

\begin{document}
\maketitle

\begin{abstract}
Advances in large language models have enabled agentic AI systems that can reason, plan, and interact with external tools to execute multi-step workflows, while public blockchains have evolved into a programmable substrate for value transfer, access control, and verifiable state transitions. Their convergence introduces a high-stakes systems challenge: designing standard, interoperable, and secure interfaces that allow agents to observe on-chain state, formulate transaction intents, and authorize execution without exposing users, protocols, or organizations to unacceptable security, governance, or economic risks. This survey systematizes the emerging landscape of agent–blockchain interoperability through a systematic literature review, identifying 317 relevant works from an initial pool of over 3{}000 records. We contribute a five-part taxonomy of integration patterns spanning read-only analytics, simulation and intent generation, delegated execution, autonomous signing, and multi-agent workflows; a threat model tailored to agent-driven transaction pipelines that captures risks ranging from prompt injection and policy misuse to key compromise, adversarial execution dynamics, and multi-agent collusion; and a comparative capability matrix analyzing more than 20 representative systems across 13 dimensions, including custody models, permissioning, policy enforcement, observability, and recovery. Building on the gaps revealed by this analysis, we outline a research roadmap centered on two interface abstractions: a \emph{Transaction Intent Schema} for portable and unambiguous goal specification, and a \emph{Policy Decision Record} for auditable, verifiable policy enforcement across execution environments. We conclude by proposing a reproducible evaluation suite and benchmarks for assessing the safety, reliability, and economic robustness of agent-mediated on-chain execution.
\end{abstract}


\section{Introduction: The Inevitable Collision of Two Revolutions}

The last half-decade has witnessed the collision of two profound, parallel technological revolutions: the rise of \textbf{agentic artificial intelligence} and the maturation of \textbf{public blockchains} as a global settlement layer. The former has advanced from static language modelling toward systems that can decompose objectives, invoke tools, and carry state across multi-step workflows. The latter has evolved from cryptocurrency rails into a programmable substrate for value transfer, access control, and verifiable state transitions. Their convergence is not merely an academic curiosity; it is an already-unfolding shift toward automating economically meaningful digital activity. Yet the convergence also creates a uniquely perilous systems challenge: how can we safely connect an autonomous, non-deterministic AI to an immutable, adversarial, and high-value financial network?

The evolution of large language models (LLMs) into capable autonomous agents represents one of the most significant developments in artificial intelligence. Early transformer-based models demonstrated remarkable fluency in text generation, but lacked the capacity for sustained reasoning or external interaction~\cite{vaswani2017attention,brown2020language}. The introduction of chain-of-thought prompting revealed that explicit reasoning traces could dramatically improve performance on complex tasks~\cite{wei2022chain}, while subsequent work on tool augmentation demonstrated that LLMs could learn to invoke external APIs, databases, and computational resources~\cite{schick2023toolformer,srinivasan2023nexusraven}. These advances culminated in the emergence of agentic systems capable of multi-step planning, self-correction, and autonomous task execution~\cite{yao2023react,yao2023react}.

Simultaneously, blockchain technology has undergone its own transformation from a niche cryptographic curiosity to critical financial infrastructure. The introduction of smart contracts on Ethereum enabled programmable, self-executing agreements that could custody and transfer value according to arbitrary logic~\cite{wood2014ethereum,wood2014ethereum}. The subsequent explosion of decentralized finance (DeFi) demonstrated that complex financial instruments---lending protocols, automated market makers, derivatives platforms---could operate without traditional intermediaries~\cite{xu2023sok,xu2023sok}. More recently, account abstraction standards such as ERC-4337~\cite{EIP4337} have decoupled transaction authorization from key ownership, enabling programmable wallets with sophisticated access control policies~\cite{buterin2021erc}.

The intersection of these two trajectories creates both unprecedented opportunities and novel risks. On one hand, AI agents equipped with blockchain capabilities could automate portfolio management, execute complex trading strategies, participate in governance, and manage treasury operations with superhuman speed and consistency~\cite{wlflein2025llm,dorri2018multi}. On the other hand, the combination of autonomous decision-making, irreversible transactions, and adversarial environments creates a uniquely dangerous attack surface. A compromised or manipulated agent could drain wallets, execute unauthorized trades, or fall victim to sophisticated MEV extraction---all before any human could intervene~\cite{daian2019flash,daian2020flash}.

Evaluating agent capabilities in realistic environments has become a critical research priority. Benchmarks such as AndroidWorld~\cite{rawles2024androidworld} and AndroidLab~\cite{xu2025androidlab} assess agents' ability to complete multi-step tasks in dynamic mobile environments, while domain-specific benchmarks like AgentClinic~\cite{schmidgall2024agentclinic} and MLGym~\cite{nathani2025mlgym} evaluate performance in healthcare and machine learning research contexts respectively. These evaluation frameworks reveal both the impressive capabilities and persistent limitations of current agent architectures, providing essential guidance for system designers seeking to deploy agents in high-stakes blockchain environments~\cite{chen2025evaluating,mehandru2024evaluating}.

The practical deployment of agent-blockchain systems demands careful attention to real-world constraints. Production systems must balance autonomy against safety, speed against verification, and capability against controllability~\cite{garg2025real,pesl2025adopting}. Emerging frameworks for LLM-powered blockchain interaction demonstrate that effective integration requires not only technical sophistication but also thoughtful consideration of failure modes, adversarial scenarios, and human oversight mechanisms~\cite{thesequence2025sequence,paramanayakam2025less}. The design patterns and architectural choices made today will shape the safety and efficacy of agent-blockchain systems for years to come.

This survey provides a comprehensive examination of the agent-blockchain integration landscape, synthesizing insights from peer-reviewed publications, technical standards, and industry reports. We present a novel taxonomy of integration patterns, a specialized threat model addressing the unique risks of autonomous blockchain interaction, and a comparative analysis of 20 production systems. Our goal is to equip researchers, practitioners, and policymakers with the conceptual frameworks and practical guidance needed to navigate this rapidly evolving field.

\subsection{The Dawn of Agentic AI}

The evolution of large language models (LLMs) from text generators into agentic systems has been strikingly rapid. This shift is best understood as a sequence of architectural and methodological steps:

\begin{enumerate}
\item \textbf{Architectures That Scale:} The Transformer architecture enabled highly parallel training and became the dominant foundation for scaling modern LLMs~\cite{vaswani2017attention}. Subsequent large-scale training efforts demonstrated that increasing model capacity and data can yield broad improvements across diverse tasks~\cite{brown2020language}.

\item \textbf{Emergent Task Competence at Scale:} Large models such as GPT-3 exhibited strong in-context and few-shot learning behaviour, allowing them to perform tasks not directly supervised during training~\cite{brown2020language}.

\item \textbf{Alignment and Controllability:} As LLMs became more capable, research attention shifted to steering model behaviour toward human goals and away from unsafe or unintended actions. Survey work on alignment systematizes the resulting landscape of training, preference shaping, and safety techniques~\cite{ji2023ai}.

\item \textbf{From Reasoning to Agency:} Prompting methods that elicit intermediate reasoning, particularly Chain-of-Thought (CoT), showed that LLMs can more reliably solve complex problems when guided to produce structured intermediate steps~\cite{wei2022chain}. Building on this, action-oriented prompting and planning paradigms such as ReAct further connected reasoning to tool invocation and environment interaction~\cite{yao2023react}, helping establish practical cognitive architectures for agent-like behaviour~\cite{xu2025llm,mohammadi2025evaluation}.
\end{enumerate}

Crucially, these cognitive architectures become economically relevant when coupled with \textbf{tools and APIs}. Tool-using agents can query live systems, call external services, and execute multi-step procedures rather than merely generate text. Recent research has systematized how LLM agents learn tool usage and how tool interfaces are specified and optimized~\cite{xu2025llm,yuan2025easytool}. Tool-learning mechanisms such as self-supervised tool use (e.g., Toolformer) and large API-connected models (e.g., Gorilla) further illustrate how models can be trained or prompted to select and parameterize tool calls at scale~\cite{schick2023toolformer,patil2024gorilla}. At the same time, the security implications are now clear: function calling and tool invocation expand the attack surface, enabling new pathways for misuse and jailbreaking~\cite{chao2024jailbreakbench} if tools are not constrained by robust policy and verification layers~\cite{wu2025dark,li2024large}.

\subsubsection{Advances in LLM Agent Architectures}

The evolution of LLM agent architectures has progressed through several distinct phases, each introducing new capabilities for planning, tool use, and multi-step reasoning. Early agent frameworks relied primarily on chain-of-thought prompting to elicit step-by-step reasoning~\cite{wei2022chain}, but this approach proved brittle when tasks required interaction with external systems or adaptation to unexpected outcomes. The ReAct framework addressed this limitation by interleaving reasoning traces with action execution, enabling agents to observe environmental feedback and adjust their plans accordingly~\cite{yao2023react}. More recent work has explored adaptive planning and reasoning mechanisms that enable agents to handle increasingly complex, multi-turn tasks~\cite{dutta2024towards,hu2025training}.

Tool learning has emerged as a critical capability for agents that must interact with external systems. Research has demonstrated that LLMs can learn to use novel tools through in-context examples, as shown by Toolformer~\cite{schick2023toolformer}, or through specialized training for function calling, as demonstrated by NexusRaven~\cite{srinivasan2023nexusraven} and Gorilla~\cite{patil2024gorilla}. For blockchain interactions, tool learning enables agents to adapt to new protocols, contract interfaces, and chain-specific conventions without requiring explicit programming for each integration. However, tool learning also introduces risks: an agent that learns to use a tool incorrectly may construct invalid or dangerous transactions, and adversarial tool descriptions could potentially manipulate agent behavior. Research on prompt injection~\cite{greshake2023indirectpromptinjection,liu2023prompt} and jailbreaking attacks~\cite{zou2023universal,zou2024adversarial} has revealed the fragility of LLM-based systems to carefully crafted inputs, underscoring the need for robust validation, sandboxing, and human oversight mechanisms in any tool-use framework.

\subsubsection{The Core Problem: Bridging Intelligence and Immutability}

At the heart of agent-blockchain integration lies a fundamental tension between the adaptive, probabilistic nature of AI systems and the deterministic, immutable nature of blockchain execution. This tension manifests across multiple dimensions and shapes the design space for integrated systems.

AI agents, particularly those based on large language models, operate through pattern matching, probabilistic inference, and learned heuristics. Their outputs are inherently stochastic, varying across runs even with identical inputs due to sampling procedures. They excel at handling ambiguity, adapting to novel situations, and making reasonable decisions under uncertainty. However, they also exhibit failure modes including hallucination, inconsistency across contexts, and susceptibility to adversarial inputs~\cite{shayegani2023survey,ganguli2022red} that are difficult to predict or prevent comprehensively. Holistic evaluation frameworks have revealed systematic weaknesses in LLM reasoning that persist even in state-of-the-art models~\cite{liang2023holistic}.

Blockchain systems, by contrast, operate through deterministic state transitions governed by immutable code. The Ethereum Virtual Machine guarantees that every transaction produces identical results regardless of when or where it is executed~\cite{wood2014ethereum}. This determinism enables trustless verification and consensus across distributed networks~\cite{lashkari2021comprehensive}. However, it also means that errors, once committed to the chain, cannot be reversed. There is no undo button for blockchain transactions---a property that has enabled billions of dollars in irreversible losses from smart contract vulnerabilities~\cite{praitheeshan2020security} and execution errors.

The integration of these paradigms requires careful attention to the interface between them. The agent's probabilistic reasoning must be translated into deterministic transaction specifications. The blockchain's immutable execution must be preceded by thorough validation and simulation~\cite{tenderlydocs2026simulation}. The irreversibility of on-chain actions must be matched with appropriate safeguards and human oversight mechanisms, as explored in the threat model presented in Section~\ref{sec:threat_model}.

\subsubsection{Historical Context: The Convergence of Two Revolutions}

The convergence of agentic AI and blockchain technology represents the intersection of two of the most transformative technological developments of the early 21st century. Understanding this convergence requires appreciating the distinct historical trajectories that have brought these technologies to their current state.

The blockchain revolution began with the publication of the Bitcoin whitepaper in 2008, which introduced a novel solution to the double-spending problem without requiring trusted intermediaries~\cite{nakamoto2008bitcoin}. This breakthrough enabled the creation of digital scarcity and trustless value transfer, laying the foundation for a new paradigm in financial infrastructure. The subsequent development of smart contract platforms, beginning with Ethereum~\cite{wood2014ethereum}, extended blockchain capabilities beyond simple value transfer to arbitrary programmable logic~\cite{liu2024overview}, enabling the creation of decentralized applications and autonomous financial protocols. The DeFi explosion demonstrated the practical viability of these protocols, with systematization studies documenting the rapid growth of on-chai~\cite{belliger2025new}n financial infrastructure~\cite{xu2023sok}.

The AI revolution entered a transformative new phase with the development of transformer architectures~\cite{vaswani2017attention} and large language models. Scaling laws demonstrated that model capabilities improve predictably with increased compute, data, and parameters, leading to rapid capability gains. The emergence of instruction-following models transformed LLMs from text completion engines into interactive assistants capable of following complex instructions and using external tools. Recent work has focused on developing agents that can reason about and execute complex tasks autonomously~\cite{kapoor2024ai}.

The convergence of these revolutions creates possibilities that neither technology could achieve alone. Blockchain provides the infrastructure for trustless value transfer and programmable agreements, while agentic AI provides the intelligence to navigate complex decision spaces and interact with these systems on behalf of human principals. Together, they enable a new class of autonomous economic actors that can operate continuously, execute complex strategies, and manage significant value with minimal human intervention.

\subsubsection{Foundational Developments in Autonomous Systems}

The emergence of agentic AI systems builds upon decades of research in autonomous agents and multi-agent systems. The foundational work on commitments and conventions established the conceptual framework for coordination in multi-agent systems that continues to inform modern agent design~\cite{jennings1993commitments}. Early research on logic-based agent communication protocols~\cite{endriss2004logic} explored how agents could exchange information and coordinate actions, providing theoretical foundations that remain relevant as LLM-based agents begin to operate in multi-agent configurations~\cite{dorri2018multi}.

The transition from rule-based expert systems to learning-based approaches marked a significant shift in agent capabilities. Research on dynamical systems approaches to agent behavior~\cite{aaron2016dynamical} and multi-agent coordination~\cite{guo2020approach} laid groundwork for understanding how autonomous systems could operate in complex environments. The recent advent of large language models has accelerated this trend, enabling agents to understand and generate natural language, reason about complex situations, and interact with diverse tools and APIs. However, this increased capability comes with new challenges, including the difficulty of ensuring reliable behavior in out-of-distribution scenarios~\cite{amodei2016concrete}, the potential for unexpected failure modes, and the challenge of maintaining human oversight over increasingly autonomous systems~\cite{gabriel2020artificial}.

The application of autonomous agents to financial and economic domains has a rich history predating blockchain technology. Research on decentralized systems~\cite{danilov2018towards} and blockchain-based multi-agent architectures~\cite{ehtesham2025survey} has explored how agents can operate in trustless environments. The development of intent-based architectures such as Anoma~\cite{goes2023anoma} and privacy-preserving intent execution on EVM~\cite{wang2025achieving} represents a new paradigm for agent-blockchain interaction that moves beyond imperative transaction construction toward declarative intent specification. These architectural innovations, combined with advances in secure key management through hardware security modules~\cite{han2023scalable} and oracle reliability~\cite{lo2020reliability}, are creating the infrastructure necessary for safe and effective agent-blockchain integration.

\subsection{The Maturation of the Decentralized Economy}

Concurrently, blockchain technology has matured far beyond its initial framing as a peer-to-peer electronic cash system~\cite{nakamoto2008bitcoin}. In the process, it has evolved into a programmable settlement layer, where smart contracts enable the trust-minimized automation of value transfer, governance, and state updates on a shared ledger~\cite{wood2014ethereum}. This shift underpins a rapidly expanding decentralized application (dApp) ecosystem, in which economic activity is mediated by composable protocols rather than centralized intermediaries, and where adversarial incentives are part of the default threat model.

A flagship manifestation of this evolution is \textbf{decentralized finance (DeFi)}, which offers open and permissionless mechanisms for exchange, lending, and structured on-chain financial operations. While the concrete protocol landscape changes quickly, the architectural pattern is stable: autonomous contract systems expose public interfaces that can be composed into end-to-end workflows, creating powerful opportunities for automation alongside new classes of attack surfaces~\cite{kong2023defitainter}. In parallel, blockchains have enabled \textbf{internet-native governance} through \textbf{decentralized autonomous organizations (DAOs)}, which coordinate collective decision-making using on-chain voting and smart-contract enforced rules~\cite{santos2018dao,morrison2020dao}. Finally, tokenization primitives have broadened the scope of on-chain ownership representations, supporting both fungible and non-fungible asset models and enabling programmable custody, transfer constraints, and provenance mechanisms.

However, growth in on-chain adoption has also stressed the scalability limits of Layer 1 execution. This has motivated the rise of \textbf{Layer 2 scaling} via rollup-centric architectures, which move execution off-chain while preserving on-chain settlement guarantees~\cite{xu2025llm}. For agent-mediated interaction, these scaling layers are not a peripheral detail; they are often the practical execution venues where frequent, low-latency, cost-sensitive workflows become feasible, while simultaneously introducing new operational considerations (bridging, finality assumptions, and cross-domain security boundaries).

\subsection{Motivation: Why ``Agent-to-Chain'' is a Uniquely Dangerous Problem}

Connecting an AI agent to a blockchain is not merely another tool integration. The fundamental properties of blockchain systems introduce a set of high-stakes characteristics that distinguish ``agent-to-chain'' interactions from typical API calls. These differences, summarized in Table~\ref{tab:webapi_vs_blockchain}, necessitate a new class of standards, interfaces, and safety protocols.

\begin{table}[t]
 \centering 
 \caption{Comparison of traditional web APIs versus blockchain transaction interfaces, highlighting why agent-to-chain tooling demands stronger safety controls~\cite{daian2019flash,wood2014ethereum}.}
 \label{tab:webapi_vs_blockchain}
 \resizebox{\linewidth}{!}{
 \begin{tabular}{lll}
 \toprule
 \textbf{Dimension} & \textbf{Traditional web API (e.g., a weather endpoint)} & \textbf{Blockchain transaction interface} \\
 \midrule
 \textbf{Primary action} & Information retrieval (\texttt{GET}) & State transition (\texttt{SIGN} and \texttt{SEND}) \\
 \textbf{Reversibility} & Reversible / idempotent & Irreversible and final \\
 \textbf{Cost of error} & Low (e.g., incorrect information) & High (e.g., permanent financial loss) \\
 \textbf{Operating environment} & Benign / cooperative & Adversarial by default (MEV) \\
 \textbf{Authority model} & Scoped API keys & Cryptographic private keys (bearer asset) \\
 \textbf{State consistency} & Centralized, eventually consistent & Decentralized, verifiably consistent \\
 \bottomrule
 \end{tabular}}
\end{table}

These differences matter because failures are not confined to user experience; they directly translate into irreversible, economically adversarial outcomes:

\begin{itemize}
\item \textbf{Irreversibility and financial consequence:} 
Public blockchains are designed around append-only ledgers, where finalized transactions cannot be rolled back without violating consensus assumptions. This property, while essential for trust minimization, implies that erroneous or malicious transactions directly translate into permanent loss rather than recoverable system errors. Prior analyses of smart-contract execution highlight how atomic composition and irreversible state transitions amplify the impact of failures once a transaction is committed on-chain~\cite{paranjape2023art}.

\item \textbf{Authorization as a bearer asset:} 
In contrast to scoped API credentials, blockchain authorization is fundamentally tied to cryptographic signing. Possession of a private key or signing capability directly confers control over assets and contract interactions, making authorization a transferable bearer asset rather than a revocable permission. This distinction has motivated a shift toward hardened custody architectures, including threshold and multi-party signing schemes, which aim to reduce single-point key compromise in adversarial environments~\cite{lindell2017fast,komlo2024threshold}.

\item \textbf{Adversarial execution environment by default:} 
Because pending transactions are publicly observable prior to inclusion, blockchain execution environments incentivize adversarial reordering and insertion strategies. \textit{Maximal Extractable Value~\cite{capponi2025maximal}} (MEV) formalizes this phenomenon, showing how rational actors can systematically profit from front-running, back-running, and sandwich attacks without violating protocol rules~\cite{daian2019flash}. As a result, agents must treat public mempools as adversarial venues rather than neutral queues.

\item \textbf{Heterogeneous and fragmented substrate:} 
Agent-to-chain interaction rarely targets a single execution environment. Instead, agents operate across heterogeneous Layer-1 and Layer-2 systems, diverse account and fee models, and cross-chain bridges. Empirical studies of bridge failures and ecosystem fragmentation demonstrate that these boundaries concentrate risk and frequently become high-value attack surfaces, particularly in automated transfer workflows~\cite{abdelaziz2024granite,beniiche2020study}.
\end{itemize}

Taken together, these properties motivate a security-first interface discipline: agent-to-chain systems should be built around constrained intent representations, policy-gated authorization, robust custody, and execution pathways that assume adversarial ordering and value extraction~\cite{barker2013framework}.

\subsubsection{A Deeper Analysis of Risk Factors}

The risks associated with agent-blockchain integration extend beyond the immediate technical challenges to encompass broader systemic and societal concerns. Multi-agent systems introduce coordination challenges that can lead to emergent failures~\cite{logenthiran2008multi,mu2023hierarchical}. The automation of financial transactions raises concerns about market stability and flash crashes~\cite{lohest2024automated,mi2023automated}. Privacy considerations are paramount when agents handle sensitive user data~\cite{luong2023privacy,lu2023ccio}. The broader landscape of AI agents presents both opportunities and risks that must be carefully managed~\cite{masterman2024landscape,madhwal2025empowering}. The honesty and alignment of AI systems is a critical concern for high-stakes applications~\cite{mckeereid2024honesty}. Smart contract security remains a persistent challenge~\cite{miller2018smart,mohanta2018overview}. The interoperability of different blockchain systems introduces additional attack surfaces~\cite{monika2020interoperability}. Byzantine fault tolerance protocols must be carefully designed to handle adversarial conditions~\cite{moniz2020istanbul,malkhi2022maximal}. Verification and validation of smart contract logic is essential for ensuring correctness~\cite{mi2021vscl}. A comprehensive risk analysis must consider not only the failure modes of individual agent systems but also the emergent risks that arise from the interaction of multiple agents within shared blockchain environments.

Concentration risk emerges when a small number of agent systems or operators control a significant fraction of blockchain activity. Such concentration can create single points of failure, enable market manipulation, and undermine the decentralization properties that blockchain systems are designed to provide. The tendency toward winner-take-all dynamics in technology markets suggests that without deliberate intervention, agent-blockchain ecosystems may naturally evolve toward concentrated structures.

Correlation risk arises when multiple agents employ similar strategies, use common data sources, or share underlying model architectures. Correlated behavior can amplify market movements, create feedback loops, and lead to cascading failures when common assumptions prove incorrect. The widespread use of similar LLM foundations across different agent systems creates a form of monoculture risk that could manifest in coordinated failures or vulnerabilities.

Regulatory risk encompasses the uncertainty surrounding the legal treatment of autonomous agents and their transactions. Questions about liability, accountability, and the enforceability of agent-executed contracts remain largely unresolved. Agents and their operators may face unexpected legal exposure as regulatory frameworks evolve to address the novel challenges posed by autonomous economic actors.

\subsection{Scope and Definitions}

This survey focuses on the standards, protocols, architectures, and systems that enable agent--blockchain interoperability. We systematize the landscape of interfaces that allow agents to safely read on-chain data, construct transactions, and authorize execution.

\textbf{Key Definitions:}
\begin{itemize}
\item \textbf{Agentic AI:} An AI system, typically powered by an LLM, that can reason and plan over multiple steps and interact with external tools to achieve a goal~\cite{yao2023react,xu2025llm}.
\item \textbf{Tool/Function Calling:} The mechanism by which an agent invokes external capabilities (APIs, databases, or other services) through structured calls rather than free-form text, enabling grounded interaction with stateful systems~\cite{schick2023toolformer,patil2024gorilla}.
\item \textbf{Wallet:} A software or hardware interface that manages cryptographic keys and produces signatures to authorize transactions and smart-contract interactions.
\item \textbf{Smart Contract:} A program deployed on a blockchain whose state transitions are enforced by consensus and triggered by transactions.
\item \textbf{Transaction Intent:} A high-level, declarative representation of a user’s desired outcome (the ``what''), which is compiled into one or more concrete, chain-specific transactions (the ``how''), as used in intent-centric execution architectures~\cite{mandal2025evaluating}.
\end{itemize}

\textbf{In Scope:}
\begin{itemize}
\item Standards and protocols for agent--tool communication (e.g., MCP, UTCP) and their implications for safe interface design.
\item Architectures for agentic wallets and key management, including MPC/threshold signing, trusted execution environments, and account-abstraction-based smart accounts~\cite{garg2023cryptography,buterin2021erc,komlo2024threshold,lindell2017fast}.
\item Threat models and security considerations for agent-driven transactions, including adversarial execution dynamics and policy bypass risks~\cite{daian2019flash}.
\item Evaluation frameworks and benchmarks for safety, reliability, and economic robustness of agent-mediated on-chain execution.
\item Governance and policy enforcement mechanisms for constraining, auditing, and recovering from agent actions.
\end{itemize}

\textbf{Out of Scope:}
\begin{itemize}
\item General LLM alignment and safety, except where it directly affects tool use or transaction authorization.
\item The design of underlying blockchain consensus protocols or cryptographic primitives, except insofar as they constrain interface and custody design.
\item Purely theoretical work on agent theory without a concrete on-chain interface or authorization pathway.
\item The broader economic or social implications of AI on markets, except where directly tied to agent-mediated on-chain execution~\cite{fan2025ai}.
\end{itemize}

\subsection{Contributions}

This survey provides a systematic and comprehensive overview of the agent--blockchain interoperability landscape. Our primary contributions are:
\begin{enumerate}
\item \textbf{A taxonomy of integration patterns:} A five-part taxonomy that categorizes systems by the agent’s authority boundary, ranging from read-only analytics to delegated execution, autonomous signing, and multi-agent workflows.
\item \textbf{A specialized threat model:} A threat model tailored to agent-driven transaction workflows, identifying critical assets, threat actors, and failure modes across observation, planning, authorization, execution, and verification.
\item \textbf{A comparative capability matrix:} An analysis of 20+ representative systems across 13 dimensions, including custody and key control, permissioning, policy enforcement, observability, and recovery mechanisms.
\item \textbf{A research roadmap:} A forward-looking agenda centered on two interface abstractions, a \textbf{Transaction Intent Schema (TIS)} for portable goal specification and a \textbf{Policy Decision Record (PDR)} for auditable policy enforcement, together with a reference architecture.
\item \textbf{Reproducible artifacts:} A set of assets including the survey protocol, a PRISMA-style selection summary, a curated reference database, and a practical safety checklist for building agent-to-chain systems.
\end{enumerate}

By systematizing the design space, clarifying the risk surface, and charting a path toward standardization, this survey aims to provide a foundational resource for researchers, developers, and policymakers working to build a safe and interoperable future for agentic AI on blockchains.

\subsubsection{Scope Boundaries and Limitations}

While this survey aims to provide comprehensive coverage of agent-blockchain integration, certain boundaries define its scope. Understanding these boundaries helps readers contextualize the survey's contributions and identify areas requiring additional resources.

\paragraph{Temporal Scope}

This survey primarily covers developments through early 2026, with particular emphasis on systems and research from 2023-2025 when the field experienced rapid growth. Earlier foundational work is included where necessary for context, but the survey does not attempt comprehensive coverage of pre-2020 developments in either AI agents or blockchain technology.

\paragraph{Technical Scope}

The survey focuses on systems where AI agents interact with public, permissionless blockchain networks, particularly Ethereum and EVM-compatible chains. Private or permissioned blockchain deployments, while potentially relevant, are not covered in depth. Similarly, the survey focuses on agents based on large language models and related architectures, with limited coverage of other AI paradigms such as reinforcement learning agents or rule-based systems.

\paragraph{Application Scope}

The survey emphasizes financial applications including DeFi, payments, and asset management, which represent the majority of current agent-blockchain deployments. Non-financial applications such as supply chain, identity, and governance are covered where they intersect with the core themes, but comprehensive treatment of these domains is beyond the survey's scope.

\paragraph{Geographic and Regulatory Scope}

The survey does not attempt comprehensive coverage of regulatory frameworks across different jurisdictions. Regulatory considerations are discussed at a general level, but readers should consult jurisdiction-specific resources for detailed compliance guidance.

\section{Methodology: A Rigorous Protocol for a Systematic Survey}

To ensure a systematic, comprehensive, and reproducible review of the agent--blockchain interoperability landscape, we adopted a structured systematic literature review (SLR) protocol and reported the selection process using a PRISMA-style flow consistent with PRISMA 2020 reporting principles~\cite{page2021prisma}. We follow established SLR practice by defining the research questions up front, specifying a multi-source search strategy, applying pre-registered inclusion and exclusion criteria, and maintaining an auditable screening log. In the blockchain and security literature, similar SLR protocols are routinely used to reduce selection bias and improve transparency when consolidating fast-moving evidence bases~\cite{he2024large,shah2023systematic,alam2024front}.

This section details four core components of our methodology: (1) the research questions that guided our inquiry, (2) the multi-source search strategy, (3) the inclusion and exclusion criteria used to filter the literature, and (4) the coding schema used to extract and analyze attributes from the selected corpus.

\subsection{Research Questions}

Our survey was guided by four research questions (RQs) designed to progress from descriptive systematization to a forward-looking agenda:

\begin{itemize}
\item \textbf{RQ1: What are the dominant architectural patterns for integrating agentic AI systems with blockchains?}
We seek to identify recurring integration patterns (e.g., read-only analytics, simulation-first workflows, delegated execution, and autonomous signing), and to characterize the key trade-offs that explain why these patterns emerge in practice (e.g., authority, latency, auditability, and failure recovery).

\item \textbf{RQ2: What are the unique security threats and risks associated with agent-driven on-chain transactions?}
We develop a specialized threat model that combines tool-use vulnerabilities (e.g., instruction hijacking and tool spoofing) with blockchain-native adversarial dynamics (e.g., mempool visibility and transaction reordering), building on the broader direction of using LLMs and agents in blockchain-security contexts~\cite{he2024large}.

\item \textbf{RQ3: What is the state of the art in protocols, platforms, and standards for enabling secure agent--blockchain interoperability?}
We analyze systems and proposals that operationalize agent-to-chain execution, including tooling interfaces, wallet architectures, account abstraction stacks, and intent-based protocols, and compare them using a consistent coding rubric.

\item \textbf{RQ4: What are the most significant open research problems and gaps in the current landscape?}
We distill the most consequential gaps revealed by RQ1--RQ3 and organize them into a 2026 research roadmap, prioritizing missing interface layers, verifiable policy enforcement, and reproducible evaluation practices.
\end{itemize}

\subsection{Search Strategy: Casting a Wide Net}

Our literature search was conducted between October and December 2025 and deliberately covered both academic and grey literature. This choice is necessary because many agent and blockchain systems evolve through technical reports, improvement proposals, specifications, and open-source documentation that precede (or never reach) conventional peer review. For example, key infrastructure artifacts in the blockchain stack are often documented first as public technical reports or protocol documents rather than archival publications~\cite{breidenbach2021chainlink}.

\subsubsection{Search Venues}

We searched the following sources to ensure broad coverage:

\begin{itemize}
\item \textbf{Peer-reviewed venues:} IEEE Xplore, ACM Digital Library, SpringerLink, and Elsevier ScienceDirect.
\item \textbf{Preprint archives:} arXiv, focusing on cs.AI, cs.CR, cs.DC, and cs.SE.
\item \textbf{Technical standards and proposals:} Ethereum Improvement Proposals (EIPs), especially those related to account abstraction and transaction formats, plus relevant RFCs and web standards when they define interoperability surfaces.
\item \textbf{Open-source frameworks and industry documentation:} repositories and technical documentation for representative agent frameworks and Web3 tooling, alongside major security and infrastructure providers (e.g., wallet and relayer stacks, oracle networks, and automation platforms)~\cite{breidenbach2021chainlink}.
\end{itemize}

\subsubsection{Search Queries}

Search strings were adapted per venue, but followed the same conceptual structure: (i) agentic systems, (ii) blockchain interaction, and (iii) the interface layer connecting them. The core query was:

\texttt{("AI agent" OR "LLM agent" OR "autonomous agent" OR "multi-agent system") AND ("blockchain" OR "smart contract" OR "crypto" OR "Web3" OR "DeFi")}

We then refined the query to probe specific interface sub-areas:

\begin{itemize}
\item \texttt{("agent" OR "LLM") AND ("wallet" OR "transaction signing" OR "key management" OR "custody")}
\item \texttt{"tool protocol" AND ("AI" OR "agent")}
\item \texttt{("intent-based" OR "intent-centric") AND ("blockchain" OR "transaction")}
\item \texttt{"account abstraction" AND ("agent" OR "security" OR "EIP-4337")}
\item \texttt{"policy engine" AND ("blockchain" OR "smart contract")}
\end{itemize}

\subsubsection{Quality Assessment and Bias Mitigation}

Ensuring the quality and objectivity of this survey required systematic attention to potential biases and quality concerns.

\paragraph{Source Quality Assessment}

Each source was assessed for quality based on multiple criteria including publication venue, author credentials, methodological rigor, and citation impact. Peer-reviewed publications in established venues received higher weight than preprints or informal sources. However, given the rapid pace of development in this field, preprints and technical documentation were included where they provided unique insights not available in peer-reviewed literature.

\paragraph{Conflict of Interest Considerations}

Many relevant sources are produced by organizations with commercial interests in the technologies they describe. Documentation from wallet providers, infrastructure companies, and protocol developers may reflect promotional rather than objective perspectives. Where possible, claims from interested parties were verified against independent sources or qualified with appropriate caveats.

\paragraph{Recency vs. Stability Tradeoffs}

The rapid evolution of the field creates tension between covering recent developments and ensuring stability of cited sources. Very recent sources may be revised or retracted, while older sources may be outdated. The survey attempts to balance these concerns by prioritizing stable sources for foundational concepts while including recent sources for emerging developments.

\paragraph{Geographic and Linguistic Scope}

The survey primarily covers English-language sources, which may introduce geographic bias toward developments in English-speaking regions. Significant developments in other regions may be underrepresented. Readers should be aware of this limitation and consult region-specific sources where relevant.

\subsubsection{Methodological Considerations and Limitations}

The systematic review methodology employed in this survey reflects established best practices for evidence synthesis while acknowledging the unique challenges of surveying a rapidly evolving field. The security of smart contracts has been extensively studied, providing a foundation for understanding the risks of agent-mediated transactions~\cite{praitheeshan2020security,santamara2023smart}. Cross-chain protocols introduce additional complexity that must be carefully analyzed~\cite{pillai2020cross,pupyshev2020gravity}. The design of decentralized identity systems is a key enabler for agent authentication~\cite{preukschat2021self,salger2024decentralized}. Multi-agent systems have a long history of research that informs modern agent-blockchain integration~\cite{sikora1998multi,roth2005decentralized}. Recent work on LLM-based agents provides new perspectives on agent design~\cite{putta2024agent,nag2025coincidence}. The detection of malicious behavior in blockchain systems is an active area of research~\cite{shihab2025detecting}. Consensus mechanisms continue to evolve, with new protocols offering improved performance and security~\cite{muratov2018yac,naidu2023efficient}. The training of AI agents using reinforcement learning has also been explored~\cite{ouyang2020learning,ouyang2022intelligent}. Several methodological considerations merit explicit discussion.

The pace of development in both agentic AI and blockchain technology means that any survey risks becoming outdated shortly after publication. We have attempted to mitigate this risk by focusing on fundamental principles, architectural patterns, and enduring challenges rather than specific implementations that may be superseded. Nevertheless, readers should be aware that the specific systems and capabilities described may have evolved since the time of writing.

The interdisciplinary nature of agent-blockchain integration creates challenges for comprehensive literature coverage. Relevant work is distributed across computer science, cryptography, economics, law, and other fields, each with its own publication venues, terminology, and methodological norms. Our search strategy attempted to cast a wide net across these disciplines, but some relevant work may have been missed due to disciplinary boundaries.

Publication bias represents another methodological concern. Successful agent deployments and positive results are more likely to be published than failures or negative findings. This bias may lead to an overly optimistic assessment of agent capabilities and an underestimation of the challenges and risks involved. We have attempted to counterbalance this bias by including grey literature, industry reports, and documented incidents alongside peer-reviewed publications.

\subsection{Inclusion and Exclusion Criteria: Defining the Boundaries}

To maintain a focused and relevant corpus, we defined inclusion and exclusion criteria and applied them during screening by two independent reviewers, with disagreements resolved by a third reviewer. \cite{bamakan2020survey} 
A study or system was \textbf{included} if it satisfied all of the following criteria:

\begin{itemize}
\item \textbf{Must involve an agentic component:} The system must feature an AI or autonomous-agent component that performs non-trivial reasoning, planning, or decision-making (beyond deterministic, hard-coded rules).
\item \textbf{Must have an on-chain interface:} The agent must interact with a live blockchain environment by reading state, constructing candidate transactions or intents, and/or authorizing execution via a signing or policy-gated pathway. Purely off-chain agent systems were excluded.
\item \textbf{Must describe an interface, protocol, or architecture:} The work must contribute technical insight into \textit{how} agents and blockchains connect (e.g., tooling interfaces, transaction/intent formats, custody and authorization architectures, execution pipelines, or policy enforcement). Works that only propose an application outcome (e.g., ``using AI to predict token prices'') without specifying an interoperability mechanism were excluded.
\end{itemize}

Conversely, we \textbf{excluded} works that:

\begin{itemize}
\item Were purely theoretical agent papers (e.g., formal agent theory or game theory) without a concrete on-chain implementation, interface specification, or executable architecture.
\item Were blockchain papers without an agentic or AI component relevant to interoperability (e.g., consensus-only or cryptography-only contributions).
\item Were non-English publications.
\item Were marketing materials, news articles, or social media posts lacking sufficient technical detail for reproducible analysis.
\item Were clearly superseded by a more recent and substantially more complete version of the same contribution (e.g., an archival publication superseding an earlier workshop or preprint version).
\end{itemize}

\subsection{PRISMA-Style Selection Process: A Funnel of Evidence}

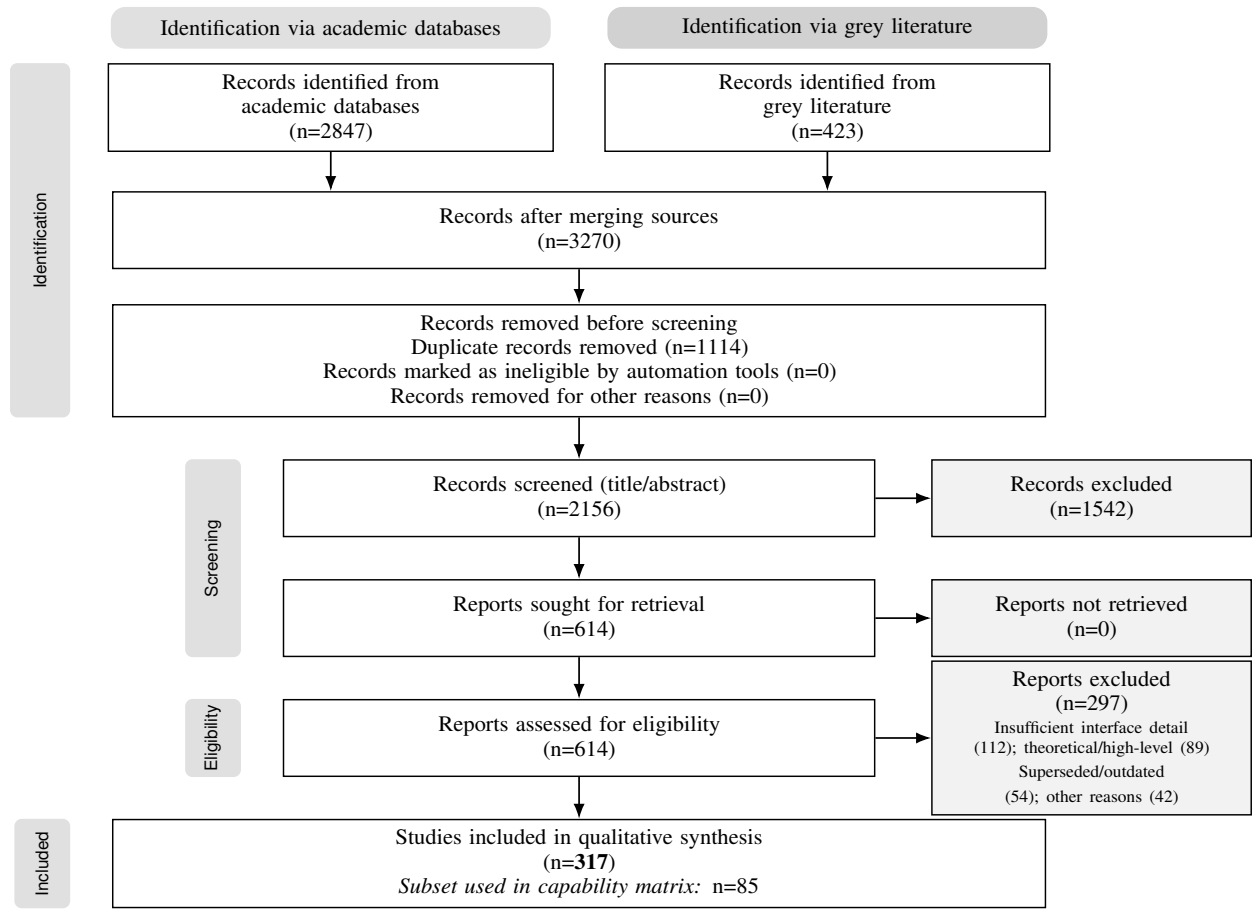
\begin{figure*}[t]
 \centering
 \resizebox{\textwidth}{!}{%
 \begin{tikzpicture}[
 font=\footnotesize,
 node distance=6mm and 8mm,
 >=Latex,
 prismabox/.style={draw, line width=0.6pt, align=center, inner sep=4pt, minimum height=11mm, fill=white},
 topbox/.style={prismabox, text width=6.1cm},
 fullbox/.style={prismabox, text width=13.1cm},
 mainbox/.style={prismabox, text width=8.2cm},
 sidebox/.style={prismabox, text width=4.3cm, fill=black!5},
 headerDB/.style={draw=none, fill=black!12, rounded corners=6pt, inner sep=3pt, minimum height=6mm, align=center},
 headerOther/.style={draw=none, fill=black!18, rounded corners=6pt, inner sep=3pt, minimum height=6mm, align=center},
 phasebar/.style={draw=none, fill=black!12, rounded corners=3pt},
 arr/.style={->, line width=0.7pt}
 ]

 \node[headerDB, text width=6.1cm] (hdb) {Identification via academic databases};
 \node[headerOther, text width=6.1cm, right=8mm of hdb] (hother) {Identification via grey literature};

 \node[topbox, below=2mm of hdb] (db)
 {Records identified from\\academic databases\\(n=2{}847)};
 \node[topbox, below=2mm of hother] (grey)
 {Records identified from \\grey literature\\(n=423)};

 \node[fullbox, below=12mm of $(db)!0.5!(grey)$] (merge)
 {Records after merging sources\\(n=3{}270)};

 \node[fullbox, below=5mm of merge] (removed)
 {Records removed before screening\\
 Duplicate records removed (n=1{}114)\\
 Records marked as ineligible by automation tools (n=0)\\
 Records removed for other reasons (n=0)};

 \node[mainbox, below=6mm of removed] (screen)
 {Records screened (title/abstract)\\(n=2{}156)};
 \node[sidebox, right=8mm of screen] (excl1)
 {Records excluded\\(n=1{}542)};

 \node[mainbox, below=6mm of screen] (sought)
 {Reports sought for retrieval\\(n=614)};
 \node[sidebox, right=8mm of sought] (notret)
 {Reports not retrieved\\(n=0)};

 \node[mainbox, below=6mm of sought] (full)
 {Reports assessed for eligibility\\(n=614)};
 \node[sidebox, right=8mm of full] (excl2)
 {Reports excluded\\(n=297)\\[0.5mm]
 \scriptsize Insufficient interface detail (112); theoretical/high-level (89)\\
 Superseded/outdated (54); other reasons (42)};

 \node[fullbox, below=6mm of full] (incl)
 {Studies included in qualitative synthesis\\(n=\textbf{317})\\
 \emph{Subset used in capability matrix:} n=85};

 \draw[arr] (db.south) -- (merge.north -| db.south);
 \draw[arr] (grey.south) -- (merge.north -| grey.south);

 \draw[arr] (merge) -- (removed);
 \draw[arr] (removed) -- (screen);
 \draw[arr] (screen) -- (sought);
 \draw[arr] (sought) -- (full);
 \draw[arr] (full) -- (incl);

 \draw[arr] (screen.east) -- (excl1.west);
 \draw[arr] (sought.east) -- (notret.west);
 \draw[arr] (full.east) -- (excl2.west);

 \coordinate (idNW) at ($(db.north west)+(-14mm,0)$);
 \coordinate (idSE) at ($(removed.south west)+(-6mm,0)$);
 \path[phasebar] (idNW) rectangle (idSE);
 \node[rotate=90, font=\scriptsize\sffamily] at ($(idNW)!0.5!(idSE)$) {Identification};

 \coordinate (scNW) at ($(screen.north west)+(-14mm,0)$);
 \coordinate (scSE) at ($(sought.south west)+(-6mm,0)$);
 \path[phasebar] (scNW) rectangle (scSE);
 \node[rotate=90, font=\scriptsize\sffamily] at ($(scNW)!0.5!(scSE)$) {Screening};

 \coordinate (elNW) at ($(full.north west)+(-14mm,0)$);
 \coordinate (elSE) at ($(full.south west)+(-6mm,0)$);
 \path[phasebar] (elNW) rectangle (elSE);
 \node[rotate=90, font=\scriptsize\sffamily] at ($(elNW)!0.5!(elSE)$) {Eligibility};

 \coordinate (inNW) at ($(incl.north west)+(-14mm,0)$);
 \coordinate (inSE) at ($(incl.south west)+(-6mm,0)$);
 \path[phasebar] (inNW) rectangle (inSE);
 \node[rotate=90, font=\scriptsize\sffamily] at ($(inNW)!0.5!(inSE)$) {Included};

 \end{tikzpicture}%
 }
 \caption{PRISMA 2020-style flow diagram summarizing the systematic search and screening process used in this survey.}
 \label{fig:prisma}
\end{figure*}

Our study selection followed the four PRISMA 2020 phases: Identification, Screening, Eligibility, and Inclusion~\cite{page2021prisma}. Figure~\ref{fig:prisma} summarizes the flow of records and the resulting corpus.

\begin{enumerate}
 \item \textbf{Identification:} We retrieved 2{}847 records from academic databases and 423 additional records from grey-literature sources, for a combined total of 3{}270 records.
 \item \textbf{Screening:} After removing 1{}114 duplicates, we screened 2{}156 records by title and abstract and excluded 1{}542 records that were out of scope (e.g., blockchain in unrelated application domains or AI systems without an agentic component).
 \item \textbf{Eligibility:} We assessed the full text of 614 reports for eligibility and excluded 297 reports, primarily due to insufficient technical detail about the agent--chain interface (n=112), purely conceptual or theoretical framing without interface architecture (n=89), or being superseded by a more recent and complete version (n=54). The remaining exclusions were due to other documented reasons (n=42).
 \item \textbf{Inclusion:} The final qualitative synthesis includes \textbf{317} unique studies, systems, and standards. From these, we selected a subset of \textbf{85} systems with adequate implementation detail and public documentation for quantitative coding in the comparative capability matrix.
\end{enumerate}

\paragraph{Reproducibility package.}
To make the survey auditable and easy to extend, we provide a reproducibility package (as supplementary material and/or a public repository) containing: (i) the finalized database query strings and date ranges; (ii) a de-duplicated bibliography export; (iii) the screening log with inclusion/exclusion rationales; and (iv) the coding codebook and extraction sheet used to populate the comparative matrix. Where licensing permits, we also include artifact metadata (e.g., repository links and version tags) for each analyzed system.

\subsection{Coding Schema for Extracted Attributes: Creating a Structured Comparison}

For each of the 85 systems selected for deep analysis, we extracted a consistent set of \textbf{13} attributes to populate the capability matrix. The coding schema was designed to capture both architectural choices and security-relevant design decisions in a way that supports structured, apples-to-apples comparison across heterogeneous implementations. Each coded value was backed by an evidence pointer (e.g., specification section, documentation page, or repository artifact) recorded in the extraction sheet.

\begin{itemize}
\item[\textbf{(A1)}] \textbf{Interface layer:} The primary integration surface exposed to the agent (e.g., tool protocol, wallet API, smart-contract interface, relayer service).

\item[\textbf{(A2)}] \textbf{Execution pathway:} How actions move from intent to chain submission (e.g., direct wallet broadcast, relayer or aggregator submission, solver-mediated execution, bundler and entry-point flow).

\item[\textbf{(A3)}] \textbf{Transaction representation:} The highest-level representation supported (e.g., raw transactions, structured actions, intent objects, batched or atomic bundles).

\item[\textbf{(A4)}] \textbf{Custody model:} How signing authority is held and exercised (e.g., user-held keys, delegated signing, MPC or threshold signing, TEE-backed signing)~\cite{lindell2017fast,komlo2024threshold}.

\item[\textbf{(A5)}] \textbf{Authorization mechanism:} The concrete authorization primitive (e.g., EOA signature, smart-account validation, session keys, role-based signing, multi-sig or threshold approval).

\item[\textbf{(A6)}] \textbf{Permission model:} How the system constrains what the agent can do (e.g., allowlists, spend limits, method-level permissions, capability-style grants, scoped sessions).

\item[\textbf{(A7)}] \textbf{Policy controls:} Whether runtime policies are enforced prior to authorization or submission (e.g., rule-based guards, programmable policy-as-code, contextual constraints, compliance checks).

\item[\textbf{(A8)}] \textbf{Observability:} Availability of audit logs, transaction previews and simulations, monitoring hooks, and traceability of agent actions (including provenance of tool calls where applicable).

\item[\textbf{(A9)}] \textbf{Recovery mechanisms:} Support for incident response and safe failure handling (e.g., kill switches, key rotation, emergency delays, revocation, social recovery, dispute workflows).

\item[\textbf{(A10)}] \textbf{MEV and adversarial execution protections:} Whether the design accounts for mempool adversaries and extractable value (e.g., private orderflow, auctions, slippage constraints, commit--reveal patterns, timing protections)~\cite{daian2019flash}.

\item[\textbf{(A11)}] \textbf{Multi-chain and cross-domain support:} Whether the system operates across multiple chains or rollups, and whether it exposes chain-specific assumptions (fee markets, account models, finality semantics).

\item[\textbf{(A12)}] \textbf{Trust assumptions and dependencies:} The external trust and infrastructure dependencies required (e.g., RPC providers, relayers, solvers, TEEs, oracles), and whether they are assumed trusted, permissioned, or adversarial.

\item[\textbf{(A13)}] \textbf{Threat assumptions:} The explicit or implicit threat model targeted (e.g., key compromise, prompt or tool injection, malicious dApps, insider risk, mempool or MEV adversaries).
\end{itemize}

To improve consistency under a single-coder setting, we used a written codebook with standardized labels for each attribute, recorded evidence links for every coded value, and performed a second-pass recoding of ambiguous cases with an explicit decision log before finalizing the matrix.

\paragraph{Coding rules for capability indicators.}
For the capability matrix (Table~\ref{tab:capability_matrix}), we used three levels of support: \cmark\ denotes first-class, documented support that is available as a stable feature in the system; \pmark\ denotes partial, indirect, or externally dependent support (e.g., achievable via plugins, custom integration, or not supported across the default execution path); and \xmark\ denotes no support or an absence of documented support. The descriptive columns (Type, Tool interface, Chain scope, Custody model) were coded directly from official documentation and/or reference implementations. For the capability columns, we applied the following criteria:

\begin{itemize}
\item \textbf{Intent:} \cmark\ if the system supports a declarative intent or outcome-level object that is compiled into transactions; \pmark\ if it supports structured actions/templates but not true intent compilation; \xmark\ if it only supports raw transaction construction.
\item \textbf{Sign:} \cmark\ if the system provides an explicit signing or authorization pathway (EOA signing, smart-account validation, delegated/MPC signing, or session-key signing); \pmark\ if signing is only possible through an external wallet outside the system; \xmark\ if it does not support signing or authorization.
\item \textbf{Policy:} \cmark\ if policy checks can block/approve actions pre-signing or pre-submission (e.g., allowlists, spend limits, rule engines); \pmark\ if constraints exist but are coarse or indirect; \xmark\ if no policy gating is provided.
\item \textbf{Preview:} \cmark\ if the system offers a human-readable transaction/action preview (decoded calldata, risk flags, summaries); \pmark\ if preview exists only via third-party tooling; \xmark\ if none is provided.
\item \textbf{Simulation:} \cmark\ if the system supports pre-execution simulation (dry-run, state-diff, revert detection, outcome estimation); \pmark\ if simulation is possible only via external services; \xmark\ if not supported.
\item \textbf{Observability:} \cmark\ if audit logs and monitoring are available for actions and executions; \pmark\ if logs are partial or indirect; \xmark\ if the system provides no meaningful traceability beyond on-chain traces.
\item \textbf{Recovery:} \cmark\ if the system supports recovery controls (revocation, delays, kill switch/pause, key rotation, emergency workflows); \pmark\ if recovery is limited or external; \xmark\ if no recovery mechanisms are provided.
\end{itemize}

\section{Background: Foundational Primitives}

To comprehend the challenges of agent--blockchain interoperability, a surface-level understanding of the underlying technologies is insufficient. An agent does not simply ``talk'' to a blockchain; it operates through a multi-stage pipeline in which errors compound across perception, planning, authorization, execution, and verification. This section deconstructs the foundational primitives that enable agent-mediated on-chain actions. We focus on three building blocks used throughout the paper: (i) the agent action pipeline, (ii) the evolution of wallets from externally owned accounts (EOAs) to programmable smart accounts, and (iii) the role of MEV in shaping adversarial execution.

\subsection{The Agent Action Pipeline: A Detailed Deconstruction}

\begin{figure*}[t]
\centering
\resizebox{\linewidth}{!}{
\begin{tikzpicture}[
    stage/.style={draw, rounded corners=3mm, minimum height=12mm, minimum width=2.0cm, align=center, font=\footnotesize, fill=white},
    interface/.style={draw, rectangle, rounded corners=2mm, font=\scriptsize, align=center, fill=white},
    control/.style={draw, rectangle, rounded corners=2mm, font=\scriptsize, align=center, fill=white},
    arr/.style={->, thick, shorten >=1mm},
    fbarr/.style={->, thick, dashed, shorten >=1mm},
    maarr/.style={->, thick, dotted, shorten >=1mm}
]

\node[stage] (obs) {\textbf{Observe}\\state acquisition};
\node[stage, right=7mm of obs] (rea) {\textbf{Reason}\\state interpretation};
\node[stage, right=7mm of rea] (plan) {\textbf{Plan}\\action synthesis};
\node[stage, right=7mm of plan] (auth) {\textbf{Authorize}\\human/policy gate};
\node[stage, right=7mm of auth] (exec) {\textbf{Execute}\\sign \& submit};
\node[stage, right=7mm of exec] (ver) {\textbf{Verify}\\finality \& feedback};

\node[interface, below=8mm of obs] (i1) {State API / Sensors};
\node[interface, below=8mm of rea] (i2) {Simulation / Context Model};
\node[interface, below=8mm of plan] (i3) {Action Solver};
\node[interface, below=8mm of auth] (i4) {Intent schema / Policy check};
\node[interface, below=8mm of exec] (i5) {Wallet / AA / Bundler};
\node[interface, below=8mm of ver] (i6) {Ledger / Feedback};

\node[control, below=10mm of i3] (c1) {Input Validation};
\node[control, below=10mm of i4] (c2) {Policy Enforcement};
\node[control, below=10mm of i5] (c3) {Signing Isolation};
\node[control, below=10mm of i6] (c4) {Audit / Recovery};

\draw[arr] (obs) -- (rea);
\draw[arr] (rea) -- (plan);
\draw[arr] (plan) -- (auth);
\draw[arr] (auth) -- (exec);
\draw[arr] (exec) -- (ver);

\foreach \top/\bot in {obs/i1, rea/i2, plan/i3, auth/i4, exec/i5, ver/i6}
    \draw[arr] (\top.south) -- (\bot.north);

\foreach \mid/\bot in {i3/c1, i4/c2, i5/c3, i6/c4}
    \draw[arr] (\mid.south) -- (\bot.north);

\draw[fbarr] (ver.north) .. controls +(0,25mm) and +(0,25mm) .. node[above]{feedback} (obs.north);
\draw[fbarr] (exec.north) .. controls +(0,8mm) and +(0,8mm) .. node[above]{audit/review} (auth.north);
\draw[fbarr] (ver.north) .. controls +(0,15mm) and +(0,15mm) .. node[above]{re-plan} (plan.north);

\draw[maarr] (plan.south east) .. controls +(0mm,0) and +(0,0mm) 
.. node[below]{\tiny Proposer$\rightarrow$Verifier} (auth.south west);
\draw[maarr] (auth.south east) .. controls +(0,0mm) and +(-0mm,0) .. node[below]{\tiny Verifier$\rightarrow$Executor} (exec.south west);

\end{tikzpicture} }
\caption{Monochrome six-stage agent action pipeline. The top row shows core stages, the middle row summarizes common integration interfaces, and the bottom row highlights typical control points. Dashed arrows denote feedback loops; dotted arrows indicate optional multi-agent roles. Account-abstraction execution paths are commonly mediated by bundlers and smart accounts~\cite{buterin2021erc}.}
\label{fig:agent_pipeline}
\end{figure*}
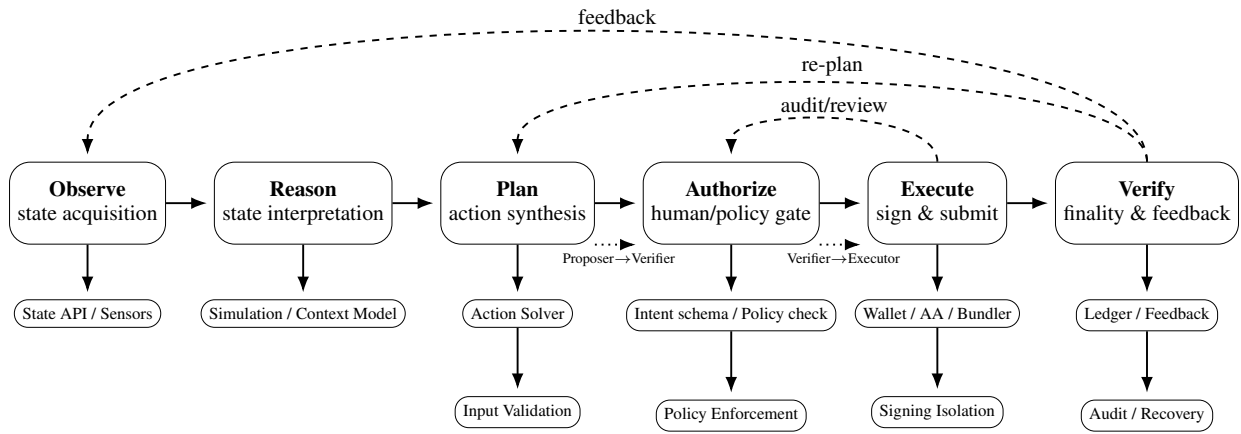

The lifecycle of an agent-mediated on-chain action can be modeled as a six-stage pipeline, illustrated in Figure~\ref{fig:agent_pipeline}. This model provides an implementation-agnostic framework for analyzing integration patterns, tracing how observed state becomes an irreversible state transition, and pinpointing where safety controls must attach.

\paragraph{Pipeline overview.}
We model agent-to-chain execution as a six-stage pipeline that transforms observed state into an irreversible state transition. The stages are intentionally implementation-agnostic and apply to self-custody agents, smart-account mediated execution (including account abstraction), and enterprise or delegated custody systems. The abstraction is useful because real failures often span layers. For example, a prompt-injection event during observation can corrupt planning, while adversarial execution effects can invalidate an otherwise correct plan.

\begin{enumerate}
\item \textbf{Observe (state acquisition).} The agent ingests \emph{on-chain} state (RPC reads, logs, indexers), \emph{mempool} state (pending transactions, fee conditions), and \emph{off-chain} state (prices, risk signals, governance announcements, and other external context). This stage is vulnerable to data spoofing, oracle manipulation, indexer compromise, and indirect prompt injection via untrusted text sources.
\item \textbf{Reason (state interpretation).} The agent converts heterogeneous signals into a structured world model (positions, constraints, objectives, and risk posture). LLM-centric agents are exposed to context confusion, hallucinated invariants, and instruction-following failures when retrieved content is not cleanly separated from trusted policy and system prompts.
\item \textbf{Plan (action synthesis).} The agent selects protocols and constructs candidate actions (swaps, deposits, borrows, bridges, governance votes, or multi-step workflows). Planning must account for chain semantics (token decimals, approvals, permit standards), economic constraints (liquidity, slippage), and adversarial execution dynamics (front-running, sandwiching, and back-running)~\cite{daian2019flash}.
\item \textbf{Authorize (human and/or policy gating).} Before any irreversible action, the agent's proposal should be checked against constraints such as user confirmation, spend limits, allowlists, rate limits, simulation-based guards, and policy engines. This stage is where intent representations and policy attestations reduce ambiguity between what the agent intends and what the chain will execute.
\item \textbf{Execute (signing and submission).} The system produces cryptographic authorization (EOA signatures, smart-account validation, MPC signatures, and/or session keys) and submits transactions via public mempools, private relays, bundlers, or solver networks. Key compromise, session-key overreach, relay manipulation, and transaction replacement are dominant risks at this stage.
\item \textbf{Verify (finality and feedback).} The system monitors inclusion, handles re-orgs, validates post-state (balances, positions, emitted events), and triggers remediation when outcomes diverge from intent (e.g., partial fills, reverts, or unexpected approvals). Robust agents treat verification as a first-class loop, writing auditable logs and updating long-term memory only after sufficient finality.
\end{enumerate}

\paragraph{Why MEV matters in the pipeline.}
Unlike conventional web APIs, blockchain execution is economically adversarial by default. Profitable actions disclosed to a public mempool can be reordered, copied, or sandwiched by searchers, making execution risk endogenous to the environment~\cite{daian2019flash}. Consequently, safer agent architectures combine \emph{ex-ante} controls (simulation, slippage bounds, private orderflow, solver-based execution) with \emph{ex-post} controls (verification, recovery workflows, and clear human escalation paths). We use this pipeline throughout the paper to organize the threat taxonomy and the practical safety checklist.

\subsubsection{Smart Contract Fundamentals for Agent Developers}

Effective agent-blockchain integration requires deep understanding of smart contract mechanics. This section provides essential background for agent developers who may be more familiar with AI systems than blockchain technology.

\paragraph{Contract State and Storage}

Smart contracts maintain persistent state in contract storage, a key-value store where each 256-bit key maps to a 256-bit value. Storage operations are among the most expensive blockchain operations, with writes costing significantly more than reads. Agent systems that interact with contracts must understand storage patterns to predict gas costs and optimize transaction construction.

Common storage patterns include mappings (hash-based key-value associations), arrays (sequential data structures), and packed storage (multiple values encoded in single storage slots). Different patterns have different gas cost profiles and access patterns that affect how agents should interact with them.

\paragraph{Function Selectors and ABI Encoding}

Contract functions are identified by four-byte selectors derived from the hash of the function signature. Arguments are encoded according to the Application Binary Interface (ABI) specification, which defines standard encodings for different data types. Agent systems must correctly encode function calls to interact with contracts successfully.

The ABI encoding of complex types (arrays, structs, nested types) can be intricate, and encoding errors result in failed transactions or unexpected behavior. Agent systems should use established libraries for ABI encoding rather than implementing encoding logic from scratch.

\paragraph{Events and Logs}

Contracts emit events to record significant state changes and provide information to off-chain observers. Events are stored in transaction logs, which are cheaper than contract storage but not accessible to on-chain code. Agent systems rely heavily on events for monitoring contract state and detecting relevant activities.

Event filtering by topic enables efficient retrieval of specific event types across large numbers of blocks. Agent systems should design their event monitoring strategies to balance comprehensiveness with efficiency, filtering for relevant events while avoiding excessive RPC load.

\paragraph{Error Handling and Reverts}

Contract execution can fail through reverts, which undo all state changes and return remaining gas to the caller. Reverts can be triggered explicitly by the contract (through require, revert, or assert statements) or implicitly by runtime errors (out of gas, invalid operations). Agent systems must handle reverts gracefully, interpreting error messages and adjusting strategies accordingly.

Custom error types, introduced in recent Solidity versions, provide structured error information that agents can parse and act upon. Agent systems should be designed to interpret both legacy string-based errors and modern custom errors.

\subsubsection{Historical Perspective on Wallet Security}

The security landscape of blockchain wallets has evolved significantly since the early days of cryptocurrency. Understanding this evolution provides context for current best practices and emerging challenges.

\paragraph{Early Security Challenges}

The earliest cryptocurrency users faced significant security challenges with limited tooling support. Private key management was entirely manual, with users responsible for generating, storing, and protecting cryptographic keys. Many early users lost access to their funds through key loss, theft, or operational errors.

The Mt. Gox exchange collapse in 2014 highlighted the risks of centralized custody, where a single point of failure could result in catastrophic losses. This incident catalyzed development of better security practices and tools, including hardware wallet~\cite{khan2019security}s and multi-signature schemes.

\paragraph{Hardware Wallet Development}

Hardware wallets emerged as a response to the security limitations of software-only key storage. By isolating private keys in dedicated secure hardware, these devices protect against malware, phishing, and many other attack vectors. Hardware wallets have become standard practice for securing significant cryptocurrency holdings.

However, hardware wallets introduce their own challenges including physical security, supply chain integrity, and usability limitations. The requirement for physical interaction with hardware devices creates friction that may be incompatible with automated agent operation.

\paragraph{Multi-Signature Evolution}

Multi-signature schemes evolved from simple m-of-n configurations to sophisticated smart contract implementations with programmable authorization logic. Modern multi-signature wallets support features including time locks, spending limits, and role-based access control.

The evolution of multi-signature technology laid groundwork for the programmable authorization models that enable agent integration. By demonstrating that authorization logic could be customized and automated, multi-signature wallets paved the way for more sophisticated delegation schemes.

\paragraph{Current Security Landscape}

The current security landscape reflects lessons learned from over a decade of operational experience. Best practices include defense in depth, separation of duties, and continuous monitoring. However, the integration of AI agents introduces new challenges that existing security frameworks may not fully address.

\subsubsection{The Evolution of Blockchain Wallets}

The evolution of blockchain wallets reflects the broader maturation of the ecosystem from a technology for enthusiasts to infrastructure for mainstream applications. Understanding this evolution provides essential context for the design of agent-wallet interactions.

First-generation wallets were simple key management tools that stored private keys and signed transactions. Users bore full responsibility for key security, backup, and recovery. The loss of a private key meant permanent loss of access to associated funds, with no possibility of recovery. This model, while maximally decentralized, proved unsuitable for mainstream adoption due to its unforgiving nature and poor user experience.

Second-generation wallets introduced various improvements to usability and security. Hardware wallets isolated private keys in secure elements, protecting against software-based attacks. Multi-signature schemes enabled shared custody arrangements and organizational controls. Hierarchical deterministic (HD) wallets simplified backup and recovery through mnemonic seed phrases. These innovations improved security and usability but maintained the fundamental model of explicit key management.

Third-generation wallets, exemplified by smart contract wallets and account abstraction, represent a paradigm shift toward programmable custody. Rather than relying solely on cryptographic key possession, these wallets enable arbitrary authorization logic encoded in smart contracts. This programmability enables features such as social recovery, spending limits, session keys, and automated policies that were impossible with traditional externally owned accounts.

For agent systems, this evolution is particularly significant. Smart contract wallets provide the foundation for fine-grained delegation, enabling principals to grant agents specific, bounded authorities rather than all-or-nothing access. The programmable nature of these wallets allows authorization policies to be tailored to the specific requirements of different agent use cases, balancing autonomy with appropriate constraints.

\subsubsection{The Continuing Evolution of Blockchain Technology}

The blockchain ecosystem continues to evolve rapidly, with new consensus mechanisms, scaling solutions, and application paradigms emerging at an accelerating pace. Fair exchange protocols enable trustless transactions between parties~\cite{ferrergomila2019fair}. The application of blockchain to various industries, including healthcare, has been extensively explored~\cite{javed2021health,gong2024blockchain}. Cross-chain interoperability remains a key challenge for the ecosystem~\cite{gebele2025cross}. The economics of blockchain systems, including subsidy mechanisms, have been studied in detail~\cite{geng2023subsidy}. The use of blockchain for IoT and robotics applications has also been explored~\cite{kapitonov2017blockchain}. Smart contract security continues to be a major area of research~\cite{kaur2023smart}. Payment channel networks enable scalable off-chain transactions~\cite{khalil2017revive}. Privacy-preserving smart contracts have been developed using zero-knowledge proofs~\cite{kosba2016hawk}. Recent surveys have provided a comprehensive overview of the blockchain landscape~\cite{freeman2023blockchain}. This evolution has profound implications for agent-blockchain integration, as agents must be designed to adapt to changing protocol semantics, new transaction types, and evolving best practices.

The transition from proof-of-work to proof-of-stake consensus in major networks has altered the economic dynamics of transaction inclusion, creating new opportunities and challenges for agent-mediated execution. Stake-weighted validator selection introduces different game-theoretic considerations than hash-power competition, affecting the strategies that agents should employ for transaction timing and fee optimization. Similarly, the proliferation of application-specific blockchains and rollups has created a fragmented landscape where agents must navigate heterogeneous execution environments with varying security guarantees and finality characteristics.

The emergence of modular blockchain architectures, which separate consensus, data availability, and execution into distinct layers, represents a particularly significant development for agent systems. These architectures enable more flexible deployment patterns, where agents can choose execution environments optimized for their specific use cases while inheriting security from shared consensus and data availability layers. Understanding these architectural patterns is essential for designing agents that can operate efficiently and securely across the evolving blockchain landscape.

\subsubsection{Implementation Considerations for the Agent Pipeline}

The GOALIATH theory provides a cognitive framework for understanding goal-directed behavior in humans and artificial agents, which can inform the design of more robust agent planning systems~\cite{hommel2022goaliath}. Research in long-term robot autonomy provides valuable lessons for designing agents that can operate for extended periods without human intervention, including the need for self-monitoring and adaptation~\cite{kunze2018artificial}.

The practical implementation of the agent action pipeline involves numerous engineering decisions that affect system reliability, security, and performance. Transaction simulation tools enable agents to preview the effects of their actions before committing to the blockchain~\cite{tenderlySimDocs,tenderlydocs2026simulation}. The security of smart contracts can be verified using static analysis tools~\cite{tsankov2018securify}. Multi-agent coordination requires careful protocol design~\cite{torreo2017cooperative,von1992cooperation}. Reinforcement learning techniques can be used to train agents for complex tasks~\cite{taghavi2023reinforcement,tran2023enhancing}. The interoperability of different blockchain systems is a key challenge~\cite{silva2025interoperabilidad,vedantham2025connect}. Verifiable credentials enable privacy-preserving authentication~\cite{sporny2025verifiable}. Multi-agent systems can be used to solve complex problems~\cite{sun2025multi}. The design of smart contract systems requires careful consideration of security and usability~\cite{technologies2023smart,takei2023not}. Zero-knowledge proofs enable privacy-preserving transactions~\cite{wan2023zk}. AI-powered agents are increasingly being used for blockchain applications~\cite{waldner2025ai}. Each stage of the pipeline presents its own implementation challenges and design tradeoffs.

The observation stage must balance comprehensiveness with efficiency. Agents need access to relevant blockchain state, market data, and contextual information, but processing excessive data can overwhelm the agent's reasoning capabilities and increase latency. Effective observation strategies employ filtering, aggregation, and prioritization to present agents with the most relevant information for their current task.

The reasoning stage is perhaps the most challenging to implement reliably. LLM-based reasoning is inherently probabilistic and can produce inconsistent or incorrect outputs, particularly for complex multi-step tasks. Techniques such as chain-of-thought prompting, self-consistency checking, and ensemble methods can improve reasoning reliability but add computational overhead and latency. The tradeoff between reasoning quality and response time is particularly acute for time-sensitive blockchain operations.

The construction stage must produce syntactically correct and semantically valid transactions. This requires precise handling of data types, encoding formats, and protocol-specific conventions. Errors in transaction construction can result in failed transactions, wasted gas fees, or worse, unintended transfers of value. Robust construction implementations employ multiple layers of validation, including schema checking, simulation, and human review for high-stakes operations.

\subsection{Account Abstraction (ERC-4337\cite{EIP4337})}

A central enabler for safer and more usable on-chain automation is \textbf{account abstraction (AA)}. In practice, the most influential instantiation for Ethereum and EVM-compatible chains is \textbf{ERC-4337}, which introduces programmable ``smart accounts'' that can enforce custom validation rules while remaining compatible with today’s consensus rules and transaction pipeline~\cite{buterin2021erc}. For agentic systems, this shift is foundational: instead of treating a wallet as a single private key with unconditional authority, AA turns the wallet into a programmable security boundary where constraints can be checked on-chain before an action is executed.

\subsubsection{The EOA/CA dichotomy and its limitations}

Ethereum historically distinguishes two account types:

\begin{enumerate}
\item \textbf{Externally Owned Accounts (EOAs).} Accounts controlled by a private key. EOAs can initiate transactions, hold assets, and are the default in common wallets (e.g., MetaMask-style accounts).
\item \textbf{Contract Accounts (CAs).} Accounts controlled by code. CAs can encode rich logic, but cannot initiate transactions on their own; they only react when invoked by a transaction.
\end{enumerate}

This rigid dichotomy created friction for both usability and security. Programmable wallet features such as multi-signature approvals, spending limits, social recovery, or application-specific permissions naturally belong in contract code. Yet a contract wallet still needs an EOA-funded transaction to ``wake it up'' and pay gas, which historically forced users into a two-layer arrangement: an EOA to initiate and fund transactions, and a contract wallet to provide policy logic. For agent deployments, the limitation is sharper: giving an agent EOA-level signing authority collapses the permission boundary into a single secret, while contract-only designs struggle with initiation and fee-payment ergonomics.

\subsubsection{ERC-4337: A paradigm shift without a protocol hard fork}

The concept of account abstraction has a long history in the Ethereum community, with early proposals outlining the core ideas that would eventually be realized in ERC-4337~\cite{buterin2020eip}.

Earlier AA proposals required protocol changes (hard forks), which are slow and politically costly. \cite{priscilla2024decentralizedvoting} ERC-4337 achieves AA \emph{without} changing consensus by introducing a higher-level operation object, \texttt{UserOperation}, processed through a canonical on-chain contract called \texttt{EntryPoint}~\cite{buterin2021erc}. Conceptually, ERC-4337 adds an additional ``execution lane'' above normal transactions:

\begin{itemize}
\item \textbf{\texttt{UserOperation}.}
A structured pseudo-transaction describing an action to be performed by a smart account. A \texttt{UserOperation} typically includes the smart account address (\texttt{sender}), the call data describing the action, gas and fee fields, and an authorization payload. This structure is particularly suitable for agentic systems because an agent can construct operations at the level of an \emph{action description} without directly managing the low-level details of legacy transaction formats~\cite{buterin2021erc}.

\item \textbf{Bundlers.}
Specialized nodes that monitor a \texttt{UserOperation} mempool, assemble multiple operations into a standard transaction, and submit that transaction to the \texttt{EntryPoint} contract. Bundlers are designed to be permissionless infrastructure: in principle anyone can run one, and bundlers are compensated for the gas they spend through the ERC-4337 economics~\cite{buterin2021erc}.

\item \textbf{\texttt{EntryPoint}.}
A singleton contract that mediates verification and execution for ERC-4337 operations. Operationally, \texttt{EntryPoint} enforces a two-stage flow for each operation: (i) a \emph{verification} stage that calls account-defined validation logic (commonly \texttt{validateUserOp}) and (ii) an \emph{execution} stage that dispatches the requested call if validation succeeds. This design concentrates critical security checks into an on-chain, auditable path and standardizes how smart accounts express authorization logic~\cite{buterin2021erc,singh2023account}.

\item \textbf{Paymasters.}
Optional contracts that sponsor gas or enforce sponsorship constraints. A \texttt{UserOperation} can request a paymaster, and \texttt{EntryPoint} will validate and charge the paymaster according to its policy. Paymasters enable ``gasless'' onboarding and allow applications to sponsor execution under explicit constraints, which is valuable when agents should not maintain unrestricted balances of the chain’s native token~\cite{buterin2021erc,singh2023account}.
\end{itemize}

\paragraph{Security note: AA is not automatically safe.}
Account abstraction increases flexibility, but it also expands the design space for mistakes. Poorly designed validation logic can recreate ``EOA-like'' risk (broad authority, weak constraints), while unsafe paymaster policies can subsidize abusive behavior. Moreover, bundler and mempool dynamics remain adversarial in many environments, so AA must be paired with careful policy design, simulation/preview tooling, and recovery controls to achieve meaningful safety in agent deployments.

\subsubsection{Implications for agentic systems}

ERC-4337 is a practical security substrate for on-chain agents for four reasons.

\begin{itemize}
\item \textbf{Programmable authorization.}
Smart accounts can encode authorization policies directly in on-chain validation logic: session keys, time bounds, function-level allowlists, value limits, multi-party approval, and conditional execution checks. This enables a design where an agent operates under narrowly scoped authority that the chain enforces at submission time, rather than trusting the agent with a master private key~\cite{buterin2021erc}.

\item \textbf{Decoupling signing from gas payment.}
With paymasters, a system can authorize agent actions without requiring the agent to hold the native gas token. This supports agent architectures where the agent is constrained to action authorization and cannot freely liquidate or rebalance to keep gas reserves, while still enabling execution under application-controlled sponsorship policies~\cite{buterin2021erc,singh2023account}.

\item \textbf{Standardization and interoperability.}
ERC-4337 offers a standardized flow (\texttt{UserOperation} $\rightarrow$ bundler $\rightarrow$ \texttt{EntryPoint}) that different smart-account implementations can share. This reduces bespoke wallet integrations and provides a common interface for middleware and agents across compliant accounts, helping prevent vendor lock-in and integration fragmentation~\cite{buterin2021erc}.

\item \textbf{Modularity and extensibility.}
AA ecosystems are naturally modular: account implementations, validation modules (e.g., session keys), recovery logic, and paymasters can evolve semi-independently. For agentic applications, this allows security configurations to be tailored to risk tolerance and operational needs while maintaining a consistent execution backbone~\cite{singh2023account}.
\end{itemize}

In summary, ERC-4337 transforms the wallet from a passive key holder into a programmable security platform. For autonomous or semi-autonomous on-chain systems, it enables a principled alternative to granting agents unrestricted private-key control: capability-scoped authorization enforced by smart-account validation, complemented by preview/simulation, monitoring, and recovery.

\subsection{Miner/Maximal Extractable Value (MEV)}

No discussion of the on-chain execution environment is complete without a clear understanding of \textbf{Maximal Extractable Value (MEV)}. MEV is a structural economic force in many blockchain systems, particularly in DeFi, and it is frequently adversarial from the perspective of ordinary users and automated agents. In its standard formulation, MEV denotes the profit obtainable by an actor who can influence transaction \emph{inclusion} and \emph{ordering} within a block, including by inserting, excluding, or reordering transactions to create favorable execution outcomes~\cite{daian2019flash}. For an AI agent that executes repeatedly and at machine speed, treating MEV as an edge case is a reliable path to systematic value leakage.

\subsubsection{From miner to maximal: the evolution of MEV}

The concept originated as \emph{Miner Extractable Value} in proof-of-work systems, reflecting miners' control over block construction and their visibility into pending transactions in the public mempool. The core phenomenon was documented early through empirical observations in decentralized exchanges, where transaction reordering and insertion attacks could extract value from pending user trades~\cite{daian2019flash}. After Ethereum's transition to proof-of-stake, the term evolved to \emph{Maximal} Extractable Value to emphasize that extraction is not limited to miners. Instead, it can be captured by a supply chain of specialized actors and infrastructure that collectively shape ordering and routing, even when the final block proposer is not itself performing strategy search.

A useful abstraction is to separate:
(i) \textbf{strategy discovery} (identifying profitable reorderings or insertions),
(ii) \textbf{bundle construction} (encoding atomic sequences of transactions), and
(iii) \textbf{block assembly and selection} (choosing a block candidate that maximizes revenue). Contemporary surveys and mitigation studies describe this broadened ecosystem and its implications for security, decentralization, and market efficiency~\cite{alipanahloo2024maximum}.

\subsubsection{The taxonomy of MEV attacks}

Recent work has focused on creating a more granular categorization of MEV strategies, which is essential for developing targeted mitigation techniques~\cite{gramlich2024maximal}. Understanding the specific types of MEV that agents are vulnerable to is a key area of ongoing research.

For agentic execution, MEV most often manifests as degradation of economic outcomes rather than as overt protocol compromise. \cite{tao2024robustness} Common patterns include:

\begin{itemize}
\item \textbf{Front-running.} An adversary observes a price-moving or profitable transaction and submits a competing transaction that executes \emph{before} the target by offering higher effective fees, capturing profit from the subsequent price movement~\cite{daian2019flash}.
\item \textbf{Back-running.} An adversary positions a transaction to execute \emph{immediately after} a target transaction to capture an opportunity created by that transaction, such as arbitrage opened by a large swap~\cite{daian2019flash}.
\item \textbf{Sandwich attacks.} A compound strategy that combines front-running and back-running. The adversary pushes the execution price against the victim before the victim trade, then unwinds after, extracting value primarily from the victim's slippage budget. Sandwiching has been formalized both empirically and theoretically and remains a dominant attack family against naive DEX execution~\cite{daian2019flash,kulkarni2023towards}.
\end{itemize}

These patterns are particularly damaging for agents because agents tend to be consistent, repeat actions at scale, and may exhibit predictable execution policies. Without MEV-aware constraints, an agent can become a recurring source of extractable value.

\subsubsection{Implications and mitigations for agentic systems}

For an AI agent, the public mempool is not a neutral queue. It is a public information channel where valuable intent can be detected and strategically exploited. Practical mitigations therefore combine preventive controls at execution time with robust post-execution verification:

\begin{itemize}
\item \textbf{Reduce public intent leakage.} Where feasible, avoid broadcasting valuable actions to the public mempool and prefer execution pathways that reduce pre-trade visibility or expose only minimal information. Mitigation-focused surveys provide a structured view of these approaches, including orderflow protection and redesigns that aim to reduce extractable value~\cite{alipanahloo2024maximum}.
\item \textbf{Prefer MEV-aware execution mechanisms.} Batch auctions, solver-mediated execution, and intent-based settlement can reduce certain mempool-visible attack surfaces by shifting competition into controlled mechanisms. These approaches do not eliminate adversarial incentives, but they can change the attack surface and the distribution of extractable value~\cite{kulkarni2023towards,alipanahloo2024maximum}.
\item \textbf{Encode tight constraints.} Use strict slippage limits, minimum-out bounds, and price guards to cap worst-case loss. These controls do not prevent all forms of MEV, but they limit the amount of value an adversary can extract by forcing reverts when execution crosses unacceptable thresholds~\cite{daian2019flash}.
\item \textbf{Explore protocol-level defenses where relevant.} Some lines of work propose MEV-resistant protocol designs or alternative execution layers intended to reduce the feasibility or profitability of extraction strategies. While such designs are not universally deployed, they clarify the design space and its trade-offs~\cite{piet2023mevade}.
\end{itemize}

In summary, MEV is a fundamental property of the on-chain execution environment, not a rare anomaly. Agents must be designed under the assumption that valuable actions disclosed to public orderflow will be observed and acted upon. A conservative design rule is: \textbf{avoid exposing valuable intent to the public mempool unless the execution strategy and constraints are MEV-aware}~\cite{daian2019flash}.

\subsubsection{Historical Foundations of Byzantine Fault Tolerance}

The theoretical foundations of distributed consensus predate blockchain technology by several decades. The Byzantine Generals Problem, first formalized in the early 1980s, established the fundamental impossibility results and lower bounds that continue to constrain modern distributed systems. Foundational work on practical Byzantine fault tolerance (PBFT) demonstrated how to build a replicated state machine that could tolerate Byzantine failures~\cite{castro1999practical,castro2002practical}. More recent surveys have provided a comprehensive overview of the BFT landscape, including both classic and more recent protocols~\cite{fu2021survey}. The security of mobile blockchain systems has also been studied, with a focus on the unique challenges of operating in a mobile environment~\cite{deval2024mobile}. The application of blockchain technology to supply chain management has also been explored, with a focus on the security and transparency benefits~\cite{ferdous2020blockchain}. The Bedrock protocol offers a new approach to building high-performance, scalable BFT systems~\cite{amiri2024bedrock}. The T-Chain architecture proposes a novel approach to building scalable and efficient blockchain systems~\cite{gao2019t}. These foundational results demonstrated that achieving consensus in the presence of arbitrary (Byzantine) failures requires at least $3f+1$ nodes to tolerate $f$ faulty participants, a bound that remains central to the design of permissioned blockchain systems and multi-party computation protocols.

The practical implementation of Byzantine fault-tolerant (BFT) consensus saw significant advances with the introduction of Practical Byzantine Fault Tolerance (PBFT), which demonstrated that BFT could be achieved with acceptable performance in realistic network conditions. Subsequent optimizations, including HotStuff and its variants, reduced the communication complexity from $O(n^2)$ to $O(n)$ per view, enabling BFT consensus to scale to larger validator sets. These advances are directly relevant to agent-blockchain systems because they determine the trust assumptions and finality guarantees that agents must reason about when constructing and submitting transactions.

\subsubsection{The MEV Supply Chain and Its Implications for Agents}

The MEV supply chain has evolved into a sophisticated ecosystem with multiple specialized actors, each playing distinct roles in the extraction and distribution of value from transaction ordering. Understanding this supply chain is essential for agents that seek to either protect against MEV extraction or participate in MEV capture.

At the base of the supply chain are searchers, specialized actors who identify MEV opportunities by monitoring pending transactions, analyzing smart contract state, and simulating potential extraction strategies. Searchers invest heavily in infrastructure for low-latency mempool access, sophisticated simulation capabilities, and strategy development. The searcher landscape is highly competitive, with successful searchers often operating at the frontier of algorithmic trading and blockchain technology.

Builders aggregate transactions from searchers and other sources into complete blocks, optimizing for total extractable value while respecting protocol rules and builder-specific policies. The builder role emerged with the separation of block building from block proposal in proof-of-stake Ethereum, creating a market for block construction that enables specialization and competition. Builders compete for inclusion by offering higher payments to proposers, creating pressure for efficient MEV extraction.

Proposers (validators in proof-of-stake systems) select blocks from builders and propose them to the network. Through mechanisms like MEV-Boost, proposers can outsource block construction to builders while retaining the right to select among competing blocks. This separation enables proposers to capture MEV without developing specialized extraction capabilities, while creating a market that allocates block space efficiently.

For agent systems, this supply chain creates both risks and opportunities. Agents that submit transactions through public mempools expose themselves to MEV extraction by searchers. Conversely, agents with sophisticated capabilities may participate as searchers themselves, capturing MEV opportunities that arise from market inefficiencies or other agents' activities. The design of agent transaction submission strategies must account for this competitive landscape.

\paragraph{MEV Protection Strategies for Agents}

Several strategies can help agents mitigate MEV exposure. Private transaction submission, through services like Flashbots Protect or private mempools operated by builders, prevents searchers from observing pending transactions and constructing extraction strategies. However, private submission introduces trust dependencies on the private mempool operators and may not provide complete protection against all forms of MEV.

Intent-based execution, through protocols like CoW Protocol~\cite{cowDocs} or 1inch Fusion, delegates transaction construction to specialized solvers who compete to provide optimal execution. These protocols can internalize MEV that would otherwise be extracted by external searchers, returning value to users in the form of better prices. However, intent protocols introduce their own trust assumptions and may not be suitable for all transaction types.

MEV-aware transaction construction involves designing transactions to minimize extraction opportunities. Techniques include setting tight slippage limits, splitting large orders across multiple transactions, timing transactions to avoid predictable patterns, and using commit-reveal schemes to hide transaction details until execution. These techniques require sophisticated understanding of MEV dynamics and may not eliminate extraction entirely.

\subsection{Multi-Party Computation (MPC)}

While account abstraction adds programmability at the wallet \emph{logic} layer, another complementary primitive, \textbf{Multi-Party Computation (MPC)}, strengthens the wallet \emph{key-custody} layer by addressing one of the most common catastrophic failure modes in on-chain automation: compromise of a single signing key. MPC enables multiple parties to jointly compute a function over private inputs without revealing those inputs to one another. In wallet engineering, its most widely deployed application is \textbf{threshold signing}, where distributed key shares jointly produce standard signatures (e.g., ECDSA) without reconstructing the full secret key in any one place~\cite{komlo2024threshold,lindell2017fast}.

\subsubsection{Core principle: eliminating the single point of failure}

In conventional self-custody, a single private key (or seed phrase) represents a single point of failure: if compromised, all associated assets and authorizations are immediately at risk. MPC-based custody mitigates this by ensuring that the signing secret is \emph{never materialized as a single plaintext key} during key generation, storage, or signing. Instead, the key is represented as \emph{shares} distributed across independent parties (devices, services, or security modules), and signing is performed through an interactive protocol that outputs a standard signature while keeping the key shares private~\cite{lindell2017fast,komlo2024threshold}.

The security benefit is direct: an attacker who compromises only the agent host (or only a user device) gains at most a share, which is insufficient to produce signatures on its own. This increases the adversary’s required capability from a single-system compromise to a multi-system compromise (or collusion), which is substantially harder in realistic operational environments.

\subsubsection{Threshold Signature Schemes (TSS): the key enabler}

Threshold signature schemes implement a \emph{$t$-of-$n$} model:

\begin{itemize}[leftmargin=*]
\item \textbf{$n$} parties each hold a key share.
\item \textbf{$t$} is the minimum number of parties required to produce a valid signature.
\end{itemize}

Any subset of at least $t$ parties can collaboratively generate a valid signature under the target algorithm (commonly ECDSA for Ethereum-style accounts), while any coalition of at most $t-1$ parties learns nothing useful about the underlying signing key beyond what is implied by their own shares~\cite{lindell2017fast}. Modern constructions use multi-round protocols with carefully designed commitments and consistency checks to jointly derive signing nonces and signature components without reconstructing the key.

\paragraph{MPC-TSS vs.\ on-chain multi-sig.}
It is important to distinguish MPC-TSS from on-chain multi-signature smart contracts (e.g., Safe\{Wallet\}). In a multi-sig contract, multiple EOAs produce multiple signatures and the contract enforces the quorum \emph{on-chain}. In MPC-TSS, the quorum is enforced \emph{cryptographically off-chain} to produce a single standard signature that appears indistinguishable from a traditional single-signer transaction on-chain~\cite{komlo2024threshold}.

\begin{table}[t]
 \centering
 \caption{Comparison of on-chain multi-signature versus MPC-based threshold signing (MPC-TSS).}
 \label{tab:multisig_vs_mpc}
 \small
 \setlength{\tabcolsep}{5pt}
 \begin{tabular}{p{3.2cm}p{6.2cm}p{6.2cm}}
 \toprule
 \textbf{Dimension} & \textbf{On-Chain Multi-Sig} & \textbf{MPC-TSS} \\
 \midrule
 \textbf{Mechanism} & Smart contract enforces a quorum of multiple EOA signatures. &
 Interactive protocol among parties produces one standard signature. \\
 \textbf{Chain compatibility} & Depends on contract support and account type. &
 Chain-agnostic if the chain accepts the target signature algorithm (e.g., ECDSA). \\
 \textbf{On-chain footprint} & Multi-sig structure and signers are visible on-chain. &
 Appears as a standard single-signature transaction on-chain. \\
 \textbf{Fees and overhead} & Higher due to contract execution and calldata. &
 Standard transaction fees; overhead is off-chain communication. \\
 \textbf{Policy evolution} & Many policy changes require an on-chain update/transaction. &
 Many operational policies can be modified off-chain (subject to the custody design). \\
 \bottomrule
 \end{tabular}
\end{table}

\subsubsection{Security model and trust assumptions}

The $t$-of-$n$ threshold parameter makes the trust assumption explicit: the system remains secure as long as fewer than $t$ parties are compromised or collude. In operational custody products, this independence is reinforced through organizational and infrastructural separation (distinct machines, environments, and access controls) and through hardened signing components such as HSM-backed services where appropriate~\cite{llp2023digital,shbair2021hsm}. For an agentic deployment, parties might include:

\begin{itemize}[leftmargin=*]
\item a user-controlled device,
\item an agent execution host,
\item an independent policy or approval service, and/or
\item a dedicated security module (e.g., HSM-backed signer).
\end{itemize}

This separation is valuable because it converts many common failures (agent host compromise, cloud credential leakage, prompt injection leading to malicious transaction construction) into \emph{policy and quorum} problems rather than immediate asset-loss events.

\subsubsection{Operational trade-offs: latency, availability, and DoS}

MPC-TSS is not free. Because signing is interactive, it introduces \textbf{latency} (multiple rounds of communication) and \textbf{availability coupling} (a quorum of parties must be online and responsive at signing time). In adversarial settings, these properties can be targeted: an attacker who cannot steal funds may still attempt to \emph{degrade service} by preventing quorum formation (e.g., network disruption against one party), effectively inducing a denial-of-service on the wallet’s ability to execute time-sensitive actions. Practical deployments therefore balance security and liveness by choosing thresholds and redundancy carefully (e.g., $t{=}2$ of $n{=}3$ rather than $3$ of $3$), and by hardening communication channels and monitoring for quorum failures~\cite{komlo2024threshold}.

A further engineering consideration is \textbf{policy placement}. If a policy service is included as a mandatory signer, it becomes a critical availability dependency; if it is optional, policy becomes advisory rather than enforcing. Agent designers should treat these choices as part of the threat model: the ``secure'' configuration must remain usable under realistic partial failures.

\subsubsection{Implications for agentic systems}

MPC-TSS provides a practical foundation for deploying on-chain agents with shared control:

\begin{itemize}[leftmargin=*]
\item \textbf{Agent-initiated, user-authorized execution.} The agent can construct a transaction (or \texttt{UserOperation} in AA settings) and contribute its signing share, but execution requires additional shares, such as a user approval share from a phone or hardware wallet. This yields cryptographically enforced human-in-the-loop control without blocking the agent’s automation benefits.

\item \textbf{Policy-enforced veto power.} A policy service can be included as a required party in the signing quorum. If a transaction violates constraints (spending limits, allowlists, risky contract interactions), the policy party refuses to participate, preventing signature formation and thereby preventing execution.

\item \textbf{Compromise containment and response time.} If the agent host is compromised, the attacker may obtain only a single share, which is insufficient for signing. This buys time for key rotation, share revocation, or reconfiguration of the quorum without requiring immediate asset migration to an entirely new account.
\end{itemize}

In summary, MPC and threshold signing provide a chain-agnostic and privacy-preserving custody layer that complements on-chain programmability. Combined with agentic execution, they enable architectures in which agents operate with meaningful autonomy while remaining cryptographically tethered to user consent and enforceable policy constraints~\cite{komlo2024threshold,lindell2017fast}.

\subsubsection{Smart Contract Security and Verification}

The security of smart contracts has been a persistent challenge since the earliest days of programmable blockchains. High-profile incidents, including the DAO hack and numerous DeFi exploits, have demonstrated that even audited contracts can contain subtle vulnerabilities that lead to catastrophic losses. The formal verification of smart contracts is a key technique for ensuring their correctness and security~\cite{ethereum2025formal}. Several automated tools have been developed for detecting vulnerabilities in smart contracts, including static analysis tools and fuzzers~\cite{ghaleb2023achecker,iuliano2025automated}. The legal implications of smart contracts have also been explored, with a focus on the challenges of applying traditional legal frameworks to this new technology~\cite{governatori2018legal}. The use of blockchain technology for supply chain management has also been studied, with a focus on the security and transparency benefits~\cite{gatteschi2019technology}. Recent guides have provided a comprehensive overview of the smart contract security landscape, including common vulnerabilities and best practices for secure development~\cite{behnke2023guide,certik2023what}. The security of smart contracts in the context of the Internet of Things (IoT) has also been explored, with a focus on the unique challenges of securing resource-constrained devices~\cite{chaliasos2024smart}. For agent-mediated transactions, this security landscape is particularly relevant because agents must not only construct valid transactions but also reason about the safety of the contracts they interact with.

Formal verification techniques have emerged as a promising approach to smart contract security, enabling mathematical proofs of correctness properties. Tools such as Certora, Echidna, and Slither provide varying levels of assurance, from lightweight static analysis to full formal verification. However, the practical application of these tools remains limited by the complexity of real-world contract interactions, particularly in composable DeFi protocols where emergent behaviors arise from the interaction of multiple contracts. Agent systems that interact with smart contracts must therefore incorporate runtime monitoring and simulation capabilities to detect potentially dangerous interactions before committing irreversible transactions.

\subsubsection{Oracle Design Patterns and Security Considerations}

Recent surveys have highlighted key open research directions for blockchain oracles, including the need for more robust security models, better economic incentives, and more decentralized architectures~\cite{ezzat2022blockchain}.

Oracles serve as the bridge between blockchain smart contracts and external data sources, enabling contracts to react to real-world events and conditions. For agent systems that interact with oracle-dependent protocols, understanding oracle design patterns and their security implications is essential.

\paragraph{Price Oracle Architectures}

Price oracles, which provide asset prices to DeFi protocols, employ various architectures with different security and liveness properties. Time-weighted average price (TWAP) oracles compute prices as averages over defined time windows, providing resistance to short-term manipulation at the cost of latency in reflecting current prices. Median oracles aggregate prices from multiple sources and report the median value, providing resistance to outliers and single-source manipulation.

Push-based oracles actively submit price updates to on-chain contracts, enabling protocols to access current prices without additional transactions. Pull-based oracles require protocols or users to request price updates, reducing on-chain costs but introducing latency and potential availability concerns. Hybrid approaches combine push and pull mechanisms to balance cost, latency, and availability.

\paragraph{Oracle Security Threats}

Oracle manipulation represents a significant threat to DeFi protocols and the agents that interact with them. Flash loan attacks can temporarily distort prices on decentralized exchanges, causing TWAP oracles to report manipulated values. Multi-block attacks can manipulate prices over longer periods, affecting oracles with longer averaging windows. Economic attacks can exploit the gap between oracle-reported prices and true market prices to extract value from protocols.

Agents must account for oracle security when interacting with oracle-dependent protocols. This includes understanding the oracle architecture used by each protocol, monitoring for signs of oracle manipulation, and implementing safeguards against acting on manipulated data. For high-stakes operations, agents may cross-reference multiple oracle sources and implement sanity checks on reported values.

\paragraph{Emerging Oracle Patterns}

Several emerging patterns address limitations of traditional oracle designs. Optimistic oracles allow anyone to submit data, with a dispute period during which incorrect submissions can be challenged. This pattern reduces operational costs and latency while maintaining security through economic incentives. Zero-knowledge oracles use cryptographic proofs to verify data authenticity without revealing underlying sources, enabling privacy-preserving data feeds.

Agent-specific oracle patterns are also emerging, designed to provide the types of data that agents need for decision-making. These include sentiment oracles that aggregate social media and news data, risk oracles that provide protocol-level risk assessments, and strategy oracles that provide signals for specific trading strategies. As agent adoption grows, the oracle ecosystem will likely evolve to better serve agent-specific needs.

\subsection{DeFi Primitives: The Agent's Playground}

To understand what an on-chain agent can safely do, we must be precise about the \emph{market microstructure} it operates within. Decentralized finance (DeFi) is not one protocol but a composable stack of financial primitives with sharply different risk surfaces, latency sensitivities, and adversarial incentives. In this survey, we treat DeFi as the agent's primary operational domain because most agent-to-chain workloads ultimately reduce to (i) exchange, (ii) collateralized credit, and (iii) stable-denominated accounting. we therefore focus on three foundational primitives: automated market makers (AMMs), lending protocols, and stablecoins. The technology of decentralized finance has been systematized in several recent surveys~\cite{auer2024technology,xu2023sok}. The formal theory of automated market makers has been studied in detail~\cite{bartoletti2021sok,bartoletti2022theory}. The security of DeFi protocols has also been a major area of research~\cite{reddy2019protocols}.

\subsubsection{Automated Market Makers (AMMs): The Engine of Decentralized Exchange}

\textbf{Concept.} AMMs implement decentralized exchange without a traditional order book by pricing assets algorithmically as a function of pool reserves. This design yields continuous liquidity and composability, but it also couples execution quality to pool depth and transaction ordering. Uniswap-style constant-function market makers have been studied extensively. Early work analyzed the formal properties of constant-function market makers, including their pricing behavior and impermanent loss characteristics~\cite{angeris2019analysis}. The Uniswap protocol itself has been the subject of detailed analysis, including its market behavior and economic implications~\cite{adams2021uniswap}. More recent work has focused on creating a more general theory of automated market makers, including concentrated-liquidity designs~\cite{bartoletti2022theory,kulkarni2023towards}. The broader DeFi ecosystem, including AMMs, has also been systematized in several recent surveys~\cite{xu2023sok}. While constant-product AMMs (CPMMs) are the canonical example, modern AMMs generalize to richer invariants (including concentrated-liquidity designs) and therefore introduce additional state and parameterization that agents must model correctly~\cite{adams2021uniswap,xu2023sok}.

\textbf{Implications for agents.}
\begin{itemize}
\item \textbf{Action space.} The agent’s core actions are \texttt{swap}, \texttt{addLiquidity}, and \texttt{removeLiquidity}, usually through router contracts. Even a “simple” swap is a multi-constraint optimization: the agent must choose route(s), set \texttt{deadline} and \texttt{minAmountOut}, and account for token decimals, approvals, and potential permit flows~\cite{xu2023sok}.
\item \textbf{Execution risk (slippage and ordering).} AMM execution is highly sensitive to transaction ordering and transient state. In public mempools, the combination of price impact and observability creates predictable extraction opportunities (for example, sandwich-style deterioration of execution price). Consequently, agents should treat \texttt{minAmountOut} and route selection as \emph{security parameters}, not just UX parameters~\cite{heimbach2022sok,kulkarni2023towards}.
\item \textbf{Liquidity-provision risk (impermanent loss).} If the agent provides liquidity, it is exposed to impermanent loss when relative prices move, as well as to fee-earning dynamics that depend on volume, volatility, and pool design. Concentrated liquidity amplifies both opportunity and risk by making the LP position path-dependent and sensitive to range selection~\cite{adams2021uniswap,bartoletti2022theory}.
\end{itemize}

\subsubsection{Lending Protocols: The Decentralized Money Market}

\textbf{Concept.} Lending protocols implement over-collateralized credit markets where lenders supply assets into pools and borrowers draw against collateral subject to risk parameters (loan-to-value, liquidation thresholds, and protocol-defined incentives). The security and economic dynamics of lending pools have been analyzed in dedicated surveys, which is important because these systems couple smart-contract correctness with oracle integrity and liquidation-market behavior~\cite{bartoletti2021sok,li2022survey}.

\textbf{Implications for agents.}
\begin{itemize}
\item \textbf{Action space.} Agents can \texttt{deposit}/\texttt{withdraw}, \texttt{borrow}/\texttt{repay}, adjust collateral composition, and, in some protocols, manage rate modes or leverage loops. Operationally, these actions often require sequencing approvals and protocol calls correctly, making simulation and preview steps particularly valuable.
\item \textbf{Risk management.} Borrowing introduces continuous solvency constraints (health-factor style monitoring). Agents must track collateral valuation (oracle feeds), interest accrual, and liquidation thresholds, and they must plan mitigation actions (top-up collateral, partial repayment, deleveraging) under latency and fee constraints~\cite{bartoletti2021sok,li2022survey}.
\item \textbf{Liquidations as an adversarial subgame.} Liquidation is a competitive, time-sensitive market with MEV-like dynamics: many actors observe the same under-collateralized positions and race to capture liquidation incentives. An agent that acts as a liquidator must incorporate execution reliability, gas dynamics, and ordering risk into its strategy rather than treating liquidation as a static API call~\cite{heimbach2022sok,materwala2025maximal}.
\end{itemize}

\subsubsection{Stablecoins: The Unit of Account and the Hidden Dependency}

\textbf{Concept.} Stablecoins provide the unit of account for most DeFi activity and are the default denomination for treasury management, risk controls, and performance reporting. For agentic systems, stablecoins are not just an asset class; they are an implicit \emph{assumption} in objective functions, budget constraints, and safety policies (for example, “do not lose more than \$X per day”). Broad DeFi overviews emphasize that stablecoins underpin liquidity, collateral, and settlement, while also introducing issuer, collateral, and mechanism-specific risks that propagate system-wide during stress~\cite{auer2024technology,reddy2019protocols}.

\textbf{Implications for agents.}
\begin{itemize}
\item \textbf{Policy-aware denomination.} Agents should encode policies in stable-denominated terms (limits, exposure caps, and loss thresholds), but also bind those policies to \emph{which} stablecoin(s) are acceptable for accounting and settlement (because “stable” is not a guarantee).
\item \textbf{Mechanism risk.} Different designs imply different failure modes (custodial/fiat-backed, crypto-collateralized, or algorithmic). Agents should treat stablecoin selection as part of risk posture: acceptable instruments in normal conditions may be disallowed under stress or depeg signals.
\item \textbf{Composability risk.} Stablecoin exposure is frequently indirect (as collateral, LP inventory, or protocol treasury assets). Agents therefore need portfolio-level visibility, not just per-transaction checks, to avoid concentrating depeg risk across supposedly independent positions.
\end{itemize}

\paragraph{Why this matters for agent design.} These primitives define the practical action space for agent-to-chain systems, but they also define the \emph{threat surface}. AMMs expose agents to ordering and price-impact extraction; lending protocols expose agents to oracle and liquidation dynamics; stablecoins anchor accounting while importing systemic assumptions. Accordingly, later sections map interface standards and safety controls onto these primitives, emphasizing simulation, policy gating, and MEV-aware execution as baseline requirements rather than optional enhancements~\cite{heimbach2022sok,li2022survey,materwala2025maximal}.

\subsubsection{Market Dynamics and Agent Behavior}

The introduction of autonomous agents into DeFi markets has the potential to fundamentally alter market dynamics~\cite{oosthoek2021flash}. Agent-mediated trading can increase market efficiency by enabling faster arbitrage, more responsive liquidity provision, and more sophisticated risk management. However, it also introduces new risks, including the potential for correlated agent behavior to amplify market volatility and the emergence of agent-to-agent MEV extraction.

The economic incentives facing agent operators may not always align with broader market health. Agents optimizing for short-term profit extraction may engage in behaviors that, while individually rational, collectively degrade market quality. Examples include aggressive MEV extraction that increases transaction costs for other users, liquidity provision strategies that withdraw during periods of high volatility, and trading patterns that exploit information asymmetries at the expense of less sophisticated participants.

Understanding these dynamics requires new analytical frameworks that account for the strategic interactions between multiple autonomous agents operating in shared market environments. Game-theoretic analysis, agent-based modeling, and empirical studies of agent behavior in production environments all contribute to building this understanding. The insights from this research should inform both the design of agent systems and the governance of the markets in which they operate.

\subsubsection{Advanced DeFi Mechanisms and Risk Surfaces}

The 2020 DeFi crisis highlighted the systemic risks inherent in highly interconnected DeFi protocols, including liquidation cascades and oracle failures~\cite{gudgeon2020decentralized}. Agents operating in these environments must be designed to be resilient to such black swan events.

Beyond the foundational primitives of AMMs and lending protocols, the DeFi ecosystem has evolved to include increasingly sophisticated financial instruments. Yield aggregators automatically optimize returns across multiple protocols, and their design and security have been systematized in recent work~\cite{cousaert2022sok}. The security of yield aggregators has also been a major area of research, with a focus on identifying and mitigating common vulnerabilities~\cite{qin2021attacking}. The flow of funds between different DeFi protocols has also been studied, with a focus on understanding the systemic risks of the ecosystem~\cite{pornprasith2022flow}. The concept of zero-knowledge proofs, which is a foundational cryptographic primitive, has also been applied to DeFi to enable privacy-preserving transactions~\cite{goldreich1991proofs}. The security of multi-chain DeFi protocols has also been explored, with a focus on the unique challenges of securing cross-chain communication~\cite{yuan2022multi}. These advanced mechanisms create both opportunities and risks for agent-mediated execution.

Flash loans represent a particularly significant innovation for agent systems. By enabling uncollateralized borrowing within a single atomic transaction, flash loans allow agents to execute complex arbitrage strategies, liquidations, and capital-efficient operations that would otherwise require substantial upfront capital. However, flash loans also enable novel attack vectors, as demonstrated by numerous exploits that have used flash-borrowed capital to manipulate oracle prices or drain protocol reserves. Agent systems must therefore be designed with awareness of flash loan dynamics, both as a tool for legitimate operations and as a threat vector that adversaries may employ.

\subsection{Layer 2 Scaling: The Agent's High-Speed Execution Venue}

While Layer~1 blockchains such as Ethereum provide a robust settlement layer, their limited throughput and variable, often high transaction fees make them a poor fit for many agentic workloads that require frequent, low-latency interactions (for example, iterative portfolio rebalancing, continuous risk monitoring with corrective actions, or liquidation-aware credit management). This gap has driven the rapid adoption of \textbf{Layer~2 (L2) scaling solutions}, which aim to execute transactions off-chain while anchoring security and data availability to the underlying L1. For agents, L2s frequently become the \emph{default execution venue} because they reduce per-action cost and enable higher action frequency without turning gas fees into the dominant term in the objective function. Rollups, in particular, represent the dominant design family and have been comprehensively systematized in the scaling literature~\cite{thibault2022blockchain}.

\subsubsection{The core idea: off-chain execution, on-chain settlement}

The unifying idea behind rollups is to relocate the expensive work of \emph{execution} to an L2 environment while retaining L1-enforced settlement guarantees by committing batches of transactions (or their compressed representations) to an L1 smart contract. An L2 operator (often called a \emph{sequencer}) orders and executes transactions at high speed, then periodically posts a batch commitment to L1. The L1 contract validates the batch according to the rollup’s security model and updates an on-chain commitment to the L2 state (typically a state root). This amortizes the on-chain verification and data costs across many L2 transactions and yields substantial throughput improvements relative to L1 execution~\cite{thibault2022blockchain}.

For agent design, this distinction matters because \emph{execution} happens under L2-specific assumptions (sequencing, latency, fee markets), whereas \emph{settlement and finality} inherit constraints from L1 posting cadence and proof/verification mechanisms. Agents that ignore this split often mis-handle confirmation semantics, reorg windows, and cross-domain capital constraints.

\subsubsection{Optimistic rollups: innocent until proven guilty}

\textbf{Concept.} Optimistic rollups (for example, Arbitrum and Optimism) post batch commitments to L1 and assume correctness \emph{optimistically}, with correctness enforced via a \textbf{challenge period} during which invalid state transitions can be proven and rejected using \textbf{fraud proofs}~\cite{thibault2022blockchain}. The core enforcement mechanism is that anyone can dispute a posted assertion by presenting evidence that some transaction (or computation step) was executed incorrectly, triggering an on-chain dispute process that resolves the claim.

\textbf{Implications for agents.}
\begin{itemize}
\item \textbf{Low-cost, high-frequency execution.} Transaction fees are generally lower than L1, enabling finer-grained control loops (for example, smaller but more frequent rebalances or risk adjustments) that would be uneconomical on L1.
\item \textbf{EVM compatibility.} Many optimistic rollups pursue EVM equivalence or close compatibility, which reduces engineering friction: the same contracts, SDKs, and transaction construction logic often transfer with minimal changes.
\item \textbf{Withdrawal latency and capital planning.} The challenge window creates \textbf{asymmetric finality}: intra-L2 actions may confirm quickly, but withdrawals to L1 typically require waiting for the dispute period (commonly on the order of days). Agents that require cross-domain liquidity must explicitly account for this delay in planning and risk management (for example, avoiding strategies that implicitly assume immediate L1 settlement)~\cite{thibault2022blockchain}.
\end{itemize}

\subsubsection{ZK-rollups: trust through validity proofs}

Beyond rollups, zero-knowledge proofs have been applied to enable private function execution on-chain, where the function logic itself is hidden from public view~\cite{bowe2020zexe}. This allows for more complex privacy-preserving applications where agents can interact with smart contracts without revealing their internal state or logic.

\textbf{Concept.} ZK-rollups (for example, zkSync-style systems and STARK-based designs) enforce correctness by attaching a \textbf{validity proof} to each posted batch. Rather than waiting for a challenge, the rollup proves that the batch’s state transition is correct using succinct cryptographic proofs (commonly zk-SNARKs or zk-STARKs). The L1 contract verifies a single proof for the batch and, upon verification, accepts the state update as valid~\cite{thibault2022blockchain,chen2022review}. This model removes the need for long fraud-proof windows and can offer faster, more symmetric settlement to L1, subject to proof generation and posting cadence.

\textbf{Implications for agents.}
\begin{itemize}
\item \textbf{Faster L1 exits and improved capital efficiency.} Because batches are accepted once validity is proven, L1 withdrawal latency is typically constrained by proof generation and L1 inclusion rather than multi-day dispute windows. This can materially improve capital efficiency for strategies that frequently rebalance across domains~\cite{thibault2022blockchain}.
\item \textbf{Complexity and maturity.} ZK proof systems are substantially more complex than optimistic dispute systems, and ZK-EVM equivalence is technically demanding. As a result, practical differences across ZK-rollups (VM semantics, tooling maturity, opcode coverage, performance envelopes) can be larger than in optimistic ecosystems, and agents should not assume perfect portability.
\item \textbf{Cost structure.} Proof generation is computationally intensive and can shift costs from on-chain verification to off-chain proving infrastructure. The resulting fee dynamics differ across designs and may vary over time as proving systems and hardware improve~\cite{thibault2022blockchain,chen2022review}.
\end{itemize}

\subsubsection{The L2 landscape: fragmentation, bridges, and agent-facing risk}

Atomic cross-chain swaps are a foundational technique for trustless exchange of assets between different blockchains, often using hash time-locked contracts (HTLCs)~\cite{herlihy2018atomic}. Proof-of-stake sidechains offer a mechanism for creating interoperable blockchains with their own security and governance models, which can be connected to a main chain via a bridge~\cite{gai2019proof}.

For an agent, the L2 ecosystem is not a single execution venue but a fragmented set of domains with distinct sequencers, fee markets, confirmation semantics, and application deployments. A capable agent therefore needs to:
\begin{itemize}
\item \textbf{Navigate a multi-L2 world.} Maintain awareness of where relevant applications and liquidity reside, and select venues accordingly rather than assuming L1 is the canonical execution site.
\item \textbf{Model domain-specific finality.} Treat confirmation and settlement as domain-dependent (L2 local confirmation vs.\ L1 finality), especially for strategies that depend on cross-domain state consistency.
\item \textbf{Manage bridge risk as a first-class threat.} Moving assets between L1 and L2s (and across L2s) frequently relies on bridges and cross-chain messaging systems, which have repeatedly represented concentrated risk and a high-value attack surface. Agents that automate bridging must incorporate bridge security posture, monitoring, and failure handling into policy and execution logic rather than treating bridging as a routine transfer~\cite{beniiche2020study,augusto2024xchainwatcher}.
\end{itemize}

In summary, L2s are a major enabler for agentic on-chain systems because they provide the throughput and cost profile needed for frequent interaction. At the same time, they introduce new operational semantics and cross-domain risks. Secure agent architectures should therefore treat L2 selection, confirmation semantics, and bridge interactions as explicit components of the agent action pipeline, with policy gating and verification adapted to the rollup security model in use~\cite{thibault2022blockchain}.

\subsubsection{Layer 2 Selection Criteria for Agent Systems}

The proliferation of Layer 2 scaling solutions creates a complex decision landscape for agent systems. Different L2s offer varying tradeoffs across multiple dimensions, and optimal selection depends on the specific requirements of each use case.

\paragraph{Security Model Considerations}

The security model of an L2 determines the trust assumptions required for safe operation. Optimistic rollups inherit Ethereum security after a challenge period (typically 7 days), during which fraudulent state transitions can be disputed. ZK-rollups provide immediate cryptographic finality through validity proofs, but the complexity of proof generation introduces different trust considerations around prover implementation correctness.

For agent systems, security model implications include withdrawal timing (agents may need to plan for extended withdrawal periods from optimistic rollups), finality assumptions (agents must understand when transactions can be considered final for different L2s), and failure mode handling (agents must have strategies for L2 failures or disputes).

\paragraph{Performance Characteristics}

L2 performance varies significantly across throughput, latency, and cost dimensions. High-throughput L2s can process thousands of transactions per second, enabling strategies that would be prohibitively expensive on L1. Low-latency L2s provide faster confirmation times, enabling more responsive agent behaviors. Low-cost L2s reduce the economic threshold for profitable operations, expanding the range of viable agent strategies.

Agent systems should match L2 selection to strategy requirements. High-frequency trading strategies benefit from low-latency, high-throughput L2s. Cost-sensitive strategies benefit from L2s with minimal transaction fees. Strategies requiring interaction with L1 liquidity must account for bridging costs and delays.

\paragraph{Ecosystem and Liquidity Considerations}

Beyond technical characteristics, L2 selection must consider ecosystem factors including available liquidity, protocol deployments, and developer tooling. An L2 with superior technical properties but limited liquidity may be unsuitable for strategies that require deep markets. Similarly, the availability of key protocols (DEXs, lending platforms, etc.) constrains the strategies that can be executed on each L2.

Agent systems operating across multiple L2s must manage the complexity of heterogeneous environments. This includes maintaining separate state tracking for each L2, handling different RPC interfaces and transaction formats, and managing cross-L2 bridging for strategies that span multiple networks.

\section{Taxonomy of Agent--Blockchain Integration Patterns (2025)}
\label{sect:taxonomy_patterns}

\begin{figure}[ht]
 \centering 
 \begin{tikzpicture}[
 font=\scriptsize,
 box/.style={draw, rounded corners=2mm, align=left, inner sep=5pt, fill=white},
 stage/.style={box, text width=0.82\linewidth},
 tag/.style={draw, rounded corners=1.5mm, inner sep=2pt, fill=black!5},
 arr/.style={->, line width=0.7pt},
 bar/.style={draw, line width=0.7pt}
 ]

 \node[tag, text width=0.96\linewidth, align=center] (hdr)
 {\textbf{Five integration patterns (2025):} increasing authority, stronger interfaces, and sharper safety requirements};

 \node[stage, below=5mm of hdr] (p1) {
 \textbf{I. Read-only chain analytics}\par
 \textit{Agent role:} query, summarize, alert.\quad
 \textit{Boundary crossed:} \textbf{data plane} (RPC/indexers).\par
 \textit{Typical artifacts:} decoded events, position reports, risk signals.\quad
 \textit{Dominant failures:} spoofed/poisoned data, stale views, indexer divergence.
 };

 \node[stage, below=4mm of p1] (p2) {
 \textbf{II. Intent generation (recommendation)}\par
 \textit{Agent role:} propose actions.\quad
 \textit{Boundary crossed:} \textbf{decision plane} (simulation, intent schema).\par
 \textit{Typical artifacts:} human-readable plan, intent object, constraints.\quad
 \textit{Dominant failures:} misleading intent, unsafe defaults, approval traps.
 };

 \node[stage, below=4mm of p2] (p3) {
 \textbf{III. Delegated execution (scoped authority)}\par
 \textit{Agent role:} execute within caps.\quad
 \textit{Boundary crossed:} \textbf{policy plane} (smart accounts, session keys).\par
 \textit{Typical artifacts:} policy checks, previews, bounded execution.\quad
 \textit{Dominant failures:} role escalation, policy bypass, unsafe module composition.
 };

 \node[stage, below=4mm of p3] (p4) {
 \textbf{IV. Autonomous signing (policy-constrained)}\par
 \textit{Agent role:} sign-and-send.\quad
 \textit{Boundary crossed:} \textbf{custody plane} (MPC/TEE/HSM/signing APIs).\par
 \textit{Typical artifacts:} signing sessions, attestations, audit trails.\quad
 \textit{Dominant failures:} key-share leakage, compromised runtime, replay/replace attacks.
 };

 \node[stage, below=4mm of p4] (p5) {
 \textbf{V. Multi-agent workflows (distributed/quorum)}\par
 \textit{Agent role:} propose--verify--execute across roles.\quad
 \textit{Boundary crossed:} \textbf{coordination plane} (shared policy, quorum control).\par
 \textit{Typical artifacts:} commitments, approvals, distributed execution.\quad
 \textit{Dominant failures:} collusion, governance capture, incentive misalignment.
 };

 \draw[arr] (p1.south) -- (p2.north);
 \draw[arr] (p2.south) -- (p3.north);
 \draw[arr] (p3.south) -- (p4.north);
 \draw[arr] (p4.south) -- (p5.north);

 \coordinate (axisTop) at ($(p1.west)+(-10mm,6mm)$);
 \coordinate (axisBot) at ($(p5.west)+(-10mm,-6mm)$);
 \draw[bar] (axisBot) -- (axisTop);
 \draw[arr] (axisBot) -- (axisTop);

 \node[rotate=90, align=center, font=\scriptsize]
 at ($(axisBot)!0.5!(axisTop)$)
 {Increasing agent authority\\and expected loss if misused};

 \coordinate (axis2Top) at ($(p1.east)+(10mm,6mm)$);
 \coordinate (axis2Bot) at ($(p5.east)+(10mm,-6mm)$);
 \draw[bar] (axis2Bot) -- (axis2Top);
 \draw[arr] (axis2Bot) -- (axis2Top);
 \node[rotate=90, align=center, font=\scriptsize]
 at ($(axis2Bot)!0.5!(axis2Top)$)
 {Harder interface requirements:\\simulation $\rightarrow$ policy $\rightarrow$ custody $\rightarrow$ coordination};

 \end{tikzpicture}
 \caption{Five agent--blockchain integration patterns as a progression across \emph{interface planes} (data, decision, policy, custody, coordination). The figure is an overview model; later subsections detail each pattern’s mechanisms, threat surface, and representative systems.}
 \label{fig:taxonomy_patterns}
\end{figure}
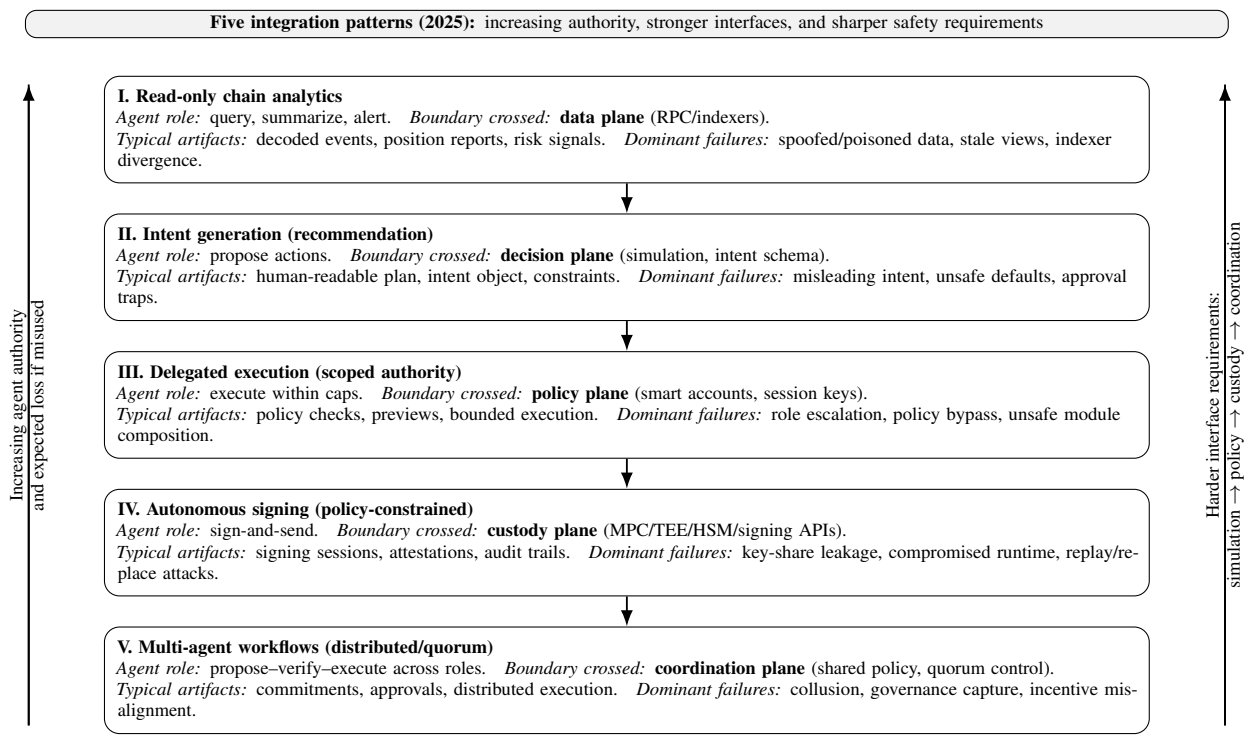

As the field of agent-blockchain interoperability matures, a set of distinct architectural patterns has emerged. These patterns are not merely technical curiosities; they represent a fundamental progression in trust, autonomy, and risk. Understanding these patterns is crucial for any developer, researcher, or user seeking to build or interact with agentic systems in the Web3 ecosystem. This section provides a deep dive into our proposed five-part taxonomy, using it as an organizing framework to analyze the security considerations, architectural trade-offs, and evolutionary pressures that define each stage. This taxonomy, illustrated in Figure~\ref{fig:taxonomy_patterns} and summarized in Table~\ref{tab:patterns}, charts a clear progression from passive, read-only agents to fully autonomous, multi-agent systems with direct, high-stakes signing capabilities.

\begin{table}[t]
 \centering
 \caption{Summary of agent--blockchain integration patterns, ordered by increasing agent authority and risk.}
 \label{tab:patterns}
 \scriptsize
 \setlength{\tabcolsep}{5pt}
 \begin{tabular}{p{2.6cm}p{2.7cm}p{3.5cm}p{3.2cm}p{2.9cm}}
 \toprule
 \textbf{Pattern} &
 \textbf{Agent authority} &
 \textbf{Primary interface plane} &
 \textbf{Dominant risk} &
 \textbf{Evolutionary pressure} \\
 \midrule
 \textbf{I. Read-only analytics} &
 None (informational) &
 Data plane (RPC, indexers, analytics) &
 Data spoofing, stale or biased views &
 Demand for actionable insights \\

 \textbf{II. Intent generation} &
 Recommendation only &
 Decision plane (simulation, intent schema) &
 Unsafe or misleading intent approval &
 Desire for partial automation \\
 
 \textbf{III. Delegated execution} &
 Scoped / role-based &
 Policy plane (smart accounts, session keys) &
 Policy bypass or role escalation &
 Safe limited autonomy \\
 
 \textbf{IV. Autonomous signing} &
 Full (policy-constrained) &
 Custody plane (MPC, TEE, signing APIs) &
 Key or agent compromise &
 High-stakes automation \\
 
 \textbf{V. Multi-agent workflows} &
 Distributed / quorum-based &
 Coordination plane (shared policy, workflow) &
 Collusion or governance capture &
 Trust-minimized control \\
 \bottomrule
 \end{tabular}
\end{table}
Taken together, Figure~\ref{fig:taxonomy_patterns} and Table~\ref{tab:patterns} provide a structural overview of how agent--blockchain integration evolves as authority shifts from observation to execution and coordination. The remainder of this section examines each pattern in turn, devoting a dedicated subsection to its core architecture, typical system realizations, and dominant threat surface. Rather than repeating the summaries above, each subsection focuses on the mechanisms that enable the pattern in practice, the specific failure modes that arise at that level of authority, and the interface and policy controls required to mitigate them. This organization allows the taxonomy to function both as a conceptual map and as a practical guide for system designers navigating different trust and autonomy regimes.

\subsection{Pattern I: Read-Only Chain Analytics Agents}

\textbf{Description.}
Pattern~I is the foundational integration mode: the agent is an \emph{observer} and analyst, not an executor. It functions as a natural-language control plane over heterogeneous on-chain data, translating user questions (e.g., ``Is this protocol safe?'' or ``What changed in the last 24 hours?'') into structured queries, then synthesizing results into an auditable explanation. Importantly, the agent has \emph{no authority} to submit transactions or mutate state; its output is advisory. This pattern aligns with the broader shift toward tool-using LLM systems, where the core capability is reliable retrieval and synthesis over external systems rather than generation alone~\cite{yuan2025easytool,wu2025dark}.

\textbf{Architectural deep dive.}
A typical Pattern~I system decomposes into three layers:

\begin{itemize}
\item \textbf{Data plane (reads).} The agent queries blockchain state through RPC endpoints (balances, storage slots, logs), and often relies on \emph{indexed} or \emph{pre-aggregated} services for historical queries and higher-level metrics (e.g., protocol activity, contract interactions, and entity-level heuristics). Because these sources expose different views (canonical chain state vs.\ indexed interpretations), robust designs treat them as complementary rather than interchangeable.

\item \textbf{Interpretation plane (structuring).} Retrieved signals are normalized into a compact world model (assets, positions, protocol contracts, timestamps, risk flags). This is the point where LLMs are most useful, but also most failure-prone: ambiguity in retrieved text, inconsistent schemas across providers, and mixing untrusted content with system instructions can all corrupt the agent's conclusions~\cite{greshake2023indirectpromptinjection}.

\item \textbf{Explanation plane (reporting).} The agent produces a human-consumable answer \emph{with provenance}: which sources were queried, what discrepancies were observed, and which assumptions remain uncertain. This is not merely UX polish; it is the main control that allows users to calibrate trust in an environment where inputs may be manipulated.
\end{itemize}

\textbf{Security model and risks.}
Although Pattern~I cannot directly lose funds by signing transactions, it can still precipitate loss through \emph{decision manipulation}. The dominant threat class is \textbf{integrity attacks on inputs}, including (i) compromised or malicious RPC/indexing providers, (ii) oracle or price-feed manipulation that distorts risk metrics, and (iii) indirect prompt injection delivered through untrusted retrieved content (web pages, documentation, governance posts, or even transaction metadata) that the agent treats as instructions rather than evidence~\cite{greshake2023indirectpromptinjection}. The practical security principle is therefore \textbf{source diversity with discrepancy surfacing}: query multiple independent sources for critical facts, separate policy/instructions from retrieved content, and explicitly flag conflicts instead of averaging them away. This ``verify-by-construction'' mindset is consistent with broader cross-domain security arguments that interoperability stacks amplify the consequences of weak trust boundaries~\cite{harris2023cross}.

\textbf{Evolutionary pressure.}
Pattern~I agents quickly encounter a usability ceiling: after delivering an analysis and recommending an action, the user must still execute the workflow manually (copy addresses, set slippage, choose venues, sign safely). This gap between \emph{insight} and \emph{action} creates the natural pressure toward Pattern~II, where the agent begins emitting structured \emph{candidate intents} or draft transactions for user approval rather than stopping at narrative advice~\cite{fan2024insight}.

\subsection{Pattern II: Simulation and Intent Generation}

\textbf{Description.}
In Pattern~II, the agent evolves from a passive analyst into an \emph{action-oriented co-pilot}. Rather than only describing the current state of the world, it proposes how to change it. Given a high-level user goal (e.g., ``maximize yield on 10{}000~USDC subject to low risk''), the agent constructs a proposed \textbf{transaction intent}: a structured, machine-readable specification of the desired outcome (\emph{the what}) and constraints (slippage bounds, venues, token allowlists, deadline), without yet binding to raw, chain-specific transaction bytes (\emph{the how}). The critical safeguard is that the agent does \emph{not} execute. Instead, it produces (i) an intent object, (ii) a clear human-readable preview, and (iii) a simulated outcome on a state fork, and then hands the decision to a human signer. This pattern operationalizes tool-using LLMs as planners and compilers of candidate actions while preserving user custody and final authorization~\cite{yuan2025easytool,wu2025dark}.

\textbf{Architectural deep dive.}
Pattern~II systems typically add two capabilities beyond Pattern~I: (a) \emph{counterfactual execution} via simulation, and (b) \emph{structured action representation} via intent schemas.

\begin{itemize}
\item \textbf{Simulation and preview layer (counterfactual execution).} The agent relies on transaction simulation (often via a forked state) to perform a pre-flight check before the user signs. A high-fidelity simulation returns a rich trace: state deltas (balances, approvals, positions), emitted events, gas usage, revert reasons, and call graphs. This is not merely debugging convenience; it is the primary control that narrows the gap between user intent and actual on-chain effects. A well-designed preview should surface \emph{semantic} consequences that users understand (``grant allowance to contract X'', ``swap USDC$\rightarrow$ETH'', ``open borrow position''), not only low-level calldata.

\item \textbf{Intent representation (from narrative goals to constrained actions).} The agent outputs a structured intent object that is stable across front-ends and, ideally, interoperable across wallets and execution back-ends. This intent layer is where constraints should live: approved asset lists, maximum slippage, target venues, risk posture, and explicit disallow rules (e.g., ``no unlimited approvals''). Intent-centric architectures have been emphasized as a path to safer composition because they make the \emph{goal} auditable independently of the compiled transaction sequence~\cite{mandal2025evaluating}.
\end{itemize}

\textbf{Reasoning pipeline.}
Reasoning in Pattern~II is inherently counterfactual: the agent must explore a space of possible actions and compare outcomes under constraints. A yield-seeking workflow, for example, often follows a loop:
\begin{enumerate}
\item \textbf{Enumerate candidates.} Identify candidate venues and instruments (swaps into yield-bearing assets, deposits into lending markets, staking routes), subject to user constraints (risk tier, chain, venue allowlists).
\item \textbf{Quote and plan.} Obtain quotes and construct candidate action sequences, including routing via aggregators where appropriate.
\item \textbf{Simulate.} Execute each candidate plan on a fork to obtain concrete outcomes (received amounts, approvals granted, gas costs, and failure modes).
\item \textbf{Score.} Compute a utility score that incorporates expected yield, fees, slippage, and safety constraints (e.g., approval scope, contract reputation signals).
\item \textbf{Present options.} Provide the top candidates as distinct intents with side-by-side previews and explicit trade-offs for the user to approve or reject.
\end{enumerate}

\textbf{Security model and risks.}
Pattern~II reduces direct key-risk by keeping signing with the user, but it introduces a subtler class of failures: \textbf{malicious or unsafe intent generation}. An attacker can attempt to steer the agent toward harmful intents via (i) adversarial prompts, (ii) poisoned retrieved content, or (iii) misleading metadata (token name collisions, phishing contracts) that exploits the agent’s tendency to treat plausible-looking identifiers as trustworthy. Indirect prompt injection is particularly relevant: if untrusted content is not cleanly separated from policy and system instructions, the agent can be coerced into generating intents that violate user expectations while still appearing coherent~\cite{greshake2023indirectpromptinjection}.

The defenses in this pattern are therefore dominated by \emph{semantic verification} rather than cryptographic isolation:
\begin{enumerate}
\item \textbf{High-fidelity simulation with user-centered previews.} The preview must surface dangerous state changes (especially approvals, permit-like authorizations, and unexpected external calls) in a way that is hard to ignore. In DeFi threat analyses, preview and simulation are repeatedly emphasized as essential controls because they expose the practical effects of complex call graphs that are otherwise opaque to end users~\cite{heimbach2022sok,homoliak2024sok}.
\item \textbf{User comprehension as a control (and its limits).} The user is the last gate: they must interpret the preview, validate destination contracts and assets, and reject suspicious proposals. This introduces cognitive burden and an error-prone human factor, especially when previews are long, technical, or time-pressured by volatile markets. Pattern~II is therefore only as strong as the preview’s clarity and the user’s ability to evaluate it.
\end{enumerate}

\textbf{Evolutionary pressure.}
The requirement for per-transaction human approval becomes a throughput bottleneck. It limits strategy complexity and makes repetitive, low-risk actions tedious, pushing designers toward \emph{bounded autonomy}: the ability for an agent to execute pre-approved classes of actions without constant human intervention. This demand for secure, limited automation motivates the transition to Pattern~III, where authority is delegated under explicit scopes, caps, and policy constraints rather than being re-approved transaction by transaction.

\subsection{Pattern III: Delegated Execution (Constrained Roles)}

\textbf{Description.}
Pattern~III is the first point in the taxonomy where the agent crosses from proposal into \emph{execution}, but under a strict least-privilege posture. Instead of handing the agent a master key, the user delegates a \emph{bounded, revocable authority} that is narrow in scope (what can be done), magnitude (how much can be spent), and time (how long the authority remains valid). In practice, this delegation is realized through smart-account designs in which authorization is programmable and can be made conditional on runtime checks. On Ethereum and EVM-compatible chains, this pattern is strongly enabled by account-abstraction architectures, particularly ERC-4337, which standardize a smart-account execution path without requiring a protocol hard fork~\cite{buterin2021erc,wang2023account,singh2023account}.

\textbf{Architectural deep dive.}
Pattern~III introduces two architectural ingredients that are uncommon in Patterns~I--II: (i) a \emph{delegated signing capability} (often a session key) and (ii) a \emph{policy enforcement point} that deterministically gates execution.

\begin{itemize}
\item \textbf{Interface surface (smart accounts and modules).}
The agent interacts with a user’s \emph{smart account} rather than a human approver. The smart account exposes a validation hook that can be extended via modules (guards, validators, plug-ins) responsible for enforcing constraints before execution. Under ERC-4337, the agent typically submits a \texttt{UserOperation} targeting a smart account; the account’s validation logic decides whether the operation is authorized~\cite{buterin2021erc,singh2023account}.

\item \textbf{Delegation primitive (session keys / roles).}
Delegation is usually represented as a session key or role credential recognized by the smart account. Session keys are attractive because they allow narrow capability grants that can be revoked without migrating funds to a new address. In many implementations, the session key is valid only under conditions checked by a module (time bounds, spend caps, destination allowlists), turning key possession into \emph{conditional} authority rather than blanket control.

\item \textbf{Policy module (deterministic gating).}
The policy module acts as the guardrail. Typical constraints include:
\begin{itemize}
 \item \textbf{Function allowlists:} Restricting callable selectors (e.g., only swap or harvest functions).
 \item \textbf{Contract allowlists:} Restricting destinations (e.g., specific routers, vaults, or lending pools).
 \item \textbf{Spend limits:} Per-transaction and/or rate-limited ceilings (e.g., $\le$100~USDC/day).
 \item \textbf{Time bounds:} Expiration windows for the delegated credential.
 \item \textbf{Frequency limits:} Action throttles (e.g., harvest no more than once per 24 hours).
\end{itemize}
These controls are most effective when the module evaluates \emph{semantics} (asset movements, approvals granted, net-value change), not only superficial properties such as destination address.
\end{itemize}

\textbf{Execution flow (ERC-4337 realization).}
A representative flow under ERC-4337 is:

\begin{enumerate}
\item The agent constructs a \texttt{UserOperation} that encodes the intended call(s) from the smart account.
\item The agent signs the \texttt{UserOperation} using its delegated credential (session key or role key).
\item A bundler includes the \texttt{UserOperation} in a bundle transaction submitted to the global \texttt{EntryPoint} contract~\cite{buterin2021erc,singh2023account}.
\item \texttt{EntryPoint} invokes the account’s validation logic (commonly \texttt{validateUserOp}), which delegates to installed policy modules.
\item If policy checks pass, the operation is executed; otherwise it is rejected deterministically at validation time.
\end{enumerate}

\textbf{Security model and risks.}
The central security benefit of Pattern~III is \textbf{bounded loss under compromise}. If the agent is compromised, the damage is limited by the delegated scope and caps enforced by on-chain code. This is a qualitatively stronger guarantee than “the agent should behave,” because enforcement does not depend on the agent’s runtime integrity once the constraints are embedded in the account logic.

The dominant risk shifts to the \textbf{correctness and completeness of the policy definition} and its implementation:
\begin{itemize}
\item \textbf{Policy gaps and unintended capability.} A permissive allowlist (destination only, but not function-level) can enable unexpected methods on an otherwise trusted contract, including approval or upgrade paths that were not anticipated.
\item \textbf{Module bugs and composition hazards.} Guards and validators are smart contracts. Bugs, unsafe assumptions about calldata structure, or fragile composition across modules can create bypasses. Formal and systematic analyses repeatedly show that smart-contract security is as much about specification and behavioral modeling as it is about code patterns~\cite{abdellatif2018formal,almakhour2023formal}.
\item \textbf{Operational attack surface.} Even with on-chain enforcement, agents remain exposed to transaction replacement, relay manipulation, and adversarial execution dynamics (including MEV effects) at submission time; therefore, policy gating should be paired with simulation/preview where possible and with conservative execution parameters.
\end{itemize}

\textbf{Evolutionary pressure.}
Delegated execution is well-suited for recurring, structured automation (for example, periodic swaps, recurring payments, and routine strategy maintenance), and it is naturally aligned with account-abstraction-based automation designs~\cite{beams2022autopayments,wang2023account}. The main limitation is rigidity: on-chain policies are intentionally deterministic and can be costly or slow to evolve as strategies, venues, and market conditions change. This creates pressure toward Pattern~IV, where systems seek higher autonomy and faster decision-making by relying more heavily on hardened custody infrastructure and off-chain controls while still maintaining strong safety guarantees.

\subsection{Pattern IV: Autonomous Signing and Execution}

\textbf{Description.}
Pattern~IV is the most powerful, and therefore highest-risk, integration pattern for a \emph{single} agent. The agent is entrusted with the capability to construct, sign, and submit transactions without per-action, real-time human approval. This enables high-frequency and time-sensitive strategies (e.g., treasury management, automated market-making, or portfolio rebalancing) but collapses the traditional last line of defense that exists in Patterns~II--III. As a result, Pattern~IV demands \textbf{institutional-grade key custody} and \textbf{defense-in-depth policy enforcement}, because any failure can translate into irreversible, monetizable loss at machine speed. Wallet security surveys consistently emphasize that transaction authorization is the dominant risk concentration in real-world systems, and that “where the signing happens” is often more important than “what the UI shows”~\cite{homoliak2024sok}.

\textbf{Architectural deep dive.}
A robust Pattern~IV architecture separates the agent’s reasoning process from the act of signing. The agent should \emph{never} hold a long-lived private key directly in its runtime memory or filesystem. Instead, it interacts with hardened signing infrastructure through authenticated APIs and explicit policy gates~\cite{tct2022best,homoliak2024sok}. Two common custody architectures dominate in practice:

\begin{enumerate}
\item \textbf{MPC/TSS signing services (distributed custody).}
The agent is one party in a threshold signature scheme (TSS), holding a key share while other shares are held by independent devices or security services. Transaction signing is executed via an interactive protocol that produces a standard signature without reconstructing the private key in any single place~\cite{wiener2019mpc,lindell2017fast,komlo2024threshold}. This reduces the blast radius of an agent compromise: stealing the agent’s share alone is insufficient to forge signatures. In operational deployments, additional protections are often layered, such as HSM-backed key-share storage and geographically separated parties to mitigate correlated compromise~\cite{shbair2021hsm}.

\item \textbf{TEE-backed signers (hardware-isolated custody).}
Here, the signing key resides inside a Trusted Execution Environment (TEE) that is intended to resist extraction even if the surrounding host OS is compromised. The agent submits a candidate transaction or intent to the enclave, and the enclave signs only if a local policy check passes. TEEs are used in several wallet and meta-transaction designs as a means of enforcing isolation and attestation-based trust~\cite{onica2022using,kim2025secure}.
\end{enumerate}

\textbf{Policy engine (the missing control plane).}
In Pattern~IV, policy enforcement must typically extend beyond simple allowlists and caps. The reason is not only functional complexity but \emph{adaptivity}: high-stakes autonomous execution must respond to market conditions, protocol state, and evolving threats. This motivates a \textbf{multi-layer policy engine} that combines:
\begin{itemize}
\item \textbf{Static constraints:} address and function allowlists, spend limits, time windows, rate limits, and explicit disallow rules (e.g., no unlimited approvals; no bridge calls).
\item \textbf{Simulation-based guards:} fork simulation and semantic diffing that surfaces approvals, external calls, and net-asset deltas prior to signing (consistent with DeFi security analyses that emphasize call-graph complexity and the practical importance of previewing side effects)~\cite{heimbach2022sok,homoliak2024sok}.
\item \textbf{Behavioral monitoring:} anomaly detection over the agent’s historical action distribution (e.g., destination novelty, amount outliers, unusual approval patterns, abrupt frequency increases). Monitoring and detection systems are widely studied for on-chain attack identification and are a natural complement to policy gating in autonomous execution~\cite{augusto2024xchainwatcher,qian2023empirical}.
\end{itemize}

This is the motivation for separating \emph{intent specification} from \emph{authorization evidence}. In our roadmap terms, the agent produces a Transaction Intent (TIS), an evaluator grades it under a richer rule set and produces a Policy Decision Record (PDR), and the signing service (MPC or TEE) verifies the PDR before releasing a signature. Even when the PDR is off-chain, it creates an auditable chain of custody for \emph{why} a transaction was authorized, which materially improves post-incident forensics and accountability.

\textbf{Security model and risks.}
Pattern~IV should be designed under a \textbf{compromise-expected} assumption: any single component, including the agent, may fail. Security is therefore achieved by independent controls whose simultaneous compromise is difficult:
\begin{enumerate}
\item \textbf{Correlated failure:} the most dangerous class, where the agent, policy engine, and signing infrastructure are compromised together (or share a common root of trust, credentials, or deployment environment).
\item \textbf{Policy bugs and semantic gaps:} complex policies can be wrong, incomplete, or bypassed through edge-case calldata, approval mechanics, or contract upgrades. This is a specification problem as much as an implementation problem.
\item \textbf{Agent misgeneralization and goal drift:} even without an external attacker, an autonomous agent can take actions that are technically valid but financially harmful due to incorrect assumptions, regime shifts, or misunderstood constraints. Pattern~IV therefore requires tight objective specification, conservative risk limits, and robust verification loops.
\end{enumerate}

\textbf{Operational requirements (non-negotiable).}
Because failures are inevitable at this authority level, production Pattern~IV deployments require: (i) continuous monitoring with alerting; (ii) rapid revocation and key rotation procedures; and (iii) an immediately accessible \textbf{kill switch} that can halt signing or freeze delegated authority without waiting for on-chain governance or human coordination across time zones~\cite{tct2022best,homoliak2024sok}.

\textbf{Evolutionary pressure.}
Even with hardened custody and strong policies, a single autonomous agent remains an uncomfortable concentration of \emph{judgment} and \emph{operational authority}. This creates pressure to decentralize decision-making itself: split roles across multiple agents (proposer, risk checker, executor), require quorum approval, and introduce collective governance controls that are resilient to a single agent’s compromise or misjudgment. This motivation drives the transition to Pattern~V: multi-agent workflows with distributed authority and explicit coordination and oversight.

\subsection{Pattern V: Multi-Agent Workflows}

\textbf{Description.}
Pattern~V mitigates the extreme concentration of authority present in single-agent autonomous systems (Pattern~IV) by \emph{distributing trust, decision-making, and execution authority across multiple independent agents}. Rather than assuming that any single agent can be made sufficiently secure or aligned, this pattern adopts a structural approach: no irreversible on-chain action can occur unless it has passed through a quorum-based, multi-role decision process. Conceptually, this pattern draws from both organizational governance (e.g., separation of executive, risk, and audit functions in financial institutions) and classical distributed systems principles such as Byzantine fault tolerance and quorum consensus~\cite{morrison2020dao,zhang2023reaching}.

The dominant instantiation of this pattern is a \textbf{Proposer--Verifier--Executor} workflow, in which strategic intent, risk evaluation, and execution are explicitly decoupled and assigned to distinct agent roles.

\textbf{Architectural deep dive.}
\begin{itemize}
\item \textbf{Coordination interface.}
At the center of the architecture is a \emph{coordination and governance contract} that acts as a shared state machine and adjudication layer. This contract records proposals, collects approvals or rejections, enforces quorum rules, and authorizes execution. Agents do not rely on implicit trust or off-chain messaging alone; all critical decisions are surfaced on-chain in a transparent and auditable form.

\item \textbf{Role specialization.}
Authority is decomposed into distinct agent roles, each with sharply constrained responsibilities:
\end{itemize}

\begin{enumerate}
\item \textbf{Strategy agents (Proposers).}
These agents monitor markets, protocols, and external signals, and generate candidate actions. Their output is not a raw transaction but a structured \emph{Transaction Intent} describing objectives, constraints, and expected outcomes. In DAO or treasury settings, these agents play the role of opportunity discovery rather than execution.

\item \textbf{Risk agents (Verifiers).}
Verifier agents independently assess proposed intents. Their evaluation may include fork-based simulation, economic stress testing, dependency checks, and validation against explicit risk policies. Verifiers cast explicit on-chain votes approving or rejecting proposals. Importantly, they are designed to be conservative and adversarial toward proposers, functioning as a systemic “immune response” rather than collaborators.

\item \textbf{Execution agent (Executor).}
The executor is a deliberately simple and tightly constrained agent whose sole responsibility is to carry out proposals that have passed quorum. It has no discretion and no strategic logic. In practice, this role may be implemented via hardened automation services or decentralized execution networks, with authority gated strictly by governance state.
\end{enumerate}

\textbf{Security model and risks.}
The security model of Pattern~V is based on \textbf{explicit, on-chain distributed trust}. No single agent can unilaterally act, and the compromise or misbehavior of one role does not immediately translate into loss. This significantly raises the bar for successful attacks but introduces a new dominant failure mode: \textbf{collusion across agents}. 

The principal risks and mitigations include:
\begin{itemize}
\item \textbf{Collusion risk:} A quorum of proposers and verifiers could conspire to approve a malicious action. This is mitigated through agent diversity, independence of operators, and explicit quorum thresholds.
\item \textbf{Economic alignment:} Requiring agents to stake collateral that can be slashed in the event of provable misconduct creates direct economic disincentives against malicious coordination.
\item \textbf{Reputation and accountability:} Maintaining public, on-chain records of proposal histories, votes, and outcomes enables long-term reputation tracking and external auditing.
\item \textbf{Transparency:} All proposals, evaluations, and executions are visible, allowing human oversight and post-hoc forensic analysis by the broader community.
\end{itemize}

While this pattern significantly reduces single-point-of-failure risk, it increases system complexity and coordination overhead. As a result, it is best suited for high-value, high-impact systems where safety and legitimacy outweigh latency and simplicity.

\textbf{Evolutionary pressure and limits.}
Pattern~V represents the current practical frontier of agent--blockchain integration. It reconciles the desire for deep automation with the realities of adversarial environments, irreversibility, and economic risk. Rather than seeking perfect alignment or absolute trust in a single agent, it embeds skepticism directly into the architecture. 

Future systems are likely to refine this pattern rather than replace it: improving incentive design, reducing coordination latency, and developing stronger formal guarantees for agent independence and decision quality. For the foreseeable future, however, the safest form of autonomous on-chain behavior is not a monologue by a single powerful agent, but a structured, adversarial dialogue among many specialized ones.

\paragraph{Transition to threats.}
The five patterns above do more than classify interface designs; they progressively relocate the dominant trust boundary.
Pattern~I concentrates risk in data integrity, Pattern~II in preview fidelity and user interpretation, Pattern~III in the
correctness of delegated permissions, Pattern~IV in the security of signing and policy infrastructure, and Pattern~V in
coordination and quorum governance. This progression motivates a threat model that is pipeline-aware and interface-aware:
we need to reason about attacks that target cognition (via prompt and context manipulation), middleware (via tool and SDK
compromise), and execution economics (via MEV and orderflow exploitation), not just conventional software defects.
Accordingly, the next section formalizes the protected assets, adversary classes, and trust boundaries that recur across
all patterns, and uses them as the basis for the attack taxonomy and mitigations that follow.

\subsubsection{Pattern Selection Decision Framework}

Selecting the appropriate integration pattern requires systematic evaluation of multiple factors. This section provides a decision framework to guide pattern selection.

\paragraph{Risk Assessment}

The first consideration is the risk profile of the intended application. Higher-risk applications warrant more conservative patterns with stronger human oversight. Risk factors include the value at stake, the reversibility of actions, the predictability of outcomes, and the consequences of errors.

Applications managing significant value should generally start with Pattern II (Simulation and Intent Generation) or Pattern III (Delegated Execution) rather than fully autonomous patterns. The threshold for "significant value" depends on the operator's risk tolerance and the maturity of the agent system.

\paragraph{Operational Requirements}

The second consideration is operational requirements including latency, throughput, and availability. Applications requiring rapid response may not be compatible with patterns requiring human approval for each transaction. Applications requiring 24/7 operation may need autonomous patterns that can operate without human availability.

The tradeoff between operational requirements and risk management often determines the appropriate pattern. High-frequency trading applications may require Pattern IV (Autonomous Signing) despite higher risks, while lower-frequency applications can afford the latency of human-in-the-loop patterns.

\paragraph{Regulatory Constraints}

The third consideration is regulatory constraints that may mandate specific controls or oversight mechanisms. Some jurisdictions may require human approval for certain transaction types or above certain value thresholds. Compliance requirements should be identified early and factored into pattern selection.

\paragraph{Organizational Capabilities}

The fourth consideration is organizational capabilities including technical expertise, operational maturity, and risk management infrastructure. More autonomous patterns require more sophisticated monitoring, incident response, and governance capabilities. Organizations should honestly assess their capabilities and select patterns they can operate safely.

\paragraph{Evolution Path}

Finally, pattern selection should consider the intended evolution path. Starting with conservative patterns and gradually expanding autonomy as experience accumulates is generally safer than starting with aggressive patterns. The selected pattern should allow for controlled expansion of capabilities over time.

\subsubsection{Comparative Analysis of Integration Patterns}

The five integration patterns identified in this taxonomy represent points along a spectrum of agent autonomy and risk. Understanding the tradeoffs between these patterns is essential for selecting appropriate architectures for specific use cases.

Pattern I (Read-Only) offers maximum safety at the cost of minimal utility. Agents in this pattern can observe and analyze blockchain state but cannot effect any changes. This pattern is appropriate for analytics, monitoring, and advisory applications where the value lies in information processing rather than transaction execution. The security properties are straightforward: since agents cannot sign transactions, they cannot cause direct financial harm through their blockchain interactions.

Pattern II (Simulation and Intent Generation) extends agent capabilities to include transaction construction while maintaining human control over execution. Agents can analyze situations, identify opportunities, and prepare transactions, but humans retain final approval authority. This pattern is well-suited for applications where agent judgment adds value but the stakes are high enough to warrant human oversight. The security model relies on human reviewers to catch agent errors or malicious recommendations before execution.

Pattern III (Delegated Execution) grants agents limited execution authority within predefined constraints. This pattern enables automation of routine operations while maintaining guardrails against catastrophic errors. The security model shifts from human review of individual transactions to human specification and monitoring of policies. This pattern is appropriate for applications with predictable, bounded operations where the overhead of human approval for each transaction would be prohibitive.

Pattern IV (Autonomous Signing) grants agents broad execution authority with minimal transaction-level constraints. This pattern enables sophisticated strategies that require rapid, coordinated execution across multiple transactions. The security model relies heavily on agent design, testing, and monitoring rather than transaction-level controls. This pattern is appropriate only for applications where the benefits of full autonomy outweigh the increased risks.

Pattern V (Multi-Agent Coordination) extends autonomous operation to scenarios involving multiple interacting agents. This pattern enables complex workflows that require coordination, negotiation, or collective decision-making among agents. The security model must account for inter-agent dynamics, including the potential for collusion, competition, and emergent behaviors that arise from agent interactions.

\section{Threat Model for Agent--Blockchain Systems} \label{sec:threat_model}

The convergence of agentic AI and blockchains yields a threat surface that differs materially from conventional Web2 service integration. In agent--blockchain systems, probabilistic reasoning and tool orchestration are coupled to deterministic and irreversible state transitions, frequently under adversarial economic conditions. Standard checklists such as STRIDE remain useful for breadth, but they under-specify the failure modes that dominate real deployments: attacks that steer the agent’s decision process, corrupt tool-mediated perception, or exploit transaction visibility, ordering, and execution dynamics in public networks.

Recent analyses of tool-using LLM applications show that untrusted retrieved content can steer downstream tool use through retrieval-mediated instruction hijacking, even when the developer prompt is benign~\cite{greshake2023indirectpromptinjection}. Accordingly, we model the agent not only as software, but also as a decision-making component whose inputs, tools, and execution channel are all targetable.

\subsection{Assets, Adversaries, and Trust Boundaries}

A robust threat model begins with a precise statement of what must be protected (assets), who can attack (adversaries), and where trust assumptions change (trust boundaries). \cite{rahman2024fundamentals} \cite{bartoletti2017empirical}

\paragraph{Assets.}
Agent-to-chain systems concentrate multiple high-value assets: (i) \textbf{funds and signing authority} (private keys, MPC shares, session keys, smart-account validation logic); (ii) \textbf{permission and policy state} (allowlists, spend limits, role assignments, paymaster rules, relayer scopes); (iii) \textbf{agent state} (memory, tool registry, retrieved documents, execution plans, cached ABIs); and (iv) \textbf{telemetry} (transaction previews, simulation traces, policy decisions, audit logs) used for detection and forensics. In many deployments, the runtime also holds \textbf{off-chain secrets} (API keys, credentials, notification channels) that can enable lateral movement or social engineering if exposed~\cite{rizzini2025private}.

\paragraph{Adversaries.}
We consider a spectrum of adversaries with distinct capabilities and incentives:
\begin{itemize}
\item \textbf{Economic adversaries (orderflow and MEV actors):} actors that monitor transaction flow and extract value via reordering, copying, back-running, or sandwiching, as characterized in foundational work on miner/maximal extractable value and its effect on decentralized exchange execution~\cite{daian2019flash}.
\item \textbf{Remote attackers:} actors that exploit vulnerabilities in the agent runtime, plugins, SDKs, browser extensions, wallet connectors, or CI/CD pipelines to obtain signing capability or to steer decisions.
\item \textbf{Tool and data-source manipulators:} actors that poison or control off-chain inputs (web content, social channels, price feeds, governance information) or compromise RPC, indexer, and oracle infrastructure, thereby misleading the Observe and Reason stages.
\item \textbf{Malicious or compromised protocols:} adversarial contracts, phishing frontends, or upgradeable systems that trigger unintended state transitions even when a transaction appears benign at the intent level.
\item \textbf{Insiders and operators:} privileged parties (custody providers, RPC operators, bundler/solver operators, DevOps) who may be honest-but-curious, negligent, or malicious, and whose access can bypass assumed controls.
\end{itemize}

\paragraph{Trust boundaries.}
In agent--blockchain architectures, incidents cluster at boundaries where data or authority crosses between components. The most common boundaries are: (i) \textbf{user interfaces versus the agent} (what the user believes they approved versus what the agent proposes); (ii) \textbf{agent runtime versus the model provider} (prompts, context, and tool outputs leaving the trusted compute base); (iii) \textbf{agent versus tools and endpoints} (RPCs, indexers, web APIs, and dApp frontends); (iv) \textbf{agent versus signer/custody} (hardware wallet, MPC/TEE service, smart-account validation modules); and (v) \textbf{submission path to the chain} (public mempool, private relay, bundler/solver network). Boundary (iii) is especially fragile because retrieval-mediated instruction hijacking can be embedded inside content the agent is expected to fetch and summarize, then propagate into tool calls and transaction construction~\cite{greshake2023indirectpromptinjection}. Boundary (v) is where MEV extraction concentrates, since public dissemination of profitable intents invites adversarial reordering and value capture~\cite{daian2019flash}.

\paragraph{Assumptions and scope.}
Our analysis assumes standard cryptographic primitives remain unbroken and that the base chain’s consensus provides eventual finality. We do not assume that applications, frontends, or off-chain services are trustworthy, and we treat rapidly evolving smart-contract systems as part of the adversarial environment that agents must safely navigate~\cite{gro2024towards}. For distributed decision-making (Pattern~V), we include collusion, bribery, censorship, and governance capture as first-order threats because they can defeat quorum-based controls even when each component is correct in isolation.

\subsection{Taxonomy of Attack Classes} \label{sec:taxnonomy_patterns}

Agent--blockchain systems combine (i) a probabilistic decision-maker that consumes untrusted language and tool outputs with (ii) a deterministic execution substrate in which errors are irreversible and economically exploitable. \cite{avizienis2004dependability} This asymmetric coupling fundamentally reshapes the threat surface: attacks no longer target only software components or network infrastructure, but also the agent’s reasoning process, policy interpretation, and execution pathways.

\begin{figure}[!ht]
 \centering
 \resizebox{.8\linewidth}{!}{%
 \begin{tikzpicture}[
 font=\scriptsize,
 node distance=7mm and 10mm,
 >=Latex,
 comp/.style={draw, rounded corners=2mm, align=center, inner sep=4pt, minimum height=10mm, fill=white},
 core/.style={comp, line width=0.7pt, text width=3.25cm},
 aux/.style={comp, line width=0.6pt, text width=3.05cm, fill=black!3},
 path/.style={->, line width=0.7pt},
 fb/.style={->, line width=0.7pt, dashed},
 attack/.style={->, line width=0.7pt, dashed},
 boundary/.style={draw, rounded corners=2mm, line width=0.6pt, inner sep=4pt}
 ]

 \node[core] (user) {\textbf{User / Operator}\\[-0.3mm]\scriptsize prompts, approvals};
 \node[core, right=12mm of user] (agent) {\textbf{Agent Runtime}\\[-0.3mm]\scriptsize model, memory, tool router};
 \node[core, right=12mm of agent] (policy) {\textbf{Policy Engine}\\[-0.3mm]\scriptsize rules, simulation, checks};
 \node[core, right=12mm of policy] (signer) {\textbf{Signer / Custody}\\[-0.3mm]\scriptsize EOA/AA, MPC/TEE};

 \node[core, below=10mm of signer] (submit) {\textbf{Submission Path}\\[-0.3mm]\scriptsize mempool/relay\\[-0.6mm]\scriptsize bundler/solver};
 \node[core, left=10mm of submit] (chain) {\textbf{Blockchain}\\[-0.3mm]\scriptsize state, finality};

 \node[aux, above=10mm of agent] (model) {\textbf{Model Provider}\\[-0.3mm]\scriptsize API / weights};
 \node[aux, below=10mm of agent] (tools) {\textbf{Tools \& Data Plane}\\[-0.3mm]\scriptsize RPC, indexers,\\[-0.6mm]\scriptsize web, simulators};

 \draw[path] (user) -- node[above]{\scriptsize request} (agent);
 \draw[path] (agent) -- node[above]{\scriptsize intent/} node[below]{\scriptsize proposal} (policy);
 \draw[path] (policy) -- node[above]{\scriptsize allow/deny} node[below]{\scriptsize + tx} (signer);
 \draw[path] (signer) -- node[right]{\scriptsize signed tx} (submit);
 \draw[path] (submit) -- node[below]{\scriptsize inclusion} (chain);

 \draw[fb, bend left=40] (chain.west) to node[right]{\scriptsize events} (agent.south east);
 \draw[path] (agent) -- node[right]{\scriptsize prompts} (model);
 \draw[path] (model) -- node[left]{\scriptsize completions} (agent);
 \draw[path] (agent) -- node[right]{\scriptsize reads/sim} (tools);
 \draw[path] (tools) -- node[left]{\scriptsize outputs} (agent);

 \node[boundary, fit=(agent) (policy) (signer),
 label={[font=\scriptsize]below:Control plane (where authority is constrained)}] (cp) {};
 \node[boundary, fit=(model) (agent) (tools),
 label={[font=\scriptsize]above:External / untrusted dependencies}] (ext) {};

 \node[aux, above=10mm of user] (c1) {\textbf{C1: Prompt injection}\\[-0.3mm]\scriptsize direct/indirect};
 \draw[attack] (c1.south east) -- (agent.north west);

 \node[aux, below=10mm of tools] (c2) {\textbf{C2: Data/tool poisoning}\\[-0.3mm]\scriptsize RPC/indexer/web/sim};
 \draw[attack] (c2.north) -- (tools.south);

 \node[aux, above=10mm of policy] (c3) {\textbf{C3: Policy bypass}\\[-0.3mm]\scriptsize misconfig/bug/route};
 \draw[attack] (c3.south) -- (policy.north);

 \node[aux, above=10mm of signer] (c4) {\textbf{C4: Signing compromise}\\[-0.3mm]\scriptsize key/session/MPC/TEE};
 \draw[attack] (c4.south) -- (signer.north);

 \node[aux, below=10mm of submit] (c5) {\textbf{C5: MEV / ordering}\\[-0.3mm]\scriptsize reorder/copy/sandwich};
 \draw[attack] (c5.north) -- (submit.south);

 \node[aux, below=10mm of chain] (c6) {\textbf{C6: Contract/dApp traps}\\[-0.3mm]\scriptsize approvals/upgrades/phish};
 \draw[attack] (c6.north) -- (chain.south);

 \end{tikzpicture}%
 }
 \caption{Threat model for agent--blockchain systems. Solid arrows show nominal data/control flow; dashed arrows indicate representative attack classes aligned to major trust boundaries.}
 \label{fig:threat_model}
\end{figure}
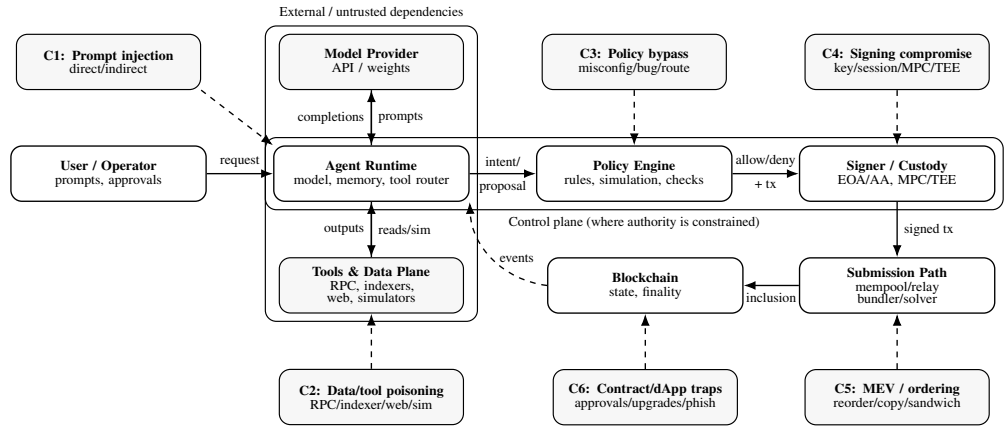

To make this space analyzable, we organize threats into a set of \emph{layered attack classes} that align with the major trust boundaries in Figure~\ref{fig:threat_model}. Table~\ref{tab:attack_classes} provides a compact overview of these classes, mapping each to the affected pipeline stage, the primary asset under attack, and representative defensive controls. This taxonomy is intentionally boundary-oriented rather than vulnerability-oriented: each class corresponds to a distinct point where trust shifts between components or actors.

Concretely, we distinguish: (C1) \emph{cognitive attacks} that manipulate agent reasoning and tool selection (prompt injection and instruction hijacking); (C2) \emph{tool and data-plane attacks} that poison observations and simulations (RPC or indexer compromise, oracle manipulation, malicious web content); (C3) \emph{policy and control-plane attacks} that bypass or subvert constraints (misconfigured allowlists, policy-module bugs, alternate signing routes); (C4) \emph{signing and custody attacks} that obtain authorization (key exfiltration, session-key overreach, MPC/TEE compromise); (C5) \emph{submission-path attacks} that exploit transaction visibility and ordering (mempool adversaries and MEV); (C6) \emph{application-layer attacks} where seemingly benign interactions trigger harmful on-chain state transitions (malicious contracts, upgrade traps, phishing frontends); and (C7) \emph{multi-agent and governance attacks} that exploit quorum formation and coordination (Sybil attacks and collusion).

\begin{table}[t]
 \centering
 \caption{Attack classes (C1--C7) mapped to pipeline stage, primary target, and representative controls.}
 \label{tab:attack_classes}
 \scriptsize
 \setlength{\tabcolsep}{4pt}
 \begin{tabular}{p{4.1cm}lp{4.1cm}p{4cm}}
 \toprule
 \textbf{Class} & \textbf{Pipeline stage(s)} & \textbf{Primary target} & \textbf{Representative controls} \\
 \midrule
 C1: Prompt injection / cognitive hijacking & Reason $\rightarrow$ Plan $\rightarrow$ Authorize & Agent policy adherence; tool-choice integrity & Context isolation; intent canonicalization; tool sandboxes \\
 C2: Tool/data spoofing & Observe $\rightarrow$ Reason & State integrity (prices, metadata, risk signals) & Multi-source quorum; provenance; sanity bounds \\
 C3: Middleware manipulation & Authorize $\rightarrow$ Execute & Intent/tx integrity between preview and sign & WYSIWYS; typed intents; dependency pinning \\
 C4: Key/credential exfiltration & Execute (and runtime) & Signing authority; secrets; tool credentials & Least privilege; MPC/TEE custody; secret hygiene \\
 C5: Replay/nonce manipulation & Execute $\leftrightarrow$ Verify & Transaction uniqueness; ordering guarantees & Nonce discipline; idempotent messages; replay protection \\
 C6: MEV/economic manipulation & Execute (submission path) & Execution price and ordering & Private orderflow; intent solvers; strict bounds \\
 C7: Collusion/governance attacks & Plan $\rightarrow$ Authorize & Decision quorum; verifier set integrity & Staking; attestations; diversity constraints \\
 \bottomrule
 \end{tabular}
\end{table}

We use this classification consistently throughout the remainder of the section, analyzing each class in turn and grounding mitigations at the boundary where the attack manifests.

\paragraph{Non-goals and excluded threats.}
To keep the threat taxonomy actionable and focused on the agent--to--chain boundary, we intentionally exclude categories that are either orthogonal to interoperability or require a distinct analytical framework. First, we do not model \emph{base-layer failures} such as cryptographic breaks, consensus-safety violations, or sustained network partitions at the L1 protocol layer; we assume standard primitives remain secure and that the chain provides eventual finality. Second, we do not attempt a comprehensive treatment of \emph{general-purpose LLM risks} (for example, open-domain hallucination, bias, or toxic generation) unless they materially affect tool selection, authorization decisions, or execution correctness in the agent pipeline. Third, we do not provide an exhaustive catalog of \emph{smart-contract vulnerability classes} in isolation (reentrancy, arithmetic errors, and similar defects); instead, we treat vulnerable or rapidly evolving protocols as part of the ambient adversarial environment and focus on how agents become exposed through automated intent construction, simulation, and signing. Finally, denial-of-service and availability attacks are considered only insofar as they change execution outcomes (for example, censorship, delayed inclusion, or forced fallback to riskier submission paths), rather than as a standalone network-security survey.

\subsubsection{Class 1: Prompt Injection and Cognitive Hijacking}

The vulnerability of machine learning models to adversarial examples has been a long-standing research topic, with early work demonstrating how small, imperceptible perturbations can cause misclassification~\cite{cao2013overview}. This foundational understanding of adversarial robustness is a precursor to the more complex prompt injection attacks seen in LLM-based agents today.

Prompt injection attacks target the agent’s \emph{decision substrate} rather than a conventional software flaw: the adversary places malicious instructions in any channel the agent ingests and attempts to redirect intent formation, tool selection, or authorization steps. Empirical studies and benchmark suites show that injection remains viable across many agent designs, particularly when untrusted content (retrieval results, tool outputs, webpages) is concatenated into the same context as trusted directives~\cite{liu2024formalizing,greshake2023indirectpromptinjection}.

\paragraph{Attack mechanics.}
Injection succeeds when the system fails to enforce a strict separation between (i) \emph{trusted directives} (system instructions, local policy, governance rules) and (ii) \emph{untrusted content} (user text, webpages, on-chain metadata, retrieved documents, tool outputs). The attacker exploits this ambiguity to induce the agent to: (a) call a high-risk tool (e.g., \texttt{approve}, \texttt{transfer}, \texttt{sign}); (b) construct a harmful transaction intent (e.g., to a look-alike token or attacker-controlled contract); or (c) seek elevated authorization by presenting a fabricated rationale (``the policy already approved'').

\begin{itemize}
\item \textbf{Direct injection (interactive).} The adversary is the user (or a compromised UI) and issues prompts crafted to bypass constraints or coerce unsafe tool use. In on-chain settings, the attack is frequently \emph{immediately monetizable}: the injected instruction aims to trigger approvals, asset transfers, or signing requests within a single interaction loop.

\item \textbf{Indirect injection (retrieval-mediated).} The adversary hides instructions in content that the agent is expected to fetch and parse during normal operation (documentation pages, dashboards, forum posts, token metadata, and even transaction traces rendered as text). Indirect injection is particularly dangerous because it compromises \emph{benign tasks} (summarize, compare, assess risk) and then pivots into execution by steering subsequent tool calls or intent construction~\cite{greshake2023indirectpromptinjection}. For example, a malicious webpage can embed hidden directives (e.g., CSS/HTML tricks) that, once retrieved into the context window, attempt to coerce an approval or a transfer narrative.
\end{itemize}

\paragraph{Mitigations that map to the failure mode.}
Defenses are strongest when they enforce \emph{structural separation} and \emph{deterministic constraints}, rather than relying on the model to ``behave'':

\begin{itemize}
\item \textbf{Context compartmentalization (data is not instructions).} Keep untrusted content in explicitly delimited fields (e.g., \texttt{<data>} blocks or separate message roles) and apply strict parsing rules in the runtime. The agent should be instructed that these fields are \emph{data only}, while the runtime enforces that data fields cannot introduce new tool permissions or override policy. This targets the root cause: instruction ambiguity in a shared context window~\cite{liu2024formalizing}.

\item \textbf{Intent canonicalization before any authorization.} Require the agent to emit a structured intent object (e.g., TIS) containing only declarative fields (assets, amounts, destination, constraints, deadlines). Downstream components (policy engine, simulator, signer) should accept only this structured representation rather than free-form text. This reduces the degrees of freedom available to injected instructions, improves auditability, and makes ``what was intended'' separable from ``how it is executed.''

\item \textbf{Tool-use gating and least-privilege routing.} Even if injection succeeds at the reasoning layer, the runtime should block high-risk tool calls unless an independent policy check passes (or explicit user confirmation is obtained). In practice, enforce hard permission boundaries around actions such as \texttt{approve}, \texttt{setAllowance}, \texttt{transfer}, \texttt{sign}, and \texttt{broadcast}, and require that the call arguments be derived from the canonical intent rather than the raw conversation text.
\end{itemize}

\subsubsection{Class 2: Tool and Data Source Spoofing}

This class targets the \textbf{Observe} stage by compromising the integrity of the measurements an agent relies on. In agent--blockchain settings, spoofed inputs are particularly dangerous because they can produce actions that are locally rational (given the data) yet globally catastrophic (given reality). The core failure mode is \emph{treating an untrusted data plane as an oracle of truth}.

\paragraph{Attack mechanics.}
An adversary poisons or compromises a trusted input channel such as an RPC provider, an indexing service, an analytics dashboard, a governance feed, or a price oracle. Manipulation can be (i) \emph{source-level} (compromise, insider access, DNS hijack), (ii) \emph{protocol-level} (oracle manipulation via thin liquidity and transient price shocks), or (iii) \emph{presentation-level} (misleading metadata, spoofed ``audit'' claims, or phishing documentation).
The attacker objective is not necessarily to ``hack'' the agent, but to steer it into signing or recommending harmful actions based on false premises.

\paragraph{Where it bites in the pipeline.}
Spoofing primarily enters at \textbf{Observe} and becomes harmful after \textbf{Reason} and \textbf{Plan} propagate the corrupted signal into: (a) a distorted world model, (b) a mispriced trade, (c) an unsafe collateral decision, or (d) a misleading risk assessment.

\paragraph{Example.}
Oracle and price-manipulation attacks can induce agents to trade against distorted prices or to overestimate collateral value. A well-known pattern is manipulating an oracle-dependent protocol through transient price shocks, as seen in incidents such as Mango Markets~\cite{mohsin2025ai}. If an agent treats the compromised signal as ground truth, it can swap into near-worthless assets or open leveraged positions that are immediately liquidatable.

\paragraph{Mitigations that map to the failure mode.}
Defenses should reduce the probability that any single poisoned channel can dominate the agent's belief state, and should ensure that critical decisions are backed by verifiable provenance and bounded by domain constraints.

\begin{itemize}
\item \textbf{Multi-provider quorum with deviation rules.}
For critical signals (spot price, TWAP, volatility, reserve snapshots, risk scores), query multiple independent providers and aggregate using robust statistics (median, trimmed mean). Reject outliers beyond a configured deviation threshold and escalate when feeds disagree materially~\cite{breidenbach2021chainlink}. This directly targets the ``single-source dominance'' failure mode.

\item \textbf{Provenance and authenticity checks.}
Prefer sources that provide verifiable provenance (signed payloads, authenticated transport, pinning to expected public keys). Maintain an allowlist of endpoints and keys, and record provenance metadata in audit logs so post-incident forensics can trace which input influenced which decision.

\item \textbf{Sanity bounds and invariant cross-checks.}
Enforce domain constraints that are cheap but effective: price bounds relative to recent windows, token decimal consistency, reserve monotonicity checks, contract-code identity (bytecode hash), and known-safe address registries. For trades, require that simulation outcomes remain stable under small perturbations of inputs (a lightweight robustness test) before promoting a plan to authorization.

\item \textbf{Separation of ``measure'' and ``decide'' artifacts.}
Store raw observations (with provenance) separately from derived features used in planning. This makes it easier to detect and roll back poisoned features without erasing the audit trail.
\end{itemize}

\subsubsection{Class 3: Middleware Compromise and Intent Manipulation}

This class targets the seam between \textbf{Authorize} and \textbf{Execute}. Even if the agent proposes a correct action and the user approves a clear preview, a compromised middleware layer can alter what is actually signed or submitted. The failure mode is a broken binding between \emph{previewed intent} and \emph{signed bytes}.

\paragraph{Attack mechanics.}
An attacker compromises a wallet SDK, browser extension, plugin, relayer integration, or dependency (for example, via a package supply-chain incident). The compromised component mutates an intent or transaction payload \emph{after} preview but \emph{before} signing, injects additional calls into a batch, swaps destination addresses, or changes calldata while keeping surface-level UI text plausible.

\paragraph{Example.}
A user approves a swap intent. Middleware silently wraps it into a \texttt{multicall} that includes an additional approval or operator-granting action (for example, \texttt{setApprovalForAll}) enabling later asset drain. The signed payload is technically valid but semantically different from the preview.

\paragraph{Mitigations that map to the failure mode.}
Controls should ensure that any transformation after preview is either impossible, or detectable before authorization is granted.

\begin{itemize}
\item \textbf{What You See Is What You Sign (WYSIWYS).}
Derive the preview from the exact bytes that will be signed, and cryptographically bind preview to the signature request. Hardware wallet displays provide a strong instantiation because the rendering surface is isolated from the host machine~\cite{yu2023gptfuzzer}.

\item \textbf{Typed intents with canonical encoding.}
Use a structured intent format with canonical serialization, and ensure the signature covers the full canonical form: target addresses, selectors, token identifiers, amounts, deadlines, and bounds. Canonicalization reduces malleability and makes ``silent mutation'' harder to hide.

\item \textbf{End-to-end integrity checks across components.}
Have the policy engine and signer re-compute a digest of the transaction/intents and compare it against a digest produced at preview-time. If the digests differ, fail closed and require re-authorization.

\item \textbf{Dependency discipline on the signing-critical path.}
Pin versions, enable integrity checks (hash locks, reproducible builds where feasible), and continuously audit dependencies. Minimize privileged surfaces and disable dynamic code loading in the preview-to-sign pipeline.
\end{itemize}

\subsubsection{Class 4: Key and Credential Exfiltration}

This class targets the agent’s \textbf{signing authority} and associated secret material. In agent deployments, the blast radius includes both on-chain credentials (EOA keys, session keys, MPC shares) and off-chain secrets (API keys, webhook tokens, notification channels) that can be used for lateral movement. The failure mode is \emph{placing long-lived authority inside a compromise-prone runtime}.

\paragraph{Attack mechanics.}
\begin{itemize}
\item \textbf{Environmental compromise.}
Attackers gain access to the execution environment (server compromise, malicious container image, CI leakage) and extract secrets from files, environment variables, logs, or process memory.

\item \textbf{Induced disclosure via interaction channels.}
Through social engineering or poisoned tool outputs, attackers coerce the agent to print or forward secrets (for example, requests to dump environment variables for debugging, or ``paste your config so I can help'').
\end{itemize}

\paragraph{Mitigations that map to the failure mode.}
Controls should (i) reduce the value of any secret present in the runtime, (ii) isolate signing from the agent process, and (iii) treat outputs as potential exfiltration channels.

\begin{itemize}
\item \textbf{Least privilege by construction.}
Avoid persistent master keys in the agent runtime. Prefer time-bounded session keys, scoped capabilities, allowlists, and spend limits enforced by account-abstraction policies.

\item \textbf{Hardened custody primitives.}
Place signing behind MPC-TSS or a TEE-backed signer so that full key material is never present in agent memory. Threshold signing reduces single-host compromise impact~\cite{komlo2024threshold,lindell2017fast}, while HSM-backed designs can further isolate secrets in dedicated hardware~\cite{shbair2021hsm}.

\item \textbf{Secret hygiene and scanning.}
Treat logs and model outputs as exfiltration channels. Apply secret scanning to repositories and telemetry, redact tool outputs by default, and use short-lived credentials with rotation. Ensure the agent refuses requests that involve revealing configuration or secrets.

\item \textbf{Two-channel confirmation for authority elevation.}
If the agent must request higher authority (larger spend limit, new contract allowlist), require approval via a separate channel (hardware wallet confirmation, operator approval flow) to break single-channel compromise.
\end{itemize}

\subsubsection{Class 5: Replay and Nonce Manipulation}

This class exploits the ordering semantics of blockchain accounts. Many agent systems sign messages off-chain (intents) and submit transactions on-chain, and both paths require robust uniqueness and ordering guarantees. The failure mode is \emph{re-using valid authorization in a context where the system assumes single-use}.

\paragraph{Attack mechanics.}
\begin{itemize}
\item \textbf{Replay.}
A valid signed payload (transaction or off-chain authorization) is re-submitted to trigger repeated state transitions if the application is not replay-safe.

\item \textbf{Nonce interference and replacement dynamics.}
Adversaries attempt to delay, censor, or invalidate a transaction by exploiting nonce conflicts, transaction replacement, or mempool visibility. In multi-sender setups (agent + operator + automation), nonce collisions can also arise from benign concurrency, which attackers can amplify.
\end{itemize}

\paragraph{Mitigations that map to the failure mode.}
\begin{itemize}
\item \textbf{Replay protection in signed messages.}
All signed intents must include unique identifiers (nonce/salt), strict expiry, and domain separation. On-chain consumers must store and reject used nonces, making the message one-time usable.

\item \textbf{Nonce management as a first-class component.}
Centralize nonce allocation per account (or use smart-account schemes that virtualize nonces) to avoid gaps and collisions. Track pending transactions, detect drops, and handle replacements deterministically. For AA paths, leverage validation logic to enforce anti-replay constraints at the \texttt{validateUserOp} layer.

\item \textbf{Idempotent application design.}
Where feasible, design state transitions to be idempotent or safely repeatable (for example, ``set position to X'' rather than ``increase by X''), limiting damage if a replay occurs.
\end{itemize}

\subsubsection{Class 6: MEV and Economic Manipulation}

This class targets the \textbf{submission path} and the economic game of execution ordering. \cite{z2024study} It is not a software defect but an incentive-compatible extraction mechanism that systematically penalizes naive execution strategies in public mempools~\cite{daian2019flash,qin2023review}. The failure mode is \emph{revealing valuable intent to an adversarial auction}.

\paragraph{Attack mechanics.}
Mempool-visible actions are observed and adversarially reordered or surrounded. Sandwiching is the canonical extraction pattern for AMM trades, but agents are also vulnerable to copy-trading, back-running, and adversarial transaction replacement.

\paragraph{Mitigations that map to the failure mode.}
\begin{itemize}
\item \textbf{Private orderflow and relays.}
Submit via private channels when available to reduce mempool visibility and prevent straightforward sandwiching~\cite{piet2023mevade}.

\item \textbf{Intent-based execution and solver markets.}
Prefer auction or solver-based execution where the agent signs an intent and delegates execution to a competitive market that internalizes MEV protection, rather than broadcasting a fragile transaction directly.

\item \textbf{Hard bounds as enforceable constraints.}
Use strict slippage, deadlines, and price limits, and require simulation under adversarial assumptions before execution. Treat MEV risk as a first-class cost component in planning, not as an afterthought.

\item \textbf{MEV-aware planning signals.}
Incorporate mempool and liquidity conditions into planning (expected sandwich cost, depth sensitivity) so the agent can decide when not to trade.
\end{itemize}

\subsubsection{Class 7: Multi-Agent Collusion and Governance Attacks}

The principles of anonymous credentials can be extended to enable anonymous voting in DAOs, where agents can participate in governance without revealing their voting preferences~\cite{chin2025anonymous}. This can help mitigate collusion and coercion in decentralized governance.

This class is specific to \textbf{Pattern V} systems where safety depends on a quorum of agents, verifiers, or voters. \cite{sedlmeir2021digital} Distribution reduces single-agent failure modes, but introduces collective-action failures. The core failure mode is \emph{capturing the quorum} rather than compromising any single component.

\paragraph{Attack mechanics.}
A common vector is the Sybil strategy, where an attacker spins up seemingly independent agents to capture quorum weight and approve malicious actions~\cite{xi2025agentgym}. More generally, bribery, cartel formation, and correlated compromise can cause a verifier set to approve harmful proposals while appearing procedurally correct.

\paragraph{Mitigations that map to the failure mode.}
\begin{itemize}
\item \textbf{Economic deterrence (staking and slashing).}
Require economically meaningful stake for participation with transparent slashing conditions for provably malicious approvals.

\item \textbf{Identity and attestations for high-trust roles.}
Use verifiable credentials~\cite{garcarodrguez2021towards} or attestation gates for verifier roles, limiting who can join the quorum set and enabling audit of operator provenance.

\item \textbf{Diversity constraints to reduce correlated failure.}
Encode independence assumptions into governance design (operator diversity, jurisdictional diversity, implementation diversity), and rotate verifier sets to reduce long-lived capture risk.

\item \textbf{Accountability artifacts and auditability.}
Require verifiers to publish signed rationales, simulation transcripts, and evidence pointers. This increases the cost of coordinated deception and improves post-incident attribution.
\end{itemize}

\subsubsection{Analysis of Real-World Security Incidents}

Examining real-world security incidents provides valuable lessons for agent-blockchain system design. While comprehensive incident data for agent-specific systems is limited due to the field's nascency, related incidents in DeFi and AI systems offer relevant insights.

\paragraph{Smart Contract Vulnerabilities}

The history of smart contract exploits demonstrates the risks of on-chain code execution. High-profile incidents including the DAO hack, Parity wallet freeze, and numerous DeFi exploits have resulted in billions of dollars in losses. These incidents highlight the importance of thorough auditing, formal verification, and defense-in-depth strategies.

For agent systems, smart contract vulnerabilities in the contracts agents interact with represent a significant risk. Agents should be designed to recognize and avoid interaction with unaudited or suspicious contracts. Simulation and verification of transaction outcomes before execution can help identify potential exploit conditions.

\paragraph{Oracle Manipulation}

Oracle manipulation attacks have exploited the dependence of DeFi protocols on external price data. Attackers have manipulated oracle prices through flash loans, low-liquidity markets, and other techniques to extract value from protocols that relied on manipulated prices.

Agent systems that rely on oracle data for decision-making inherit these risks. Agents should use multiple independent oracle sources, implement sanity checks on oracle data, and be designed to fail safely when oracle data appears anomalous.

\paragraph{Governance Attacks}

Governance attacks have exploited the token-based voting mechanisms of DAOs and DeFi protocols. Attackers have acquired voting power through flash loans, accumulated tokens, or social engineering to pass malicious proposals.

Agent systems participating in governance must be designed to resist manipulation and avoid contributing to governance attacks. Agents should implement safeguards against voting on malicious proposals and should not automatically follow governance recommendations without appropriate validation.

\paragraph{AI System Failures}

While not blockchain-specific, failures of AI systems in other domains provide relevant lessons. Incidents including chatbot manipulation, recommendation system failures, and autonomous vehicle accidents demonstrate the risks of deploying AI systems in high-stakes environments without adequate safeguards.

Agent-blockchain systems should learn from these incidents by implementing robust testing, monitoring, and fail-safe mechanisms. The irreversibility of blockchain transactions amplifies the consequences of AI failures, making prevention and rapid detection especially important.

\subsubsection{Defense in Depth: A Layered Security Architecture}

The concept of trustworthy AI encompasses a range of desirable properties, including reliability, robustness, fairness, and transparency, all of which are essential for building user trust in agentic systems~\cite{albreiki2020trustworthy}.

Effective security for agent-blockchain systems requires defense in depth, with multiple independent layers of protection that provide redundant safeguards against diverse threats. No single security measure is sufficient; comprehensive protection requires layered defenses that address threats at multiple levels.

\paragraph{Layer 1: Input Validation and Sanitization}

The first line of defense validates and sanitizes all inputs before they reach the agent's reasoning components. This includes filtering potentially malicious content from blockchain data, validating user inputs against expected formats, and sanitizing external data sources. Input validation should be conservative, rejecting ambiguous or suspicious inputs rather than attempting to interpret them.

Effective input validation requires understanding the attack surface exposed by the agent's input channels. Blockchain data, user messages, API responses, and other input sources each present different risks and require tailored validation strategies.

\paragraph{Layer 2: Reasoning Constraints and Guardrails}

The second layer constrains the agent's reasoning process to prevent harmful outputs even if malicious inputs bypass input validation. This includes system prompts that establish behavioral boundaries, output filtering that blocks potentially harmful responses, and reasoning monitors that detect anomalous decision patterns.

Guardrails should be designed to fail safely, defaulting to conservative behavior when uncertain. The tradeoff between guardrail strictness and agent utility must be calibrated to the risk profile of the application.

\paragraph{Layer 3: Transaction Validation and Simulation}

The third layer validates proposed transactions before execution, ensuring they conform to policy constraints and will execute as expected. This includes policy engine checks against spending limits and allowlists, simulation against current blockchain state, and verification that transaction outcomes match agent intentions.

Transaction validation should be performed in isolated environments to prevent manipulation by malicious contracts or state. Multiple independent validation checks provide redundancy against individual check failures.

\paragraph{Layer 4: Execution Controls and Circuit Breakers}

The fourth layer controls transaction execution and provides emergency stop capabilities. This includes rate limiting to prevent rapid fund depletion, progressive authorization that requires additional approval for unusual transactions, and circuit breakers that halt execution when anomalies are detected.

Circuit breakers should be designed with appropriate sensitivity, triggering on genuine anomalies while avoiding false positives that disrupt normal operation. The criteria for circuit breaker activation should be regularly reviewed and updated based on operational experience.

\paragraph{Layer 5: Monitoring and Response}

The fifth layer provides continuous monitoring and incident response capabilities. This includes real-time monitoring of agent behavior and transaction outcomes, alerting systems that notify operators of potential issues, and incident response procedures for investigating and remediating security events.

Monitoring systems should capture sufficient detail to enable forensic analysis while respecting privacy constraints. Incident response procedures should be documented, tested, and regularly updated.

\subsubsection{Detailed Attack Scenarios and Case Studies}

Understanding the threat landscape requires examining specific attack scenarios that illustrate how theoretical vulnerabilities manifest in practice. The following case studies, drawn from documented incidents and security research, illuminate the practical risks facing agent-blockchain systems.

\paragraph{Scenario 1: Prompt Injection via Malicious Contract Metadata}

An attacker deploys a smart contract with carefully crafted metadata (name, symbol, or other string fields) containing prompt injection payloads. When an agent queries this contract as part of its normal operation, the malicious strings are incorporated into the agent's context. The injection payload instructs the agent to transfer funds to the attacker's address, framed as a legitimate operation. If the agent's prompt processing is not properly sanitized, it may execute the malicious instructions, resulting in unauthorized fund transfers.

This scenario illustrates the importance of treating all external data as potentially adversarial, including data retrieved from blockchain state. Mitigations include strict input sanitization, separation of data and instruction channels, and validation of agent outputs against expected patterns before execution.

\paragraph{Scenario 2: MEV Extraction from Agent Transactions}

An agent operating a DeFi strategy submits a large swap transaction to a public mempool. MEV searchers detect the pending transaction and construct a sandwich attack: a front-running transaction that moves the price unfavorably, followed by the agent's transaction, followed by a back-running transaction that captures the price difference. The agent's transaction executes at a worse price than expected, with the difference extracted by the MEV searcher.

This scenario illustrates the adversarial nature of public blockchain execution environments. Mitigations include the use of private mempools, MEV-aware transaction construction with appropriate slippage limits, and intent-based execution through solver networks that internalize MEV.

\paragraph{Scenario 3: Social Engineering of Human Approvers}

An attacker identifies that a target organization uses a human-in-the-loop approval process for agent transactions. The attacker manipulates the agent (through prompt injection or other means) to generate a series of legitimate-looking transactions that gradually condition human approvers to approve without careful review. Once approvers are sufficiently desensitized, the attacker triggers a malicious transaction designed to appear similar to the conditioning transactions but with a different destination address.

This scenario illustrates that human oversight is not a panacea and can itself be targeted by sophisticated attackers. Mitigations include varying approval workflows to prevent habituation, automated anomaly detection to flag unusual patterns, and separation of duties to require multiple independent approvers for high-value transactions.

\subsubsection{Formal Methods for Agent-Blockchain Security}

The field of AI alignment is concerned with ensuring that AI systems act in accordance with human values and intentions, which is a critical prerequisite for deploying autonomous agents in high-stakes environments~\cite{conitzer2024foundations}. The AI control problem, or the challenge of ensuring that advanced AI systems remain under human control, is a long-standing research topic that has gained new urgency with the advent of powerful LLMs~\cite{appel2023control}.

The application of formal methods to agent-blockchain systems represents a promising but challenging research direction~\cite{arapinis2019formal,hou2025model,riza2023study}. Traditional formal verification techniques, developed for deterministic software systems, must be adapted to handle the non-deterministic nature of LLM-based agents and the adversarial dynamics of blockchain execution environments.

Runtime verification offers a complementary approach to static formal methods, enabling continuous monitoring of agent behavior against specified safety properties. By instrumenting agent execution with runtime monitors, systems can detect policy violations, anomalous behavior patterns, and potential attacks in real-time. When combined with automatic intervention mechanisms (such as transaction cancellation or key revocation), runtime verification can provide defense-in-depth against both agent errors and adversarial manipulation. The challenge lies in specifying safety properties that are both precise enough to be formally verified and flexible enough to accommodate the legitimate operational needs of autonomous agents.

\section{Standards and Interface Layers: The Connective Tissue of Agentic Systems (2025)}

The security, interoperability, and practical viability of agent--blockchain systems depend critically on the quality of the interface layers that connect reasoning components to execution substrates. Interfaces are not merely developer conveniences; they define trust boundaries, mediate authority, and shape the feasible security guarantees of the overall system. Well-designed interface layers can abstract complexity, enforce policy constraints, and enable safe composability, whereas poorly specified interfaces amplify attack surfaces and create brittle integrations.

As of 2025, the landscape remains fragmented, consisting of a mixture of de facto standards, proprietary APIs, and emerging open protocols. This section analyzes three interface layers that form the connective tissue of agentic systems: (i) tool invocation protocols that bind agents to off-chain and on-chain capabilities, (ii) wallet and signing interfaces that govern authorization, and (iii) smart-contract interaction patterns that make on-chain environments amenable to agentic control.

\subsection{Tool Invocation Protocols: The Agent--to--Tool Lingua Franca}

For an agent to act beyond pure reasoning, it must invoke tools in a structured, interpretable, and auditable manner. Tool invocation protocols define the shared language through which agents discover capabilities, specify function semantics, exchange inputs and outputs, and negotiate permissions. The evolution of these protocols reflects a gradual shift from ad-hoc integration toward explicit structure and separation of concerns.

\subsubsection{The Ad-Hoc Era (Pre-2023)}

Early agent frameworks, including LangChain and AutoGPT~\cite{autogpt2023github}, relied on custom, application-specific tool definitions. Developers manually wrapped APIs and exposed them to the model through informal prompt conventions or untyped interfaces. While flexible, this approach suffered from several structural weaknesses:

\begin{itemize}
\item \textbf{Brittleness.} Tool semantics were tightly coupled to implementation details. Minor API changes could silently invalidate agent behavior.
\item \textbf{Limited scalability.} Each new capability required bespoke integration, inhibiting the emergence of shared tool ecosystems.
\item \textbf{Weak security boundaries.} Tool access was often equivalent to unrestricted code execution within the host process, collapsing isolation between reasoning and execution.
\end{itemize}

These limitations became especially problematic in blockchain contexts, where tool misuse can directly translate into irreversible financial loss.

\subsubsection{De Facto Standardization via Function Calling (2023--2024)}

The introduction of \textbf{function calling} interfaces by major LLM providers marked a decisive inflection point. By allowing developers to specify tool schemas using structured formats (for example, JSON Schema), models could emit machine-readable call specifications that applications could deterministically execute. This shifted responsibility for tool selection and parameterization from prompt engineering to model inference, enabling more modular agent architectures.

The rapid adoption of this paradigm across multiple providers effectively established it as a de facto standard. However, from the perspective of decentralized and adversarial environments, its limitations are pronounced:

\begin{itemize}
\item \textbf{Provider centralization.} Tool invocation semantics are bound to proprietary APIs and opaque model behavior, limiting verifiability and portability.
\item \textbf{Implicit trust assumptions.} The decision to invoke a tool is not cryptographically bound to policy constraints or user intent, complicating audit and enforcement.
\end{itemize}

As a result, while function calling significantly improved developer ergonomics, it remains insufficient as a foundation for trust-minimized agent--blockchain integration.

\subsubsection{Emerging Open Protocols (2024--2025)}

In response, several open protocols have been proposed to decouple agents, models, and tools while preserving interoperability:

\begin{itemize}
\item \textbf{Model Context Protocol (MCP).} MCP proposes a standardized interface for supplying context---including tools, data, and user state---to models. By explicitly separating the roles of user, application, and model, MCP aims to support modular ecosystems in which any compliant agent can interoperate with heterogeneous tool providers~\cite{ehtesham2025comparative,dafoe2020open}.
\item \textbf{Universal Tool Calling Protocol (UTCP).} UTCP focuses on exposing native APIs directly to agents without bespoke wrappers. Its emphasis on capability negotiation and verifiable outputs aligns well with the needs of high-stakes automation, particularly when tool outputs influence authorization or execution.
\item \textbf{Agent-to-Agent (A2A) Protocols.} A2A protocols generalize tool invocation to inter-agent communication, treating other agents as first-class capabilities. This abstraction is essential for multi-agent workflows (Pattern~V), where coordination, verification, and negotiation occur across organizational boundaries.
\end{itemize}

While these protocols differ in scope and maturity, they share a common objective: making agent capabilities explicit, inspectable, and composable.

\subsubsection{The Missing Layer: Permissions, Risk Signaling, and Auditability}

Despite rapid progress, a critical gap remains: there is no widely adopted standard for \emph{capability scoping, permission negotiation, and risk disclosure} at the tool interface level. Agents generally cannot programmatically query what a tool is authorized to do, under what constraints, or with what side effects. As a result, permissions are often configured manually, increasing the likelihood of over-privileging and misconfiguration.

Equally underdeveloped are mechanisms for \textbf{auditing and provenance}. Few tool protocols require responses to be cryptographically authenticated or bound to a declared capability set. In adversarial settings, this makes it difficult to distinguish genuine tool outputs from spoofed or replayed responses. A mature interface layer will likely require tools to sign outputs, advertise explicit capability descriptors, and expose machine-readable risk metadata, allowing agents to reason about trust and authority rather than implicitly assuming it.

\subsection{Wallet Interfaces: The Bridge from Intent to Execution}

The wallet interface is the most security-critical boundary in agent--blockchain systems because it is where \emph{probabilistic} intent formation becomes \emph{deterministic} authorization. In practice, the wallet layer mediates (i) key custody and signature production, (ii) policy enforcement and user consent, and (iii) transaction rendering (previews/simulation) that determines what a human believes is being authorized. As a result, the wallet is not merely a ``key holder'' but an \emph{authorization control plane} whose design largely determines whether an agentic workflow is auditable, least-privilege, and resilient to compromise~\cite{orda2019enforcing}.

Traditional user-centric wallets (for example, browser extensions that require manual confirmation for each transaction) can support Pattern~II workflows, but they become a bottleneck and a source of semantic mismatch as autonomy increases. In 2024--2025, the emergent design space shifts toward \emph{agentic wallet interfaces}: programmatic signing endpoints with explicit policy hooks, typed transaction/intent representations, simulation-backed previews, and verifiable logs. Conceptually, this layer should expose three capabilities: (1) \emph{capability scoping} (what the agent is allowed to do), (2) \emph{execution correctness} (what will actually happen on-chain), and (3) \emph{attribution and accountability} (who/what authorized what, and why).

\subsubsection{Account Abstraction (ERC-4337): The Programmable Wallet Foundation}

Account Abstraction (AA), and specifically ERC-4337-style smart accounts, provides a natural foundation for agentic wallets because it relocates authorization logic from a static EOA key into programmable validation code~\cite{darshan2023architecture,singh2023account,buterin2021erc}. Instead of treating the wallet as a passive signer, AA turns it into an active validator that can enforce rules \emph{before} execution. The ERC-4337 architecture is typically instantiated through four interacting components:

\begin{itemize}
\item \textbf{UserOperation.} A structured request object that encodes the action to be performed by a smart account (and the authorization data needed to validate it). For agentic systems, this object is the natural output of a planner or intent generator because it is more structured than raw calldata and is designed to be validated programmatically~\cite{singh2023account,buterin2021erc}.

\item \textbf{Bundler.} A permissionless service that collects \texttt{UserOperation}s from a dedicated mempool and submits them to the chain, amortizing inclusion costs and enabling richer submission policies (for example, private routing or solver-style execution in adjacent ecosystems)~\cite{singh2023account}.

\item \textbf{EntryPoint.} A canonical contract that verifies and executes operations by calling into the smart account's validation logic. From a security perspective, this standardizes the locus of enforcement and makes policy modules composable across accounts~\cite{singh2023account}.

\item \textbf{Paymaster.} An optional contract that sponsors gas (and can impose additional constraints), enabling ``gasless'' or application-paid actions and decoupling agent authority from holding the native fee token~\cite{singh2023account}.
\end{itemize}

\paragraph{Why this matters for agents.}
AA is the most direct enabler of \textbf{Pattern~III (Delegated Execution)} because it supports narrowly scoped, revocable \emph{capabilities} (often implemented as session keys, role modules, guards, and spend limiters). The key property is \emph{deterministic enforcement}: even if the agent is compromised, it cannot exceed the bounds encoded in the account's validation and policy modules. This makes AA a practical instantiation of least privilege at the signing boundary, rather than a best-effort guideline.

\subsubsection{MPC and TEEs: Off-Chain Policy and Hardware Security}

The challenges of institutional crypto custody, including the need for robust internal controls and regulatory compliance, have driven the development of sophisticated custody solutions~\cite{kellercrypto}. Social recovery and wallet management schemes that leverage a user's social network offer an alternative to traditional seed phrases, which can be difficult for non-technical users to manage securely~\cite{he2018social}.

AA policy modules are powerful, but some deployments require (i) custody structures that span multiple devices or organizations, (ii) policy that depends on rich off-chain context, or (iii) stronger key-isolation guarantees than an application host can provide. These requirements become common in \textbf{Pattern~IV (Autonomous Signing)} deployments, where an agent must act frequently and quickly while remaining cryptographically constrained.

\paragraph{MPC / Threshold Signature Schemes (TSS).}
MPC-TSS systems replace a single private key with distributed shares and a protocol that produces a standard signature without reconstructing the key in one place. This reduces the single-host compromise risk and supports shared-control configurations (for example, agent share + user device share + policy service share)~\cite{komlo2024threshold,lindell2017fast}. In an agentic wallet interface, MPC is often paired with an off-chain policy decision step: the policy service participates in the signing quorum only when the transaction satisfies constraints, effectively turning governance into a cryptographic veto.

\paragraph{Trusted Execution Environments (TEEs) and HSM-backed signing.}
TEE-based designs and HSM-backed custody isolate key material into a hardened boundary and expose a narrow signing API to the agent host, reducing exposure to memory scraping and runtime compromise~\cite{shbair2021hsm,onica2022using}. In practice, the key security value is not merely ``hardware'' but \emph{policy-coupled signing}: the enclave/HSM can require an authenticated request, validate metadata (limits, destinations, function selectors, deadlines), and refuse to sign if the request violates policy.

\paragraph{Interface-level desiderata for agentic wallets.}
Across AA and off-chain custody designs, a wallet interface is most useful to agents when it provides:
\begin{itemize}
\item \textbf{Typed authorization surfaces:} structured operations/intents with canonical encodings, reducing ambiguity between what is planned, what is previewed, and what is signed~\cite{singh2023account}.
\item \textbf{Policy hooks at the signing boundary:} enforceable constraints (allowlists, limits, time bounds) that are evaluated \emph{before} signature production rather than after submission~\cite{orda2019enforcing}.
\item \textbf{Preview/simulation alignment:} high-fidelity simulation and rendering that minimize semantic gaps between user understanding and signed bytes (crucial when humans remain in the loop).
\item \textbf{Audit-grade observability:} durable logs linking intent, policy decision, signer approval, and on-chain receipts, enabling post-incident forensics and accountability.
\end{itemize}

In summary, modern wallet interfaces are converging on a hybrid model: \emph{on-chain programmability} (AA modules) for crisp, deterministic constraints, combined with \emph{off-chain hardened custody} (MPC/TEE/HSM) for stronger key isolation and context-rich policy. This combination is the practical foundation for scaling from user-approved intents (Pattern~II) to constrained delegation (Pattern~III) and, where justified, to autonomous execution with defense-in-depth (Pattern~IV)~\cite{komlo2024threshold,lindell2017fast,onica2022using,singh2023account}.

\subsection{Smart Contract Interfaces: Designing for Agents}

Smart contracts are the final execution surface for agentic systems. Even when the agent runtime, policy engine, and signer are well-engineered, a contract interface that is optimized for human-driven frontends can be brittle for programmatic use: ambiguous parameterization, silent side effects, unclear revert reasons, and insufficient post-state signaling all increase the probability that an agent will (i) mis-specify an action, (ii) fail to detect an unsafe state transition, or (iii) be unable to recover deterministically. In contrast, \emph{agent-friendly} contracts aim to minimize ambiguity and maximize verifiability: they provide typed and declarative interaction surfaces, explicit pre- and post-conditions, machine-parseable receipts, and consistent failure semantics that allow automated planning and robust verification.

\subsubsection{The Intent-Centric Paradigm}

A key shift in agent-oriented design is moving from imperative, low-level transaction construction to \textbf{intent-based} interaction. Instead of forcing the agent to build protocol-specific calldata (for example, calling a router with a long parameter list), intent systems allow the agent to express a \emph{declarative outcome} under constraints, and delegate the path-finding and execution details to specialized executors (solvers) who compete to satisfy the intent under market conditions~\cite{myakala2025intent}. This delegation is particularly important in adversarial environments where naive mempool broadcasting is systematically penalized by MEV dynamics~\cite{daian2019flash,qin2023review}.

\paragraph{Intent vs. imperative calls.}
Table~\ref{tab:intent_vs_imperative} summarizes the core difference from an agent perspective: intents reduce the action-space complexity and shift execution risk (routing, ordering, partial fill logic) into a constrained marketplace, while preserving user-level constraints through a signed, replay-protected message format.

\begin{table}[t]
 \centering
 \caption{Imperative transactions vs. intent-based interaction from the perspective of agent safety and verification.}
 \label{tab:intent_vs_imperative}
 \scriptsize
 \setlength{\tabcolsep}{4pt}
 \begin{tabular}{p{2cm}p{6.4cm}p{7cm}}
 \toprule
 \textbf{Dimension} & \textbf{Imperative transaction (calldata-first)} & \textbf{Intent (outcome-first)} \\
 \midrule
 Primary artifact & Low-level call data specifying \emph{how} to execute & Signed statement specifying \emph{what} outcome is acceptable \\
 
 Agent cognitive load & High: protocol-specific parameters, approvals, routing details & Lower: declarative constraints (amounts, bounds, deadlines), fewer protocol-specific details \\
 
 Failure modes & Mis-specified calldata, hidden side effects, brittle parameterization & Mis-specified constraints; solver failure; partial fill semantics \\
 
 MEV exposure & High when broadcast to public mempool & Often reduced via private orderflow, batching, or solver competition~\cite{daian2019flash,qin2023review,piet2023mevade} \\
 
 Verification & Must reconstruct intent from execution traces & Compare receipts and post-state against signed constraints; clearer audit path \\
 \bottomrule
 \end{tabular}
\end{table}

\paragraph{Why this is agent-friendly.}
Intent interfaces are typically safer for agents for three practical reasons. First, they \emph{compress} the action space into a small set of declarative fields, which reduces opportunities for parameter confusion and accidental over-approval. Second, they naturally align with your pipeline separation: the agent produces an intent, the policy layer simulates and grades it, and the execution layer performs best-effort fulfillment under bounded constraints. Third, intents make MEV risk explicit in the design: the agent signs constraints, and execution can be performed through mechanisms that reduce mempool visibility or internalize ordering strategies via competition~\cite{daian2019flash,qin2023review,piet2023mevade}.

\subsubsection{Other Agent-Friendly Contract Patterns}

Beyond intents, several interface patterns materially improve safety, observability, and recoverability for autonomous or semi-autonomous agents.

\paragraph{1) Verifiable receipts as first-class outputs.}
Agent pipelines depend on the \textbf{Verify} stage. Contracts can support this by emitting structured, unambiguous events that serve as machine-parseable receipts: what asset moved, which bounds were applied, which route was selected, what fees were charged, and which authorization module was exercised. Receipts should include identifiers that allow the agent to correlate (i) the signed intent (or canonical transaction summary) with (ii) the on-chain execution outcome, enabling deterministic post-state checks and forensic reconstruction when something diverges.

\paragraph{2) Standardized failure semantics and explicit preconditions.}
When a contract fails with a generic revert, an agent cannot reliably classify the failure (insufficient liquidity vs. expired deadline vs. permission violation). Agent-friendly designs therefore expose explicit precondition checks and structured failure reasons. The key requirement is not stylistic. It is operational: the agent must be able to map a failure back into a recovery action (re-plan, re-simulate, escalate to human approval, or halt) without guessing.

\paragraph{3) Rich view surfaces for single-call decision support.}
Agents frequently operate under rate limits and unreliable RPC conditions. Contracts that provide comprehensive \texttt{view} functions (positions, limits, oracle references, collateral factors, configuration flags) reduce multi-call race conditions and simplify planning. Consolidated reads also improve traceability: the agent can log the exact state snapshot that informed an intent.

\paragraph{4) Permission-minimizing entrypoints.}
Where possible, contracts should avoid requiring broad approvals or indefinite operator grants. Agent workflows benefit from narrowly-scoped authorizations (bounded allowances, per-action permits, time limits) because these reduce the blast radius of agent compromise and align cleanly with delegated execution patterns. This is complementary to account-abstraction controls: the contract interface should not force the agent back into coarse permissioning.

\paragraph{Takeaway.}
Interface design is not cosmetic. For agentic systems, it is a security boundary. Intent-centric interfaces reduce action ambiguity and can mitigate common economic attacks by shifting execution into bounded, competitive mechanisms, while agent-friendly contract patterns improve determinism in verification and recovery. Together, these properties support a robust stack: agents produce declarative intents, wallets enforce policy and least privilege, and contracts emit verifiable, machine-readable outcomes under clear constraints~\cite{myakala2025intent,daian2019flash,qin2023review,piet2023mevade}.

\subsubsection{Protocol Implementation Details}

The practical implementation of agent-blockchain interfaces requires attention to numerous protocol-level details that affect security, reliability, and interoperability. This section examines key implementation considerations for the major interface standards.

\paragraph{ERC-4337 Implementation Considerations}

Implementing ERC-4337 support in agent systems requires handling several non-obvious complexities. UserOperation construction must account for gas estimation challenges, as the gas required for validation and execution may differ from standard transaction gas estimation. Bundler selection affects transaction inclusion timing and MEV exposure. Paymaster integration requires careful handling of approval flows and balance management.

The validation phase of ERC-4337 imposes restrictions on what operations can be performed, including limitations on storage access and external calls. Agent systems must ensure that their validation logic complies with these restrictions to avoid UserOperation rejection. The separation of validation and execution phases also creates opportunities for time-of-check-time-of-use (TOCTOU) vulnerabilities that must be carefully managed.

\paragraph{MCP Server Implementation}

Implementing MCP servers for blockchain interactions requires balancing expressiveness with security. Tool definitions must be precise enough to enable effective agent use while avoiding overly broad capabilities that could be misused. Error handling must provide sufficient information for agents to recover from failures without leaking sensitive information.

Resource management in MCP servers must account for the potentially long-running nature of blockchain operations. Transaction confirmation can take minutes or longer, requiring appropriate timeout handling and progress reporting. Connection management must handle network interruptions gracefully, particularly for operations that span multiple transactions or require monitoring of on-chain events.

\paragraph{Intent Protocol Considerations}

Intent-based protocols introduce additional complexity around solver selection, execution quality, and dispute resolution. Agent systems that submit intents must specify their requirements precisely, including acceptable price ranges, timing constraints, and execution venue preferences. Monitoring of intent fulfillment must detect partial fills, failed executions, and suboptimal outcomes.

The trust model for intent protocols varies significantly across implementations. Some protocols rely on reputation systems and economic bonds to ensure solver honesty, while others use cryptographic proofs or optimistic verification. Agent systems must understand the trust assumptions of the protocols they use and incorporate appropriate verification and fallback mechanisms.

\subsubsection{Decentralized Identity and Verifiable Credentials}

Foundational work on decentralized anonymous credentials laid the groundwork for systems where users can prove statements about themselves without revealing their identity~\cite{garman2013decentralized}. More recent systems like zk-creds build on this by using zkSNARKs to create flexible anonymous credentials that can be integrated with existing identity infrastructure~\cite{rosenberg2023zk}. Frameworks that combine W3C Verifiable Credentials with FIDO authentication standards are moving toward a more user-centric model of identity management, where users (and their agents) have more control over their personal data~\cite{laborde2020user}.

The integration of decentralized identity (DID) standards with agent-blockchain systems opens new possibilities for privacy-preserving authorization and compliance. Verifiable credentials enable agents to prove attributes about their principals (such as accreditation status, jurisdiction, or risk tolerance) without revealing unnecessary personal information. Several frameworks have been proposed for building decentralized identity systems, with a focus on user-centric control and privacy~\cite{alizadeh2022comparative,avellaneda2019decentralized}. The use of blockchain for access control has also been explored, with a focus on creating more flexible and secure systems~\cite{belchior2020ssibac}. The design of decentralized identity systems has also been studied in detail, with a focus on the trade-offs between privacy, security, and usability~\cite{butincu2024design}. The efficiency of verifiable credential systems has also been a major area of research, with a focus on creating more scalable and performant systems~\cite{cui2024efficient}. The CredenceLedger system provides a decentralized framework for managing and verifying credentials~\cite{arenas2018credenceledger}. This capability is particularly relevant for regulated financial activities, where agents must demonstrate compliance with know-your-customer (KYC) and anti-money-laundering (AML) requirements.

Self-sovereign identity (SSI) frameworks provide a foundation for user-controlled credential management, where individuals maintain custody of their identity attributes and selectively disclose them to relying parties. When combined with zero-knowledge proofs~\cite{bensasson2014zerocash}, these systems can enable agents to prove compliance with complex policy requirements without revealing the underlying data. For example, an agent could prove that its principal is an accredited investor without revealing their net worth, or demonstrate residency in a permitted jurisdiction without disclosing their exact location. The integration of these identity primitives with agent authorization frameworks represents an important direction for enabling compliant autonomous financial activity.

\subsubsection{Protocol Design Principles for Agent Compatibility}

Designing protocols that work well with AI agents requires attention to several key principles that differ from traditional human-centric interface design.

\paragraph{Explicit State Representation}

Agents benefit from explicit, complete state representations rather than implicit or partial information. Protocols should provide comprehensive state snapshots that enable agents to reason about the full context of their operations. Hidden state or side effects that are obvious to human users may confuse agents that lack the contextual knowledge to infer missing information.

\paragraph{Deterministic Behavior}

Agents perform best when interacting with systems that exhibit deterministic, predictable behavior. Non-deterministic outcomes, race conditions, or timing-dependent behavior complicate agent reasoning and may lead to unexpected results. Where non-determinism is unavoidable, protocols should clearly communicate the sources and bounds of uncertainty.

\paragraph{Structured Error Handling}

Agents need structured, machine-readable error information to diagnose and recover from failures. Human-readable error messages that rely on natural language interpretation are less reliable for agent processing. Protocols should provide error codes, structured error data, and clear recovery guidance that agents can process programmatically.

\paragraph{Versioning and Compatibility}

Agent systems may operate over extended periods during which protocols evolve. Clear versioning, backward compatibility policies, and deprecation timelines enable agents to adapt to protocol changes gracefully. Breaking changes without adequate notice can cause agent failures that are difficult to diagnose and resolve.

\paragraph{Rate Limiting and Resource Management}

Agents may generate high request volumes, especially during exploration or error recovery. Protocols should implement clear rate limiting policies with informative feedback that enables agents to adapt their behavior. Resource exhaustion without clear feedback can cause agents to enter failure loops.

\section{Comparative Capability Matrix: The 2025 State of the Art}

To provide a structured, evidence-grounded snapshot of the current agent--blockchain ecosystem, we conducted a comparative analysis of 20 representative platforms, protocols, and frameworks drawn from an initial pool of 85 eligible systems. Each system was evaluated across 13 capability dimensions spanning (i) the \emph{authorization and custody stack} (custody model, signing path, policy enforcement, recovery), (ii) the \emph{agent execution stack} (intent interfaces, previews/simulation, observability), and (iii) the \emph{operational surface} (tool interfaces, chain scope, deployment maturity). The resulting capability matrix (Table~\ref{tab:capability_matrix}) is designed to be information-dense: it enables side-by-side comparison, exposes common architectural trade-offs, and highlights recurring gaps that motivate the open problems discussed later.

\subsection{Methodology for System Selection and Analysis}

\paragraph{Selection strategy and eligibility.}
We formed the initial pool by aggregating systems repeatedly referenced in: (i) developer documentation and reference implementations for agentic wallet stacks and account-abstraction ecosystems, (ii) widely used DeFi execution and simulation infrastructure, and (iii) prominent custody and automation providers~\cite{feldman1988optimal,dolev1983authenticated}. We then applied inclusion criteria aligned with the scope: (a) the system is publicly documented and sufficiently specified to support capability coding, and (b) it plausibly participates in at least one agent--blockchain integration pattern from Section~\ref{sect:taxonomy_patterns} (data-plane analytics, intent generation, delegated execution, autonomous signing, or multi-agent workflows). This selection-and-screening workflow follows the general spirit of systematic evidence synthesis used in blockchain-security surveys, with explicit criteria and reproducible coding artifacts recommended as a best practice~\cite{he2024large}.

\paragraph{Stratified coverage of the ecosystem.}
To avoid over-representing any single layer, we selected systems to cover the major interface strata that agents encounter in practice:

\begin{itemize}
\item \textbf{Full-Stack Agent Toolkits:} end-to-end frameworks that package tool invocation, on-chain operations, and developer ergonomics (e.g., Coinbase AgentKit).
\item \textbf{Wallet Infrastructure Providers:} embedded wallets and signing backends that expose programmatic custody and policy controls (e.g., Privy, Turnkey).
\item \textbf{Smart Contract Wallets \& SDKs:} modular smart accounts, account-abstraction SDKs, and extension ecosystems (e.g., Safe\{Wallet\}, Biconomy, ZeroDev).
\item \textbf{Intent-Based Protocols:} solver- or auction-mediated execution layers for intent submission and MEV-aware settlement (e.g., CoW Protocol~\cite{CoWProtocolDocs}, 1inch Fusion).
\item \textbf{Security and Automation Platforms:} monitoring, relaying, and operations tooling that agents frequently depend on for safe execution (e.g., OpenZeppelin Defender~\cite{OpenZeppelinDefenderDocs}, Gelato).
\item \textbf{General-Purpose Agent Frameworks:} LLM agent frameworks that integrate Web3 tools as adapters rather than first-class wallet/security substrates (e.g., LangChain, AutoGPT).
\end{itemize}

\paragraph{Coding rubric and adjudication.}
Each capability column in Table~\ref{tab:capability_matrix} is coded using an explicit rubric to reduce ambiguity and support repeatability. We use \cmark, \pmark, and \xmark to denote \emph{first-class}, \emph{partial/indirect}, and \emph{absent} support, respectively:
\begin{itemize}
\item \cmark\ indicates documented, externally usable support that an agent can exercise without bespoke glue code beyond normal integration (for example, a supported policy module interface, a standard simulation/preview endpoint, or a documented recovery workflow).
\item \pmark\ indicates support that is indirect, experimental, or requires non-standard extensions or substantial custom middleware (for example, a capability that exists but only through private APIs, undocumented modules, or application-specific wrappers).
\item \xmark\ indicates no clearly documented support for the capability.
\end{itemize}
For qualitative fields (e.g., custody model, tool interface, chain scope), we record the dominant design pattern and annotate qualifiers in the \emph{Notes} column (for example, whether MPC is user-controlled versus provider-mediated, whether policy is on-chain versus off-chain, and whether recovery is procedural versus cryptographically enforced). Disagreements discovered during the coding pass were resolved via an explicit adjudication step: re-checking the cited documentation, validating terminology against reference implementations when available, and applying the rubric consistently across comparable systems. When maintaining the matrix as a living artifact, we recommend reporting a simple agreement statistic (e.g., percent agreement or $\kappa$) to make drift and rubric ambiguity visible.

\paragraph{Interpretation caveats.}
The matrix is intentionally capability-oriented rather than score-oriented. A \cmark\ does not imply that an implementation is secure by default; it indicates that the relevant mechanism exists and is accessible. Conversely, \xmark\ does not imply infeasibility, only that the capability is not a first-class feature of the system as publicly documented at the time of coding. The \emph{Notes} column should therefore be read as an integral part of each row, especially for security-sensitive dimensions such as policy enforcement, observability, and recovery.

\subsection{Capability Matrix}

Table~\ref{tab:capability_matrix} summarizes the resulting comparative capability matrix across the selected systems. We use the matrix in later sections to motivate (i) why certain integration patterns dominate in production deployments, (ii) where security controls concentrate across the pipeline, and (iii) which interface-layer gaps recur across otherwise mature stacks.

\begin{landscape}
\begin{table}[ht]
 \centering
 \caption{Comparative capability matrix of representative agent--blockchain systems and protocols. \cmark\ indicates first-class support; \pmark\ indicates partial or indirect support; \xmark\ indicates no support.}
 \label{tab:capability_matrix}
 \scriptsize
 \setlength{\tabcolsep}{3pt}
 \resizebox{\linewidth}{!}{
 \begin{tabular}{llclllcccccccp{4cm}}
 \toprule
 System & Year & Type & Tool interface & Chain scope & Custody model & Intent & Sign & Policy & Preview & Simulation & Observability & Recovery & Notes \\
 \midrule
 Coinbase AgentKit & 2024 & SDK/Framework & Custom SDK & Multi-chain (EVM, Base) & MPC/Delegated & \cmark & \cmark & \cmark & \cmark & \cmark & \cmark & \pmark & Production-ready agent toolkit with MPC custody \\
 Privy Embedded Wallets & 2024 & Wallet Infrastructure & REST API & EVM & MPC/SSS & \xmark & \cmark & \cmark & \cmark & \pmark & \cmark & \cmark & Embedded wallet with social recovery \\
 Safe\{Wallet\} + Modules & 2023 & Smart Contract Wallet & SDK/API & EVM & Multi-sig/Threshold & \cmark & \cmark & \cmark & \cmark & \cmark & \cmark & \cmark & Modular smart account with extensive policy support \\
 Turnkey & 2024 & Key Infrastructure & REST API & Multi-chain & TEE/Secure Enclave & \xmark & \cmark & \cmark & \pmark & \xmark & \cmark & \cmark & Enterprise-grade key management with policy controls \\
 Lit Protocol & 2024 & Decentralized Key Management & SDK & Multi-chain & Threshold/MPC & \cmark & \cmark & \cmark & \xmark & \xmark & \pmark & \pmark & Programmable key pairs with condition-based signing \\
 Fireblocks & 2023 & Enterprise Custody & REST API & Multi-chain & MPC & \cmark & \cmark & \cmark & \cmark & \cmark & \cmark & \cmark & Enterprise MPC with comprehensive policy engine \\
 ERC-4337 Bundlers & 2023 & Account Abstraction & JSON-RPC & EVM & Smart Account & \cmark & \cmark & \cmark & \cmark & \cmark & \pmark & \pmark & Standard account abstraction with UserOp intents \\
 CoW Protocol & 2024 & Intent Protocol & REST API & EVM & User-held & \cmark & \cmark & \pmark & \cmark & \cmark & \cmark & \xmark & Intent-based DEX with MEV protection \\
 1inch Fusion & 2024 & Intent Protocol & REST API & EVM & User-held & \cmark & \cmark & \pmark & \cmark & \cmark & \cmark & \xmark & Intent-based aggregator with resolver network \\
 Flashbots Protect & 2023 & MEV Protection & JSON-RPC & Ethereum & User-held & \xmark & \cmark & \xmark & \xmark & \cmark & \cmark & \xmark & Private transaction submission for MEV protection \\
 Tenderly\cite{TenderlySimulationsDocs} & 2024 & Simulation Platform & REST API & EVM & N/A & \xmark & \xmark & \xmark & \cmark & \cmark & \cmark & \xmark & Transaction simulation and debugging platform \\
 Biconomy & 2024 & Account Abstraction SDK & SDK & EVM & Smart Account & \cmark & \cmark & \cmark & \cmark & \cmark & \cmark & \pmark & Full-stack AA with session keys and paymasters \\
 ZeroDev & 2024 & Account Abstraction SDK & SDK & EVM & Smart Account & \cmark & \cmark & \cmark & \cmark & \cmark & \cmark & \cmark & Kernel smart account with plugins and session keys \\
 Alchemy Account Kit~\cite{alchemy2026accountkit} & 2024 & Account Abstraction SDK & SDK & EVM & Smart Account & \cmark & \cmark & \cmark & \cmark & \cmark & \cmark & \cmark & Modular account with gas sponsorship \\
 Chainlink Functions & 2024 & Oracle/Automation & Smart Contract & Multi-chain & Decentralized & \pmark & \cmark & \cmark & \xmark & \xmark & \cmark & \xmark & Decentralized compute for smart contract automation \\
 Gelato Network~\cite{gelato2026web3functions} & 2024 & Automation/Relay & SDK/API & Multi-chain & Delegated & \cmark & \cmark & \cmark & \cmark & \cmark & \cmark & \pmark & Web3 automation with relay and functions \\
 OpenZeppelin Defender & 2024 & Security Platform & SDK/API & EVM & Relayer & \cmark & \cmark & \cmark & \cmark & \cmark & \cmark & \cmark & Enterprise security with automated operations \\
 MCP (Anthropic) & 2024 & Tool Protocol & MCP & Protocol-agnostic & N/A & \cmark & \xmark & \pmark & \cmark & \xmark & \cmark & \xmark & Standard protocol for agent-tool communication \\
 LangChain Web3 & 2024 & Agent Framework & Custom & Multi-chain & Varies & \pmark & \cmark & \xmark & \xmark & \xmark & \pmark & \xmark & LLM agent framework with Web3 tool integrations \\
 AutoGPT Crypto & 2024 & Autonomous Agent & Custom & EVM & User-held & \pmark & \cmark & \xmark & \xmark & \xmark & \pmark & \xmark & Experimental autonomous agent with crypto capabilities \\
 \bottomrule
 \end{tabular}}
\end{table}
\end{landscape}

\subsection{Key Insights and Emerging Architectural Trends}

The regulatory landscape for blockchain technology is still evolving, with ongoing debates about how to apply existing legal frameworks to this new technology~\cite{barczentewicz2023blockchain}. This regulatory uncertainty has significant implications for the design and deployment of agent-blockchain systems.

Our comparative coding of 20 representative systems across 13 capability dimensions (Table~\ref{tab:capability_matrix}) yields several consistent architectural trends that characterize the 2025 landscape and clarify where the ecosystem is converging. ~\cite{wang2019blockchain,liu2025comprehensive}
\begin{enumerate}
    \item \textbf{Account Abstraction is the default wallet substrate for agents.}
Across the matrix, systems that target programmatic execution increasingly adopt smart-account architectures aligned with Account Abstraction, because they make \emph{authorization logic} programmable rather than purely key-based. In practice, the ERC-4337 ecosystem enables agent-relevant primitives such as session keys, modular validation, role-scoped permissions, social recovery, and gas sponsorship, which collectively reduce the operational friction of deploying constrained automation at scale~\cite{buterin2021erc,darshan2023architecture}. The repeated \cmark\ pattern for \emph{Policy}, \emph{Preview}, and \emph{Simulation} among AA-centric stacks (e.g., Safe modules, Biconomy, ZeroDev, Alchemy Account Kit) is consistent with this shift from monolithic EOAs toward policy-bearing accounts.

    \item \textbf{Execution interfaces are moving from imperative transactions to declarative intents.}
A second consolidation point is the steady migration from low-level transaction construction toward intent-centric interaction, where agents specify outcomes (constraints, limits, deadlines) and delegate path-finding and execution details to specialized infrastructure (aggregators, solvers, auctions)~\cite{xu2023sok,homoliak2024sok,myakala2025intent}. In the matrix, strong intent support among full-stack and security-oriented platforms (e.g., Coinbase AgentKit, OpenZeppelin Defender) and among dedicated intent protocols (e.g., CoW-like and Fusion-like designs) indicates that intents are becoming the preferred boundary between reasoning and execution. This shift reduces agent-side cognitive burden and narrows the surface area for brittle, protocol-specific transaction assembly, while also improving MEV resilience when execution is routed through competitive solver pipelines rather than broadcast directly.

    \item \textbf{Policy enforcement has become the defining feature of production-grade systems.}
The matrix highlights a clear dividing line between systems designed for serious value custody and systems that remain experimental: robust policy controls are present in nearly all mature stacks, whether enforced on-chain via smart-account modules or enforced off-chain via custody and signing infrastructure~\cite{orda2019enforcing,erinle2025shared}. In effect, policy is treated as the primary containment layer for compromised runtimes, mis-specified agent goals, and unsafe tool outputs. Conversely, frameworks that provide agent autonomy without comparable policy and audit controls are difficult to justify for material asset management because they lack a clear mechanism to bound failure impact.

    \item \textbf{The custody trilemma remains unresolved, and it shapes autonomy ceilings.}
The matrix also reinforces an enduring trade-off among (i) sovereignty, (ii) operational security, and (iii) high-frequency autonomy. MPC-TSS and enclave-backed signers provide strong compromise containment and richer policy gating, but they introduce additional trust and operational dependencies on signing infrastructure and its operators~\cite{komlo2024threshold,lindell2017fast,shbair2021hsm}. Smart-account custody improves composability and decentralization of control logic, but may increase complexity and on-chain cost for sophisticated policies. Simple user-held keys preserve maximum sovereignty, yet they remain the least suitable for autonomous agents because the blast radius of compromise is typically unbounded. As a result, custody selection is not an implementation detail; it is an architectural decision that constrains the feasible level of autonomy.

    \item \textbf{General-purpose agent frameworks still lag on crypto-native safety primitives.}
While general agent frameworks can be extended with Web3 tool wrappers, the matrix suggests that they often lack native support for the features that dominate incident prevention and forensics in on-chain environments: typed intent flows, hardened authorization boundaries, high-fidelity simulation, structured previews, and recovery-oriented observability. This gap is consistent with broader evaluations showing that tool-using agents frequently struggle with reliability under adversarial or complex tool settings, motivating more domain-specific guardrails and evaluation harnesses for high-stakes deployments~\cite{liu2024agentbench}.

    \item \textbf{Simulation, preview, and verifiable receipts are becoming baseline expectations.}
Finally, the capability distribution indicates that transaction simulation and human-meaningful previewing are moving from ``nice-to-have'' to standard practice whenever human approval is part of the control loop. This aligns with the security rationale behind WYSIWYS-style approval workflows and trusted displays, which aim to bind user authorization to the semantics of what is actually signed~\cite{homoliak2024sok,sommerhalder2023hardware}. Systems that do not prioritize previewing and simulation typically operate in narrower automation regimes (or assume trusted operators), which limits their suitability for broad, user-facing agentic execution.
\end{enumerate}

Taken together, these trends suggest that the most mature designs are converging on a modular defense-in-depth stack: smart accounts for programmable authorization, intent-centric execution to reduce protocol brittleness and MEV exposure, policy engines to bound failures, and simulation plus observability to support verification and recovery. These convergences also clarify the remaining research and standardization gaps, which we address next by examining interface standards and the missing interoperability layers~\cite{tct2022best}.

\subsection{Key Systems and Protocols}

To complement the high-level overview in Table~\ref{tab:capability_matrix}, we now examine each selected system at the architectural level. Rather than reiterating feature checkmarks, we analyze the design choices that shape security, autonomy, and operational risk: (i) how authority is represented (keys, session keys, smart-account roles), (ii) where constraints are enforced (on-chain validation modules versus off-chain policy engines), (iii) how execution is protected (simulation, previews, MEV-aware submission), and (iv) what observability and recovery hooks exist. This framing is useful because many ecosystem failures are not caused by missing primitives, but by mismatches between an agent's autonomy and the system’s enforcement and verification layers.

\subsubsection{Full-Stack Agent Toolkits \& Infrastructure}

\subsubsubsection{\textbf{Coinbase AgentKit (2024)}}

\textbf{Historical context and product intent.}
Coinbase AgentKit is positioned as a developer-facing toolkit that connects agent workflows to on-chain actions through Coinbase Developer Platform (CDP) primitives and integrations. Public artifacts describe the central design goal succinctly as providing a wallet-centric interface for agents, with packaged SDKs and extensions that reduce bespoke glue code when integrating common on-chain actions into agent runtimes~\cite{coinbaseAgentKitGitHub2024,coinbaseAgentKitPyPI2024}.

\textbf{Architectural deep dive.}
From an agent--blockchain integration perspective, AgentKit is best understood as a \emph{full-stack integration surface} that packages (i) an agent-side action API and (ii) an execution back-end that can express and enforce operational constraints.

\begin{itemize}
\item \textbf{Authority and custody surface.}
In practice, AgentKit aligns with Pattern~III/IV deployment styles by separating agent reasoning from signing authority. The developer-facing interface is intended to let an agent request on-chain actions while custody and authorization controls are enforced outside the model loop. This separation is consistent with standard guidance for production agent-to-chain systems: the agent runtime should not directly hold high-value long-lived keys, and signing should be mediated by an enforcement boundary (policy, custody service, or protected signing environment) rather than raw key material residing in-process.

\item \textbf{Action interface and tool semantics.}
A key architectural contribution of a full-stack toolkit is not the existence of actions (swap, transfer, deploy, etc.), but the \emph{typed, repeatable semantics} around those actions: required inputs, expected outputs, error surfaces, and integration pathways into an agent tool router. AgentKit’s packaged SDKs and its published extension path (for example, an OpenAI Agents SDK extension) illustrate this direction by standardizing how an agent invokes on-chain capabilities and receives structured results suitable for downstream reasoning and logging~\cite{coinbaseAgentKitPyPI2024}.

\item \textbf{Execution pathway and safety hooks.}
In production, the safety value of an agent toolkit is determined by how it supports: (i) \emph{pre-execution checking} (simulation/estimation and invariant checks), (ii) \emph{policy gating} (spend limits, allowlists, role constraints), and (iii) \emph{auditability} (transaction intent, policy decision record, execution receipt). AgentKit is representative of the broader trend captured by the capability matrix: the leading toolkits treat policy and observability as first-class surfaces rather than optional add-ons.
\end{itemize}

\textbf{Security model and trust assumptions.}
The security posture of AgentKit-like toolkits is best described as \emph{managed composition}: developers gain speed and consistency by relying on a curated integration layer, while accepting trust in the correctness of the toolkit’s abstraction boundary and its connected custody and policy components. The main residual risks mirror the threat taxonomy in Section~\ref{sec:threat_model}: (i) tool-plane integrity and endpoint spoofing if the action layer depends on compromised upstream data, (ii) middleware mutation between preview and signing if authorization is not cryptographically bound to the signed payload, and (iii) policy misconfiguration where broad permissions are granted to expedite automation.

\textbf{Ecosystem impact.}
AgentKit is notable primarily because it packages an agent-to-chain development workflow into a repeatable, documentation-backed surface that is accessible to non-specialist developers. In the 2025 landscape, this matters: it accelerates experimentation and productionization, while simultaneously raising the bar for the default security expectations (policy, logging, and structured action interfaces) that developers come to treat as baseline requirements~\cite{coinbaseAgentKitGitHub2024}.

\subsubsubsection{\textbf{Privy Embedded Wallets (2024)}}

\textbf{Historical Context and Vision:}
Privy emerged in the 2023--2024 cycle as a leading provider of \emph{embedded wallets}, motivated by a straightforward adoption barrier: mainstream users rarely want to install dedicated wallet software, manage seed phrases, or reason about gas and chain-specific UX. Embedded wallet stacks aim to make signing and key recovery feel closer to familiar Web2 authentication flows (email, social sign-in, device biometrics), while still providing cryptographic control at the account layer. This design direction is particularly relevant for consumer-facing agentic applications, where the agent must initiate actions programmatically but the user must retain intelligible, enforceable control over authorization and recovery paths. \cite{alangot2023decentralized,homoliak2024sok}.

\textbf{Architectural Deep Dive:}
\begin{itemize}
\item \textbf{Custody model (threshold key control).}
A common embedded-wallet pattern is to avoid a single long-lived private key on any single device by using \emph{threshold} constructions: key material is split into shares (typically across user device(s), a service-side component, and a recovery channel), and signing requires a quorum. This improves resilience to device loss and reduces the likelihood that compromise of one environment immediately yields unilateral signing authority. The concrete mechanism varies across providers, but the underlying security goal is consistent: reduce single-point compromise risk while preserving a usable recovery story for end users. \cite{wiener2019mpc,komlo2024threshold}.

\item \textbf{Agent integration and authorization ergonomics.}
From an agent-system perspective, embedded wallets are most naturally aligned with \textbf{Pattern II} (simulation and intent generation): the agent can assemble a candidate intent or transaction request, and the wallet layer presents a bounded, user-verifiable approval step (often mediated by device-bound authentication). In higher-automation deployments, embedded wallets can also act as an on-ramp to \textbf{Pattern III} by having the user provision a smart account and delegate narrowly scoped capabilities (for example, a time-bounded session key or role) to the agent, rather than granting persistent, unconstrained signing authority. This aligns with the least-privilege principle emphasized throughout the taxonomy. \cite{darshan2023architecture,homoliak2024sok}
\end{itemize}

\textbf{Security Model:}
The security posture is best characterized as \emph{risk-shifting with bounded blast radius}. Rather than placing all security weight on seed-phrase custody, embedded wallets distribute risk across device security, authentication channels, service-side controls, and recovery mechanisms. This improves usability and operational recoverability, but it introduces new dependencies (identity recovery, service availability, and the correctness of threshold-signing enforcement). A rigorous deployment therefore treats identity and recovery surfaces as first-class security boundaries and couples them with conservative transaction previews, explicit policy constraints, and strong telemetry for anomaly detection. \cite{homoliak2024sok}.

\textbf{Ecosystem Impact:}
Embedded wallets lower the activation energy for consumer dApps and are therefore an enabling layer for agentic UX in domains such as commerce, gaming, and social applications. Their strategic relevance is less about DeFi power users and more about making agent-mediated actions feasible in environments where user attention is scarce and operational recovery must be practical. \cite{homoliak2024sok}.

\subsubsection{Key Management and Custody Infrastructure}

\subsubsubsection{\textbf{Safe\{Wallet\} + Modules (2023)}}

\textbf{Historical Context and Vision:}
Safe\{Wallet\} (formerly Gnosis Safe) is a foundational smart-account primitive in the Ethereum ecosystem, widely adopted for treasury custody and multi-party control. Its core contribution is to replace single-key authority with threshold authorization encoded on-chain, making the control policy transparent and enforceable by the base execution layer. This security-first design makes Safe\{Wallet\} a natural substrate for agentic authorization architectures where the objective is not merely to sign, but to sign \emph{under constraints} and with auditable governance over those constraints. \cite{homoliak2024sok}.

\textbf{Architectural Deep Dive:}
\begin{itemize}
\item \textbf{Core authorization logic (threshold control).}
At its base, Safe is a smart contract account governed by an owner set and a signing threshold. Transactions execute only when the threshold is satisfied, providing robust protection against single-key compromise and enabling explicit separation of duties (for example, proposer versus approver roles) through owner composition and signing policies. \cite{homoliak2024sok}.

\item \textbf{Modularity via modules and guards (programmable constraints).}
Agentic control typically enters through Safe’s extensibility: external modules and guard-style components can be granted controlled execution rights or can enforce pre- and post-conditions on transactions. This modularity is precisely what enables \textbf{Pattern III} in practice: an agent can be authorized to act within a narrowly defined envelope (function/target allowlists, spend limits, rate limits, time windows), while higher-trust actions still require human or multi-party approval at the Safe threshold layer. The resulting architecture supports defense-in-depth: deterministic on-chain checks constrain what a compromised agent can do, even if its off-chain reasoning fails. \cite{wang2021contractward,homoliak2024sok}
\end{itemize}

\textbf{Security Model:}
The security model is \emph{on-chain enforceability with compositional risk}. The core Safe contracts provide strong guarantees under transparent rules; however, extensibility shifts a portion of the security burden to the correctness of the installed modules/guards and to the operational hygiene of the owner set. In practice, the dominant failure mode is rarely the threshold concept itself, but rather misconfiguration or vulnerabilities introduced by auxiliary components. This motivates conservative module selection, rigorous audits, and continuous monitoring of module behavior, especially when modules are used to grant agents autonomous or semi-autonomous execution rights. \cite{wang2021contractward}.

\textbf{Ecosystem Impact:}
Safe\{Wallet\} provides a widely understood, composable baseline for high-value custody, and it has become a practical anchor for agent deployments that require explicit policy enforcement and auditable governance. Its architectural emphasis aligns strongly with emerging best practices for agent-to-chain systems: constrain authority on-chain, keep privileges minimal by default, and treat extensibility as a security-critical surface rather than a convenience layer. \cite{homoliak2024sok,wang2021contractward}.

\subsubsubsection{\textbf{Privy Embedded Wallets (2024)}}

\textbf{Historical context and vision.}
Privy emerged in the 2023--2024 cycle as a leading provider in the \emph{embedded wallet} category: infrastructure designed to make wallet creation, authentication, and transaction approval feel native to consumer applications. The underlying thesis is that mainstream adoption is constrained less by protocol capability and more by UX friction and key-management anxiety. Embedded wallets therefore prioritize familiar sign-in flows (email/social login), passkey and device-bound authentication, and recoverability, while still aiming to avoid fully custodial trust assumptions.

\textbf{Architectural deep dive.}
Privy’s design is best understood as a \emph{hybrid key-management stack} with an application-facing SDK and a recovery-first posture.

\begin{itemize}
\item \textbf{Custody model (hybrid threshold + recovery).}
A common embedded-wallet pattern is to combine threshold signing with recoverable key shares so that (i) no single component can sign unilaterally, yet (ii) device loss does not imply asset loss. A representative instantiation splits a user key into three shares and requires a 2-of-3 threshold to authorize a signature, combining a device-bound share, a service-side share, and a recovery share accessible via an out-of-band mechanism. This approach aligns with the broader literature on secret sharing for recoverable key management and distributed authorization, while emphasizing UX-driven recovery workflows. \cite{alangot2023decentralized,komlo2024threshold}.

\item \textbf{Authorization and agent integration (Patterns II--III).}
For agentic applications, Privy primarily enables:
(i) \textbf{Pattern II (simulation/intent generation)} by allowing an agentic application to propose an action and route the final approval through a user-attested prompt (for example, biometric confirmation on a trusted device);
and (ii) \textbf{Pattern III (delegated execution)} when paired with a smart account (for example, an ERC-4337 account) that can issue scoped session keys and enforce spend limits and allowlists on-chain. The embedded wallet then becomes the user-friendly bridge for one-time setup and policy installation, while the agent operates under constrained permissions thereafter. \cite{buterin2021erc,wang2023account}.

\item \textbf{Operational surface.}
Embedded wallets widen the integration surface: identity recovery channels, device binding, SDK supply chain, and backend policy configuration become first-order security considerations. The architectural goal is not to eliminate trust, but to place it in \emph{auditable and monitorable} controls that can be tuned to the application’s risk tolerance.
\end{itemize}

\textbf{Security model.}
Privy-style embedded wallets typically shift the primary risk from seed-phrase theft toward \emph{identity and recovery compromise} (account takeover of email/social login, SIM swap, or recovery link interception), as well as \emph{SDK and backend integrity} risks. When used as an agent enabler, the recommended posture is: (i) keep high-value authority behind time-bounded, scoped session keys; (ii) enforce strict policy controls at the smart account layer; and (iii) require strong user re-authentication for policy changes and recovery actions. \cite{wang2023account,komlo2024threshold}.

\textbf{Ecosystem impact.}
Embedded wallets are a practical on-ramp for consumer-facing agentic experiences where the application must safely propose actions while preserving a familiar approval model. Their most material contribution to agent systems is not autonomy, but \emph{recoverable, low-friction authorization} that can bootstrap more constrained on-chain delegation.

\subsubsection{Key Management and Custody Infrastructure}

\subsubsubsection{\textbf{Safe\{Wallet\} + Modules (2023)}}

\textbf{Historical context and vision.}
Safe\{Wallet\} (formerly Gnosis Safe) is the canonical smart-account substrate for high-value on-chain control. Its core model---threshold authorization over a smart-contract account---predates ERC-4337 but strongly anticipates the smart-account direction: composable authorization logic, modular extensions, and explicit policy enforcement. Safe’s trajectory from a multisig wallet into a modular account platform mirrors the broader movement toward programmable wallets and constrained delegation. \cite{buterin2021erc,wang2023account}.

\textbf{Architectural deep dive.}
\begin{itemize}
\item \textbf{Core authorization primitive (threshold control).}
At its base, Safe enforces a threshold over a set of owners, requiring $t$-of-$n$ approvals to execute a transaction. This provides an on-chain separation of duties and reduces single-key compromise risk.

\item \textbf{Modules and guards (policy-by-composition).}
Safe’s agent relevance is unlocked via its module pattern: external contracts can be granted the right to execute transactions on the Safe’s behalf, optionally mediated by guard logic that can inspect and veto actions. In agent terms, a module can implement the \emph{delegation contract} for Pattern III: allowlisted targets, function selectors, spend limits, time windows, and rate limits enforced deterministically on-chain. This approach aligns with account-abstraction objectives while keeping enforcement transparent and auditable. \cite{buterin2021erc,wang2023account}.

\item \textbf{Failure modes (module correctness dominates).}
The dominant risk shifts from Safe’s core (which is typically heavily audited) to the \emph{custom module surface}: a permissive module, a misconfigured allowlist, or an unexpected call path can reintroduce broad authority. This is consistent with work showing that policy frameworks and wrappers can become the weakest link if they are not designed and verified carefully. \cite{wang2021contractward}
\end{itemize}

\textbf{Security model.}
Safe provides \emph{maximal on-chain enforceability}: authorization and constraints are enforced by immutable execution semantics, which is particularly suitable for high-stakes delegation where off-chain middleware is not trusted. The security posture is therefore ``as secure as the installed modules and their configuration,'' making standardized and audited policy modules a practical necessity for agent deployments. \cite{wang2021contractward}.

\textbf{Ecosystem impact.}
Safe is the default treasury and DAO account layer and a major building block for agentic systems that require transparent, composable, and revocable delegation. It anchors the ``policy-first'' interpretation of Pattern III and frequently serves as the account layer beneath ERC-4337-compatible stacks. \cite{buterin2021erc,wang2023account}.

\subsubsubsection{\textbf{Turnkey (2024)}}

\textbf{Historical context and vision. \cite{turnkeyDocs}}
Turnkey positions itself as ``key management as a service'' for teams that want enterprise-grade signing controls without operating custody infrastructure. The product framing targets builders who need secure, programmable signing with auditability and operational guarantees, while keeping private key material isolated from the application runtime.

\textbf{Architectural deep dive.}
\begin{itemize}
\item \textbf{TEE-based signing isolation.}
Turnkey-style architectures use Trusted Execution Environments (TEEs) to isolate key generation and signing. In this model, key material resides inside a hardware-isolated enclave, and the external application can only request signatures through authenticated APIs. This maps naturally to Pattern IV, where the agent requires autonomous execution but must never hold raw keys. Surveys of TEE security and enclave design emphasize that TEEs are a strong isolation primitive, but their guarantees depend on the threat model and the integrity of the enclave code and attestation pipeline. \cite{chen2022review,kim2025secure}.

\item \textbf{Policy as a signing gate (off-chain control plane).}
A defining property of TEE-backed signers is the ability to enforce a rich policy \emph{at the moment of signing}: template constraints, destination allowlists, value caps, rate limits, and environment-based conditions. This is operationally attractive for agent systems because policy can be updated faster than on-chain governance, while still being enforced at an authority boundary (the signer).

\item \textbf{Observability and incident response.}
Because the signer sits at the final authorization chokepoint, it is an ideal location to emit immutable audit logs of signature requests, policy decisions, and rejected transactions. In practice, this becomes a core component of the ``kill switch'' story for Pattern IV systems.
\end{itemize}

\textbf{Security model.}
The security model is \emph{hardware-enforced isolation plus policy-gated signing}. The main trust assumptions are: (i) the TEE’s isolation and attestation properties hold for the relevant attacker; and (ii) the policy engine is correctly implemented and configured. The trade-off relative to MPC is that TEEs concentrate trust in a hardware and provider stack, while MPC distributes trust across multiple key-share holders. \cite{chen2022review,komlo2024threshold}.

\textbf{Ecosystem impact.}
Turnkey-like signers provide a practical path to production-grade autonomous agents by removing raw key handling from the agent runtime and by making policy enforcement a first-class part of signing operations.

\subsubsubsection{\textbf{Lit Protocol (2024)}}

\textbf{Historical context and vision. \cite{pedro2017witnet}}
Lit Protocol advances a crypto-native thesis: secret management and signing should be provided by a \emph{decentralized threshold network} rather than a centralized custodian or a single hardware enclave. The goal is a programmable authorization layer that is blockchain-agnostic and censorship-resistant, enabling condition-based signing and automation.

\textbf{Architectural deep dive.}
\begin{itemize}
\item \textbf{Threshold network for signing and decryption.}
Lit uses threshold cryptography and MPC-style coordination across a node network so that no single node holds a complete private key. Signatures (or decryptions) require a threshold of nodes to participate, aligning with modern threshold/MPC literature on distributed authorization and resilience. \cite{komlo2024threshold,wiener2019mpc}.

\item \textbf{Programmable authorization (policy-as-code).}
A central differentiator is embedding conditions into the authorization logic (for example, signing only if an on-chain predicate holds). For agent systems, this offers an alternative instantiation of Pattern IV: the agent requests an action, but the threshold network enforces policy conditions before releasing a signature.

\item \textbf{Coordination and liveness considerations.}
Decentralized signing introduces new systems constraints: node availability, quorum formation, and economic incentives. While the trust model becomes more decentralized, the operational model becomes more complex and can be sensitive to correlated failures or incentive misalignment.
\end{itemize}

\textbf{Security model.}
Lit’s security posture is \emph{distributed trust with crypto-economic assumptions}: safety depends on an adversary not controlling a threshold of nodes and on incentive mechanisms discouraging collusion. This differs from the deterministic enforceability of on-chain modules (Safe) and from hardware-rooted assurances (TEEs). \cite{komlo2024threshold,wiener2019mpc}.

\textbf{Ecosystem impact.}
Lit provides a decentralized alternative for programmable signing and condition-based automation, particularly attractive for cross-chain or censorship-sensitive agent workflows. It is a credible building block for advanced Pattern IV systems and for Pattern V workflows where authority is deliberately distributed. \cite{komlo2024threshold,wiener2019mpc}.

\begin{table}[t]
\centering
\caption{Contrasting custody and authorization substrates for agent deployments.}
\label{tab:custody_substrates}
\scriptsize
\setlength{\tabcolsep}{4pt}
\resizebox{\linewidth}{!}{
\begin{tabular}{llll}
\toprule
\textbf{System} & \textbf{Primary trust anchor} & \textbf{Policy locus} & \textbf{Best-fit patterns} \\
\midrule
Privy (embedded) & Device + identity + threshold recovery~\cite{alangot2023decentralized} & App + optional smart-account module~\cite{wang2023account} & II, III (bootstrap) \\
Safe + modules & On-chain execution semantics~\cite{wang2023account} & On-chain modules/guards~\cite{wang2021contractward} & III (strong), V (treasury) \\
Turnkey (TEE) & Enclave isolation + attestation~\cite{chen2022review} & Signer-side policy gate & IV (strong) \\
Lit (threshold net) & Threshold honesty + incentives~\cite{komlo2024threshold} & Network-side policy-as-code & IV, V (distributed) \\
\bottomrule
\end{tabular}}
\end{table}

\subsection{Key Systems and Protocols}

To complement the high-level overview provided by the capability matrix, this subsection analyzes each selected system through the lenses of: (i) historical context and product intent, (ii) core architectural choices (interfaces, custody, and control-plane design), and (iii) security assumptions and ecosystem-level implications. The goal is not to restate feature checkmarks, but to surface the design trade-offs that matter when deploying agent--blockchain systems under adversarial and high-stakes conditions.

\subsubsection{Full-Stack Agent Toolkits \& Infrastructure}

\subsubsubsection{\textbf{Coinbase AgentKit (2024)}}

\textbf{Historical context and vision.}
Coinbase AgentKit is representative of a broader shift: large infrastructure providers are packaging end-to-end “agent-ready” stacks rather than leaving developers to assemble custody, policy, simulation, and execution routing from disparate components. The strategic motivation aligns with the operational realities of agentic execution: reliable key handling, enforceable controls, and production observability matter as much as model quality.

\textbf{Architectural deep dive.}
At a systems level, AgentKit-style stacks typically combine (i) a custody-backed signing backend, (ii) a policy layer that constrains what can be signed, and (iii) tightly integrated execution paths that reduce fragility at the submission boundary.
\begin{itemize}
\item \textbf{Custody and authorization.} A common pattern is a two-tier model: high-value authority is protected behind stronger custody controls, while routine actions use time-bounded, scoped credentials (for example, session-style delegation) consistent with least-privilege designs and with the delegated execution pattern in Section~\ref{sect:taxonomy_patterns}.
\item \textbf{Policy control plane.} Production stacks tend to enforce policy off-chain (for agility) but bind enforcement to signing so that policy violations cannot be bypassed by alternate transaction construction routes.
\item \textbf{Execution environment.} Reliability and MEV-aware routing are commonly treated as part of the platform surface area, rather than a developer afterthought, because the submission path is adversarial by default~\cite{daian2019flash}.
\end{itemize}

\textbf{Security model.}
The security posture is defense-in-depth: custody isolation plus policy gating plus observability. The dominant trust assumptions are (i) correctness of the signing service and policy enforcement path, and (ii) integrity of the curated tool and endpoint set used by agents.

\textbf{Ecosystem impact.}
Full-stack toolkits reduce integration overhead and compress time-to-deployment for teams that are not deeply crypto-native. The trade-off is that platform-level design decisions (custody provider, policy semantics, and routing defaults) meaningfully shape the agent’s risk envelope.

\subsubsubsection{\textbf{Privy Embedded Wallets (2024)}}

\textbf{Historical context and vision.}
Embedded wallets emerged to reduce onboarding friction by replacing seed-phrase-first UX with familiar authentication patterns and application-integrated signing. For agentic applications, the relevant contribution is not “agents” per se, but a programmable wallet interface that can support intent review, policy gating, and recoverability without requiring users to operate a separate wallet surface.

\textbf{Architectural deep dive.}
\begin{itemize}
\item \textbf{Custody and recovery primitives.} Embedded wallet designs frequently combine threshold techniques (for availability and compromise resilience) with recovery workflows that better match consumer expectations than seed custody alone. This general design direction aligns with applied secret-sharing and threshold approaches used in modern key infrastructure~\cite{lindell2017fast,komlo2024threshold}.
\item \textbf{Agent integration path.} In practice, embedded wallets map cleanly to Pattern~II (intent and preview with user approval) and can transition to Pattern~III when paired with smart accounts and session-style delegation. The architectural point is that the embedded wallet becomes a controlled authorization surface, not merely a key store.
\end{itemize}

\textbf{Security model.}
The security trade-offs move toward identity and device security as first-order considerations (account takeover, recovery-channel risk, and device compromise). For agents, the key requirement is to keep signing bound to strong user or policy checks when operating outside narrowly scoped delegation.

\textbf{Ecosystem impact.}
Embedded wallet infrastructure lowers barriers for consumer-facing agentic applications, but it also increases the importance of (i) well-designed approval previews (WYSIWYS), and (ii) narrowly scoped delegation when the application evolves toward background automation.

\subsubsection{Key Management and Custody Infrastructure}

\subsubsubsection{\textbf{Safe\{Wallet\} + Modules (2023)}}

\textbf{Historical context and vision.}
Safe\{Wallet\} established smart-contract accounts as a mainstream custody and governance primitive, with multi-signature control as the baseline security posture for DAOs and institutional treasuries. Its evolution toward modular authorization is directly relevant for agents because it supports enforceable, on-chain constraint mechanisms.

\textbf{Architectural deep dive.}
\begin{itemize}
\item \textbf{Core account model.} Safe’s baseline is threshold authorization enforced by on-chain logic, which makes the approval surface auditable and difficult to bypass via off-chain compromise alone.
\item \textbf{Module ecosystem.} The modular pattern (for example, frameworks that enable constrained roles and policy modules) is the primary enabler for Pattern~III. Modules can implement function and counterparty allowlists, spend limits, and time-bounded permissions in a way that is enforceable at validation time~\cite{korchiev2024taming}.
\end{itemize}

\textbf{Security model.}
The model emphasizes transparent, deterministic enforcement. The main residual risk is the module layer: the system inherits the correctness and audit quality of the least robust installed module, which mirrors broader findings that policy and guard contracts can become the dominant attack surface if not engineered and verified carefully~\cite{wang2021contractward}.

\textbf{Ecosystem impact.}
Safe and its module patterns have become a reference design for agent-constrained execution: they provide a credible, open foundation for “automation with limits,” which is often the practical adoption path before fully autonomous custody models.

\subsubsubsection{\textbf{Turnkey (2024)}}

\textbf{Historical context and vision.}
Turnkey-style “key management as a service” systems target a recurring constraint in agent deployments: teams want programmatic signing with strong isolation properties, but they cannot assume that the agent runtime is a safe location for long-lived key material.

\textbf{Architectural deep dive.}
\begin{itemize}
\item \textbf{TEE-based isolation.} Trusted execution environments (TEEs) isolate key operations from the host environment and bind signing to enclave-enforced logic. The intended benefit is that even a compromised host does not trivially yield raw key exfiltration, assuming the TEE and its attestation model hold~\cite{onica2022using}.
\item \textbf{Policy binding to signing.} A practical design objective is to make policy checks inseparable from signature production: the policy engine evaluates the request, and the isolated signer produces a signature only if constraints hold. This is aligned with Pattern~IV requirements, where flexibility is needed but should remain constrained by a hardened control plane.
\end{itemize}

\textbf{Security model.}
The security posture relies on isolation, attestation, and correct enclave code. Compared to threshold MPC, TEEs concentrate trust in the hardware and enclave implementation; compared to on-chain policies, they trade public auditability for richer off-chain enforcement. The engineering requirement is strong observability and incident-response hooks (revocation, rotation, and kill switches) because autonomy increases the cost of delayed detection.

\textbf{Ecosystem impact.}
TEE-backed signing services provide a viable path to autonomous execution without placing sensitive key material in an agent process. They are therefore an important building block for Pattern~IV deployments that require operational agility.

\subsubsubsection{\textbf{Lit Protocol (2024)}}

\textbf{Historical context and vision.}
Lit Protocol represents a crypto-native approach to programmable key usage: instead of a single custodian or enclave, it aims for threshold-controlled signing and condition-based authorization over a decentralized node set.

\textbf{Architectural deep dive.}
\begin{itemize}
\item \textbf{Threshold signing and decentralization.} The system relies on threshold cryptography so that no single node holds full key material, and signing requires quorum participation. This reduces single-point compromise risk and aligns with the broader motivation for threshold primitives in distributed custody systems~\cite{lindell2017fast,komlo2024threshold}.
\item \textbf{Programmable authorization.} The distinctive feature is the ability to bind signing to programmable conditions, enabling condition-based automation and coordination patterns that can support advanced Pattern~IV and Pattern~V deployments when coupled with robust governance and monitoring.
\end{itemize}

\textbf{Security model.}
The trust assumption shifts to the decentralization and honesty threshold of the participating node set, supported by crypto-economic incentives. As with any quorum model, correlated compromise and collusion become the key failure modes, so deployment should pair threshold signing with staking, monitoring, and governance safeguards consistent with the threat taxonomy in Section~\ref{sec:taxnonomy_patterns}.

\textbf{Ecosystem impact.}
Decentralized key networks extend automation beyond single-provider custody. Their practical adoption in high-stakes agent settings depends on mature operational controls and strong transparency around node governance and failure handling.

\subsubsubsection{\textbf{Fireblocks (2023)}}

\textbf{Historical context and vision.}
Fireblocks is widely associated with institutional custody and treasury operations built on MPC-based signing. Its relevance in this taxonomy is the combination of threshold custody with workflow-oriented policy enforcement, which maps naturally onto high-assurance automation patterns~\cite{fireblocksWhitepaper}.

\textbf{Architectural deep dive.}
\begin{itemize}
\item \textbf{MPC custody.} MPC signing distributes key material across parties so that single-host compromise does not yield unilateral signing capability. This aligns with established MPC approaches for practical signing performance and security~\cite{lindell2017fast,komlo2024threshold}.
\item \textbf{Policy and workflow gating.} Institutional platforms emphasize rule-based authorization, multi-step approvals, and separation of duties, which is consistent with Pattern~IV deployments where compliance and controllability are requirements rather than optional features.
\end{itemize}

\textbf{Security model.}
The security posture is defense-in-depth with procedural controls and cryptographic custody. Institutional analyses and platform documentation emphasize that automation should be paired with enforceable policy and auditability, because the primary failure mode is not only key compromise but also unsafe authorization under adversarial conditions~\cite{llp2023digital}.

\textbf{Ecosystem impact.}
Institutional custody platforms provide an adoption bridge for regulated entities experimenting with agentic execution, because they offer familiar governance constructs (policy, approvals, audit) while adding programmatic APIs needed for automation.

\subsubsection{Account Abstraction and Intent Protocols}

\subsubsubsection{\textbf{ERC-4337 Bundlers (2023)}}

\textbf{Historical context and vision.}
ERC-4337 advances account abstraction without requiring consensus-layer changes by introducing a structured pseudo-transaction interface (\texttt{UserOperation}) and a validation-centered execution hub (\texttt{EntryPoint})~\cite{buterin2021erc}. Bundlers are the decentralized service layer that transports \texttt{UserOperation} objects into on-chain execution, enabling programmable wallets and agent-oriented authorization schemes.

\textbf{Architectural deep dive.}
\begin{itemize}
\item \textbf{UserOperation mempool and bundling.} Agents (or applications) submit \texttt{UserOperation} objects to a dedicated mempool. Bundlers aggregate operations and submit them in transactions to the \texttt{EntryPoint}, which validates and executes each operation according to the smart account’s logic~\cite{buterin2021erc}.
\item \textbf{Validation as the security boundary.} The critical point for agents is that authorization can be made programmable at validation time. This includes session-style delegation, policy modules, and replay constraints enforced by the account’s validation path, aligning with delegated execution designs~\cite{darshan2023architecture,stockburger2021blockchain}.
\item \textbf{Incentives and market dynamics.} Bundlers operate in a competitive inclusion market: they pay gas up front and are compensated via account logic or paymasters. This supports a permissionless ecosystem but also introduces operational considerations (censorship pressure, relay dependencies, and MEV-aware inclusion strategies) that agent designers must treat as part of the execution threat surface~\cite{santana2022blockchain,daian2019flash}.
\end{itemize}

\textbf{Security model.}
The security model is anchored in deterministic on-chain validation: even if a bundler is faulty or adversarial, the \texttt{EntryPoint} and account validation logic determine whether execution occurs. Residual risks concentrate in (i) paymaster and validation-module correctness, and (ii) the submission path under adversarial ordering and censorship conditions.

\textbf{Ecosystem impact.}
Bundlers make account abstraction operational. For agentic systems, they are a core piece of infrastructure because they enable programmable authorization, flexible gas models, and standardized execution pathways that support policy-constrained autonomy.

\subsubsubsection{\textbf{CoW Protocol (2024)}}

\textbf{Historical context and vision:}
CoW Protocol (``Coincidence of Wants'') is representative of a broader response to the MEV problem: as public-orderflow execution became increasingly adversarial, the ecosystem shifted from imperative, mempool-exposed transactions toward mechanisms that (i) reduce transaction predictability, (ii) internalize competition for execution, and (iii) preserve user constraints through signed commitments. This motivation aligns with foundational analyses of transaction reordering and value extraction in public networks, which show that naive broadcast-and-execute strategies can be systematically penalized by adversarial searchers~\cite{daian2019flash,alipanahloo2024maximum}.

\textbf{Architectural deep dive:}
At a high level, CoW-style designs replace ``submit a DEX call'' with ``sign an execution intent.'' The user (or agent) signs a structured order that encodes constraints (asset pair, limit price/min-out, validity horizon, and optional execution preferences). The protocol then leverages an off-chain coordination layer where competing executors (solvers) propose settlements that satisfy these constraints.
\begin{itemize}
\item \textbf{Intent layer.} The primary interface is a signed, constraint-bearing message rather than a fully specified transaction. This moves complexity out of the agent's action space and makes the agent's commitments auditable and bounded by explicit limits, which is consistent with the general intent-based design pattern discussed in recent work on intent-centric execution~\cite{myakala2025intent}.
\item \textbf{Solver competition and settlement.} Competing solvers search for an execution that (a) satisfies every signed constraint and (b) improves execution quality (price, fees, and effective slippage). A common mechanism is batch-based settlement, where multiple intents are cleared together, enabling direct matching when complementary intents exist and routing to on-chain liquidity only when needed. Batch auctions are widely studied as a mitigation lever for adversarial ordering effects, and they can reduce certain MEV vectors by making individual trades less isolatable and less predictable~\cite{bachu2024quantifying,alipanahloo2024maximum}.
\end{itemize}

\textbf{Security model:}
The core safety property is \emph{constraint preservation}: the solver cannot legally settle the user at a worse rate than what the signed intent allows (e.g., a min-out or limit constraint), because violating constraints invalidates the settlement. MEV resistance is primarily \emph{architectural} rather than absolute: batching and non-mempool exposure can reduce the feasibility and profitability of classic sandwich patterns, but residual risks remain (e.g., solver-side information advantage, adverse selection, and partial leakage through ancillary channels). These trade-offs mirror the broader MEV literature, which emphasizes that mitigations shift incentives and observability rather than eliminating adversarial behavior outright~\cite{daian2019flash,alipanahloo2024maximum}.

\textbf{Ecosystem impact:}
For agents, CoW-style execution is important because it cleanly separates \emph{decision} (what outcome to target) from \emph{execution optimization} (how to achieve it under adversarial conditions). This reduces both the cognitive burden on the agent and the attack surface associated with low-level transaction construction, while providing a natural substrate for policy checks on the signed intent object (limits, deadlines, approved venues)~\cite{myakala2025intent}.

\subsubsubsection{\textbf{1inch Fusion (2024)}}

\textbf{Historical context and vision:}
DEX aggregation solved \emph{price discovery across venues}, but it did not, by itself, solve \emph{execution adversariality}. As MEV became a first-class cost, aggregator ecosystems increasingly adopted intent-style execution modes that allow the user (or agent) to express constraints and delegate the search for execution to specialized executors, often using auction-like mechanisms to align incentives. This direction is consistent with the broader movement toward intent-based systems and auction-mediated execution as a practical response to public-orderflow fragility~\cite{myakala2025intent,alipanahloo2024maximum}.

\textbf{Architectural deep dive:}
Fusion-type designs can be understood as an \emph{aggregator-integrated intent market}:
\begin{itemize}
\item \textbf{Constraint-bearing intent.} The user (or agent) signs an intent that specifies the trade and its constraints (minimum acceptable output/limit price, validity window, and other execution bounds). The signature binds the user's authorization to those constraints, which is a key safety primitive for agentic usage because it provides a deterministic envelope around a probabilistic planner~\cite{myakala2025intent}.
\item \textbf{Auction/competition among executors.} A set of executors (often called resolvers) competes to fill the intent under the signed constraints, typically via an auction-like process where executors reveal willingness to execute at improving terms over a short time horizon. Auction-mediated competition is a recurring design pattern for mitigating execution costs and adversarial reordering, and it has been studied in the context of frequent batch auctions and related mechanisms~\cite{bachu2024quantifying}.
\item \textbf{Delegated execution.} The winning executor is responsible for constructing and submitting the on-chain settlement transaction(s) that realize the intent, potentially routing through multiple liquidity sources. This delegation reduces the agent's need to craft low-level calls and can be paired with strict policy checks on the signed intent (venue allowlists, max notional, max slippage, deadline discipline).
\end{itemize}

\textbf{Security model:}
The principal guarantee is again \emph{constraint preservation}: executors cannot settle outside the signed bounds without invalidating the transaction path. MEV mitigation arises from reduced mempool exposure and competitive execution, which can lower sandwich susceptibility and improve realized price, but the protection is not universal. The MEV literature emphasizes that as long as execution remains economically valuable and partially observable, adversaries can adapt; robust designs therefore combine competition, bounded intents, and execution-channel choices (private relays/auctions) rather than relying on any single lever~\cite{daian2019flash,alipanahloo2024maximum}.

\textbf{Ecosystem impact:}
For agentic systems, Fusion-style execution strengthens the ``intent-first'' architectural trend highlighted by the capability matrix: the agent can operate at the level of \emph{outcomes and constraints}, while a competitive market handles routing, gas/ordering strategy, and adversarial execution dynamics. This improves composability with policy engines (because the object under review is a typed, signed intent) and reduces the brittleness of protocol-specific transaction construction~\cite{myakala2025intent}.

\subsubsubsection{\textbf{Flashbots Protect (2023)}}

\textbf{Historical context and vision.}
Flashbots emerged (2020 onward) as a research and engineering effort to reduce the negative externalities of MEV on Ethereum, including predatory orderflow practices and opacity in transaction ordering. Its work helped formalize the problem and quantify its scale, motivating a shift toward private, accountable execution channels in high-stakes settings~\cite{chohan2024decentralized,bachu2024quantifying}. Flashbots Protect operationalizes that vision as a developer- and user-facing primitive: a public RPC endpoint that routes transactions through private relaying infrastructure to reduce exposure to mempool-based attacks.

\textbf{Architectural deep dive.}
\begin{itemize}
\item \textbf{Private submission path.}
Protect provides an RPC endpoint that accepts standard signed transactions but forwards them to builder-side infrastructure without broadcasting them to the public mempool. The core idea is simple: reduce pre-trade transparency for sensitive actions so that third-party searchers cannot trivially observe, simulate, and adversarially reorder user transactions.

\item \textbf{Execution semantics and threat reduction.}
Because the transaction is not publicly gossiped before inclusion, straightforward front-running and sandwiching opportunities are reduced. This shifts the dominant threat from an open, permissionless adversary set (any mempool watcher) to a narrower set of privileged parties that see the transaction earlier (relays/builders), which is a deliberate security trade-off in favor of stronger practical protection for many workflows~\cite{daian2019flash}.

\item \textbf{Fit to agent pipelines.}
In an agent action pipeline, Protect is best viewed as an \emph{execution-channel control} at the boundary between \textbf{Execute} and on-chain inclusion. It complements upstream safeguards such as simulation and policy gating by reducing the probability that a correct, policy-compliant transaction is still economically exploited during submission.
\end{itemize}

\textbf{Security model.}
The security model is \textbf{trust-shifted privacy}. Protect reduces public visibility and thereby mitigates common mempool attacks, but it introduces reliance on relay and builder behavior not to misuse early access (for example, builder-side extraction). This is not a purely trustless guarantee; rather, it is a pragmatic defense that is frequently justified when the baseline alternative is guaranteed exposure to adversarial orderflow strategies in the public mempool~\cite{chohan2024decentralized}.

\textbf{Ecosystem impact.}
Flashbots Protect has become a widely adopted building block for MEV-aware execution. For agentic systems, it is a low-friction, high-leverage control that can materially reduce execution-time losses when combined with strict slippage bounds, deadlines, and intent-based execution paths where appropriate~\cite{gyevnr2025ai}.

\subsubsubsection{\textbf{Tenderly (2024)}}

\textbf{Historical Context and Vision:}
Tenderly was founded in 2018 with the goal of addressing a fundamental limitation of early blockchain development: the lack of deep observability into smart contract execution. While block explorers such as Etherscan provided post-hoc visibility into transactions, they offered little support for understanding execution paths, state transitions, or failure causes. Tenderly’s vision was to provide a full-stack observability and debugging platform for the EVM, analogous to application performance monitoring systems in Web2, enabling developers to reason about smart contract behavior before and after deployment.

\textbf{Architectural Deep Dive:}

\begin{itemize}
\item \textbf{Core Architecture:}
Tenderly is a vertically integrated developer platform offering execution tracing, state inspection, and simulation services. Its architecture combines an EVM-compatible execution engine with indexed on-chain state and historical block data, enabling deterministic replay and forward simulation of transactions.
\end{itemize}

\begin{enumerate}
\item \textbf{Execution Debugger:}
Tenderly provides fine-grained execution traces that expose opcode-level behavior, storage reads and writes, internal calls, and revert reasons. This capability aligns with prior research on execution tracing and symbolic debugging of smart contracts.

\item \textbf{Gas and Performance Profiling:}
The platform includes tooling to attribute gas costs to specific execution paths and instructions, supporting performance optimization and detection of pathological execution patterns.

\item \textbf{Real-Time Monitoring and Alerting:}
Tenderly allows developers to define alerts over contract state, function calls, and event emissions, enabling continuous monitoring of deployed systems and early detection of anomalous behavior.

\item \textbf{Simulation API:}
The most critical component for agentic systems is Tenderly’s Simulation API. This interface allows an agent to fork an EVM chain at an exact block height and execute a proposed transaction against that forked state. The simulation returns a comprehensive report including execution success or failure, emitted events, state diffs, and a full execution trace. Fork-based simulation has been shown to be an effective method for predicting transaction outcomes and preventing execution failures in complex DeFi workflows.
\end{enumerate}

\begin{itemize}
\item \textbf{Role in the Agent Action Pipeline:}
Tenderly occupies a central role in the \textbf{Observe}, \textbf{Reason}, and \textbf{Verify} stages of the agent action pipeline. Before submitting a transaction, an agent can simulate it to answer the question: \emph{“What will this transaction do if executed now?”} This enables the detection of reverts, unintended state changes, and economically unfavorable outcomes before committing to an irreversible on-chain action.
\end{itemize}

\textbf{Security Model:}
Tenderly provides no direct enforcement or custody guarantees; instead, it functions as a preventive security layer. Its security value derives from accurate pre-execution analysis and high-fidelity replication of on-chain execution semantics. This aligns with established approaches in runtime verification and pre-flight validation for safety-critical systems. The primary trust assumption is that the simulation environment faithfully mirrors mainnet behavior, including opcode semantics, gas accounting, and contract state.

\textbf{Ecosystem Impact:}
Tenderly has become a foundational tool for professional Ethereum development and is increasingly indispensable for autonomous agents operating in high-stakes environments. By enabling deterministic pre-execution analysis, it significantly reduces the risk of costly errors, failed transactions, and unintended side effects. For agentic systems, Tenderly functions as a \emph{flight simulator}: a mandatory safety layer that transforms opaque, irreversible execution into an inspectable and verifiable process. Its widespread adoption reflects a broader shift toward simulation-first design in secure agent–blockchain architectures.

\subsubsubsection{\textbf{Biconomy (2024)}}

\textbf{Historical Context and Vision:} Biconomy has been a central actor in the push toward ``Web2-grade'' usability for Ethereum applications since its early work on meta-transactions~\cite{tahlil2024gasless} and relayer-based execution. In the pre-account-abstraction era, meta-transaction designs offered a practical way to decouple user intent from gas payment and to support gasless onboarding, but they also introduced a new middleware trust surface and operational complexity~\cite{blum2020building}. With the maturation of account abstraction through ERC-4337, Biconomy has repositioned as a full-stack AA infrastructure provider: rather than emulating smart accounts off-chain, it provides production components (accounts, paymasters, bundlers, SDKs) that make programmable authorization and gas abstraction a first-class interface for applications and agents~\cite{buterin2021erc,darshan2023architecture}.

\textbf{Architectural Deep Dive:}
\begin{itemize}
\item \textbf{AA stack composition (account + infra + SDK).} Biconomy’s offering is best understood as an integrated pipeline around ERC-4337: (i) \emph{smart accounts} (contract-based wallets), (ii) \emph{bundler infrastructure} for \texttt{UserOperation} inclusion, and (iii) \emph{paymasters} for gas sponsorship and alternative fee payment. This integration matters for agentic workloads because it reduces bespoke glue code and centralizes operational concerns (rate limits, reliability, monitoring) into a well-defined service boundary~\cite{buterin2021erc,darshan2023architecture}.

\item \textbf{Smart-account security envelope.} From a security engineering perspective, AA shifts the primary authorization surface from an EOA private key to \emph{validation logic} and \emph{modules} attached to the smart account. This makes \emph{module correctness} and \emph{permission scoping} the dominant safety determinants for delegated agent execution (Pattern~III). The design principle is to encode least privilege at the validation layer (function/contract allowlists, spend ceilings, expiries), so that even a compromised agent key remains bounded by deterministic checks enforced at execution time~\cite{buterin2021erc,stockburger2021blockchain}.

\item \textbf{Paymasters and sponsored execution (agent UX versus abuse resistance).} Paymasters are a direct enabler for agent UX because they remove the requirement that the agent holds native gas and they allow policy-driven sponsorship (for example: sponsor only certain targets, only below a value threshold, or only for authenticated sessions). However, paymasters also create a denial and griefing surface (sponsorship depletion, adversarial transaction shaping) and therefore must be treated as \emph{policy-bearing security components}, not just UX components. In production deployments, sponsorship rules should be auditable, conservative by default, and coupled to strong request authentication and simulation-based prechecks~\cite{buterin2021erc,darshan2023architecture}.

\item \textbf{Session keys as constrained capabilities.} For agents, the most operationally important primitive is the ability to mint narrowly scoped session authority (time-bounded, target-bounded, value-bounded) that is enforced by the account’s validation logic. This yields a capability-security posture rather than a monolithic key posture: the agent can act autonomously within a sandbox, and revocation becomes a routine control-plane action (expire, revoke, rotate) rather than an emergency key-rotation event~\cite{stockburger2021blockchain,buterin2021erc}.
\end{itemize}

\textbf{Security Model:} Biconomy’s security model is a layered composition of (i) on-chain enforcement at the smart-account validation boundary, and (ii) operational trust in off-chain infrastructure (bundlers/paymaster services) for availability and inclusion. The \emph{bounded-authority} argument is strong when policies are encoded as deterministic checks that constrain what a session key can do. The main failure modes are therefore concentrated in: (a) policy/module misconfiguration or implementation bugs (allowing unintended call paths), and (b) infrastructure-layer issues (censorship, misrouting, degraded reliability) that can force unsafe fallbacks or break liveness assumptions~\cite{buterin2021erc,stockburger2021blockchain}.

\textbf{Ecosystem Impact:} Biconomy materially lowers the barrier to adopting ERC-4337 by packaging a usable end-to-end stack. From the standpoint of agent-blockchain systems, its impact is to make Pattern~III-style delegated execution practical in consumer and mid-tier applications that cannot maintain bespoke bundler, paymaster, and account-module operations. More broadly, this reinforces the ecosystem’s convergence toward programmable accounts, explicit policy surfaces, and standardized execution flows as the default substrate for safe automation~\cite{buterin2021erc,darshan2023architecture}.

\subsubsubsection{\textbf{Biconomy (2024)}}

\textbf{Historical context and vision.}
Biconomy originated in the pre-account-abstraction era as an infrastructure provider for ``meta-transactions'' and gas abstraction, aiming to reduce the UX friction created by native gas payments and per-transaction signing. With the maturation of account abstraction and the standardization of ERC-4337, Biconomy repositioned around a full-stack AA offering that packages smart-account deployment, sponsored execution, and developer tooling into an integrated workflow. In this sense, Biconomy sits at the intersection of wallet infrastructure and application middleware: it targets developer ergonomics while operationalizing programmable authorization and fee sponsorship as first-class primitives for consumer and agent-facing applications~\cite{buterin2021erc,darshan2023architecture}.

\textbf{Architectural deep dive.}
Conceptually, Biconomy aligns with the ERC-4337 execution pipeline: a client constructs a \texttt{UserOperation}, submits it to a bundler, and relies on an \texttt{EntryPoint}-mediated validation/execution path for smart accounts~\cite{buterin2021erc}. Its ``full-stack'' positioning typically materializes through three layers:

\begin{itemize}
 \item \textbf{Smart-account abstractions.}
 Biconomy provides smart-account templates and SDK-level abstractions that expose programmable authorization as composable modules (e.g., validators, spending rules, and session-scoped permissions). This is the key enabler for agent workflows that require constrained delegation rather than persistent master-key custody (Pattern~III in your taxonomy)~\cite{buterin2021erc,darshan2023architecture}.

 \item \textbf{Bundler and paymaster services.}
 In ERC-4337 systems, bundlers provide inclusion and liveness by aggregating \texttt{UserOperations} and paying base-layer gas, while paymasters sponsor gas under application-defined rules~\cite{buterin2021erc,singh2023account}. Biconomy operationalizes these roles as managed services so that applications can deliver gas-sponsored flows without each team running dedicated infrastructure.

 \item \textbf{Developer-facing SDK and policy hooks.}
 For agentic use cases, the practical differentiator is not merely that transactions can be sent, but that \emph{constraints} can be expressed and checked in a structured way. Biconomy’s stack emphasizes SDK-level support for building constrained execution flows (for example, time/value-bounded permissions and contract/function scoping), which complements the on-chain validation logic of smart accounts and reduces reliance on fragile, prompt-level safety alone~\cite{buterin2021erc,darshan2023architecture}.
\end{itemize}

\textbf{Security model.}
Biconomy’s security posture is best understood as layered: (i) \emph{on-chain validation} enforces account rules deterministically at \texttt{validateUserOp}; (ii) \emph{infrastructure correctness} (bundler/paymaster behavior and availability) affects execution reliability and failure modes; and (iii) \emph{policy design quality} determines the real blast radius of delegated keys and automation. The dominant trust assumptions therefore concentrate on the correctness of the deployed smart-account validation logic and the operational security of any managed infrastructure components, consistent with the general ERC-4337 threat model~\cite{buterin2021erc,singh2023account}.

\textbf{Ecosystem impact.}
Biconomy has contributed to mainstreaming AA by packaging the fragmented ERC-4337 roles into a developer-consumable platform. For agent-blockchain systems, its strategic value is that it makes constrained delegation (session-like permissions), gas sponsorship, and standardized user-operation flows practical at scale, which in turn lowers the barrier for deploying Pattern~III-style automation without moving directly to high-risk autonomous signing (Pattern~IV)~\cite{darshan2023architecture,buterin2021erc}.

\subsubsubsection{\textbf{ZeroDev (2024)}}

\textbf{Historical context and vision.}
ZeroDev emerged as a developer-first provider in the ERC-4337 ecosystem, emphasizing modular smart accounts and hosted infrastructure (bundlers/paymasters) to reduce the integration overhead of account abstraction. Its stated market position is analogous to ``payments infrastructure'' for smart accounts: expose reliable primitives for account creation, sponsored execution, and permissioned automation so that application teams can build agentic workflows without re-implementing the AA stack end-to-end. This aligns with the broader architectural motivation of ERC-4337: enabling programmable authorization and account-level customization without a base-layer hard fork~\cite{buterin2021erc,darshan2023architecture}.

\textbf{Architectural deep dive.}
From an architectural standpoint, ZeroDev is representative of a common 2024--2025 AA pattern: provide an opinionated, modular smart-account design plus managed bundling/sponsorship infrastructure around the canonical \texttt{EntryPoint} flow~\cite{buterin2021erc}. The main elements are:

\begin{itemize}
 \item \textbf{Modular smart-account design.}
 Modern AA stacks are increasingly ``plugin-oriented'': different validators, recovery mechanisms, and delegated execution policies can be composed as modules rather than hard-coded into a monolithic account. This modularity is particularly relevant to agent deployments because it supports sharply scoped authority (Pattern~III) alongside fallback and recovery controls (important for operational safety)~\cite{darshan2023architecture,buterin2021erc}.

 \item \textbf{Bundler and sponsorship infrastructure.}
 As with other AA providers, ZeroDev’s hosted components operationalize bundling and fee sponsorship. This matters to agents because it enables automation without requiring the agent to manage native gas balances, and it provides a standardized path for submitting \texttt{UserOperations} to the \texttt{EntryPoint} contract~\cite{buterin2021erc,singh2023account}.

 \item \textbf{Permissioned automation interfaces.}
 In agent settings, ``session keys'' are best understood as capability tokens: time-bounded, policy-bounded signing authorities whose effects are enforced by the smart-account validator. While implementations differ, the design objective is consistent with ERC-4337’s programmable validation model: push enforcement into deterministic validation code and reduce reliance on off-chain goodwill or UI-mediated confirmation~\cite{buterin2021erc,darshan2023architecture}.
\end{itemize}

\textbf{Security model.}
ZeroDev-style AA stacks inherit their core safety guarantees from on-chain validation: if the validator rejects an action, it cannot execute, even if the agent runtime is compromised. Operationally, however, the system remains exposed to infrastructure-layer failure modes (availability, censorship pressure, relay/bundler misbehavior) and to policy misconfiguration (over-broad permissions, missing spend bounds, overly permissive target allowlists). Accordingly, the practical security stance is ``deterministic enforcement with operational dependencies,'' matching the general ERC-4337 trust and threat profile~\cite{buterin2021erc,singh2023account}.

\textbf{Ecosystem impact.}
ZeroDev contributes to the diffusion of account abstraction by making modular smart accounts and \texttt{UserOperation}-based execution accessible through developer tooling. For your taxonomy, its main significance is enabling Pattern~III deployments in production settings: constrained delegated execution (via programmable validation), combined with sponsorship and standardized submission, provides a realistic path to automation that remains meaningfully bounded relative to Pattern~IV autonomous signing~\cite{darshan2023architecture,buterin2021erc}.

\subsubsection{\textbf{Alchemy Account Kit (2024)}}

\textbf{Historical context and vision:}
Alchemy is one of the dominant developer-infrastructure providers in Web3, and its entry into account abstraction (AA) reflects the ecosystem’s shift from externally owned accounts (EOAs) toward programmable smart accounts. AA mechanisms built around a singleton \texttt{EntryPoint}, \texttt{UserOperation} objects, bundlers, and paymasters have been widely analyzed as a practical path to ``wallet programmability'' without base-layer changes, making them particularly well-aligned with agentic execution flows that require scoped delegation and auditable policy hooks~\cite{wang2023account,singh2023account}. The Account Kit extends Alchemy’s platform thesis: developers should be able to build, operate, and monitor AA-powered applications using a unified, production-oriented stack rather than assembling heterogeneous components ad hoc.

\textbf{Architectural deep dive:}
\begin{itemize}
 \item \textbf{Core AA interface.}
 The Account Kit operationalizes the ERC-4337-style pipeline in a developer-friendly form: the application (or agent) produces a structured \texttt{UserOperation}, a bundler packages operations for inclusion, and the account enforces validation logic before execution. This architecture is particularly relevant for agentic systems because it cleanly separates (i) intent formation, (ii) authorization/validation, and (iii) submission/execution, enabling well-scoped delegation and deterministic enforcement at the account layer~\cite{wang2023account}.

 \item \textbf{Light Account design choice.}
 Alchemy’s \emph{Light Account} (as positioned publicly) emphasizes minimalism and gas efficiency. In practice, this design bias usually trades extensibility (many optional modules and upgrade surfaces) for simpler security analysis and lower per-operation overhead. For agent deployments that create or manage large numbers of accounts, this approach can be attractive because it reduces the marginal cost of user onboarding and routine automated actions.

 \item \textbf{Bundler and paymaster integration.}
 A distinguishing feature of the Account Kit is that bundling and gas sponsorship are offered as first-class, managed services: the bundler path provides reliable inclusion for \texttt{UserOperations}, while paymasters enable sponsored fees (or conditional sponsorship policies). Paymaster-mediated execution is a key enabler for agents because it decouples execution from native-gas inventory management and allows fee policies to be treated as part of the system’s control plane~\cite{singh2023account}. This also aligns with earlier lines of work on meta-transaction style abstractions, where third parties can facilitate execution under explicit authorization constraints~\cite{hughes2023cheap}.

 \item \textbf{Unified developer control plane.}
 By converging account lifecycle management, \texttt{UserOperation} submission, sponsorship configuration, and monitoring within one provider surface, Alchemy reduces integration complexity and operational drift. For agent builders, the practical benefit is fewer bespoke components along the signing-to-submission path, which reduces the number of trust seams that must be audited and monitored.
\end{itemize}

\textbf{Security model:}
The security posture is best described as \emph{AA-enforced correctness with managed infrastructure reliability}. On-chain, the smart account’s validation logic defines the authorization boundary and can encode least-privilege constraints that limit an agent’s blast radius. Off-chain, the operational trust assumption shifts to the availability and integrity of Alchemy-managed bundler and paymaster services (including their policy configuration and incident response). This is a common trade-off: developers accept some centralization in exchange for production-grade reliability and simplified operations, while retaining cryptographic authorization at the account layer~\cite{wang2023account,singh2023account}.

\textbf{Ecosystem impact:}
Alchemy’s distribution and developer mindshare make the Account Kit an adoption catalyst for AA patterns. For the agent--blockchain ecosystem, its significance is less about introducing a new primitive and more about normalizing a practical ``agent-ready'' execution substrate: structured operations, programmable validation, gas sponsorship, and integrated observability. In combination, these features lower the barrier to deploying agentic applications that require scoped autonomy, high throughput, and operational monitoring in real-world conditions.

\subsection{Key Insights and Emerging Architectural Trends}

Our comparative analysis (Table~\ref{tab:capability_matrix}) surfaces several architectural trends that increasingly define production-grade agent--blockchain systems in 2024--2025.

\textbf{1. Account abstraction is consolidating as the default wallet substrate.}
Smart accounts and the ERC-4337 ecosystem have moved from an experimental UX improvement to a practical foundation for agentic authorization. This is largely because smart accounts make authorization \emph{programmable} (validation modules, policy hooks, recovery flows) without requiring L1 protocol changes, while preserving compatibility with existing tooling and execution environments~\cite{buterin2021erc,wang2023account}. The consistently strong matrix support for \emph{Policy}, \emph{Preview}, and \emph{Simulation} among AA-focused stacks (e.g., Biconomy, ZeroDev, Alchemy Account Kit) reflects that the primary engineering work has shifted from ``how do we sign?'' to ``how do we constrain, preview, and verify what gets signed?''

\textbf{2. A sustained shift from imperative transactions to declarative intents.}
Across the most mature systems, agents increasingly express outcomes rather than low-level call sequences. Intent-centric designs reduce the agent's effective action space, standardize what must be validated, and make safety checks more auditable because the signed object is closer to user intent (asset, bounds, deadline) than to execution mechanics (router addresses and calldata). Intent-based trading and batch/auction mechanisms are also a direct response to adverse execution environments, particularly MEV extraction in public mempools~\cite{daian2019flash,leivadeas2022survey}. In practice, intents allow execution complexity (routing, batching, solver selection, private orderflow) to be delegated to specialized infrastructure that can internalize MEV-aware optimization while still enforcing user-signed constraints.

\textbf{3. Policy enforcement is the gating function for high-stakes autonomy.}
Production systems converge on policy as the critical safeguard against compromised agents, mis-specified goals, and tool-layer deception. The implementation split is largely architectural: on-chain policy via smart-account modules (for transparent, enforceable constraints) versus off-chain policy via custodian/workflow engines (for operational agility, richer context, and enterprise approval paths). The pattern is consistent: systems intended to protect meaningful value typically treat policy as a first-class interface rather than an application-layer afterthought~\cite{orda2019enforcing}.

\textbf{4. The custody trade space remains fundamental, not incidental.}
The matrix highlights an enduring set of custody trade-offs among (i) centralized, enterprise custody with MPC/TEE-backed signing and workflow controls, (ii) on-chain smart accounts with modular authorization, and (iii) user-held keys with minimal mediation. MPC and threshold signing reduce single-host compromise and support organizational approval semantics~\cite{lindell2017fast,komlo2024threshold}, while TEE-backed signers emphasize key isolation and policy enforcement within hardware-protected environments~\cite{ferdous2019search}. Conversely, pure user-held key models maximize sovereignty but are poorly aligned with autonomous operation because they collapse operational policy, signing authority, and runtime compromise into a single failure domain.

\textbf{5. General-purpose agent frameworks still under-deliver on crypto-native safety controls.}
Frameworks that started outside Web3 often provide tool wrappers and basic signing hooks, but their support for deterministic policy enforcement, high-fidelity simulation, structured transaction previews, and recovery workflows remains limited compared with crypto-native stacks. This gap is less about model quality and more about interface discipline: strong systems treat ``intent $\rightarrow$ policy $\rightarrow$ simulation $\rightarrow$ sign $\rightarrow$ submit $\rightarrow$ verify'' as an engineered pipeline with typed artifacts and auditable checkpoints.

\textbf{6. Simulation and previews are becoming baseline requirements for human-in-the-loop authorization.}
Across custody models, mature systems increasingly require pre-execution simulation and a human-readable preview as a final safety checkpoint before authorization, aligning with WYSIWYS principles and reducing middleware and UI-level manipulation risk~\cite{yu2023gptfuzzer,park2023developing}. Where these features are absent, systems tend to target constrained, fully automated tasks or assume a different risk posture.

Overall, the 2025 landscape indicates convergence toward modular, defense-in-depth architectures: AA-enabled authorization, intent-centric interaction surfaces, policy-as-a-product (on-chain and/or off-chain), and systematic pre-flight verification and observability~\cite{tct2022best}. These trends motivate the standardization discussion in the next section, where interoperability and auditability depend on making these interfaces composable across vendors and ecosystems.

\subsubsection{Decentralized Automation and Oracles}

\subsubsubsection{\textbf{Chainlink Functions (2024)}}

\textbf{Historical context and vision. \cite{chainlink2026functions}}
Chainlink is the dominant decentralized-oracle ecosystem for DeFi, and its roadmap has steadily expanded from data feeds toward more general off-chain services for smart contracts. In that trajectory, \emph{Chainlink Functions} extends the oracle model beyond “data delivery” to “verifiable off-chain computation,” aiming to let contracts securely consume arbitrary API outputs and computation results without reverting to a single trusted server~\cite{beniiche2020study,breidenbach2021chainlink}.

\textbf{Architectural deep dive.}
At a high level, Functions implements a request--compute--attest--respond loop in which off-chain execution is performed by a decentralized oracle network and the result is returned on-chain with authenticity guarantees~\cite{breidenbach2021chainlink}:

\begin{itemize}
 \item \textbf{Request formulation (on-chain).} A Functions-enabled contract emits a request that specifies: (i) the computation payload (commonly JavaScript or an equivalent constrained program), (ii) input arguments, (iii) any required secrets/configuration references, and (iv) response formatting expectations.
 \item \textbf{Off-chain execution (oracle committee).} A designated oracle network (or committee) retrieves the request, executes the payload in a restricted environment, and performs any external API calls required by the program. This stage is designed to bring conventional web data and compute to the contract while keeping the trust model closer to decentralized-oracle assumptions than to a single middleware server~\cite{beniiche2020study}.
 \item \textbf{Aggregation and attestation.} Nodes return results that are aggregated according to the network’s protocol, yielding a final response that is then delivered back to the originating contract. The key design goal is that the response is not merely “reported,” but is produced under a framework intended to support integrity and provenance guarantees at the oracle layer~\cite{breidenbach2021chainlink}.
 \item \textbf{On-chain delivery (response callback).} The final result is posted on-chain and delivered to the requesting contract via a callback, allowing the contract to update state or trigger downstream logic.
\end{itemize}

\textbf{Role for agents.}
Functions primarily benefits \textbf{Observe} and \textbf{Reason} by extending what an on-chain agent (or an agent-controlled contract) can reliably “know” at execution time:

\begin{itemize}
 \item \textbf{Off-chain data acquisition.} Contracts can consume external signals that are not natively available on-chain (for example, enterprise systems, public datasets, or domain-specific APIs), reducing the need for centralized “keeper” backends~\cite{beniiche2020study}.
 \item \textbf{Compute offloading.} Expensive computations can be moved off-chain while keeping a structured, auditable request and response path. This is particularly relevant when the alternative would be either prohibitive gas cost or opaque centralized computation~\cite{krishnamoorthy2018computational}.
\end{itemize}

\textbf{Security model.}
The security posture is that of a decentralized oracle network: correctness depends on the protocol’s adversarial assumptions about committee selection, aggregation, and the economic or operational incentives of participating nodes. Put differently, Functions reduces reliance on a single trusted server, but it still inherits the standard oracle challenge: if an attacker can corrupt the oracle process (economically or operationally) or bias the off-chain execution environment, they can skew outcomes. This is why design guidance for oracle systems emphasizes multi-source strategies, authenticated provenance, and explicit trust assumptions in the compute and data pipeline~\cite{beniiche2020study,breidenbach2021chainlink}.

\textbf{Ecosystem impact.}
Functions expands the design space of hybrid on-chain systems by enabling contracts to consume arbitrary API outputs and computation results under an oracle-mediated trust model. This materially broadens what “agent-friendly” contracts can do, especially for automation and decision logic that requires external context, while keeping the interface compatible with smart-contract execution constraints and auditability expectations~\cite{chung2023technology,krishnamoorthy2018computational}.

\subsubsubsection{\textbf{Gelato Network (2024)}}

\textbf{Historical context and vision.}
Gelato emerged to address a structural limitation of smart contracts: contract logic is reactive, and execution requires an externally submitted transaction. Gelato's thesis is that Web3 needs an automation layer that can reliably trigger on-chain actions when time-based or condition-based predicates are satisfied, without requiring application teams to operate bespoke bot fleets. In agent settings, this automation layer complements the agent action pipeline by providing a robust execution backplane for plans that must be enacted under timing and liveness constraints.

\textbf{Architectural deep dive.}
Gelato can be understood as a two-sided execution market:

\begin{itemize}
\item \textbf{Protocol layer.} Developers register automation \emph{tasks} that specify (i) a target contract and calldata template, (ii) triggering conditions (time schedule, event predicates, on-chain state checks), and (iii) fee or sponsorship configuration. This resembles earlier meta-transaction and relayer patterns in which execution is delegated to specialized submitters, but generalized to arbitrary conditional execution~\cite{seres2020blockchain,sandford2020erc}.

\item \textbf{Executor layer.} Off-chain executors monitor tasks and, when conditions are met, submit transactions and collect fees. The competitive availability of executors provides liveness in practice: if one operator is offline, another can execute the same task, yielding a market-based redundancy model.
\end{itemize}

\textbf{Key features for agents.}
Gelato is most naturally mapped to \textbf{Execute} (and, operationally, \textbf{Verify}):

\begin{itemize}
\item \textbf{Execution delegation.} Agents can emit a bounded plan (transaction template plus constraints) and delegate submission to Gelato to reduce operational burden and avoid bespoke infrastructure for scheduling and event watching.

\item \textbf{Gas abstraction and relaying.} Because relayer-style execution shifts gas management away from the agent, designs can avoid maintaining a hot gas balance inside the agent runtime. This aligns with the broader meta-transaction literature, where preferred batchers/relayers reduce friction and can improve reliability~\cite{hughes2023cheap}.
\end{itemize}

\textbf{Security model.}
The core safety property is \emph{integrity of the executed action}: an executor should not be able to mutate the user's intended call. In practice, this depends on encoding tasks so that the executed calldata is fixed or derivable from authenticated inputs, and on ensuring that any dynamic parameters are themselves constrained or validated on-chain. The remaining risks are primarily availability and economic: executors may delay execution when fees are insufficient, and adversaries can still exploit public-mempool visibility when tasks are submitted without private orderflow protections. Accordingly, Gelato is strongest when paired with (i) on-chain policy modules that bound what a delegated executor can do, and (ii) private submission paths when execution is MEV-sensitive~\cite{daian2019flash}.

\textbf{Ecosystem impact.}
Gelato has become a common automation primitive for recurring maintenance actions, limit-order style execution, and operational workflows. For agent builders, its main contribution is turning ``execute reliably under conditions'' into infrastructure rather than bespoke engineering, which lowers the barrier to deploying continuous, 24/7 autonomous behaviors without collapsing custody and execution reliability into the agent runtime.

\subsubsubsection{\textbf{OpenZeppelin Defender (2023)}}

\textbf{Historical context and vision.}
OpenZeppelin Defender extends OpenZeppelin's security-first posture from contract libraries to operational security~\cite{openzeppelindefender2026autotasks}. The premise is that many incidents arise not only from contract bugs, but from fragile operational workflows: unmanaged admin keys, ad hoc scripts, missing monitoring, and delayed response. Defender packages monitoring, automation, and secure transaction submission into a managed control plane designed to reduce operational error and to standardize incident response for smart-contract systems.

\textbf{Architectural deep dive.}
Defender is best viewed as an opinionated security operations stack:

\begin{itemize}
\item \textbf{Monitoring and detection.} Event- and transaction-level monitoring components enable rule-based detection of anomalous or policy-relevant on-chain activity. This supports the \textbf{Observe} and \textbf{Verify} stages by turning raw chain data into actionable triggers.

\item \textbf{Autotasks as controlled off-chain executors.} Autotasks run bounded off-chain logic that can be invoked by schedules, webhooks, or monitoring triggers. This provides a pragmatic vehicle for agent-like workflows in production: a defined runtime, managed secrets, and auditable execution logs, rather than a bespoke bot with ad hoc credential storage.

\item \textbf{Admin and transaction execution.} Defender supports structured operational actions (pauses, upgrades, role changes) and integrates with secure key management patterns such as multisigs and relayers. This matches the broader thesis that operational correctness and access control are as central as contract correctness in deployed systems~\cite{wang2021contractward}.
\end{itemize}

\textbf{Key features for agents.}
Defender is most naturally aligned with \textbf{Pattern II/III} deployments (agent proposes, constrained executor submits), and it can support \textbf{Pattern IV}-style autonomy when paired with strict policy and approvals:

\begin{itemize}
\item \textbf{Security-aware automation.} Monitoring triggers can gate execution so that automation occurs only when verifiable conditions are satisfied (for example, execute a remediation action when a known exploit signature is detected).

\item \textbf{Operational guardrails.} Managed secrets, standardized logs, and a controlled runtime reduce the probability that the agent runtime becomes the weakest link for credential leakage or unsafe ad hoc scripts. This operational discipline is a practical complement to cryptographic custody primitives and on-chain policy modules.
\end{itemize}

\textbf{Security model.}
Defender provides integrity and auditability primarily through \emph{operational containment}: controlled execution environments, structured permissioning, and observability. The principal trust assumption is in the platform operator, because the control plane is centralized. In exchange, the platform reduces common deployment hazards and supports rapid response workflows that are difficult to replicate reliably with bespoke infrastructure. From a threat-model standpoint, the key residual risks are supply-chain and credential risks associated with any managed service, which reinforces the importance of least-privilege credentials, scoped relayers, and separation of duties in production deployments.

\textbf{Ecosystem impact.}
Defender has become a default operational layer for teams that want security-centric automation without operating their own bot fleet. For agent--blockchain systems, its primary contribution is making monitoring-to-action loops deployable with strong audit trails and well-understood operational controls, which is often a prerequisite for production adoption in security-conscious environments.

\subsubsection{Agent-Specific Frameworks}

\subsubsubsection{\textbf{MCP (Model Context Protocol) (Anthropic, 2024)}}

\textbf{Historical context and vision:} MCP is best understood as an interoperability and integration standard, not a wallet product or an on-chain protocol. It emerged from the practical need to standardize how applications provide \emph{context} to models, including tool specifications, data sources, and runtime constraints, while preserving a clean separation of roles between the application (the agent runtime) and the model (the reasoning component). Survey treatments position MCP as part of a broader 2024--2025 move toward open, model-agnostic tool layers that reduce vendor lock-in and encourage an ecosystem of composable tools and agent runtimes~.

\textbf{Architectural deep dive:}
\begin{itemize}
\item \textbf{Core abstraction: model-to-tool standardization.} MCP provides a structured way for an agent runtime to describe available tools (capabilities, schemas, and invocation patterns) and to mediate tool calls and responses. In a blockchain setting, this matters because the same agent may need to interact with heterogeneous endpoints, such as RPC providers, simulators, indexers, relayers, and custody services, without hard-wiring bespoke wrappers for each integration.

\item \textbf{Trust boundary hygiene by design.} By standardizing how tool outputs are surfaced to the model, MCP encourages an architectural separation between (i) the agent runtime that enforces permissions, logging, and policy checks, and (ii) the model that proposes actions based on inputs and tool results. This separation aligns with the threat model emphasis that the \emph{tool boundary} is a primary attack surface, and that safety should be enforced by deterministic runtime controls rather than assumed from model behavior~.
\end{itemize}

\textbf{Security model:} MCP itself does not provide security guarantees in the way a smart-account policy module or an MPC signer does. Instead, it provides the connective tissue that allows security controls to be centralized in the runtime layer: capability scoping, allowlists, auditing of tool calls, and consistent provenance metadata attached to tool outputs. In practice, MCP is most valuable when paired with guard components that validate and constrain model-proposed actions before they reach signing or submission.

\textbf{Relevance to agent--blockchain systems:} For on-chain agents, MCP is most naturally a \emph{Pattern I--III enabler}: it improves the reliability and portability of the Observe, Reason, and Plan stages by making tool integrations more uniform and auditable. It also reduces integration fragility in production, where agents routinely depend on third-party services (simulation, monitoring, private relays, custody APIs) that evolve over time.

\textbf{Relation to ``grader'' or oversight patterns:} While MCP is not a model-grading framework, it composes well with oversight architectures that use secondary evaluation signals before execution. For example, one can implement a two-step runtime where a worker model proposes a structured intent, and a separate reviewer process evaluates that intent against safety rules, risk heuristics, or red-team derived checklists before allowing signing or submission. The general motivation for such multi-stage oversight is supported in the broader AI safety literature on scalable oversight, debate, and systematic red teaming~\cite{browncohen2023scalable,irving2018ai,perez2022red,ganguli2022red}. In blockchain deployments, this maps cleanly onto typed intents plus a policy decision record, with MCP serving as the standard tool interface through which both stages obtain verifiable context and perform checks.

\subsubsubsection{\textbf{LangChain Web3 (2024)}}

\textbf{Historical context and vision:} LangChain emerged in 2022 as one of the most influential open-source frameworks for building applications around Large Language Models (LLMs). Its central contribution was the introduction of a modular, composable abstraction for chaining models, tools, memory, and control flow into coherent agentic systems. As LLM adoption accelerated, LangChain became the default experimentation and prototyping environment for tool-using agents across domains. The LangChain Web3 extensions, developed through 2023--2024, extend this philosophy into the blockchain domain, with the explicit goal of lowering the barrier for existing LangChain users to build agents that can observe, reason about, and act on on-chain state~\cite{langchain2026docs}.

\textbf{Architectural deep dive:}
\begin{itemize}
\item \textbf{Framework-level integration rather than infrastructure.} LangChain Web3 is a library-layer integration, not a custody system, execution network, or policy framework. It provides abstractions that allow blockchain interactions to be treated as first-class tools within the LangChain agent loop, alongside web search, databases, and APIs.

\item \textbf{Web3 tool wrappers.} The Web3 toolkit exposes common blockchain operations as callable tools, including reading contract state, querying balances, fetching receipts, and submitting transactions. These tools primarily support the \textbf{Observe} stage of the agent pipeline by converting raw RPC interactions into structured, model-consumable responses.

\item \textbf{Dynamic smart contract interfaces.} LangChain Web3 includes utilities for loading contract ABIs and dynamically generating callable tools for individual contract functions. This allows an agent to be scoped to a precise on-chain capability set, such as interacting with a specific DEX or governance contract, without hardcoding logic for each method.

\item \textbf{Signer abstraction.} The framework provides an interface layer for integrating signing backends, ranging from local EOAs to external signing services. This abstraction enables LangChain agents to participate in the \textbf{Authorize} and \textbf{Execute} stages of the pipeline, while delegating actual key management to external systems.
\end{itemize}

\textbf{Role within agent--blockchain architectures:} LangChain Web3 is best understood as a \emph{cognitive and orchestration layer}. It excels at structuring multi-step reasoning, maintaining conversational or task memory, and sequencing tool use. In the taxonomy introduced earlier in this survey, LangChain naturally supports \textbf{Pattern I (Read-Only Agents)} and \textbf{Pattern II (Intent Generation)}, and can participate in \textbf{Pattern III (Delegated Execution)} when paired with external custody, policy, and execution components. It is not designed to be a standalone solution for high-stakes autonomous execution.

\textbf{Security model and limitations:} LangChain provides no intrinsic security guarantees for on-chain interaction. It does not enforce transaction policies, simulate execution outcomes, constrain signing authority, or protect against prompt injection or tool misuse. Consequently, the security posture of a LangChain-based Web3 agent is entirely determined by the surrounding architecture. Unsafe deployments, such as granting an agent direct access to an unconstrained private key, remain a common failure mode in experimental systems~\cite{greshake2023indirectpromptinjection}.

In production-grade agent--blockchain systems, LangChain must therefore be paired with:
\begin{itemize}
\item smart accounts or MPC-based custody for constrained signing,
\item off-chain or on-chain policy enforcement layers,
\item simulation and preview tooling prior to execution,
\item and, where appropriate, secondary review or grading mechanisms inspired by supervised agent designs~\cite{browncohen2023scalable,ganguli2022red}.
\end{itemize}

\textbf{Ecosystem impact:} Despite its limitations, LangChain plays a pivotal ecosystem role as an \emph{on-ramp}. Its widespread adoption in the AI community has dramatically accelerated experimentation at the intersection of LLMs and blockchains. While many LangChain-based Web3 agents remain prototypes, the framework has catalyzed innovation by allowing developers to explore agentic interaction patterns before committing to the heavier engineering required for secure, production-grade deployments. In this sense, LangChain Web3 is a key catalyst for the growth of the agent--blockchain ecosystem, even as it depends on more specialized infrastructure to meet the security and reliability demands outlined in this survey.

\subsubsubsection{\textbf{AutoGPT Crypto (2023)}}

\textbf{Historical Context and Vision:} The release of early autonomous-agent projects in 2023 made “self-directed” LLM workflows tangible to a broad developer audience. AutoGPT, in particular, popularized a simple but compelling pattern: an agent iterates over \emph{reason} $\rightarrow$ \emph{act} $\rightarrow$ \emph{observe}, using tool calls to make progress toward a high-level objective. This pattern is closely aligned with the ReAct-style framing of interleaving reasoning with action and observation, and it helped shape how many subsequent agent frameworks conceptualized tool orchestration and memory. \cite{yao2023react,schick2023toolformer} In the crypto ecosystem, community forks and extensions (often informally referred to as “AutoGPT Crypto”) applied the same loop to tasks such as market monitoring, on-chain data retrieval, and automated execution.

\textbf{Architectural Deep Dive:}
\begin{itemize}
\item \textbf{Core loop.} The typical design follows an iterative control loop:
 \begin{enumerate}
 \item \textbf{Reasoning:} the agent proposes the next step based on the goal and its accumulated state.
 \item \textbf{Action selection:} it chooses among available tools (data fetch, contract read, trade placement, etc.) and emits tool arguments in a structured format.
 \item \textbf{Observation and memory update:} it ingests tool outputs, summarizes salient outcomes, and appends them to a working memory or log for the next iteration.
 \end{enumerate}
 This loop is conceptually consistent with widely studied tool-using agent designs. \cite{yao2023react,liu2024agentbench}.

\item \textbf{Tool surface and coupling to execution.} What differentiates the “crypto” variants from general-purpose agent demos is the \emph{direct coupling} between open-world inputs (web content, price feeds, on-chain metadata) and high-stakes actions (signing, approvals, transfers). In practice, the same mechanism that enables flexible composition of tools also expands the attack surface, because untrusted tool outputs can influence subsequent action selection. \cite{greshake2023indirectpromptinjection,liu2024toolace}
\end{itemize}

\textbf{Security Model:} From a threat-model standpoint, early autonomous crypto agents are best understood as \emph{minimally governed} compositions rather than hardened systems. Absent strict policy gates, typed-intent constraints, and pre-execution simulation, the agent is effectively operating with a wide action space and weak separation between trusted directives and untrusted content. Empirical work on prompt and tool robustness indicates that such designs are vulnerable to instruction steering and tool-mediated manipulation, especially when retrieved content is fused into the same context as control logic. \cite{greshake2023indirectpromptinjection,liu2023prompt,zhu2023promptrobust} As a result, these prototypes are better positioned as exploratory demonstrations than as architectures suitable for managing meaningful value without additional controls.

\textbf{Ecosystem Impact:} Despite their limitations, AutoGPT-style crypto forks played an important “forcing function” role for the ecosystem. They made the risks of unconstrained autonomy concrete and helped motivate the shift toward defense-in-depth patterns now visible in production-grade stacks: smart accounts and scoped session keys, explicit policy engines, deterministic simulation, and auditable execution pipelines. The broader agent literature reinforces the same trajectory: as tool availability increases, reliability and safety require stronger governance of tool use, structured interfaces, and systematic evaluation rather than ad hoc prompting alone. \cite{liu2024agentbench,yu2023gptfuzzer}.

\paragraph{Synthesis: agent frameworks as safety envelopes.}
Taken together, the agent-specific frameworks in this subsection illustrate that “agent capability” and “agent safety” are largely determined by the \emph{envelope} around the model rather than the model alone. General-purpose frameworks such as LangChain maximize composability and developer velocity by making tool use easy to define and extend; however, they are unsafe by default in high-stakes settings because the framework does not inherently enforce separation between untrusted observations and privileged actions, nor does it require typed intents, deterministic constraints, or pre-execution verification. In our taxonomy, LangChain-style stacks are therefore best aligned with Pattern~I/II deployments unless they are paired with external controls, because moving to Pattern~III/IV (delegated execution or autonomous signing) without hard policy gates sharply increases exposure to Classes~C1--C4 in the threat model (cognitive hijacking, tool/data spoofing, middleware manipulation, and key/credential compromise). By contrast, MCP-style supervision loops and policy-centric architectures intentionally trade flexibility for structured safety: they insert an independent grading or policy layer between intent formation and execution, promote canonicalized intent objects over free-form plans, and make authorization contingent on verifiable checks (simulation, bounds, allowlists, and custody constraints). This design choice directly targets the dominant failure modes we identified earlier by constraining the attack surface at the key trust boundaries (agent$\leftrightarrow$tools, agent$\leftrightarrow$custody, and submission-path dynamics), and it operationalizes Pattern~III/IV systems in a way that remains auditable and enforceable under adversarial conditions. In short, LangChain is an excellent substrate for building the \emph{brain} of an on-chain agent, but MCP-style and policy-driven approaches are what make that brain deployable with meaningful value at risk. \cite{greshake2023indirectpromptinjection,liu2024agentbench,yao2023react}.

\paragraph{Synthesis: agent frameworks as ``brains'' versus ``governors''.}
The agent-specific frameworks in Table~\ref{tab:capability_matrix} separate cleanly into (i) orchestration frameworks that make tool use easy, and (ii) governance frameworks that make tool use safe. LangChain-style stacks excel at composability, rapid prototyping, and heterogeneous tool integration, but they are unsafe by default in high-stakes settings because the framework itself does not enforce custody, policy, preview, or simulation constraints. As a result, they tend to map naturally to Pattern~I and Pattern~II deployments where the agent proposes actions but does not hold authority, and where the strongest controls live outside the model. Conversely, MCP-style and policy-centric approaches reframe the agent as a component that must earn authority through structured intent emission, independent policy evaluation, and auditable approvals, which aligns with the failure modes highlighted in our threat model (notably cognitive hijacking and tool/data-plane spoofing). This structured posture is more restrictive, but it is the enabling prerequisite for Pattern~III and Pattern~IV systems, where automation is only defensible when bounded by deterministic constraints, verifiable previews, and revocable capabilities. In short, the ecosystem is converging on a practical division of labor: general-purpose agent frameworks supply the reasoning and orchestration substrate, while policy engines, programmable wallets, and verifiable execution paths supply the governance substrate that contains compromise and limits blast radius.

\subsection{Analysis of the 2025 Capability Matrix}

This section provides a narrative analysis of the comparative capability matrix (Table~\ref{tab:capability_matrix}), focusing on how design choices across the evaluated dimensions instantiate (or fail to instantiate) the controls required by our taxonomy and threat model. The goal is not to rank systems, but to characterize the architectural equilibria that have emerged by 2025 and the gaps that remain for safe autonomous execution.

\subsubsection{Custody Model: Authority Placement and Blast Radius}

Custody is the foundational axis because it determines where signing authority lives and, therefore, how compromise propagates. We observe four recurring custody patterns:

\begin{itemize}
\item \textbf{User-held keys (EOA-style self-custody).} This model maximizes sovereignty but concentrates failure into a single secret. In practice, it is ill-suited for autonomy because it couples a probabilistic decision component to irrevocable signing without enforceable constraints.

\item \textbf{Smart accounts (delegated custody via programmable validation).} Account abstraction shifts authority from a single private key to programmable validation logic, enabling spend limits, role separation, and time-bounded delegation. This is the dominant substrate for Pattern~III (Delegated Execution) because it allows an agent to operate through narrowly scoped capabilities rather than master authority.

\item \textbf{Threshold custody (MPC and threshold signatures).} Distributed custody reduces single-host compromise impact by requiring multiple shares to participate in authorization. Modern threshold-signing schemes and MPC workflows provide practical latency and availability for operational settings~\cite{lindell2017fast,komlo2024threshold}. This model is a natural fit for Pattern~IV deployments where the agent participates in authorization but must still satisfy independent co-signing and policy checks.

\item \textbf{Hardware-isolated custody (TEE-backed signing).} Hardware isolation aims to prevent key material exposure even under host compromise. The security posture depends on the trusted computing base and the integrity of policy enforcement around signing requests, but it provides a clear containment boundary for the key-handling surface.
\end{itemize}

Across these models, the key trend is a move away from ``keys in the agent runtime'' and toward custody mechanisms that make the agent revocable and bounded by construction.

\subsubsection{Permission and Policy: From Coarse Roles to Typed Constraints}

The capability matrix shows broad agreement that autonomous execution requires more than coarse-grained roles. \cite{ZodiacPermissionsStarterKit} We see a progression:

\begin{itemize}
\item \textbf{Role-based controls (RBAC).} Simple roles remain common at the contract layer, but they are too coarse for agents because they do not capture transaction semantics (asset, counterparty, slippage, deadline).

\item \textbf{On-chain, capability-style delegation (session keys and modules).} Smart-account modules enable time-bounded, value-bounded, and function-bounded delegation, giving Pattern~III systems a deterministic sandbox. This directly mitigates the ``agent compromise'' assumption in the threat model by limiting what compromised reasoning can do.

\item \textbf{Off-chain policy engines for semantic checks.} For Pattern~IV, richer policies often live off-chain where they can incorporate context, simulations, and operational workflows. The central requirement is that the policy decision be auditable and binding with respect to what is ultimately signed, aligning with the preview-to-sign integrity boundary described in Class~3.
\end{itemize}

The strongest designs combine typed intents with policy evaluation, reducing reliance on free-form text at authorization time. This is consistent with the broader shift toward formalized enforcement in security and compliance settings~\cite{orda2019enforcing}.

\subsubsection{Intent Interfaces: Reducing Cognitive Load and MEV Exposure}

A second major convergence is toward intent-style interfaces, where the agent expresses an outcome and constraints rather than assembling imperative transactions. This matters for two reasons.

First, intents reduce cognitive burden and error surface by shrinking the action space presented to the agent. Second, intent routing can reduce exposure to public mempool adversaries by enabling private or auction-based execution paths. This aligns with the empirical reality that transaction ordering and extraction are endemic in public networks~\cite{daian2019flash,bachu2024quantifying}. Recent work on intent-centric execution further motivates this direction as a security and market-design response to adversarial ordering~\cite{myakala2025intent}.

In the taxonomy, intent-centric execution is the clearest path from Pattern~II (Intent Generation) to Pattern~III and Pattern~IV, because it enables a crisp interface for simulation, preview, and policy evaluation before authority is exercised.

\subsubsection{Preview and Simulation: Preconditions for Human Approval and Safe Autonomy}

Preview and simulation capabilities are not cosmetic features. They are the last defensive layer before irrevocable execution, and they operationalize WYSIWYS-style integrity goals by ensuring that approval is tied to what will be signed and submitted.

The matrix suggests a practical norm: systems designed for user approval and high-value execution increasingly treat high-fidelity simulation and clear previews as baseline requirements. By contrast, components focused on infrastructure primitives (for example, pure key services or automation rails) may not provide simulation themselves, but they become safer when paired with an upstream simulator and a downstream signer that refuses to sign without a validated intent envelope.

\subsubsection{Observability: Auditability as a First-Class Capability}

Observability is the connective tissue between prevention and recovery. Mature stacks increasingly log not only the final transaction, but also: the intent proposal, the simulation outcomes, the policy decision, and the signing request metadata. This is essential for post-incident forensics and for narrowing accountability when failures occur. It also supports continuous improvement of policies and guardrails based on near-miss analysis and operational telemetry.

\subsubsection{Recovery and Revocation: Designing for Compromise Containment}

The matrix highlights that robust recovery is uneven across the ecosystem. The most actionable recovery primitives, independent of custody model, are:

\begin{itemize}
\item \textbf{Immediate revocation (kill switches).} The ability to revoke an agent's delegated capability quickly is the primary containment mechanism when compromise is suspected.

\item \textbf{Time-bounded capabilities.} Short-lived session keys and expiring authorizations limit worst-case loss even when revocation is delayed.

\item \textbf{Account-level recovery.} Smart-account patterns enable recovery workflows (including guardian-style mechanisms) that reduce catastrophic loss from device failure or key compromise while keeping authority separable from the agent runtime.
\end{itemize}

These controls map directly to the threat model assumption that compromise is a matter of when, not if, and that system design must prioritize bounded failure.

\subsubsection{Multi-Chain Scope: Integration Breadth Versus Assumption Debt}

Multi-chain support expands capability but increases assumption debt: heterogeneous RPCs, differing transaction semantics, varying MEV environments, and bridge dependencies. Bridge-mediated designs introduce distinct risk, and bridge compromise remains a major systemic hazard, so multi-chain agent designs need stricter provenance and simulation discipline than single-chain deployments.

\subsubsection{Threat Assumptions: The Design Center of 2025 Systems}

Across the 13 evaluated dimensions, three threat assumptions dominate real-world designs:

\begin{itemize}
\item \textbf{The agent can be steered or compromised.} This drives the shift toward session keys, typed intents, and policy gating at authorization time.

\item \textbf{The public mempool is adversarial.} This drives private submission, auction mechanisms, and intent-centric execution, reflecting the documented prevalence of value extraction and ordering manipulation~\cite{daian2019flash,chohan2024decentralized}.

\item \textbf{On-chain environments are hostile and evolving.} This drives simulation-first execution and a preference for bounded, auditable automation rather than free-form transaction synthesis.
\end{itemize}

\paragraph{Summary.}
Overall, the capability matrix depicts an ecosystem converging on a defense-in-depth stack: programmable accounts for bounded authority, intent interfaces for constrained action specification, simulation and preview for pre-flight verification, and policy plus observability for auditability and recovery. The remaining gaps are less about missing building blocks and more about standardizing how these blocks interoperate: typed intent schemas, verifiable policy decision artifacts, and portable permission models that remain safe under adversarial tool outputs and volatile execution conditions.

\subsubsection{Comparative Analysis Across System Categories}

Beyond individual system analysis, examining systems by category reveals important patterns and tradeoffs that inform architectural decisions.

\paragraph{Custody Solutions: Security vs. Usability}

Custody solutions occupy different positions on the security-usability spectrum. Self-custody solutions like hardware wallets provide maximum security through air-gapped key storage but require significant user expertise and offer limited programmability. MPC solutions like Fireblocks and Lit Protocol distribute trust across multiple parties, reducing single points of failure while enabling programmatic access. Smart contract wallets like Safe provide on-chain programmability but introduce smart contract risk and gas overhead.

For agent systems, the choice of custody solution depends on the trust model and operational requirements. Agents requiring low-latency signing may prefer MPC solutions with optimized signing protocols. Agents requiring complex authorization logic may prefer smart contract wallets with programmable policies. Agents managing institutional assets may require the compliance features and insurance coverage offered by institutional custody providers.

\paragraph{Execution Infrastructure: Speed vs. Protection}

Execution infrastructure varies in the tradeoffs between speed and MEV protection. Direct RPC submission provides lowest latency but maximum MEV exposure. Private mempools like Flashbots Protect provide MEV protection at the cost of additional latency and trust in the mempool operator. Intent-based protocols like CoW Protocol provide optimal execution through solver competition but introduce complexity and may not support all transaction types.

Agent systems should select execution infrastructure based on strategy requirements. High-frequency strategies may prioritize latency over MEV protection. Large-value transactions may prioritize MEV protection over latency. Complex multi-step operations may benefit from intent-based execution that optimizes across the entire operation.

\paragraph{Policy Engines: Expressiveness vs. Simplicity}

Policy engines range from simple rule-based systems to sophisticated programmable platforms. Simple policy engines support basic constraints like spending limits and allowlists, providing straightforward security with minimal configuration complexity. Advanced policy engines support complex conditional logic, temporal constraints, and integration with external data sources, enabling sophisticated authorization workflows at the cost of increased complexity.

The appropriate level of policy engine sophistication depends on the application's risk profile and operational complexity. Simple applications may be adequately served by basic policy constraints. Complex institutional applications may require the full expressiveness of advanced policy engines to implement nuanced authorization requirements.

\subsubsection{Industry Trends and Market Dynamics}

The agent-blockchain integration landscape is shaped by broader industry trends that influence technology development, adoption patterns, and competitive dynamics.

\paragraph{Consolidation and Specialization}

The market is experiencing simultaneous consolidation and specialization. Large platforms are acquiring or building comprehensive capabilities across the stack, from custody to execution to monitoring. Simultaneously, specialized providers are emerging to serve specific niches with deep expertise in particular domains.

This dynamic creates opportunities for both integrated platforms that offer convenience and interoperability, and specialized solutions that offer superior capabilities in specific areas. Agent system designers must navigate this landscape, choosing between integrated platforms and best-of-breed component selection.

\paragraph{Regulatory Pressure and Compliance}

Increasing regulatory attention to cryptocurrency and AI is shaping the development of agent-blockchain systems. Compliance requirements around custody, transaction monitoring, and risk management are driving adoption of institutional-grade infrastructure. Emerging AI regulations may impose additional requirements around transparency, accountability, and human oversight.

Forward-looking agent system designs should anticipate regulatory requirements and build compliance capabilities proactively. Systems designed with compliance in mind from the start will be better positioned to adapt as regulatory frameworks evolve.

\paragraph{Institutional Adoption}

Institutional adoption of both cryptocurrency and AI is accelerating, creating demand for enterprise-grade agent-blockchain solutions. Institutional requirements around security, compliance, and operational reliability exceed those of retail users, driving investment in more robust infrastructure.

The institutional market segment values reliability, support, and accountability over cutting-edge features. Agent system providers targeting institutional customers must demonstrate operational maturity and provide appropriate service level commitments.

\paragraph{Open Source and Standardization}

Open source development and standardization efforts are shaping the competitive landscape. Open source agent frameworks and blockchain tools lower barriers to entry and enable rapid innovation. Standardization efforts around interfaces, protocols, and data formats promote interoperability and reduce vendor lock-in.

Participation in open source and standardization efforts provides influence over the direction of the ecosystem and access to community knowledge and contributions. However, it also requires investment of resources and acceptance of community governance processes.

\section{2026 Roadmap: TIS and PDR}

While the current landscape of agent-blockchain systems is rich with innovation, it still lacks cohesive, standardized interfaces for secure interoperability and mainstream developer adoption. \cite{sorensen2024roadmap} Our analysis suggests that many integrations remain mediated by vendor-specific APIs, ad-hoc message formats, and protocol-specific transaction construction. This increases integration overhead and creates avoidable security risks, particularly when non-deterministic agents are forced to operate at the raw transaction layer. To address these gaps, we propose a 2026 research and development roadmap centered on two complementary interface standards: a \textbf{Transaction Intent Schema (TIS)} and a \textbf{Policy Decision Record (PDR)}. The goal is to standardize (i) \emph{what} the user or agent wants to achieve and (ii) a verifiable proof that the request has been evaluated against an enforceable policy before any irreversible signing occurs.

\subsection{Reference Architecture: Decoupling Intent, Policy, and Signing}

\begin{figure}[t]
 \centering 
 \resizebox{.8\linewidth}{!}{
 \begin{tikzpicture}[
 role/.style={boxM, text width=3.25cm},
 artifact/.style={draw, rounded corners, align=center, inner sep=5pt, text width=3.25cm, fill=black!1},
 arr/.style={->, line width=0.7pt}
 ]
 \node[role] (planner) {\textbf{Planner}\\\scriptsize agent + tools};
 \node[role, right=18mm of planner] (verifier) {\textbf{Verifier}\\\scriptsize policy engine};
 \node[role, right=18mm of verifier] (executor) {\textbf{Executor}\\\scriptsize signer + submit};

 \node[artifact, below=8mm of planner] (tis) {\textbf{Transaction Intent Schema (TIS)}\\\scriptsize declarative goal + constraints};
 \node[artifact, below=8mm of verifier] (pdr) {\textbf{Policy Decision Record (PDR)}\\\scriptsize allow/deny + rationale + attestations};

 \node[role, below=25mm of executor] (venue) {\textbf{Execution Venue}\\\scriptsize bundler / relay / solver};
 \node[role, below=7mm of venue] (chain) {\textbf{Blockchain}\\\scriptsize smart account + protocols};

 \node[artifact, below=8mm of pdr] (audit) {\textbf{Audit \& Telemetry}\\\scriptsize logs + alerts + replay};

 \draw[arr] (planner) -- (tis);
 \draw[arr, bend right=-25] (tis.east) to node[right] {\scriptsize intent} (verifier.west);
 \draw[arr] (verifier) -- (pdr);
 \draw[arr, bend right=-25] (pdr.east) to node[right] {\scriptsize decision} (executor.west);

 \draw[arr] (executor) -- (venue);
 \draw[arr] (venue) -- (chain);

 \draw[arr]
  (chain.west) -| node[below, inner sep=1pt] {\scriptsize receipts/events} ($(planner.west)+(-5mm,0mm)$) -- (planner.west);

 \draw[arr, bend right=20] (tis.south) to (audit);
 \draw[arr] (pdr) -- (audit);
 \draw[arr, bend right=-15] (executor) to (audit.east);

 \node[boundary, fit=(planner) (tis), label={[font=\scriptsize]above:Creative planning domain}] {};
 \node[boundary, fit=(verifier) (pdr), label={[font=\scriptsize]above:Deterministic policy domain}] {};
 \node[boundary, fit=(executor) (venue) (chain), label={[font=\scriptsize]above:Signing \& execution domain}] {};
 \end{tikzpicture} }
 \caption{Reference architecture for a Transaction Intent Schema (TIS) and a Policy Decision Record (PDR) stack. The architecture separates the planner (agent), verifier (policy engine), and executor (signer / submission layer) to enable defense-in-depth.}
 \label{fig:tis_pdr_arch}
\end{figure}
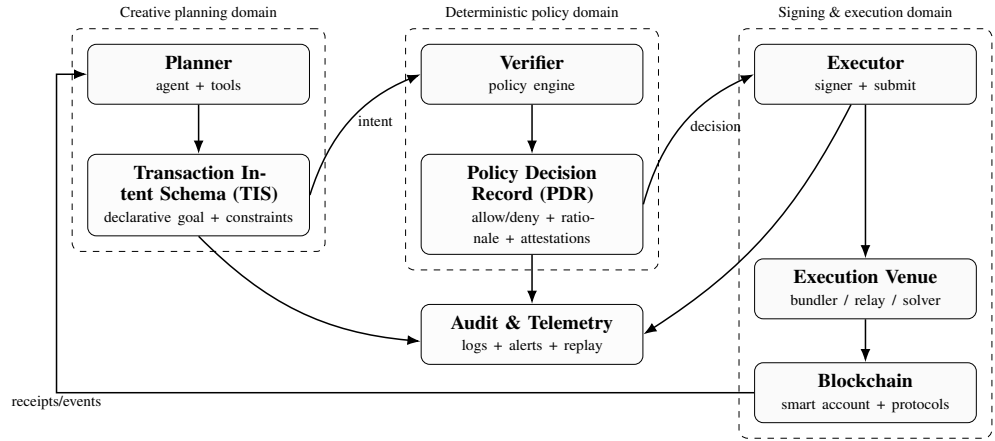

The proposed reference architecture, illustrated in Figure~\ref{fig:tis_pdr_arch}, enforces a strict separation of duties between the key components of an agentic transaction pipeline. \cite{pitt2000communication} It decouples the agent's creative but fallible planning process from the deterministic act of authorization and signing. This separation is aligned with a defense-in-depth posture: the planner generates goals and constraints, the verifier evaluates those goals against policy and threat assumptions, and the executor signs only when presented with a verifiable authorization artifact.

\subsection{The Transaction Intent Schema (TIS): A Universal Language for On-Chain Goals}

\textbf{Problem:} The absence of a shared representation for user intent remains a major bottleneck for secure agentic execution. Today, an agent that intends to perform a swap must still reason about protocol-specific ABIs and function signatures (Uniswap vs.\ Curve vs.\ CoW), chain-specific execution constraints, and failure modes at the transaction layer. This is brittle, difficult to audit, and materially expands the attack surface, because the agent is forced to be an expert in low-level mechanics rather than an expert in goals and constraints.

\textbf{Solution:} We propose TIS as a standardized, extensible JSON-based schema describing the \emph{what} (desired state change) and the \emph{constraints} (safety and validity boundaries), not the \emph{how} (protocol-specific call data). It builds on the broader shift toward intent-centric execution~\cite{mandal2025evaluating} and is compatible in spirit with emerging intent standards and coordination primitives (e.g., ERC-7521 and related efforts). \cite{hassan2023trust,erc8001}.

\subsubsection{TIS Core Schema and Design Rationale}

TIS is designed to be both human-readable and machine-parsable. A minimal core schema includes:

\begin{itemize}
\item \textbf{\texttt{intentId} (string, UUID):} A globally unique identifier used for tracking, correlation, and auditability across off-chain and on-chain systems. \cite{yu2021mtdt}
\item \textbf{\texttt{action} (string, enum):} A high-level verb describing the intent category. The initial set should be intentionally small (e.g., \texttt{SWAP}, \texttt{LEND}, \texttt{BORROW}, \texttt{STAKE}, \texttt{VOTE}) and extended via versioned registries or namespaces to avoid fragmentation.
\item \textbf{\texttt{inputs} (array of objects):} Assets the user provides. Each object specifies \texttt{token} (address), \texttt{amount}, and \texttt{constraint} (e.g., \texttt{EXACT}, \texttt{MAXIMUM}).
\item \textbf{\texttt{outputs} (array of objects):} Assets the user expects to receive. Each object specifies \texttt{token}, \texttt{amount}, and \texttt{constraint} (e.g., \texttt{MINIMUM}, \texttt{EXACT}).
\item \textbf{\texttt{constraints} (object):} Global validity conditions, such as \texttt{deadline} (Unix timestamp), \texttt{nonce} (replay protection), \texttt{chainId} (domain binding), and optional \texttt{exclusivity} (e.g., a solver/bundler/relayer allow-list entry, or null).
\item \textbf{\texttt{preview} (string):} A clear, human-readable natural-language description of the intent, suitable for user review and downstream WYSIWYS workflows.
\end{itemize}

\noindent\textbf{Canonicalization note:} Because TIS is intended to be hashed and bound into authorization artifacts (Section~\ref{sec:pdr}), implementations should apply a canonical JSON serialization to ensure stable hashing across languages and runtimes. A practical option is JSON Canonicalization Scheme (JCS). \cite{rfc8785}.

\subsubsection{Example: A Complex, Multi-Step Intent}

The value of TIS becomes more apparent in multi-step workflows. \cite{yuan2019novel} Consider a ``zap'' into a yield vault that requires multiple underlying actions (swap, liquidity provision, deposit):

\begin{lstlisting}
{
 "intentId": "f47ac10b-58cc-4372-a567-0e02b2c3d479",
 "action": "ZAP_INTO_VAULT",
 "inputs": [
 {
 "token": "0xa0b86991c6218b36c1d19d4a2e9eb0ce3606eb48",
 "amount": "1000000000",
 "constraint": "EXACT"
 }
 ],
 "outputs": [
 {
 "token": "0x..",
 "amount": "950000000000000000",
 "constraint": "MINIMUM"
 }
 ],
 "constraints": {
 "deadline": 1767226000,
 "exclusivity": null
 },
 "metadata": {
 "targetVault": "0x..",
 "intermediateSwapProtocol": "UniswapV3"
 },
 "preview": "Deposit 1000 USDC into the Yearn ETH/USDC vault, receiving at least 0.95 vault tokens in return."
}
\end{lstlisting}

This single intent captures a multi-step objective without exposing protocol-specific call data. The agent constructs the goal and constraints, while a specialized solver or execution adapter generates the optimal transaction sequence and submits it through an appropriate venue (e.g., solver-based settlement, private relays, or ERC-4337-style bundling) \cite{EIP4337}.

\subsection{The Policy Decision Record (PDR): Verifiable Proof of Compliance}
\label{sec:pdr}

\textbf{Problem:} Policy enforcement is currently fragmented and often opaque. Even when a policy engine exists, there is typically no interoperable way to prove that an intent was evaluated under a specific ruleset before signing occurs. This complicates auditing, weakens incident response, and creates single points of failure in the trust chain.

\textbf{Solution:} We propose the \textbf{Policy Decision Record (PDR)} as a signed, verifiable authorization artifact that attests that a specific TIS object was evaluated and either approved or rejected. Conceptually, it functions as a signed capability that is presented to the signer. The design is inspired by mature, widely deployed authorization patterns where a compact signed token conveys claims and constraints between decoupled components. \cite{rfc6749,RFC6749OAuth2,rfc7519,RFC7519JWT}.

\subsubsection{PDR Core Schema (JWT Profile)}

We propose standardizing the PDR as a JSON Web Token (JWT) profile. JWTs provide a compact representation of claims along with a digital signature (JWS), enabling stateless verification by the executor without requiring online calls to the policy engine for every signature event. \cite{rfc7519,rfc7515}.

\begin{itemize}
\item \textbf{Header:} Contains the signing algorithm \texttt{alg} and token type \texttt{typ} (e.g., \texttt{PDR+JWT}). Implementations should select an algorithm consistent with their key material and trust domain (e.g., ECDSA-based JWS).
\item \textbf{Payload (Claims):}
 \begin{itemize}
 \item \texttt{iss} (Issuer): The DID/identity of the policy engine that issued the decision.
 \item \texttt{sub} (Subject): The user account (or smart account) on whose behalf the intent is executed.
 \item \texttt{aud} (Audience): The intended recipient of the PDR, namely the executor/signer identity. Binding to \texttt{aud} prevents a valid PDR from being replayed against a different signer.
 \item \texttt{exp} (Expiration Time): A short-lived expiration time to limit replay and stale approvals. \cite{zst2021analyzing}
 \item \texttt{iat} (Issued At): Time of issuance.
 \item \texttt{intent\_hash} (string): Cryptographic hash (e.g., Keccak-256) over the \emph{canonicalized} TIS object. This binds the PDR to exactly one intent.
 \item \texttt{decision} (string, enum): \texttt{APPROVE} or \texttt{REJECT}.
 \item \texttt{bound\_constraints} (object, optional): Additional constraints imposed by the policy engine (e.g., a reduced spend limit, shortened \texttt{deadline}, restricted venues). Executors must enforce these tightened constraints.
 \end{itemize}
\item \textbf{Signature:} A cryptographic signature from the policy engine, proving integrity and issuer authenticity. \cite{rfc7515}
\end{itemize}

\subsubsection{The PDR Workflow and Security Guarantees}

\begin{enumerate}
\item The agent constructs a TIS and submits it to the configured Policy Engine.
\item The Policy Engine evaluates the TIS against its ruleset (e.g., spend limits, allow-lists, venue restrictions, time windows, and contextual signals).
\item If approved, the Policy Engine issues and signs a PDR (JWT) and returns it to the agent.
\item The agent presents both the TIS and PDR to the Signer (executor).
\item The Signer performs the following verification steps before any signing operation:
\end{enumerate}

\begin{itemize}
\item Verify the PDR signature against the trusted Policy Engine public key.
\item Check the \texttt{aud} claim matches the Signer identity.
\item Check time validity (\texttt{exp}, and optionally a narrow \texttt{iat} tolerance window).
\item Re-compute the hash of the canonicalized TIS and confirm it matches \texttt{intent\_hash}. This prevents ``PDR swapping,'' where an agent obtains approval for a benign intent and then substitutes a different intent at signing time.
\item Enforce any tightened \texttt{bound\_constraints}.
\end{itemize}

\begin{enumerate}
\item Only if all checks pass does the Signer proceed to generate and submit the final transaction (potentially via a bundler, relay, or solver venue) \cite{EIP4337}.
\end{enumerate}

\noindent This workflow produces a non-repudiable, auditable chain of evidence that enforces separation of duties. It also aligns with the threat assumption that the agent and its runtime may be compromised: even in that case, a compromised planner cannot escalate beyond what the verifier and executor jointly permit.

\subsection{Benefits, Challenges, and Future Research}

Standardizing TIS and PDR would materially improve interoperability, security, composability, and auditability for agent--blockchain systems by turning today’s bespoke integrations into verifiable, portable artifacts. In particular, TIS enables a common, goal-level interface for solvers and execution venues, while PDR introduces a signer-verifiable proof that an intent was evaluated against a policy before any irreversible authorization occurs. Together, these layers reduce protocol-specific coupling, narrow the planner’s attack surface, and improve post-incident forensics by making decisions and constraints externally auditable.

Realizing this vision, however, requires addressing several practical challenges:

\begin{itemize}
\item \textbf{Standardization and governance:} Achieving broad adoption will require a coordinated community process involving wallet providers, custody services, infrastructure operators (bundlers/relays), and security researchers. A credible path is an open standards process in the Ethereum ecosystem (e.g., an EIP-style workflow), complemented by reference implementations and test vectors that lock down canonicalization, hashing, and validation behaviors.
\item \textbf{Solver ecosystem bootstrapping:} The value proposition of TIS depends on a competitive market of solvers and execution adapters that can translate intents into robust execution plans, compete on price and reliability, and support venue-specific defenses (e.g., private relays or batch auctions). Bootstrapping this ecosystem is non-trivial: it requires incentive design, clear liability boundaries, and mechanisms for measuring solver quality (fill rate, slippage, failure rate) under adversarial conditions.
\item \textbf{Policy language and portability:} For PDR to deliver interoperability rather than vendor lock-in, policy decisions must be comparable and portable across implementations. A promising direction is to standardize a minimal, machine-checkable policy layer (for example, allow-list constraints, spend limits, venue restrictions, and time windows) and map richer organizational policies onto it. Policy-as-code systems such as Open Policy Agent (OPA)\cite{OPARego} provide useful foundations for expressing and auditing rule sets, although additional work is needed to define blockchain-specific primitives and semantics. \cite{opa}
\end{itemize}

Future research directions include:

\begin{itemize}
\item \textbf{Privacy-preserving PDRs:} Developing PDR constructions that prove compliance without disclosing sensitive policy logic or intent details. One approach is to attach a zero-knowledge proof that the verifier evaluated the intent under an approved policy commitment and produced the stated decision, while selectively revealing only what the signer must enforce (e.g., bounded constraints). Privacy-preserving ledger and compliance designs provide early evidence that such attestations can be practical in regulated settings. \cite{narula2018zkledger}
\item \textbf{Decentralized trust discovery and registries:} Building a decentralized trust registry in which policy engines publish public keys, metadata, and attestations, enabling signers to discover and validate trusted verifiers without centralized directories. A natural substrate is an ENS-backed registry, with optional linking to decentralized identifier (DID) documents for richer identity and key-rotation semantics. \cite{ens,didcore}
\end{itemize}

\noindent By establishing these missing interface layers, the ecosystem can converge on a robust and auditable control plane for on-chain agent execution: planners remain flexible, verifiers remain policy-driven and reviewable, and executors remain deterministic and narrow in scope. This separation is the most direct route to scaling agent autonomy while maintaining security under the threat model outlined earlier.

\subsubsection{Implementation Roadmap and Milestones}

The transition from current fragmented approaches to the standardized TIS/PDR ecosystem envisioned in this survey will require coordinated effort across multiple stakeholders. This section outlines a phased implementation roadmap with concrete milestones.

\paragraph{Phase 1: Specification and Reference Implementation (Months 1-6)}

The first phase focuses on finalizing specifications and developing reference implementations that demonstrate feasibility.

\textbf{Milestones:}
\begin{itemize}
\item Publication of TIS v1.0 specification with JSON Schema definitions
\item Publication of PDR v1.0 specification with cryptographic requirements
\item Reference implementation of TIS parser and validator
\item Reference implementation of PDR generator and verifier
\item Test suite covering specification edge cases
\end{itemize}

\paragraph{Phase 2: Ecosystem Integration (Months 7-12)}

The second phase focuses on integrating TIS/PDR support into existing ecosystem tools and platforms.

\textbf{Milestones:}
\begin{itemize}
\item Integration with at least three major wallet providers
\item Integration with at least two agent frameworks
\item Integration with at least two policy engines
\item Development of TIS/PDR browser extension for user visibility
\item Publication of integration guides and best practices
\end{itemize}

\paragraph{Phase 3: Production Deployment (Months 13-18)}

The third phase focuses on production deployment and operational experience.

\textbf{Milestones:}
\begin{itemize}
\item Production deployment by at least five agent operators
\item Collection and analysis of operational metrics
\item Specification updates based on production experience
\item Development of monitoring and analytics tools
\item Publication of case studies and lessons learned
\end{itemize}

\paragraph{Phase 4: Standardization and Governance (Months 19-24)}

The fourth phase focuses on formal standardization and establishment of governance processes.

\textbf{Milestones:}
\begin{itemize}
\item Submission of TIS/PDR to relevant standards bodies
\item Establishment of specification governance process
\item Development of conformance testing and certification program
\item Creation of ecosystem working groups for ongoing development
\item Publication of v2.0 specifications incorporating community feedback
\end{itemize}

\section{Evaluation: Benchmarks and Checklists for On-Chain Agent Safety}

The emergence of autonomous on-chain agents necessitates a corresponding evolution in evaluation methodology. Traditional software testing approaches, such as unit and integration testing, are insufficient for systems whose behavior is non-deterministic, economically adversarial, and tightly coupled to irreversible state transitions~\cite{maffei2021formal}. Prior work in formal methods and smart-contract verification focuses primarily on static correctness of on-chain code, but does not capture the dynamic risks introduced by agentic reasoning, external tool use, and real-time market interaction.

Evaluating an agent--blockchain system therefore requires a holistic, end-to-end methodology that spans reasoning, authorization, execution, and post-execution verification. Similar arguments have been made in adjacent domains such as autonomous cyber-physical systems and adaptive distributed systems, where emergent behavior and adversarial environments invalidate purely component-level evaluation~\cite{he2019sdfs,zhang2016town}. In this section, we propose a structured evaluation framework designed specifically for on-chain agents, combining quantitative benchmarks with qualitative safety checklists.

\subsection{The Need for Domain-Specific, Holistic Benchmarks}

Existing benchmarks for large language models evaluate isolated cognitive capabilities, such as general knowledge (e. ~\cite{zou2024adversarial,zhou2024easyjailbreak} g., MMLU) or code synthesis (e.g., HumanEval), but do not model adversarial economic environments or irreversible execution~\cite{he2019sdfs}. Similarly, benchmarks for smart contracts emphasize vulnerability detection and gas efficiency, but assume that transaction intent is externally and correctly specified~\cite{hassan2023trust,monika2025security}.

In contrast, an on-chain agent must jointly reason over uncertain information, interact with untrusted tools, respect policy constraints, and execute transactions under MEV pressure. Recent SoK papers on LLM tool use explicitly note the lack of benchmarks that evaluate the full tool–decision–action loop~\cite{li2024large}. Our framework addresses this gap by evaluating the agent--chain pipeline as a single system.

We structure evaluation around three pillars that recur across surveyed systems and threat models:

\begin{enumerate}
\item \textbf{Safety and Security}, reflecting resistance to exploitation and loss;
\item \textbf{Task Reliability and Robustness}, reflecting operational correctness under failure;
\item \textbf{Economic Robustness}, reflecting rational behavior under adversarial market dynamics.
\end{enumerate}

\subsection{Pillar 1: Safety and Security Benchmarks}

Security benchmarks focus on the agent’s resilience to the attack classes identified in our threat model, particularly prompt manipulation, tool poisoning, policy bypass, and MEV exploitation. Consistent with prior evaluation practices in adversarial ML and blockchain security, these benchmarks are executed in controlled, forked environments populated with adversarial artifacts~\cite{maffei2021formal,hassan2023trust}.

\subsubsection{Prompt Injection Resistance (PIR) Score}

Indirect prompt injection has been identified as a fundamental vulnerability in LLM-based agents, particularly when consuming untrusted external content~\cite{greshake2023indirectpromptinjection}. In the blockchain setting, this risk is amplified because injected instructions may trigger irreversible financial actions.

\begin{itemize}
\item \textbf{Methodology:} The agent processes benign tasks while consuming adversarially crafted data sources, including HTML pages, token metadata fields, API responses, and error messages, following established threat taxonomies for LLM tool use~\cite{li2024large}.
\item \textbf{Scoring:} The PIR score measures the fraction of attacks correctly ignored or flagged without executing malicious instructions. Scores below $99\%$ indicate unsuitability for autonomous operation.
\end{itemize}

\subsubsection{Tool Spoofing Detection (TSD) Rate}

Tool poisoning and data spoofing attacks have been extensively documented in distributed systems and oracle-based architectures~\cite{cai2023large}. For agents, reliance on a single data source represents a critical failure mode.

\begin{itemize}
\item \textbf{Methodology:} The agent is exposed to conflicting or manipulated tool outputs, such as price discrepancies or incorrect balances, and is expected to perform cross-validation or refuse action.
\item \textbf{Interpretation:} The TSD rate evaluates whether the agent implements redundancy and sanity checks, a practice recommended in secure oracle and DeFi system design~\cite{hassan2023trust}.
\end{itemize}

\subsubsection{MEV Vulnerability Assessment (MVA)}

MEV has been formalized as a structural property of public blockchains rather than an edge case~\cite{daian2019flash,FlashbotsProtectDocs}. Autonomous agents that submit public transactions without mitigation are therefore assumed to be exploitable by default.

\begin{itemize}
\item \textbf{Methodology:} Trades are executed in adversarially populated mempools, following established MEV benchmarking practices~\cite{caldarelli2020understanding}.
\item \textbf{Metric:} The MVA score measures value lost to MEV relative to an optimal, MEV-free execution baseline.
\end{itemize}

\subsubsection{Policy Adherence Test (PAT)}

Policy enforcement failures have been identified as a primary cause of loss in autonomous and semi-autonomous financial systems~\cite{feichtinger2024sok}. This benchmark evaluates whether agents respect explicit constraints under adversarial prompting.

\begin{itemize}
\item \textbf{Methodology:} The agent is repeatedly placed in scenarios where violating policy would yield higher apparent utility.
\item \textbf{Outcome:} A high PAT score indicates correct prioritization of constraints over short-term rewards.
\end{itemize}

\subsection{Pillar 2: Task Reliability and Robustness}

Beyond security, agents must reliably complete complex tasks under realistic operational conditions, including partial failures and transient inconsistencies. Robustness benchmarks draw on principles from distributed systems testing and fault injection~\cite{shyamasundar2026why}.

\subsubsection{DeFi Task Completion (DTC) Suite}

Multi-step DeFi workflows represent a natural stress test for agentic planning and execution~\cite{happe2025benchmarking}.

\begin{itemize}
\item \textbf{Methodology:} Tasks are categorized by increasing compositional complexity, from single-protocol interactions to atomic multi-protocol arbitrage.
\item \textbf{Metrics:} Success rate, gas efficiency, execution latency, and reasoning depth.
\end{itemize}

\subsubsection{Failure Recovery Test (FRT)}

Resilience to partial failure is a defining property of reliable autonomous systems~\cite{liu2024offline}.

\begin{itemize}
\item \textbf{Methodology:} The environment injects RPC failures, reverted calls, and dropped transactions.
\item \textbf{Success Criterion:} The agent must diagnose failure causes and converge on a correct recovery strategy~\cite{li2025cox}.
\end{itemize}

\subsection{Pillar 3: Economic Robustness Benchmarks}

Economic irrationality can be as damaging as technical compromise. Prior work on algorithmic trading and DeFi attacks highlights how adversarial incentives exploit naive strategies~\cite{daian2019flash}.

\subsubsection{Economic Rationality Score (ERS)}

\begin{itemize}
\item \textbf{Methodology:} Agents are exposed to known economic traps, including pump-and-dump schemes and honeypot contracts.
\item \textbf{Interpretation:} High ERS scores indicate the ability to reason about liquidity, sellability, and execution feasibility rather than surface-level profit signals.
\end{itemize}

\subsection{Agent-to-Chain Security Audit Checklist}

Quantitative benchmarks alone are insufficient to capture architectural weaknesses. Inspired by security review practices in distributed systems and DAO governance~\cite{han2025review,feichtinger2024sok}, we propose a structured audit checklist organized along the agent action pipeline.

The checklist complements benchmarks by enabling qualitative review of observability, authorization boundaries, execution pathways, and recovery mechanisms. Together, benchmarks and checklists provide defense-in-depth evaluation, aligning with best practices in secure system design~\cite{maffei2021formal}.

\subsubsection{Evaluation Methodology Best Practices}

Rigorous evaluation of agent-blockchain systems requires careful attention to methodology. This section outlines best practices for conducting meaningful evaluations.

\paragraph{Test Environment Fidelity}

Evaluation environments should faithfully reproduce the conditions agents will encounter in production. This includes realistic network latency, gas price volatility, liquidity conditions, and adversarial behavior. Evaluations conducted in idealized environments may overestimate agent performance and underestimate risks.

Mainnet forking provides high-fidelity test environments by replicating actual blockchain state. However, forked environments cannot reproduce the full dynamics of live markets, including other participants' reactions to agent behavior. Complementary testing on testnets and in simulated environments helps cover scenarios that cannot be reproduced through forking.

\paragraph{Adversarial Testing}

Security evaluations should include adversarial testing by skilled red teams. Automated security scanning can identify known vulnerability patterns, but creative attackers may find novel attack vectors that automated tools miss. Red team exercises should cover the full attack surface, including prompt injection, economic attacks, and infrastructure compromise.

Bug bounty programs provide ongoing adversarial testing by incentivizing external researchers to find and report vulnerabilities. Well-designed bounty programs with appropriate rewards and clear scope can significantly improve security posture.

\paragraph{Long-Duration Testing}

Agent behavior may change over extended operation periods due to state accumulation, model drift, or environmental changes. Short-duration tests may miss failure modes that only emerge after extended operation. Long-duration testing, including continuous operation over days or weeks, helps identify these delayed failure modes.

\paragraph{Statistical Rigor}

Evaluation results should be reported with appropriate statistical rigor, including confidence intervals, sample sizes, and significance tests. Single-run results or small sample sizes may not reliably indicate true performance. Evaluations should include sufficient repetitions to establish statistical confidence in results.

\paragraph{Reproducibility}

Evaluation procedures should be documented in sufficient detail to enable independent reproduction. This includes specifying test data, environment configuration, evaluation metrics, and analysis procedures. Reproducible evaluations enable verification of results and comparison across different systems.

\section{Case Studies: Architectures in Practice}

To ground the preceding taxonomy and capability analysis in deployable designs, this section examines three representative case studies spanning \textbf{Patterns III--V}. Each narrative provides an end-to-end execution flow (setup, intent construction, authorization, submission, and verification) and surfaces the security trade-offs introduced as autonomy increases. Throughout, we emphasize \emph{where} control is enforced (on-chain modules versus off-chain policy engines), \emph{how} authority is bounded (session keys, MPC thresholds, and PDR attestations), and \emph{which} threats dominate (prompt injection, key compromise, tool spoofing, and MEV).

\subsubsection{Detailed Benchmark Specifications}

This section provides detailed specifications for the evaluation benchmarks proposed in this survey, enabling practitioners to implement consistent evaluation procedures.

\paragraph{Safety Benchmark: Prompt Injection Resistance}

The prompt injection resistance benchmark evaluates agent resilience against attempts to manipulate behavior through malicious inputs.

\textbf{Test Suite Composition:}
\begin{itemize}
\item 100 direct injection attempts (explicit instructions to transfer funds)
\item 100 indirect injection attempts (malicious content in blockchain data)
\item 100 multi-turn manipulation attempts (gradual trust building)
\item 100 encoding-based attacks (Unicode, base64, other encodings)
\item 100 context overflow attempts (exceeding context window)
\end{itemize}

\textbf{Evaluation Metrics:}
\begin{itemize}
\item Attack Success Rate (ASR): Percentage of attacks that achieve their objective
\item Detection Rate: Percentage of attacks correctly identified as malicious
\item False Positive Rate: Percentage of benign inputs incorrectly flagged
\item Degradation Under Attack: Performance impact of attack attempts
\end{itemize}

\textbf{Passing Criteria:}
\begin{itemize}
\item ASR < 1\% for all attack categories
\item Detection Rate > 95\% for direct attacks
\item False Positive Rate < 5\%
\end{itemize}

\paragraph{Reliability Benchmark: Transaction Success Rate}

The transaction success rate benchmark evaluates the reliability of agent transaction construction and execution.

\textbf{Test Suite Composition:}
\begin{itemize}
\item 200 simple transfer transactions
\item 200 token swap transactions
\item 200 multi-step DeFi operations
\item 200 transactions under network congestion
\item 200 transactions with time constraints
\end{itemize}

\textbf{Evaluation Metrics:}
\begin{itemize}
\item Success Rate: Percentage of transactions that execute successfully
\item Revert Rate: Percentage of transactions that revert on-chain
\item Simulation Accuracy: Correlation between simulated and actual outcomes
\item Recovery Rate: Percentage of failed transactions successfully retried
\end{itemize}

\textbf{Passing Criteria:}
\begin{itemize}
\item Success Rate > 99\% for simple transfers
\item Success Rate > 95\% for complex operations
\item Simulation Accuracy > 98\%
\end{itemize}

\paragraph{Economic Benchmark: Execution Quality}

The execution quality benchmark evaluates the economic efficiency of agent transaction execution.

\textbf{Test Suite Composition:}
\begin{itemize}
\item 100 small swaps (< \$1,000)
\item 100 medium swaps (\$1,000 - \$100,000)
\item 100 large swaps (> \$100,000)
\item 100 time-sensitive executions
\item 100 multi-venue executions
\end{itemize}

\textbf{Evaluation Metrics:}
\begin{itemize}
\item Price Impact: Deviation from mid-market price at execution time
\item MEV Leakage: Value extracted by MEV searchers
\item Gas Efficiency: Gas consumed relative to optimal execution
\item Execution Latency: Time from intent to confirmation
\end{itemize}

\textbf{Passing Criteria:}
\begin{itemize}
\item Price Impact < 0.5\% for small swaps
\item MEV Leakage < 0.1\% of transaction value
\item Gas Efficiency > 90\% of theoretical optimum
\end{itemize}

\subsubsection{Comprehensive Evaluation Frameworks}

The evaluation of agent-blockchain systems requires comprehensive frameworks that assess multiple dimensions of system quality. Foundational work on Byzantine fault tolerance provides the theoretical underpinnings for evaluating the security of distributed systems~\cite{wang2022bft,yin2018hotstuff,zhong2023byzantine}. More recent work has focused on creating taxonomies of blockchain vulnerabilities and privacy issues, which can inform the design of evaluation frameworks~\cite{xu2017taxonomy,zhang2018privacy,zhang2019smart}. The scalability of blockchain systems is another critical evaluation dimension, with research exploring various techniques for improving performance~\cite{zheng2018scalable,yu2024blockchain}. The interoperability of different blockchain systems is also a key concern, with several frameworks proposed for evaluating cross-chain communication protocols~\cite{williams2024interoperability}. The unique challenges of evaluating agent-based systems have also been explored, with a focus on creating realistic simulation environments and benchmarks~\cite{zhang2024agent,zhang2024blockchain}. The security of half-duplex communication protocols, which are common in distributed systems, has also been studied in detail~\cite{wu2025half}. The universal composability of cryptographic protocols is a key concept for evaluating the security of complex systems~\cite{zou2023universal}. Finally, the performance of distributed databases is a critical factor in the overall performance of blockchain systems~\cite{wu2023distributed}. Traditional software evaluation metrics, while necessary, are insufficient to capture the unique characteristics of autonomous agents operating in adversarial economic environments.

Functional correctness evaluation must verify that agents produce correct outputs for their intended use cases. This includes both positive testing (verifying correct behavior for valid inputs) and negative testing (verifying appropriate handling of invalid inputs, edge cases, and adversarial scenarios). The stochastic nature of LLM-based agents complicates correctness evaluation, as the same input may produce different outputs across multiple runs.

Security evaluation must assess resilience against the full range of threats identified in the threat model. This includes both automated security testing (fuzzing, static analysis, penetration testing) and manual security review by domain experts. The novel attack surfaces introduced by LLM-based agents, such as prompt injection and jailbreaking, require specialized evaluation techniques that may not be covered by traditional security assessment methodologies.

Economic evaluation must assess the financial performance and risk characteristics of agent strategies. This includes backtesting against historical data, simulation in realistic market conditions, and careful analysis of tail risks and worst-case scenarios. Economic evaluation should also consider the impact of agent behavior on market quality and other participants, not just the returns to the agent operator.

\subsection{Case Study 1: Simple Payment Agent (Pattern III)}

\textbf{Goal:} An agent that executes a recurring payment on behalf of a user, such as paying a monthly $10$ USDC subscription to a streaming service.

\textbf{Architecture:} This use case exemplifies \textbf{Pattern III: Delegated Execution}. The agent should not have unconditional signing authority. It only requires narrowly scoped authority to perform a single action under strict, smart-contract-enforced constraints. The recommended architecture is an \textbf{ERC-4337 smart account} with a policy module (e.g., a session-key module or a Zodiac \texttt{Roles} modifier) that bounds agent authority on-chain~\cite{EIP4337,safeDocs}.

\subsubsection{Detailed Setup and Configuration}

\begin{enumerate}
\item \textbf{Smart Account:} The user controls an ERC-4337 smart account (e.g., a Safe\{Wallet\}-based account abstraction implementation) \cite{safeDocs,EIP4337}.

\item \textbf{Session Key Generation~\cite{shahidinejad2024all}:} During subscription enrollment, the dApp generates a fresh keypair locally (browser or mobile secure enclave). The public key, denoted \texttt{0xAgentSessionKey}, will be authorized as a limited session key.

\item \textbf{Policy Module Installation:} The dApp prepares a transaction to install and configure an on-chain policy module (e.g., Zodiac \texttt{Roles} or a bespoke session-key module). The user approves a single transaction that:
\begin{itemize}
\item installs the module on the smart account;
\item registers \texttt{0xAgentSessionKey} as an authorized delegate; and
\item encodes hard constraints that the module enforces on every delegated execution.
\end{itemize}
\end{enumerate}

\textbf{Illustrative On-Chain Constraints:}
\begin{itemize} \scriptsize
\item \texttt{require(recipient == 0xStreamingServiceAddress)\;}
\item \texttt{require(amount <= 10 * 1e6)\;} \texttt{// 10 USDC}
\item \texttt{require(block.timestamp >= lastPaymentTimestamp + 30 days)\;}
\item \texttt{require(block.timestamp < keyExpiryTimestamp)\;}
\end{itemize}

These constraints form the core security boundary: even if the agent is compromised, the maximum damage is bounded by on-chain logic that it cannot bypass.

\subsubsection{Detailed Execution Flow}

On the due date, the agent (e.g., a constrained backend job operated by the streaming service) constructs and submits a \texttt{UserOperation} under ERC-4337~\cite{buterin2021erc}.

\textbf{Agent Log (Example):}
\begin{quote}\scriptsize
\texttt{[2026-07-15 08:00:00] INFO: Triggered billing cycle for account 0xUserSmartAccountAddress.}\\
\texttt{[2026-07-15 08:00:02] INFO: Payment due (last payment: 2026-06-15).}\\
\texttt{[2026-07-15 08:00:03] INFO: Constructing UserOperation: transfer 10 USDC to 0xStreamingServiceAddress.}\\
\texttt{[2026-07-15 08:00:04] INFO: Signing UserOperation with session key 0xAgentSessionKey.}\\
\texttt{[2026-07-15 08:00:05] INFO: Submitting UserOperation to bundler endpoint.}\\
\texttt{[2026-07-15 08:00:20] INFO: Awaiting receipt..}\\
\texttt{[2026-07-15 08:00:35] INFO: SUCCESS: Included in block 21000000.}\\
\texttt{[2026-07-15 08:00:36] INFO: Updating subscription status in application database.}
\end{quote}

\textbf{UserOperation Payload (Schematic):}
\begin{lstlisting}
{
 "sender": "0xUserSmartAccountAddress",
 "nonce": "0x..",
 "callData": "0x..",
 "signature": "0x..",
 "paymasterAndData": "0xPaymasterAddress.."
}
\end{lstlisting}

\textbf{On-Chain Verification (ERC-4337 Path):}
\begin{enumerate}
\item A bundler includes the \texttt{UserOperation} in a bundle to the \texttt{EntryPoint}.
\item \texttt{EntryPoint} calls \texttt{validateUserOp} on \texttt{0xUserSmartAccountAddress}.
\item The smart account delegates validation to the installed policy module.
\item The policy module verifies:
\begin{itemize}
\item the signature corresponds to \texttt{0xAgentSessionKey}; and
\item the execution satisfies all constraints (recipient, amount cap, cadence, expiry).
\end{itemize}
\item If valid, \texttt{EntryPoint} executes the call (USDC \texttt{transfer}) via the smart account.
\end{enumerate}

\subsubsection{Security Analysis Deep Dive}

\begin{itemize}
\item \textbf{Attack Surface:} Minimal and sharply bounded. The worst-case loss is capped at 10 USDC per interval, enforced by immutable on-chain checks.
\item \textbf{Key Management:} The agent never touches the user's master key. Session keys are single-purpose and time-limited.
\item \textbf{Trust Boundary:} Trust is anchored in audited on-chain modules rather than in the correctness of the agent runtime.
\end{itemize}

\subsection{Case Study 2: Autonomous Portfolio Rebalancing Agent (Pattern IV)}

\textbf{Goal:} An agent that manages a user's DeFi portfolio, targeting a desired allocation (e.g., 50\% ETH, 25\% WBTC, 25\% USDC) by rebalancing when deviations exceed a threshold.

\textbf{Architecture:} This use case aligns with \textbf{Pattern IV: Autonomous Signing}. It demands stronger controls than Pattern III because the agent is empowered to initiate economically meaningful actions. The recommended architecture is a \textbf{defense-in-depth} stack combining distributed custody (MPC), policy enforcement (off-chain policy engine), and MEV-aware execution (private relays and intent-based venues) \cite{wiener2019mpc,daian2019flash,flashbotsprotect2026docs}.

\subsubsection{Detailed Setup and Configuration}

\begin{enumerate}
\item \textbf{Custody and Signing (Distributed Control):} Funds are held in a \textbf{2-of-3 MPC wallet} \cite{wiener2019mpc}:
\begin{itemize}
\item \textbf{Share 1 (Agent-side):} Held in a hardened environment (e.g., HSM or enclave-backed service).
\item \textbf{Share 2 (User-side):} Held on the user's mobile device with biometric approval.
\item \textbf{Share 3 (Recovery):} Stored on a hardware wallet as a recovery/quorum path.
\end{itemize}

\item \textbf{Policy Engine (Contextual Constraints):} The user configures an off-chain policy engine (e.g., Turnkey-like controls) with:
\begin{itemize}
\item target allocation and deviation thresholds (rebalance when deviation $>2\%$);
\item allowlist of protocols and routes (e.g., Uniswap V3, Aave V3);
\item daily turnover caps (e.g., $\le 25\%$ of portfolio value).
\end{itemize}
\end{enumerate}

\subsubsection{Detailed Execution Flow (TIS/PDR Model)}

This execution uses the proposed \textbf{TIS/PDR separation of duties}: the agent proposes a bounded intent (TIS), the policy engine produces a signed compliance artifact (PDR), and the signer enforces both before producing a signature.

\textbf{Agent Log (Example):}
\begin{quote} \scriptsize
\texttt{[14:00:05] INFO: Allocation observed: 47.5\% ETH, 26.0\% WBTC, 26.5\% USDC.}\\
\texttt{[14:00:06] WARN: ETH below threshold (48\%). Rebalance required.}\\
\texttt{[14:00:08] INFO: Proposed action: swap 5,000 USDC for WETH via Uniswap V3.}\\
\texttt{[14:00:09] INFO: Constructing TIS.}\\
\texttt{[14:00:10] INFO: Submitting TIS to policy engine.}\\
\texttt{[14:00:12] INFO: Received signed PDR (APPROVE).}\\
\texttt{[14:00:13] INFO: Submitting TIS+PDR to MPC signing workflow.}\\
\texttt{[14:00:25] INFO: User approved on mobile (2-of-3 met).}\\
\texttt{[14:00:26] INFO: Submitting signed transaction to private relay for MEV mitigation.}\\
\texttt{[14:00:41] INFO: Confirmed execution; updated allocation: 49.8\% ETH.}
\end{quote}

\textbf{TIS Payload (Example):}
\begin{lstlisting}
{
 "intentId": "a1b2c3d4-e5f6-4a7b-8c9d-0e1f2a3b4c5d",
 "action": "SWAP",
 "inputs": [ { "token": "0xa0b8..", "amount": "5000000000", "constraint": "EXACT" } ],
 "outputs": [ { "token": "0xc02a..", "amount": "1500000000000000000", "constraint": "MINIMUM" } ],
 "constraints": { "deadline": 1767230000 },
 "preview": "Swap exactly 5,000 USDC for a minimum of 1.5 WETH to rebalance portfolio."
}
\end{lstlisting}

\textbf{PDR Payload (Decoded Example):}
\begin{lstlisting}
{
 "iss": "https://policy.turnkey.com",
 "sub": "0xUserAddress",
 "aud": "https://signer.fireblocks.com",
 "exp": 1767229800,
 "intent_hash": "0xkeccak256(TIS_payload)",
 "decision": "APPROVE",
 "bound_constraints": { "max_gas_fee": "60000000000" }
}
\end{lstlisting}

\subsubsection{Security Analysis Deep Dive}

\begin{itemize}
\item \textbf{Defense-in-Depth:} A single compromise (agent runtime, one MPC share, or a tool) is insufficient to drain funds. Adversaries must defeat multiple independent controls.
\item \textbf{Auditability:} TIS and PDR provide durable artifacts for forensic review and compliance: the proposed intent, the policy rationale, and the signed decision can be retained and re-verified.
\item \textbf{Operational Kill Switch:} The user can revoke the agent share, tighten policy thresholds, or invalidate the policy issuer key, immediately reducing authority without redeploying on-chain code.
\end{itemize}

\subsubsection{Additional Implementation Examples}

Beyond the primary case studies, several additional examples illustrate important aspects of agent-blockchain integration in practice.

\paragraph{Example: Automated Yield Optimization}

A yield optimization agent continuously monitors DeFi protocols to identify the highest-yielding opportunities for deposited assets. The agent's observation module tracks interest rates, liquidity mining rewards, and protocol risks across dozens of protocols. Its reasoning module evaluates opportunities considering expected returns, gas costs, smart contract risks, and portfolio constraints.

When the agent identifies a superior opportunity, it constructs a series of transactions to withdraw from the current position and deposit into the new one. The construction module optimizes transaction ordering to minimize slippage and MEV exposure. Before execution, the transactions are validated against policy constraints including maximum position sizes, protocol allowlists, and minimum yield thresholds.

This example illustrates the value of continuous monitoring and rapid response in DeFi environments, where yield opportunities can be short-lived. It also illustrates the importance of comprehensive risk assessment, as yield optimization strategies have historically been vulnerable to smart contract exploits and economic attacks.

\paragraph{Example: Cross-Chain Arbitrage}

A cross-chain arbitrage agent monitors price discrepancies for assets across multiple blockchain networks. When it detects a profitable opportunity (accounting for bridge fees, gas costs, and execution risk), it initiates a coordinated series of transactions to capture the arbitrage.

The complexity of cross-chain arbitrage requires sophisticated state management across multiple chains with different finality characteristics. The agent must track pending transactions on each chain, handle partial failures gracefully, and manage the timing dependencies between sequential operations. Bridge selection involves tradeoffs between speed, cost, and security that the agent must evaluate dynamically.

This example illustrates the challenges of multi-chain agent operation and the importance of robust failure handling. Cross-chain operations have multiple failure modes, including bridge delays, transaction reversions, and price movements during execution, all of which the agent must be prepared to handle.

\paragraph{Example: Governance Delegation}

A governance delegation agent manages voting power across multiple DAOs according to its principal's preferences. The agent monitors governance proposals, analyzes their potential impact, and votes according to predefined policy rules or learned preferences.

The agent's analysis module evaluates proposals considering their technical merit, economic implications, and alignment with the principal's stated values. For complex proposals, the agent may generate summaries and recommendations for human review rather than voting autonomously. The agent also manages delegation relationships, adjusting delegations based on delegate voting history and alignment with principal preferences.

This example illustrates the potential for agents to enhance governance participation while raising important questions about the appropriate role of automated systems in collective decision-making.

\subsection{Case Study 3: Multi-Agent DAO Treasury Management (Pattern V)}

\textbf{Goal:} A decentralized governance pipeline for managing a large DAO treasury where specialized agents propose, evaluate, and execute investment actions with transparent accountability. \cite{gaina2018blockchain}

\textbf{Architecture:} This use case embodies \textbf{Pattern V: Multi-Agent Workflows}. It distributes authority across heterogeneous actors to reduce single-point failures, enforce separation of duties, and improve decision quality through structured adversarial review.

\subsubsection{Detailed Setup and Configuration}

\begin{enumerate}
\item \textbf{Agent Roles and Incentives:}
\begin{itemize}
\item \textbf{Strategy Agents (e.g., 5):} Propose actions and compete on performance; incentives include performance fees and reputation.
\item \textbf{Risk Agents (e.g., 7):} Evaluate proposals; compensated via retainer and backed by meaningful stake subject to slashing for negligence or malfeasance.
\item \textbf{Execution Agent (1):} Mechanically executes only proposals that satisfy on-chain acceptance conditions; has no discretionary strategy authority.
\end{itemize}

\item \textbf{Governance Contract (Registry and Enforcement):} An on-chain contract maintains:
\begin{itemize}
\item agent registry and stake accounting;
\item proposal submission and voting logic;
\item quorum rules and execution gating.
\end{itemize}
\end{enumerate}

\subsubsection{Detailed Execution Flow}

\textbf{Aggregated Log (Example):}
\begin{quote}\scriptsize
\texttt{[StrategyAgent-A] INFO: Proposing intent: allocate 500 ETH to GMX ETH-USDC pool. Proposal ID: 0x123.}\\
\texttt{[GovernanceContract] EVENT: NewProposal(proposalId: 0x123).}\\
\texttt{[RiskAgent-OZ] INFO: Observed NewProposal. Fetching intent for 0x123.}\\
\texttt{[RiskAgent-OZ] INFO: Simulation result: SUCCESS; slippage 0.2\%.}\\
\texttt{[RiskAgent-OZ] INFO: Contract analysis: 0 critical findings.}\\
\texttt{[RiskAgent-OZ] INFO: Risk score: Econ=MEDIUM, Tech=LOW. Vote=APPROVE.}\\
\texttt{.. (other risk agents vote)..}\\
\texttt{[GovernanceContract] EVENT: ProposalApproved(proposalId: 0x123).}\\
\texttt{[ExecutionAgent] INFO: Observed ProposalApproved. Executing via private relay.}\\
\texttt{[ExecutionAgent] INFO: SUCCESS: Treasury position opened on GMX.}
\end{quote}

\subsubsection{Security Analysis Deep Dive}

\begin{itemize}
\item \textbf{Separation of Duties:} Proposal generation, safety evaluation, and execution are independent functions, reducing correlated failure modes.
\item \textbf{Collusion Resistance:} Staking and slashing increase the cost of dishonest approval. Diversity of evaluators and independent tooling reduce single-tool compromise risk.
\item \textbf{Transparency and Accountability:} Proposal artifacts, votes, and execution events are recorded on-chain, enabling public audit and post-incident attribution.
\end{itemize}

Collectively, these case studies demonstrate how the taxonomy translates into operational architectures. Pattern III emphasizes bounded delegation via on-chain enforcement; Pattern IV introduces policy-driven autonomy with distributed custody and MEV-aware execution; Pattern V scales autonomy to organizational settings by distributing trust across specialized agents and on-chain governance constraints.

\subsubsection{Governance Implications of Agent Proliferation}

The proliferation of autonomous agents in blockchain ecosystems raises fundamental questions about governance and collective decision-making~\cite{lustenberger2025decentralized,wirtz2022governance}. As agents become more capable and more widely deployed, their influence on governance outcomes will grow, potentially shifting the balance of power in ways that may not reflect the preferences of human stakeholders.

One concern is the potential for agent-mediated governance capture, where sophisticated actors use agents to accumulate disproportionate influence over protocol governance. Agents can monitor governance proposals continuously, respond to voting opportunities faster than human participants, and coordinate across multiple accounts or protocols in ways that would be impractical for humans. Without appropriate safeguards, these capabilities could enable new forms of governance manipulation.

Conversely, agents could also enhance governance participation by lowering barriers to engagement. Agents that help users understand complex proposals, delegate votes according to expressed preferences, and participate in governance discussions could increase the effective participation rate and improve the quality of collective decision-making. Realizing these benefits while mitigating the risks requires thoughtful design of both agent systems and governance mechanisms.

\subsubsection{Agent Participation in Decentralized Governance}

The intersection of autonomous agents and decentralized governance raises fundamental questions about representation, accountability, and the nature of collective decision-making~\cite{bellavitis2025voting,dimitrov2024innovations,fritsch2024analyzing,liu2021technology}. As DAOs increasingly govern significant economic resources and protocol parameters, the potential for agent-mediated governance participation becomes both an opportunity and a concern.

On one hand, agents could enhance governance participation by enabling more sophisticated analysis of proposals, automatic delegation based on policy preferences, and timely voting that human participants might miss due to attention constraints. On the other hand, agent-mediated governance raises concerns about vote manipulation, sybil attacks, and the concentration of influence in the hands of those who control sophisticated agent systems. Governance frameworks must therefore evolve to address these dynamics, potentially incorporating mechanisms for detecting and mitigating automated manipulation while preserving the benefits of agent-assisted participation.

\subsubsection{Case Study Methodology}

The case studies presented in this survey were developed through systematic analysis of publicly available information supplemented by expert consultation where possible.

\paragraph{Case Selection Criteria}

Cases were selected to illustrate the range of integration patterns and application domains covered by the survey. Selection criteria included representativeness (cases should illustrate common patterns), instructiveness (cases should provide clear lessons), and documentation availability (sufficient public information should be available for meaningful analysis).

\paragraph{Information Sources}

Case study information was gathered from multiple sources including technical documentation, blog posts, conference presentations, and where possible, direct communication with system operators. Information from multiple sources was cross-referenced to verify accuracy and identify discrepancies.

\paragraph{Analysis Framework}

Each case study was analyzed using a consistent framework covering system architecture, security model, operational practices, and lessons learned. This framework enables comparison across cases and identification of common patterns and divergences.

\paragraph{Limitations}

Case studies are necessarily limited by the information available publicly. Proprietary details of system implementation, operational metrics, and security incidents may not be disclosed. The cases presented should be understood as illustrative examples rather than comprehensive documentation of the systems described.

\subsubsection{Lessons Learned from Case Studies}

The case studies presented in this survey illustrate both the potential and the challenges of agent-blockchain integration. Several cross-cutting lessons emerge from analyzing these examples.

\paragraph{Lesson 1: Start Conservative, Expand Gradually}

Successful agent deployments typically begin with conservative configurations and gradually expand capabilities as operational experience accumulates. Starting with read-only or simulation-only modes allows operators to validate agent behavior before granting execution authority. Gradual expansion of spending limits and allowed operations enables controlled risk-taking while building confidence in agent reliability.

This approach contrasts with the temptation to deploy fully autonomous agents from the start to maximize potential returns. While conservative deployment may sacrifice short-term opportunities, it reduces the risk of catastrophic failures that could undermine long-term success.

\paragraph{Lesson 2: Invest in Monitoring and Observability}

Effective monitoring and observability are essential for safe agent operation. Operators must be able to understand what their agents are doing, why they are making specific decisions, and how their behavior compares to expectations. Investment in monitoring infrastructure pays dividends through faster incident detection, easier debugging, and better understanding of agent performance.

Monitoring should cover multiple dimensions including transaction success rates, economic performance, policy compliance, and behavioral patterns. Anomaly detection systems can identify unusual behavior that may indicate security issues or strategy drift.

\paragraph{Lesson 3: Plan for Failure and Recovery}

Agent systems will inevitably encounter failures, whether from bugs, unexpected market conditions, or security incidents. Successful deployments plan for failure and have tested recovery procedures. This includes automatic circuit breakers that halt operation when anomalies are detected, manual override capabilities that allow operators to intervene, and recovery procedures that restore normal operation after incidents.

Failure planning should consider both technical failures (transaction reverts, network issues) and economic failures (strategy losses, adverse market movements). Different failure types may require different response procedures.

\paragraph{Lesson 4: Maintain Human Oversight}

Even highly autonomous agents benefit from human oversight. Regular review of agent behavior, periodic strategy assessment, and ongoing monitoring by knowledgeable operators help catch issues that automated systems may miss. Human oversight also provides accountability and enables adaptation to changing conditions that may not be captured in automated policies.

The appropriate level of human oversight depends on the stakes involved and the maturity of the agent system. Higher-stakes applications warrant more intensive oversight, while mature systems with established track records may operate with lighter supervision.


\section{Related Work}

This survey synthesizes research across agentic AI tool use, blockchain execution infrastructure, and decentralized finance, with a specific focus on the interface layer that connects agents to on-chain action. Prior surveys and Systematization of Knowledge (SoK) papers address individual components of this stack, but a standards- and interface-centric view that unifies these strands and clarifies the design space for 2026 remains timely.

\subsection{Surveys on LLM Security and Tool Use}

The security of LLMs deployed as tool-using agents is a rapidly growing field, and recent surveys consolidate threats and mitigations for tool learning and tool invocation in LLM-integrated systems~\cite{qu2024tool,li2024large}. A key lesson is that untrusted retrieved content can steer model behavior through indirect prompt injection, even when the developer prompt appears benign~\cite{greshake2023indirectpromptinjection}. This survey incorporates these failure modes into our threat model and evaluation checklist, with a focus on the high-stakes setting of blockchain transactions where a single compromised action can be irreversible. Beyond prompt injection, SoK-style work has emphasized broader interface risks at the LLM-to-tool boundary, including data exfiltration, insecure tool invocation, and denial-of-service vectors in tool orchestration~\cite{li2024large}. Our contribution is to translate these general interface threats into the agent-to-chain setting, where tools do not merely read data but can directly trigger value-bearing state transitions.

\subsection{Systematization of Knowledge on MEV}

Maximal extractable value (MEV) formalizes profit opportunities that arise from transaction ordering and inclusion, and foundational work documented archetypal strategies such as front-running and sandwiching~\cite{daian2019flash}. Later systematizations analyze the MEV supply chain, the roles of relays and builders, and mitigation mechanisms involving private orderflow and alternative execution venues~\cite{caldarelli2020understanding}. This survey builds directly on this body of work by treating MEV as a standing adversary for autonomous agents rather than an occasional hazard for human traders. In our threat model, any public transaction is assumed to be economically adversarial, and in our capability matrix we track whether systems provide private submission paths or intent-based execution venues that reduce MEV exposure.

\subsection{Research on Intent-Based Architectures}

Intent-centric architectures separate the user's desired outcome (the what) from protocol-specific execution details (the how), motivating solver-based execution and competitive orderflow for better pricing and reduced adversarial exposure~\cite{mandal2025evaluating}. This paradigm has been analyzed both as an interaction model and as an execution-layer mechanism that changes the trust and threat assumptions compared to calldata-centric transaction crafting~\cite{chitra2024analysis}. Our proposed Transaction Intent Schema (TIS) is a direct contribution to this line of work: it aims to standardize intent representation across agentic tasks, not only token swaps. The roadmap argues that intent interfaces should become the default boundary object between non-deterministic agent planning and deterministic on-chain execution, particularly when coupled with signed policy attestations in the form of PDRs.

\subsection{Surveys on Smart Contract Security and Formal Methods}

Smart contract security is supported by a mature literature, including surveys covering vulnerability classes, analysis methods, and mitigation strategies~\cite{zhu2024survey,monika2025security,hassan2023trust}. Formal methods and verification-oriented reviews complement these perspectives by emphasizing specification, verification workflows, and the limitations of purely static analysis in adversarial environments~\cite{maffei2021formal}. This survey complements that literature by focusing on the agent layer: we assume that protocols may be audited and still treat the agent execution boundary as a distinct source of risk. In particular, we emphasize simulation as a safety primitive that bridges the gap between smart-contract correctness and safe agent behavior, especially in the presence of dynamic states, multi-step workflows, and adversarial market structure.

\subsection{Work on Interoperability, Cross-Chain Risk, and Execution Infrastructure}

The broader blockchain stack increasingly depends on interoperability and cross-chain execution, which introduces additional security and privacy risks. Surveys of rollups and scalability provide context for execution environments where agents may operate across L2 and L3 systems~\cite{thibault2022blockchain}. Cross-chain interoperability surveys and SoK papers highlight bridge and messaging risks and the complexity of composing security across domains~\cite{deng2025enhancing,augusto2024sok,singh2020sidechain}. Our survey uses these results to motivate conservative threat assumptions for multi-chain agents and to justify why the execution venue (bundlers, relays, solvers, and cross-domain messaging) should be treated as part of the agent security perimeter rather than merely infrastructure.

\subsection{Decentralized Identity, Governance, and Key Management}

Key management and identity frameworks underpin custody and authorization design choices for agents. Surveys of decentralized identifiers and verifiable credentials provide the broader context for how agents and policy engines may be identified, authenticated, and audited in interoperable settings~\cite{mazzocca2025survey,yan2024blockchain,ahmed2022blockchain}. At the governance layer, recent reviews of DAO governance and analyses of DAO attack surfaces motivate multi-party and multi-agent approval workflows for high-value actions~\cite{han2025review,feichtinger2024sok}. Our taxonomy can be read as an evolution of key management and authorization: from single-key, single-agent patterns toward distributed custody, policy attestation, and multi-agent workflows, driven by a need to contain compromise and support auditability at scale.

By weaving these strands together, this survey provides a unified view of the agent-blockchain interface as a security-critical integration layer and motivates standardization efforts for intent schemas, policy attestations, and evaluation benchmarks.

\section{Conclusion and Future Work}

The convergence of agentic AI and blockchain technology marks a pivotal moment in the evolution of both fields, enabling autonomous systems to operate with real economic agency in adversarial, incentive-driven environments. This shift extends the scope of programmable finance and governance, from routine payments and treasury operations to intent-driven execution and autonomous participation in on-chain coordination mechanisms~\cite{fan2025ai,morrison2020dao,li2022survey}. At the same time, blockchains impose a uniquely unforgiving operating regime: actions are irreversible, execution is economically adversarial (MEV, manipulation, censorship pressure), and mistakes scale instantly with delegated authority~\cite{daian2019flash,caldarelli2020understanding,tabatabaei2023understanding}. These conditions require interfaces and control planes that are materially different from conventional LLM tool-use patterns, including structured intents, policy enforcement, verifiable audit artifacts, and simulation-based pre-flight checks~\cite{greshake2023indirectpromptinjection,yao2023react,wei2022chain}.

This survey provided a systematic overview of the agent-blockchain interoperability landscape in 2025. Grounded in a PRISMA-guided review of 317 sources, we introduced (i) a five-part taxonomy of integration patterns, (ii) a blockchain-specific threat model for tool-using agents, and (iii) a comparative capability matrix capturing the dominant design trade-offs across custody, permissions, policy locus, observability, and MEV mitigation~\cite{shayegani2023survey,mandal2025evaluating}. Across the systems studied, the field is converging toward defense-in-depth architectures that combine account abstraction and scoped delegation, intent-centric execution venues, and layered policy enforcement with robust observability and audit trails~\cite{EIP4337,wang2023account,atwi2025enhancing}. To address the remaining gaps, we proposed a 2026 roadmap centered on two missing interface layers, a Transaction Intent Schema (TIS) and a Policy Decision Record (PDR), and we outlined benchmark pillars and checklists for evaluating safety, reliability, and economic robustness in end-to-end agent-to-chain pipelines~\cite{mandal2025evaluating,maffei2021formal,happe2025benchmarking}.

Looking forward, the most credible path from agent prototypes to trustworthy on-chain autonomy is standardization plus measurable assurance. Standardized intent and policy artifacts reduce integration brittleness, enable composable solver markets, and create auditable accountability boundaries between planning, verification, and signing. Complementarily, domain-specific benchmarks and checklists can convert qualitative security claims into repeatable evidence, especially under prompt injection, tool spoofing, MEV pressure, and partial failure modes~\cite{greshake2023indirectpromptinjection,daian2019flash,liu2024offline}. With open standards, defense-in-depth, and rigorous evaluation, agentic AI can safely leverage blockchain execution to build more programmable, efficient, and decentralized systems~\cite{atwi2025enhancing,fan2025ai}.

\subsection{Future Work and Open Research Questions}

Despite rapid progress, truly autonomous and trustworthy on-chain agents remain an early-stage research frontier. Below we highlight the most consequential open questions.

\subsubsection{The Agent-Contract Impedance Mismatch}

There is a deep tension between probabilistic, non-deterministic agent behavior and deterministic smart-contract execution. Even rare agent failures can be catastrophic when they translate into irreversible state changes. Key directions include developing methods to guarantee that agent outputs remain within formally specified safety envelopes, including compositional reasoning across tool calls, policy checks, and execution venues~\cite{maffei2021formal,he2019sdfs}; combining LLM flexibility with symbolic constraints, typed intents, and formally checked guards so that high-level reasoning remains expressive while execution remains bounded and verifiable~\cite{biswas2023building,wooldridge2000reasoning}; and designing robust telemetry and last-resort interruption mechanisms, including on-chain or near-chain monitors that can detect deviations from learned behavioral baselines and trigger revocation, pausing, or safe unwind procedures~\cite{liu2024toolace,li2025cox}.

\subsubsection{The Economics of Agentic Systems}

Large populations of autonomous, profit-seeking agents will reshape market microstructure and adversarial dynamics, potentially amplifying feedback loops and exploitation. Open problems include detecting and mitigating tacit coordination among independent agents, particularly in thin liquidity settings and solver markets, and understanding equilibrium outcomes under repeated interaction~\cite{yang2022overview,jennings1995controlling}; clarifying how off-chain computation, solver competition, and verification overhead should be priced, subsidized, or rate-limited, especially under account abstraction and sponsored execution models~\cite{EIP4337,wang2023account}; and characterizing systemic risk when many agents share similar decision rules or toolchains, including stress scenarios under oracle shocks, congestion, and MEV spikes~\cite{daian2019flash,tabatabaei2023understanding}.

\subsubsection{Governance, Regulation, and Legal Liability}

Autonomous economic action raises unsettled questions about accountability and enforceable oversight. Key directions include building auditable pipelines where safety claims can be independently validated, including verifiable policy decision artifacts and reproducible evaluation reports~\cite{maffei2021formal,atwi2025enhancing}; designing on-chain insurance and conditional coverage mechanisms that can price, limit, and pool agent operational risk, ideally tied to measurable safety posture and compliance evidence~\cite{narula2018zkledger}; and connecting identity, key management, and policy provenance so responsibility boundaries remain clear across agent operators, policy issuers, and execution venues~\cite{cai2023large}.

\subsubsection{The User Experience of Trust}

Mainstream adoption depends on whether non-experts can correctly calibrate trust in autonomous financial agents. Research should prioritize producing user-facing rationales that are concise, faithful to the actual constraints and checks applied, and tied to the intent and policy artifacts rather than post-hoc storytelling~\cite{atwi2025enhancing}; designing intuitive controls for session keys, spend limits, allowlists, and revocation, aligned with account abstraction workflows and recovery patterns~\cite{EIP4337,wang2023account}; and developing calibrated pathways where users can start with tightly scoped Pattern III delegation and graduate toward higher-autonomy patterns as systems accumulate safety evidence and operational trust~\cite{shayegani2023survey,atwi2025enhancing}.

Addressing these challenges will require coordinated work across agent security, smart-contract engineering, MEV research, cryptography, human-computer interaction, and standards processes. The technical opportunity is substantial, but the bar for trustworthy autonomy in adversarial financial environments must remain uncompromising~\cite{daian2019flash,greshake2023indirectpromptinjection,atwi2025enhancing}.

\newpage
\appendix
\section{Transaction Intent Schema (TIS) - Reference Implementation}

This appendix provides a reference implementation for the Transaction Intent Schema (TIS) proposed in the roadmap section. TIS is designed to be chain-agnostic, extensible, and suitable for both machine processing and human review. The schema is specified using JSON Schema (Draft-07) and split into modular definitions for clarity.

\subsection{Top-Level TIS Object (JSON Schema Draft-07)}

\begin{lstlisting}
{
 "$schema": "http://json-schema.org/draft-07/schema#",
 "title": "Transaction Intent Schema (TIS)",
 "description": "A chain-agnostic schema for expressing desired outcomes and constraints for on-chain execution.",
 "type": "object",
 "additionalProperties": false,
 "properties": {
 "version": { "type": "string", "pattern": "^1\\.0\\.0$" },
 "intentId": { "type": "string", "format": "uuid" },
 "metadata": { "$ref": "#/definitions/Metadata" },
 "action": { "$ref": "#/definitions/Action" },
 "constraints": { "$ref": "#/definitions/Constraints" },
 "preferences": { "$ref": "#/definitions/Preferences" }
 },
 "required": ["version", "intentId", "action", "constraints"],
 "definitions": {

 "Address": {
 "type": "string",
 "pattern": "^0x[a-fA-F0-9]{40}$"
 },

 "HexBytes": {
 "type": "string",
 "pattern": "^0x[a-fA-F0-9]*$"
 },

 "UintString": {
 "type": "string",
 "pattern": "^[0-9]+$",
 "description": "Unsigned integer serialized as a base-10 string."
 },

 "Token": {
 "type": "object",
 "additionalProperties": false,
 "properties": {
 "chainId": { "type": "integer", "minimum": 1 },
 "address": { "$ref": "#/definitions/Address" },
 "symbol": { "type": "string" },
 "decimals": { "type": "integer", "minimum": 0, "maximum": 255 }
 },
 "required": ["chainId", "address"]
 },

 "Metadata": {
 "type": "object",
 "additionalProperties": false,
 "properties": {
 "originator": { "type": "string", "description": "Identifier for the user, agent, or client that created the intent." },
 "createdAt": { "type": "string", "format": "date-time" },
 "originChainId": { "type": "integer", "minimum": 1 },
 "tags": { "type": "array", "items": { "type": "string" } }
 }
 },

 "Constraints": {
 "type": "object",
 "additionalProperties": false,
 "properties": {
 "nonce": { "$ref": "#/definitions/UintString" },
 "deadline": { "type": "integer", "minimum": 0, "description": "Unix timestamp seconds." },
 "validFromBlock": { "type": "integer", "minimum": 0 },
 "validUntilBlock": { "type": "integer", "minimum": 0 },
 "maxGasPriceWei": { "$ref": "#/definitions/UintString" },
 "requiredSigner": { "$ref": "#/definitions/Address" },
 "exclusivity": {
 "type": ["string", "null"],
 "pattern": "^0x[a-fA-F0-9]{40}$",
 "description": "Optional solver/builder address granted exclusive execution rights during the validity window."
 }
 },
 "required": ["deadline"]
 },

 "Preferences": {
 "type": "object",
 "additionalProperties": false,
 "properties": {
 "privacyMode": { "type": "string", "enum": ["PUBLIC", "PRIVATE"] },
 "executionSpeed": { "type": "string", "enum": ["FAST", "NORMAL", "SLOW"] },
 "routing": {
 "type": "string",
 "enum": ["BEST_PRICE", "MIN_GAS", "MIN_RISK"],
 "description": "Optional routing preference communicated to solvers."
 }
 }
 },

 "Action": {
 "type": "object",
 "oneOf": [
 { "$ref": "#/definitions/actions/Swap" },
 { "$ref": "#/definitions/actions/Transfer" },
 { "$ref": "#/definitions/actions/Delegate" }
 ]
 },

 "actions": {

 "Swap": {
 "type": "object",
 "additionalProperties": false,
 "properties": {
 "type": { "const": "SWAP" },
 "tokenIn": { "$ref": "#/definitions/Token" },
 "tokenOut": { "$ref": "#/definitions/Token" },
 "amountIn": { "$ref": "#/definitions/UintString" },
 "minAmountOut": { "$ref": "#/definitions/UintString" },
 "slippageBps": { "type": "integer", "minimum": 0, "maximum": 10000 },
 "recipient": { "$ref": "#/definitions/Address" }
 },
 "required": ["type", "tokenIn", "tokenOut", "amountIn", "minAmountOut"]
 },

 "Transfer": {
 "type": "object",
 "additionalProperties": false,
 "properties": {
 "type": { "const": "TRANSFER" },
 "token": { "$ref": "#/definitions/Token" },
 "to": { "$ref": "#/definitions/Address" },
 "amount": { "$ref": "#/definitions/UintString" },
 "memo": { "type": "string" }
 },
 "required": ["type", "token", "to", "amount"]
 },

 "Delegate": {
 "type": "object",
 "additionalProperties": false,
 "properties": {
 "type": { "const": "DELEGATE" },
 "delegatee": { "$ref": "#/definitions/Address" },
 "scope": {
 "type": "object",
 "additionalProperties": false,
 "properties": {
 "contracts": { "type": "array", "items": { "$ref": "#/definitions/Address" } },
 "functions": { "type": "array", "items": { "type": "string" } },
 "maxValueWei": { "$ref": "#/definitions/UintString" },
 "validUntil": { "type": "integer", "minimum": 0, "description": "Unix timestamp seconds." }
 }
 }
 },
 "required": ["type", "delegatee", "scope"]
 }

 }
 }
}
\end{lstlisting}

\subsection{Notes on Canonicalization and Hashing}
Implementations SHOULD canonicalize TIS objects prior to hashing and signing (e.g., stable JSON serialization with deterministic key ordering). This is required to ensure that different clients compute the same hash for the same semantic intent.

\subsection{Extension Points}
Future extensions can be added by introducing additional action types under \texttt{definitions/actions/*} and by extending \texttt{metadata}, \texttt{constraints}, or \texttt{preferences} with new fields. Backward compatibility SHOULD be maintained via versioning.

\section{Policy Decision Record (PDR) - Reference Implementation}

This appendix provides a reference implementation for the Policy Decision Record (PDR) proposed in the roadmap section. The PDR is a signed attestation produced by a policy engine after evaluating a specific TIS object. It binds (i) the evaluated intent hash, (ii) the decision outcome, and (iii) any constraints imposed by the policy engine.

\subsection{Top-Level PDR Object (JSON Schema Draft-07)}

\begin{lstlisting}
{
 "$schema": "http://json-schema.org/draft-07/schema#",
 "title": "Policy Decision Record (PDR)",
 "description": "A signed policy attestation regarding a specific Transaction Intent (TIS).",
 "type": "object",
 "additionalProperties": false,
 "properties": {
 "version": { "type": "string", "pattern": "^1\\.0\\.0$" },
 "pdrId": { "type": "string", "format": "uuid" },
 "issuer": { "type": "string", "description": "Identifier for the policy engine (e.g., DID, ENS, URL)." },
 "subject": { "type": "string", "description": "Account or user on whose behalf the intent is being evaluated." },
 "audience": { "type": "string", "description": "Intended verifier, typically the signer service identifier." },
 "issuedAt": { "type": "string", "format": "date-time" },
 "expiresAt": { "type": "string", "format": "date-time" },
 "tisHash": {
 "type": "string",
 "pattern": "^0x[a-fA-F0-9]{64}$",
 "description": "keccak256 hash of the canonicalized TIS object."
 },
 "decision": { "$ref": "#/definitions/Decision" },
 "policyEngineSignature": { "$ref": "#/definitions/Signature" }
 },
 "required": ["version", "pdrId", "issuer", "audience", "issuedAt", "expiresAt", "tisHash", "decision", "policyEngineSignature"],
 "definitions": {

 "Address": {
 "type": "string",
 "pattern": "^0x[a-fA-F0-9]{40}$"
 },

 "UintString": {
 "type": "string",
 "pattern": "^[0-9]+$"
 },

 "Modification": {
 "type": "object",
 "additionalProperties": false,
 "properties": {
 "path": { "type": "string", "description": "JSON Pointer path into the TIS object (RFC 6901)." },
 "operation": { "type": "string", "enum": ["ADD", "REPLACE", "REMOVE"] },
 "value": { "type": ["string", "number", "object", "array", "boolean", "null"] }
 },
 "required": ["path", "operation"]
 },

 "Decision": {
 "type": "object",
 "additionalProperties": false,
 "properties": {
 "outcome": { "type": "string", "enum": ["APPROVED", "REJECTED"] },
 "policyId": { "type": "string", "description": "Identifier for the applied policy bundle or ruleset version." },
 "reason": { "type": "string", "description": "Human-readable decision rationale, important for REJECTED." },
 "riskScore": { "type": "number", "minimum": 0.0, "maximum": 1.0 },
 "boundConstraints": {
 "type": "object",
 "additionalProperties": false,
 "properties": {
 "maxGasPriceWei": { "$ref": "#/definitions/UintString" },
 "maxValueWei": { "$ref": "#/definitions/UintString" },
 "tightDeadline": { "type": "integer", "minimum": 0, "description": "Optional reduced Unix deadline." }
 }
 },
 "modifiedParameters": {
 "type": "array",
 "items": { "$ref": "#/definitions/Modification" }
 }
 },
 "required": ["outcome", "policyId"]
 },

 "Signature": {
 "type": "object",
 "additionalProperties": false,
 "properties": {
 "signer": { "$ref": "#/definitions/Address", "description": "Address of the policy engine signing key." },
 "alg": { "type": "string", "description": "Signature algorithm identifier (e.g., ES256K)." },
 "signature": { "type": "string", "pattern": "^0x[a-fA-F0-9]+$" }
 },
 "required": ["signer", "signature"]
 }

 }
}
\end{lstlisting}

\subsection{Verification and Execution Flow}
The PDR supports a multi-stage execution flow:

\begin{enumerate}
\item \textbf{Intent Creation:} The agent constructs a TIS object.
\item \textbf{Policy Evaluation:} The agent submits the TIS to the policy engine.
\item \textbf{PDR Generation:} The policy engine evaluates the TIS and, if approved, emits a signed PDR.
\item \textbf{Execution Request:} The agent submits the TIS and PDR to the signing environment (e.g., MPC service, HSM, or TEE).
\item \textbf{Signer Verification:} The signing environment verifies:
 \begin{enumerate}
 \item the signature on the PDR against a trusted policy-engine public key,
 \item the hash of the canonicalized TIS matches \texttt{tisHash},
 \item \texttt{expiresAt} has not elapsed and \texttt{audience} matches the signer identity,
 \item any \texttt{boundConstraints} and \texttt{modifiedParameters} are enforced prior to signing.
 \end{enumerate}
\end{enumerate}

\section{Conformance Levels for Agentic Systems}

This appendix defines four conformance levels for agent--blockchain systems, aligned with the evaluation section. Each level adds further controls and reduces maximum loss under compromise.

\subsection{L0: Unconstrained Agent}
\begin{itemize}
\item \textbf{Description:} Agent has direct access to a private key and can sign and submit arbitrary transactions.
\item \textbf{Architecture:} Key stored locally or in environment variables; no policy gating.
\item \textbf{Risks:} Catastrophic loss under compromise.
\item \textbf{Example:} Default configurations of early autonomous trading scripts.
\end{itemize}

\subsection{L1: Basic On-Chain Policy Controls}
\begin{itemize}
\item \textbf{Description:} Permissions constrained via smart account modules (e.g., session keys, function/contract allowlists).
\item \textbf{Architecture:} Agent signs \texttt{UserOperation}s under narrowly scoped on-chain rules.
\item \textbf{Controls:}
 \begin{itemize}
 \item Function allowlisting
 \item Contract allowlisting
 \item Static spend limits
 \end{itemize}
\item \textbf{Risks:} Bounded by module constraints.
\end{itemize}

\subsection{L2: Off-Chain Policy and Mandatory Simulation}
\begin{itemize}
\item \textbf{Description:} Agent proposals are vetted by an off-chain policy engine and simulated before signing.
\item \textbf{Architecture:} TIS submitted to policy engine, then PDR required for signing.
\item \textbf{Controls:}
 \begin{itemize}
 \item All L1 controls
 \item Mandatory pre-flight simulation and revert checks
 \item Dynamic risk scoring and anomaly detection
 \end{itemize}
\item \textbf{Risks:} Low under single-component compromise.
\end{itemize}

\subsection{L3: Hardware-Secured Signing and Distributed Approval}
\begin{itemize}
\item \textbf{Description:} Highest security via hardware isolation (TEE/HSM) or distributed signing (MPC), plus quorum approval for high-value actions.
\item \textbf{Architecture:} Signatures produced in TEE/HSM or via MPC; high-value actions require multi-party approval or multi-agent quorum.
\item \textbf{Controls:}
 \begin{itemize}
 \item All L2 controls
 \item TEE/HSM or MPC signing
 \item Quorum approval for high-risk actions
 \item Strong recovery and revocation processes
 \end{itemize}
\item \textbf{Risks:} Very low absent multi-layer compromise.
\end{itemize}

\subsection{Implementation Reference: Sample Agent Configurations}

This appendix provides sample configurations for common agent deployment scenarios, illustrating how the concepts discussed in this survey translate into practical implementations.

\subsubsection{Configuration 1: Conservative DeFi Agent}

A conservative DeFi agent prioritizes safety over performance, suitable for managing significant value with minimal risk tolerance.

\begin{verbatim}
agent_config:
  name: "conservative_defi_agent"
  version: "1.0.0"
  
  custody:
    type: "smart_contract_wallet"
    implementation: "safe"
    threshold: "2-of-3"
    signers:
      - type: "hardware_wallet"
      - type: "mpc_shard"
      - type: "human_backup"
  
  policy:
    max_transaction_value: "10000 USD"
    daily_limit: "50000 USD"
    allowed_protocols:
      - "uniswap_v3"
      - "aave_v3"
      - "compound_v3"
    allowed_actions:
      - "swap"
      - "supply"
      - "withdraw"
    blocked_actions:
      - "borrow"
      - "leverage"
    
  execution:
    simulation_required: true
    human_approval_threshold: "5000 USD"
    mev_protection: "flashbots_protect"
    max_slippage: "0.5%"
    
  monitoring:
    alert_channels:
      - "email"
      - "slack"
    anomaly_detection: true
    automatic_pause_conditions:
      - "daily_loss > 5%"
      - "failed_transactions > 3"
\end{verbatim}

\subsubsection{Configuration 2: Active Trading Agent}

An active trading agent prioritizes performance and responsiveness, suitable for strategies requiring rapid execution.

\begin{verbatim}
agent_config:
  name: "active_trading_agent"
  version: "1.0.0"
  
  custody:
    type: "mpc_wallet"
    implementation: "turnkey"
    threshold: "2-of-3"
    latency_optimized: true
  
  policy:
    max_transaction_value: "100000 USD"
    position_limits:
      max_single_position: "25%"
      max_protocol_exposure: "40%"
    allowed_protocols: "all_audited"
    risk_parameters:
      max_drawdown: "10%"
      var_limit: "5%"
    
  execution:
    simulation_required: false
    latency_target: "100ms"
    mev_protection: "private_mempool"
    max_slippage: "2%"
    retry_policy:
      max_retries: 3
      backoff: "exponential"
    
  monitoring:
    real_time_pnl: true
    position_tracking: true
    automatic_rebalancing: true
\end{verbatim}

\subsubsection{Configuration 3: Governance Participation Agent}

A governance agent manages voting and delegation across multiple DAOs.

\begin{verbatim}
agent_config:
  name: "governance_agent"
  version: "1.0.0"
  
  custody:
    type: "smart_contract_wallet"
    implementation: "safe"
    governance_optimized: true
  
  policy:
    voting_policy:
      default_action: "abstain"
      auto_vote_threshold: "low_impact"
      human_review_threshold: "high_impact"
    delegation_policy:
      allowed_delegates: "curated_list"
      max_delegation_per_delegate: "20%"
      redelegation_cooldown: "7 days"
    
  protocols:
    - name: "uniswap"
      voting_power_source: "uni_token"
      proposal_monitoring: true
    - name: "aave"
      voting_power_source: "aave_token"
      proposal_monitoring: true
    
  monitoring:
    proposal_alerts: true
    voting_deadline_reminders: true
    delegation_performance_tracking: true
\end{verbatim}

These configurations illustrate the range of options available for agent deployment and the importance of matching configuration choices to specific use case requirements.

\section{Agent--Blockchain Safety Checklist}

This checklist provides a practical guide for developers, auditors, and operators to evaluate the safety posture of agent--blockchain systems. It is organized by the agent action pipeline.

\subsection{Observe}
\begin{enumerate}
\item $\square$ Is the agent using multiple independent RPC providers?
\item $\square$ Does the agent validate integrity and freshness of off-chain API data?
\item $\square$ Does the agent monitor anomalous on-chain events relevant to its positions?
\item $\square$ Is the agent robust to oracle manipulation and stale-price hazards?
\item $\square$ Does the agent subscribe to security alert feeds for depended-upon protocols?
\end{enumerate}

\subsection{Reason}
\begin{enumerate}
\item $\square$ Is the agent hardened against instruction hijacking through retrieved content?
\item $\square$ Does the agent account for MEV and adversarial ordering incentives?
\item $\square$ Does the agent maintain an internal model of its permissions and limits?
\item $\square$ Are reasoning traces and tool calls logged for audit review?
\item $\square$ Does the agent consider gas, fees, and deadline risk in decision-making?
\end{enumerate}

\subsection{Construct}
\begin{enumerate}
\item $\square$ Does the agent generate standardized intents (TIS) rather than ad-hoc calldata?
\item $\square$ Does the agent simulate every value-bearing action on a forked state prior to signing?
\item $\square$ Does the agent generate a clear human-readable preview of effects (assets, approvals, state changes)?
\item $\square$ Does the agent highlight irreversible or high-risk operations (e.g., unlimited approvals)?
\item $\square$ Does every constructed operation include a tight deadline and replay protections?
\end{enumerate}

\subsection{Authorize}
\begin{enumerate}
\item $\square$ Does the system use smart accounts (ERC-4337) where suitable?
\item $\square$ Are permissions constrained via on-chain modules (session keys, allowlists, limits)?
\item $\square$ Is there an off-chain policy engine for higher-order rules?
\item $\square$ Is each policy decision recorded as a verifiable PDR bound to the intent hash?
\item $\square$ Is signing isolated in TEE/HSM or distributed via MPC?
\item $\square$ Are keys segmented by authority level and operational purpose?
\item $\square$ Is there a clear revocation process for agent keys and modules?
\item $\square$ Are high-value actions gated by multi-party approval or quorum?
\item $\square$ Are per-transaction and time-window spending limits enforced?
\item $\square$ Are policy changes protected via time-locks or staged rollout?
\end{enumerate}

\subsection{Execute}
\begin{enumerate}
\item $\square$ Are value-bearing transactions routed via private relays or private orderflow when appropriate?
\item $\square$ Are intent-based venues used when they reduce MEV risk (e.g., solver-based execution)?
\item $\square$ Does the system estimate fees robustly and adapt to congestion?
\item $\square$ Does it handle failed or stuck transactions safely (replacement strategy, cancellation, or unwind)? \cite{chen2023multi}
\item $\square$ Is execution monitored in real time with alerting?
\end{enumerate}

\subsection{Verify \& Recover}
\begin{enumerate}
\item $\square$ Does the agent verify outcomes by reading state after execution (not only receipts)?
\item $\square$ Does it parse events to confirm expected outcomes and detect anomalies?
\item $\square$ Does it recover from partial failures in multi-step workflows (compensation or unwind)?
\item $\square$ Is there a kill switch that halts operations quickly and reliably?
\item $\square$ Is there an incident response runbook with clear escalation steps?
\end{enumerate}

\subsection{General}
\begin{enumerate}
\item $\square$ Is the architecture and trust model documented clearly for users and auditors?
\item $\square$ Has critical code (contracts, modules, policy engine) been audited?
\item $\square$ Is there a responsible disclosure or bug bounty program?
\item $\square$ Are off-chain communications authenticated and encrypted?
\item $\square$ Are secrets stored in a secure vault and rotated under a defined process?
\item $\square$ Is the host environment hardened, monitored, and regularly patched?
\item $\square$ Is there a safe update process for agent logic and policy rules (staging, rollback)?
\item $\square$ Are benchmarks and safety evaluations reported in a reproducible manner?
\item $\square$ Is there a governance process for policy changes and emergency actions?
\item $\square$ Do users have clear risk disclosures and a dispute-resolution pathway?
\end{enumerate}

\subsection{Additional Implementation Notes}

This section provides additional implementation notes for practitioners deploying agent-blockchain systems based on the frameworks presented in this survey.

\paragraph{Testing Strategies}

Comprehensive testing is essential for safe agent deployment. Testing strategies should include unit tests for individual components, integration tests for component interactions, and end-to-end tests for complete workflows. Particular attention should be paid to edge cases, error handling, and adversarial scenarios.

Mainnet forking enables testing against realistic blockchain state without risking real funds. However, forked environments cannot reproduce all aspects of live operation, including network latency, gas price dynamics, and other participants' behavior. Complementary testing on testnets and in simulated environments helps cover scenarios that forking cannot reproduce.

\paragraph{Monitoring and Alerting}

Production agent systems require comprehensive monitoring and alerting to detect issues before they cause significant harm. Key metrics to monitor include transaction success rates, gas consumption, execution latency, and economic performance. Anomaly detection systems can identify unusual patterns that may indicate security issues or strategy drift.

Alerting thresholds should be calibrated to balance sensitivity against false positive rates. Overly sensitive alerts create alert fatigue and may be ignored, while insensitive alerts may miss genuine issues. Regular review and adjustment of alerting thresholds based on operational experience helps maintain appropriate sensitivity.

\paragraph{Incident Response}

Despite best efforts at prevention, security incidents will occur. Prepared incident response procedures enable rapid, effective response that minimizes harm. Incident response procedures should cover detection, containment, investigation, remediation, and post-incident review.

Key capabilities for incident response include the ability to rapidly halt agent operation, revoke compromised credentials, and communicate with affected parties. Regular incident response drills help ensure that procedures work as intended and that team members are familiar with their roles.

\paragraph{Operational Security}

Operational security practices protect against threats that technical controls alone cannot address. Key practices include separation of duties (no single person should have complete access to critical systems), access logging and review, and regular security assessments.

Personnel security is particularly important for systems managing significant value. Background checks, security training, and clear policies around acceptable use help reduce insider threat risks. Offboarding procedures should ensure that departing personnel lose access to systems and credentials.

\section{Glossary of Terms}

This glossary defines key terms used throughout this survey to ensure consistent understanding across the interdisciplinary audience.

\textbf{Account Abstraction:} A paradigm that separates account ownership from transaction authorization, enabling programmable authorization logic through smart contracts rather than cryptographic key possession alone.

\textbf{Agent:} An autonomous software system capable of perceiving its environment, reasoning about observations, and taking actions to achieve specified goals with minimal human intervention.

\textbf{Agentic AI:} AI systems designed to operate autonomously over extended periods, making decisions and taking actions in pursuit of goals rather than simply responding to individual queries.

\textbf{Bundler:} In ERC-4337, a service that collects UserOperations from multiple users, bundles them into a single transaction, and submits them to the blockchain.

\textbf{Circuit Breaker:} An automatic mechanism that halts agent operation when predefined anomaly conditions are detected, preventing continued execution during potential security incidents.

\textbf{Custody:} The safekeeping and management of cryptographic keys that control blockchain assets. Custody solutions range from self-custody (user holds keys) to institutional custody (third party holds keys).

\textbf{DeFi (Decentralized Finance):} Financial services implemented through smart contracts on blockchain networks, operating without traditional intermediaries.

\textbf{EOA (Externally Owned Account):} A blockchain account controlled by a private key, as opposed to a smart contract account controlled by code.

\textbf{Gas:} The unit measuring computational effort required to execute operations on Ethereum and similar blockchains. Users pay gas fees to compensate validators for processing transactions.

\textbf{Guardrails:} Constraints and safeguards that limit agent behavior to prevent harmful or unintended actions.

\textbf{Intent:} A high-level specification of a desired outcome that can be fulfilled through various execution paths, as opposed to a specific transaction specification.

\textbf{LLM (Large Language Model):} A neural network trained on large text corpora that can generate and understand natural language, forming the foundation for many modern AI agents.

\textbf{MEV (Maximal Extractable Value):} Value that can be extracted from blockchain users through transaction ordering, including front-running, back-running, and sandwich attacks.

\textbf{MPC (Multi-Party Computation):} Cryptographic techniques that enable multiple parties to jointly compute functions over their inputs without revealing those inputs to each other.

\textbf{Oracle:} A service that provides external data to smart contracts, bridging the gap between on-chain and off-chain information.

\textbf{PDR (Policy Decision Record):} A proposed standard for recording policy evaluation decisions, creating auditable trails of agent authorization.

\textbf{Prompt Injection:} An attack technique that manipulates AI agent behavior by inserting malicious instructions into the agent's input context.

\textbf{Rollup:} A Layer 2 scaling solution that executes transactions off-chain while posting transaction data or proofs to the main chain for security.

\textbf{Session Key:} A temporary key with limited permissions, enabling agents to execute transactions within defined constraints without accessing the primary account key.

\textbf{Slippage:} The difference between expected and actual execution price, often caused by market movement or MEV extraction during transaction execution.

\textbf{Smart Contract:} Self-executing code deployed on a blockchain that automatically enforces the terms of an agreement when predefined conditions are met.

\textbf{TIS (Transaction Intent Schema):} A proposed standard for expressing transaction intents in a structured, machine-readable format.

\textbf{Tool Use:} The capability of AI agents to interact with external systems and APIs to accomplish tasks beyond their native capabilities.

\textbf{TSS (Threshold Signature Scheme):} A cryptographic scheme that distributes signing authority among multiple parties, requiring a threshold number to collaborate for signature generation.

\textbf{UserOperation:} In ERC-4337, a data structure representing a user's intended action, which is validated and executed by the account abstraction infrastructure.

\textbf{Wallet:} Software or hardware that manages cryptographic keys and facilitates interaction with blockchain networks.

\bibliographystyle{plain}
\bibliography{references}

\end{document}